%% file: main.tex
\documentclass[10pt,twocolumn,letterpaper]{article}

\usepackage[pagenumbers]{cvpr}              %

\input{preamble}

\definecolor{cvprblue}{rgb}{0.21,0.49,0.74}
\usepackage[pagebackref,breaklinks,colorlinks,allcolors=cvprblue]{hyperref}

\def\paperID{8724} %
\def\confName{CVPR}
\def\confYear{2026}

\title{GLOW: Global Illumination-Aware Inverse Rendering of Indoor Scenes Captured with Dynamic Co-Located Light \& Camera}

\author{
Jiaye Wu$^1$ \quad
Saeed Hadadan$^1$ \quad
Geng Lin$^1$ \quad
Peihan Tu$^1$ \\
Auguste Gezalyan$^1$ \quad
Matthias Zwicker$^1$ \quad
David Jacobs$^1$ \quad
Roni Sengupta$^2$ \\
$^1$University of Maryland, College Park \quad
$^2$University of North Carolina at Chapel Hill
}

\begin{document}
\twocolumn[{%
\renewcommand\twocolumn[1][]{#1}%
\maketitle
\includegraphics[width=\linewidth]{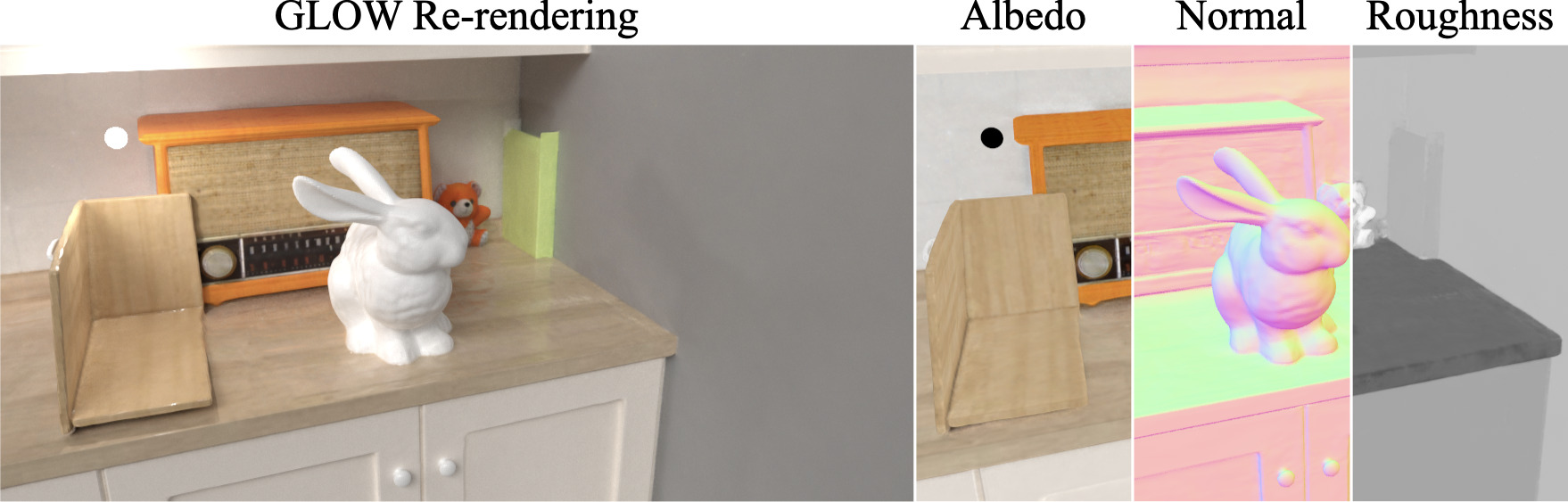}
\captionof{figure}{We perform inverse rendering of a scene with non-trivial global illumination from multiple images captured with a co-located light and camera. The image on the left shows a rendering of the reconstructed scene under different illumination, next to the recovered geometry and material properties on the right.}
\label{fig:teaser}
\vspace{1em}
}]
\input{sections/0_abstract}    
\input{sections/introduction_roni}

\input{sections/related_works}

\input{sections/methods}

\input{sections/experiments_alt}

\input{sections/conclusion}

{
    \small
    \bibliographystyle{ieeenat_fullname}
    \bibliography{main}
}
\input{sections/supp}

\end{document}

%% file: preamble.tex
\usepackage{bbm}
\usepackage[dvipsnames]{xcolor}
\usepackage{multirow}
\usepackage{multicol}
\usepackage[accsupp]{axessibility} 
\usepackage{amsthm}
\newtheorem{theorem}{Theorem}[section]

\newtheorem{lemma}[theorem]{Lemma}

\newcommand {\omegain}{\mathbf{\omega_{\mathrm{i}}}}
\newcommand {\omegaout}{\mathbf{\omega_{\mathrm{o}}}}

\makeatletter
\newcommand{\providelength}[1]{%
  \@ifundefined{\expandafter\@gobble\string#1}
   {%
    \typeout{\string\providelength: making new length \string#1}%
    \newlength{#1}%
   }
   {%
   }%
}
\makeatother

%% file: sections/0_abstract.tex
\begin{abstract}

Inverse rendering of indoor scenes remains challenging due to the ambiguity between reflectance and lighting, exacerbated by inter-reflections among multiple objects. While natural illumination-based methods struggle to resolve this ambiguity, co-located light–camera setups offer better disentanglement as lighting can be easily calibrated via Structure-from-Motion. However, such setups introduce additional complexities like strong inter-reflections, dynamic shadows, near-field lighting, and moving specular highlights, which existing approaches fail to handle. We present GLOW, a Global Illumination-aware Inverse Rendering framework designed to address these challenges. GLOW integrates a neural implicit surface representation with a neural radiance cache to approximate global illumination, jointly optimizing geometry and reflectance through carefully designed regularization and initialization. We then introduce a dynamic radiance cache that adapts to sharp lighting discontinuities from near-field motion, and a surface-angle-weighted radiometric loss to suppress specular artifacts common in flashlight captures. Experiments show that GLOW substantially outperforms prior methods in material reflectance estimation under both natural and co-located illumination.

\end{abstract}

%% file: sections/introduction_roni.tex
\vspace{-1em}
\section{Introduction}

Inverse Rendering (IR) aims to decompose images of a scene into its intrinsic components—geometry, material reflectance, and lighting. Accurate IR enables many downstream applications in AR/VR, robotics, and computational photography, including relighting and material editing~\cite{zhuLearningbasedInverseRendering2022, liInverseRenderingComplex2020, senguptaNeuralInverseRendering2019}. However, this task remains highly ill-posed and challenging due to the inherent ambiguity between material and illumination.

Recent IR approaches~\cite{liuNeRONeuralGeometry2023,wangInverseRenderingGlossy2024,hasselgrenShapeLightMaterial,zhangNeILFInterReflectableLight2023,sunNeuralPBIRReconstructionShape2023,guIRGSInterReflectiveGaussian2025} primarily rely on images captured under uncontrolled natural lighting. While such setups are flexible, they struggle to disentangle reflectance from illumination, particularly in complex indoor scenes with multiple inter-reflecting objects. A complementary line of work constrains the lighting by using a co-located light–camera setup, where a light source (e.g., a phone’s flashlight) is mounted near the camera~\cite{chengWildLightInthewildInverse2023, zhangIRONInverseRendering2022a,higoHandheldPhotometricStereo2009,chungDifferentiableInverseRendering,schmittJointEstimationPose,georgoulisTacklingShapesBRDFs2014}. This configuration is easy to calibrate using SfM and can, in principle, improve reflectance estimation. Yet, these methods are largely limited to single objects and fail to generalize to multi-object scenes due to strong inter-reflections, dynamic shadows, near-field lighting, and moving specular highlights.

To overcome these challenges, we introduce GLOW, a Global Illumination-aware Inverse Rendering framework for multi-object indoor scenes captured under co-located lighting without ambient illumination. We begin by combining and adapting physically based rendering with neural implicit surface representations~\cite{yarivVolumeRenderingNeural2021, wangNeuSLearningNeural2021}, and model inter-reflection with a learned radiance cache trained with the radiometric prior loss proposed by invNeRAD~\cite{hadadanInverseGlobalIllumination2023}. Geometry and reflectance are jointly optimized via reconstruction losses. However, this baseline still fails under co-located lighting, where static caches cannot capture radiance variations from moving lights, and dynamic specular highlights degrade reconstruction.

GLOW introduces three key innovations to address these issues, arranged in order of importance: \textit{(I) Dynamic radiance modeling:} We theoretically show that indirect bounce radiance is continuous with respect to flashlight position and design a dynamic radiance cache that captures sharp radiance discontinuities from cast shadows under moving near-field lights (\cref{sec:separ-light-radi}). \textit{(II) Specular highlight suppression:} We propose a surface-angle-weighted loss tailored for co-located captures that effectively suppresses flashlight-induced specular artifacts, which performs better than existing losses \cite{sunNeuralPBIRReconstructionShape2023, geRefNeuSAmbiguityReducedNeural2023} to suppress specular highlights~ (\cref{sec:saw}). \textit{(III) Robust optimization:} We stabilize the ill-posed joint optimization of shape and reflectance using tailored initialization and regularization for geometry and reflectance, bootstrapping geometry with a neural radiance network guided by statistical priors (\cref{sec:glow-base}).

Experimental evaluations are performed on multi-object indoor scenes comprising three synthetic scenes, ray-traced with co-located lighting, and four real scenes captured with a co-located light-camera. We observe that GLOW significantly outperforms state-of-the-art IR techniques that also rely on a co-located light-camera setup, IRON \cite{zhangIRONInverseRendering2022a} and WildLight \cite{chengWildLightInthewildInverse2023}, in material reflectance estimation by a relative improvement of 91\% and 82\% in albedo and 93\% and 81\% in roughness, respectively, without degrading shape reconstruction quality. Compared to IR techniques that rely on natural illumination, NeRO \cite{liuNeRONeuralGeometry2023} (neural implicit representation) \& IRGS \cite{guIRGSInterReflectiveGaussian2025} (3DGS), GLOW also improves material reflectance estimation by a similar margin of $\sim$90\%, without degrading shape. While this performance gain comes at the cost of requiring dense captures in a dark environment using a phone camera with its flashlight enabled, GLOW offers a $\sim10$x improvement in reflectance accuracy, making it ideal for applications where capture constraints can be traded for substantial quality gains. The code and datasets are available at \url{https://glow-inverse-rendering.github.io/}.

%% file: sections/related_works.tex
\section{Related Works}
\label{sec:related_work}
\noindent \textbf{Role of capture setup.} Recently researchers have focused on developing \textit{at-home} capture setups that can produce high-quality geometry and material reflectance without requiring any \textit{high-end} devices like a LightStage \cite{pandeyTotalRelightingLearning2021,guoRelightablesVolumetricPerformance2019, zhangNeuralLightTransport2021} with calibrated lights and cameras. These \textit{at-home} capture setups can use unconstrained natural illumination~\cite{zhangNeRFactorNeuralFactorization2021, zhangModelingIndirectIllumination2022, zhangNeILFInterReflectableLight2023, liuNeRONeuralGeometry2023,zhangPhySGInverseRendering2021a, munkbergExtractingTriangular3D2022, hasselgrenShapeLightMaterial, wuNeFIIInverseRendering2023, jinTensoIRTensorialInverse, bossNeRDNeuralReflectance2021, zhangNeMFInverseVolume2023, bossNeuralPILNeuralPreIntegrated2021, bossSAMURAIShapeMaterial, engelhardtSHINOBIShapeIllumination2024, wuFactorizedInversePath2023,guIRGSInterReflectiveGaussian2025,wangInverseRenderingGlossy2024}, a co-located light and camera~\cite{chengWildLightInthewildInverse2023, biDeep3DCapture2020, luanUnifiedShapeSVBRDF2021a, namPracticalSVBRDFAcquisition2018, zhangIRONInverseRendering2022a, schmittJointEstimationPose2020,higoHandheldPhotometricStereo2009,georgoulisTacklingShapesBRDFs2014,chungDifferentiableInverseRendering} or a moving camera and a separate flashlight~\cite{liMultiViewPhotometricStereo2020,zhaoMVPSNetFastGeneralizable2023,zhouMultiviewPhotometricStereo2013,cuiPolarimetricMultiViewStereo}. While a natural illumination capture setup is most popular due to its simplicity,  such methods often cannot extract accurate reflectance, especially in multi-object scenes due to the fundamental ambiguity between geometry, lighting, and reflectance. On the other hand, a co-located setup produces high-quality inverse rendering results but requires a dark room that limits its application. Our method uses a co-located light and camera setup, but we handle inter-reflection that is more common in multi-object scenes.

\noindent\textbf{Relightable Field.} Some prior works~\cite{zengRelightingNeuralRadiance2023, lyuNeuralRadianceTransfer2022a} attempt to learn a relight-able scene representation without explicitly reconstructing geometry or reflectance, but they are limited to relighting within their training light distribution. However, full inverse rendering enables physically based rendering and relighting the scene with illumination outside of the training distribution.

\noindent \textbf{Neural Radiance Cache.}
Neural Radiance cache~\cite{mullerRealtimeNeuralRadiance2021} is a neural network that stores the radiance values of the scene.  Radiance values stored in the neural network are used to accelerate rendering. Radiance cache allows approximation of multi-bounce global illumination by querying the neural network.  Some previous inverse rendering methods~\cite{zhangModelingIndirectIllumination2022,zhangNeILFInterReflectableLight2023,liuNeRONeuralGeometry2023} that targeted natural illumination estimate global illumination by leveraging a radiance field trained to fit to image observations. However, such approaches are not suitable for a co-located setup, as the light source is moving together with the camera. Each input image only observes a small subset of the scene for each light source location, which leads to an under-constrained radiance cache.

%% file: sections/methods.tex
\section{Method}

We capture images using a smartphone with a flashlight turned on in a dark room. Such a setup provides multiple images with varying but known illuminations, which helps resolve ambiguity between illumination, geometry and material properties that occur in a natural illumination setup. While this means our method requires a dark environment, it produces significantly better BRDF without requiring any sophisticated hardware or calibration.

We first provide  the foundation of GLOW in Sec.~\ref{sec:glow-base}. GLOW integrates and adapts multiple existing components across neural scene representation, physically-based rendering, radiance cache, radiometric priors and roughness priors. To tackle the unique challenges of co-located light and camera setup, we introduce two novel components, a dynamic light radiance cache for handling dynamic illumination in Sec.~\ref{sec:separ-light-radi} and a surface angle weighting loss to handle strong reflections in Sec.~\ref{sec:saw}. We provide a review of neural implicit rendering methods, physically based rendering, and the light transport operator in the supplementary for readers not familiar with the background.

\subsection{Foundation of GLOW}
\label{sec:glow-base}

As far as we are aware, there is no existing system tailored for inverse rendering of a multi-object scene under co-located light and camera capture setup. Therefore, we created GLOW, a neural implicit surface rendering system, by combining and adapting existing components.

Our geometry is represented with a signed distance field represented \( \operatorname{S_{\Theta_S}}(\mathbf{x}) \). With our co-located capture setup, the lighting is just the flashlight, which we model as a point light source. Our material properties are represented by principled SV-BRDF~\cite{burleyPhysicallyBasedShading}, which can fit various materials, similar to prior works~\cite{zhangNeILFInterReflectableLight2023, chengWildLightInthewildInverse2023,hadadanInverseGlobalIllumination2023, liuNeRONeuralGeometry2023, guIRGSInterReflectiveGaussian2025}. We keep most parameters of principled BRDF frozen and only optimize albedo \(\phi_a\) and roughness \(\phi_r\). Therefore, our scene parameters \(\phi\) include: \(\phi=\{ \Theta_S, \phi_a, \phi_s \}\). In practice, all parameters of the scene \(\phi\) are represented with different heads of the same neural network. 

 Our system contains three stages to address the inherent difficulties presented by global illumination in optimization for co-located light and camera inverse rendering. In the first stage, our system uses statistical priors such as the implicit prior of neural rendering and keypoints estimated in structure-from-motion~\cite{lindenbergerPixelPerfectStructureFromMotionFeaturemetric}. In the second stage, we jointly optimize geometry and material under global illumination. Finally, we perform a dedicated material property optimization step to further improve material qualities.

\subsubsection{Geometry Initialization}
At the start of training, the scene geometry is unknown, and global illumination cannot be accurately inferred. To bootstrap the learning process, we initially perform neural rendering with a learned color network \(\operatorname{C_{\Theta_C}}\).  Given that the flashlight is co-located with the camera and acts as a point light source, the irradiance at a surface point follows an inverse-square falloff, i.e., \(\frac{1}{t^2}\), where \(t\) denotes the distance from the flashlight to the queried point. To account for this near-field lighting effect, we augment the NeuS~\cite{wangNeuSLearningNeural2021} color network with an additional input parameter \(\frac{1}{t^2}\).

 We can then render the pixels of a view $\operatorname{P}(\mathbf{o}, \mathbf{v})$ through volume rendering, with \(\mathbf{o}\) as camera position, t as distance along the camera ray, \(\mathbf{p}(t)\) as 3D positions along the camera ray, \(\mathbf{n}\) as surface normals (gradient of SDF \(S_{\Theta_S}(\mathbf{x})\)), \(\mathbf{f}\) as the feature vector from the geometry network. The opacity function \(w_{\theta_S}(t)=w(S_{\Theta_S}(\mathbf{x}))\) defined by NeuS converts SDFs \(S_{\Theta_S}(\mathbf{x})\) into weights.

\begin{equation}
  \hat{y}_i = \operatorname{P}(\mathbf{o}, \mathbf{v})=\int_0^{+\infty} w_{\theta_S}(t) \operatorname{C_{\Theta_C}}(\mathbf{p}(t), \mathbf{n}, \mathbf{v}, \mathbf{f}, \frac{1}{t^2}) \text{d}t
\end{equation}

We then optimize the scene parameters with a re-rendering loss w.r.t.~captured image $y$ in linear RGB space. We use the linearized log loss introduced by RawNeRF~\cite{mildenhallNeRFDarkHigh2022}, as shown in Eq. \ref{eq:linearized_tone}, to better learn radiance in dark regions. Let \(sg(\cdot)\) denote the stop gradient operation.

\begin{equation}
    \label{eq:linearized_tone}
    \mathcal{L}_\text{recons}(\hat{y_i}, y_i) =  \left(\frac{\hat{y_i}-y_i}{sg(\hat{y_i}) + \epsilon}\right)^2.
  \end{equation}

In Sec~\ref{sec:saw}, we further introduce surface angle weighting, which is applied on reconstruction loss \(\mathcal{L}_\text{recons}\) to improve geometry reconstruction. 

\noindent\textbf{Non-Convexity of Co-Located Photometric Stereo} We found that co-located photometric stereo suffers from non-convexities. We provide further discussion in the supplementary material. Prior works in co-located light and camera often initialize geometry with depth sensors~\cite{schmittJointEstimationPose}, multi-view stereo techniques~\cite{higoHandheldPhotometricStereo2009}, or SfM points~\cite{georgoulisTacklingShapesBRDFs2014}. 

To mitigate this problem, we leverage the structure-from-motion (SfM) points, which are already used in prior inverse rendering systems for camera pose estimation. More specifically, after obtaining camera poses through SfM, we encourage the neural SDF towards zero at the SfM point locations \(X_\mathrm{sfm}\). The loss pushes the solution near correct coarse geometry, and mitigates the non-convexity.

\begin{equation}
  \label{eq:sfm_reg}
   \mathcal{L}_{\mathrm{sfm}} =  \| \operatorname{S_{\Theta_S}}(\mathbf{X_\mathrm{sfm}})\| 
\end{equation}

\subsubsection{Physically Based Rendering}
\label{sec:method_pbr}
 In our second stage, we perform physically based rendering to recover material properties under global illumination while  jointly refining geometry.  Multi-object indoor scenes exhibit strong inter-reflections, which existing direct illumination-based neural inverse rendering frameworks struggle to handle~\cite{zhangIRONInverseRendering2022a,chengWildLightInthewildInverse2023}.  Modeling multi-bounce global illumination is a difficult problem. Path tracing algorithms are computationally inefficient for inverse rendering. Instead, we rely on the radiance cache~\cite{mullerRealtimeNeuralRadiance2021} \(L_{\theta}\), a neural network that maps from scene locations \(\mathbf{x}\) and outgoing directions \(\mathbf{v}\) to scene radiance \(L\). The radiance cache significantly reduces the computational cost of modeling multi-bounce light paths by tracing only a single bounce and querying the radiance cache.

By combining neural implicit surface volume rendering and radiance cache, we can render the color of each pixel \(\operatorname{P}(\mathbf{o}, \mathbf{v})\) using volume rendering.  We perform physically based rendering by applying the light transport operator \(\mathcal{T}_\phi\) on a frozen copy of the neural radiance cache \(L_\theta^{\mathrm{frozen}}\). Each pixel \(\hat{y_i}\) will then be rendered using the following.
\begin{multline}
\label{eq:vol_render}
\hat{y_i}=\operatorname{P}(\mathbf{o}, \mathbf{v})=\int_0^{+\infty} w_{\theta_S}(t) 
\mathcal{T_\phi}(L_{\theta}^{\mathrm{frozen}}) (\mathbf{p(t)}, \omegaout)\,dt
\end{multline}
All scene parameters are jointly optimized with reconstruction loss \(\mathcal{L}_\text{recons}\) as in Eq.~\ref{eq:linearized_tone}.

Note the equation is dependent on not only albedo \(\phi_a\) and roughness \(\phi_s\), but also the geometry network \( \operatorname{S_{\Theta_S}}(\mathbf{x}) \) through volume rendering weight \(w_{\theta_S}\), and surface normal (gradient of \(\operatorname{S_{\Theta_S}}\)). Therefore, we can compute the gradient and jointly optimize all parameters.

\noindent \textbf{Frozen Mesh and Radiance Cache} Recall that the light transport operator \(\mathcal{T_\phi}\) involves the ray tracing operator $r(x,\omegain)$, which returns the closest surface intersection. As our system is based on volume rendering, we haven't defined the notion of a surface yet. The ray casting operator is evaluated against a frozen surface mesh, which is re-extracted from the geometry network every \(K\) steps. A frozen copy of the radiance cache \(L_\theta^{\mathrm{frozen}}\)  is also made every $K$ steps as the live copy of the radiance cache \(L_\theta\) might drift away from the frozen mesh during training. 

  \noindent\textbf{Radiance Cache with Radiometric Prior}
  Many prior works~\cite{zhangModelingIndirectIllumination2022,zhangNeILFInterReflectableLight2023,guIRGSInterReflectiveGaussian2025} train radiance caches by directly minimizing their discrepancy with training images. However, under co-located light-camera setup, each view corresponds to a unique lighting condition, as the light source moves with the camera. As a result, only a limited portion of the scene is observed for each light position, which leads to the radiance cache being under-constrained. InvNeRAD~\cite{hadadanInverseGlobalIllumination2023} introduces a self-supervised loss that encourages the radiance cache to obey the rendering equation everywhere in the scene, without relying on input images. So far, the radiometric prior is only applied in natural illumination settings~\cite{hadadanInverseGlobalIllumination2023}. We incorporate this loss to co-located light \& camera settings, which we denote as \(\mathcal{L}_\text{prior}\).
  
  We found our naive attempt to train a radiance cache with a radiometric prior loss for co-located light and camera failed. Therefore, we propose a dynamic radiance cache, which is further described in Sec.~\ref{sec:separ-light-radi}.

\noindent \textbf{Roughness Regularization.} So far, we designed a new inverse rendering system GLOW, which is tailored for co-located camera and light of multi-object scenes. However, we found that such a system suffers from noisy roughness, as surface shininess tends to be inherently sparse. Such a problem is common in the inverse rendering literature~\cite{wuFactorizedInversePath2023,liuNeRONeuralGeometry2023,zhangNeILFInterReflectableLight2023}. We apply the regularization techniques of FIPT~\cite{wuFactorizedInversePath2023} to improve GLOW roughness estimation. Unlike FIPT, which uses MaskFormer for semantic segmentation, we employ the state-of-the-art Segment Anything (SAM) 2 model~\cite{raviSAM2SEGMENT2025} to obtain the part-level segmentation masks, which we found to be more accurate.

Our final optimization objective becomes the following.
{

\begin{equation}
\label{eq:objective}
\mathcal{L} = \mathcal{L}_{\mathrm{recons}} + \mathcal{L}_{\mathrm{prior}} + \mathcal{L}_{\mathrm{rough}} + \mathcal{L}_{\mathrm{sfm}}
\end{equation}
}
\subsubsection{Material Optimization}
Our physically based rendering stage in Sec.~\ref{sec:method_pbr} already recovers high quality material properties. Nevertheless, the material properties representation might be constrained during stage 2, limiting achievable quality in practice. For example, we found using separate high frequency capable albedo and roughness networks often leads to ambiguity in volume rendering and degraded reconstruction. Therefore, in the third stage, we simply apply the existing inverse rendering technique invNeRAD~\cite{hadadanInverseGlobalIllumination2023} on mesh geometry extracted from our stage 2, but with our dynamic radiance cache. Further implementations details of the technique can be found in supplementary material or invNeRAD.

\subsection{Dynamic Light Radiance Cache}
\label{sec:separ-light-radi}

 With our co-located dynamic light-capture setup, lighting varies per view, so we additionally condition the radiance cache on the flashlight position \(\mathbf{x_l}\): \(L_{\theta}(\mathbf{x}, \omegaout, \mathbf{x_l})\). However, the flashlight often creates hard cast shadows, causing the intensity at affected scene points to change abruptly as the flashlight moves. Neural radiance caches often struggle to model such discontinuities w.r.t. flashlight position.

We will start by examining the sources of discontinuity in radiance by looking at the rendering equation. To study the smoothness against flashlight position \(\mathbf{x_l}\), we include it as a parameter.
\begin{align*}
  L(\mathbf{x}, \omegaout, \mathbf{x_l}) = E(\mathbf{x}, \omegaout, \mathbf{x_l}) &+ \mathcal{T_\phi}(L)(\mathbf{x}, \omegaout, \mathbf{x_l}) \\
  = E(\mathbf{x}, \omegaout, \mathbf{x_l}) &+ \mathcal{T_\phi}(E)(\mathbf{x}, \omegaout, \mathbf{x_l}) \\
                                                      &+ \sum_{k=2}^{\infty} \mathcal{T_\phi}^{k}(E) (\mathbf{x}, \omegaout, \mathbf{x_l}) \nonumber
\end{align*}

 Let \(\mathbbm{V}{(x_l\leftrightarrow\mathbf{x})}\) be a visibility function of whether the flashlight position \(\mathbf{x_l}\) is visible to the scene point \(x\). \(\theta\) is the angle between surface normal \(\mathbf{n}\) and light direction. SV-BRDF is \(F(\mathbf{x}, \omegain, \omegaout)\). With a co-located flashlight, we can further expand the first bounce term. 

\vspace{-1em}
\begin{equation*}
  \mathcal{T_\phi}(E)(\mathbf{x}, \omegaout, \mathbf{x_l}) =\mathbbm{V}{(\mathbf{x_l}\leftrightarrow\mathbf{x})} \frac{  F(\mathbf{x}, \omegain, \omegaout) \operatorname{cos}(\theta)}{\|\mathbf{x_l} - \mathbf{x}\|^2}  
\end{equation*}

As long as the  SV-BRDF \(F(\mathbf{x}, \omegain, \omegaout)\) is smooth, it is easy to notice that \(E(\mathbf{x}, \omegaout)\) and \(\frac{  F(\mathbf{x}, \omegain, \omegaout) \operatorname{cos}(\theta)}{\|\mathbf{x_l} - \mathbf{x}\|^2}\) are smooth functions with regard to flashlight position \(\mathbf{x_l}\). As we will formally prove in supplementary, \( \sum_{k=2}^{\infty} \mathcal{T_\phi}^{k}(E) (\mathbf{x}, \omegaout)\) is also continuous w.r.t. flashlight position.

Therefore, the real problematic source of discontinuity is \(\mathbbm{V}{(\mathbf{x_l}\leftrightarrow\mathbf{x})}\) or direct illumination cast shadow. To tackle such problems, we propose a dynamic light radiance cache, where our radiance cache \(L_\theta\) will be represented with the sum of two neural networks \(L_{\theta}=\mathbbm{V}{(\mathbf{x_l}\leftrightarrow\mathbf{x})}L_\theta^{\textrm{direct}} + L_\theta^{\textrm{indirect}}\), where \(L_\theta^{\textrm{direct}}\) and \(L_\theta^{\textrm{indirect}}\) are representing the continuous direct component  \(\frac{ F(\mathbf{x}, \omegain, \omegaout) \operatorname{cos}(\theta)}{\|\mathbf{x_l} - \mathbf{x}\|^2}\) and continuous indirect component \(\sum_{k=2}^{\infty} \mathcal{T_\phi}^{k}(E) (\mathbf{x}, \omegaout)\) respectively. Such a formulation avoids baking the highly discontinuous visibility function \(\mathbbm{V}{(\mathbf{x_l}\leftrightarrow\mathbf{x})}\) into the radiance cache, which impairs training.

We train the radiance cache with two separate radiometric prior losses, which encourages both radiance caches to be close to their respective component. During inference, we use the frozen mesh extracted from neural sdf to determine visibility \(\mathbbm{V}{(\mathbf{x_l}\leftrightarrow\mathbf{x})}\).

\begin{align}
\label{eq:seperated_radiometric_prior}
  &\mathcal{L}_{\mathrm{prior}}^{\mathrm{direct}}(\theta) = \| L_{\theta}^{\mathrm{direct}}(\mathbf{x},\omega_o, \mathbf{x_l}) - (E(\mathbf{x}, \omegaout) + \mathcal{T_\phi}(E)(\mathbf{x}, \omegaout))\| \nonumber \\ 
  &\mathcal{L}_{\mathrm{prior}}^{\mathrm{indirect}}(\theta) = \| L_{\theta}^{\mathrm{indirect}}(\mathbf{x},\omega_o, \mathbf{x_l}) - \sum_{k=2}^{\infty} \mathcal{T_\phi}^{k}(E) (\mathbf{x}, \omegaout) \| \nonumber \\ 
  &\mathcal{L}_{\mathrm{prior}}(\theta) = \mathbbm{V}{(\mathbf{x_l}\leftrightarrow\mathbf{x})}\mathcal{L}_{\mathrm{prior}}^{\mathrm{direct}}(\theta) + \mathcal{L}_{\mathrm{prior}}^{\mathrm{indirect}}(\theta)
\end{align}

\subsection{Surface Angle Weighting for Specularity.} 
\label{sec:saw}
We observe that strong reflections from the flashlight heavily deteriorate geometry predictions. Existing approaches~\cite{sunNeuralPBIRReconstructionShape2023,geRefNeuSAmbiguityReducedNeural2023} propose loss weighting schemes that downweight outlier observations but they tend to fail in reconstructing our complex scenes under co-located light and camera capture setup. Our idea is based on the physics of co-located light and camera. We propose to downweight very small angles (perpendicular direction) that produce strong specular reflections and very large angles (grazing direction) that cause the diffuse component to become minimal, while specular inter-reflection becomes even stronger due to Fresnel effects.

We volume render the surface normal $\mathbf{n}$ using the geometry network $S_{\Theta_S}(\cdot)$ into image and compute the angle \(\theta=\operatorname{arccos}(\mathbf{n}\cdot \mathbf{d}_L)\) between normal \(\mathbf{n}\) and light ray orientation \(\mathbf{d}_l\). We propose to weight the reconstruction loss \(\mathcal{L}_\text{recons}\) of each pixel by:

\begin{equation}
\label{eq:surface_angle}
    W_{a,b}(\theta) = \left\{
\begin{array}{ll}
      \operatorname{max}(\operatorname{cos}(a(\theta-\frac{\pi}{4})), 0) & \theta \leq \frac{\pi}{4} \\
      \operatorname{max}(\operatorname{cos}(b(\theta-\frac{\pi}{4})),0) & \theta > \frac{\pi}{4} \\
\end{array} 
\right.
\end{equation}

This function has two desirable properties. One is that it decreases slowly in the middle near \(\frac{\pi}{4}\), where neither specularity nor Fresnel effects are prominent, and decreases rapidly at the two extremes. The other is its asymmetry, as retro-reflective (when the outgoing direction of reflected light coincides with the incoming direction of light) specular highlights are only observable from a very narrow range of angles, but mirror like Fresnel effects are prominent over a larger range of angles. We use \(a=2\), \(b=4\) for our geometry initialization stage, but we use \(a=2\), \(b=0\) for GLOW after geometry initialization, as GLOW physically based rendering is capable of handling mirror like inter-reflection.

%% file: sections/experiments_alt.tex
\input{figures/qualitative_figure_synthetic}

\input{figures/qualitative_figure_real}

\input{figures/main_metrics_table}
\input{figures/synthetic_geometry_metrics}

\input{figures/merged_render_geometry}

\section{Evaluation}
Inverse rendering of a multi-object scene using co-located light and camera is an under-explored topic. Therefore, our evaluations will be focused on validating our two key claims. Firstly, we demonstrate, by using active illumination, our method can significantly constrain the solution space of ambiguous inverse rendering and achieve better performance on reflectance estimation. Moreover, existing co-located light and camera techniques are unable to handle multi-object scenes with significant inter-reflections, highlighting challenges associated with the problem. We perform qualitative evaluation on real data and quantitative and qualitative evaluation on synthetic data. 

\noindent \textbf{Synthetic data.} We created three scenes, \textit{bedroom}, \textit{shelf}, and \textit{kitchen counter} based on openly available Blender and Mitsuba scenes. To create natural illumination renders, for each scene, we download HDR environment maps from CC0-licensed Poly Heaven\cite{tuytelPolyHavenPoly} with corresponding environment. (i.e. bedroom, living room, and kitchen environment map respectively.) While real world in-the-wild illuminations tend to be more complicated than what environment maps can represent, we show in Fig.~\ref{fig:qualitative_figure_synthetic} and Tab.~\ref{tab:main_metrics} that existing natural illumination methods still perform poorly with such simplified lighting scenario. We render the natural illumination dataset with Mitsuba~\cite{wenzeljakobMitsuba3Renderer} under path-tracing renderer. The co-located dataset is created similarly, but with a moving co-located point light and camera instead of environment map. Each synthetic scene contains 1000 training images and 50 validation images with randomly generated camera poses that are shared between natural illumination and co-located renderings.

\noindent\textbf{Real data.} We captured four real scenes using co-located light and camera in a darkroom, \textit{window sill}, \textit{table}, \textit{shoe rack}, and \textit{coffee table} that contain 1006, 704, 564, and 650 images respectively. For the \textit{shoe rack} and \textit{coffee table}, we redo the capture under natural illumination while keeping the scene content fixed. We do our best to keep camera viewpoint distribution similar, but slight variation might exist due to human factors. \textit{shoe rack} ends up containing 756 images while \textit{coffee table} contains 653 images. We reserve 5\% of the images of each scene as the validation split.

\noindent\textbf{Dataset Remarks.}  Our dataset contains 3 synthetic scenes and 4 real scenes, which is diverse among similar works. For example, Chung. et. al~\cite{chungDifferentiableInverseRendering} is tested on 4 synthetic scenes and 1 real scene, while FIPT~\cite{wuFactorizedInversePath2023} is tested on 4 synthetic scenes and 2 real scenes. 

\noindent \textbf{Natural Illumination Comparison.} We compare against natural illumination methods, including NeRO~\cite{liuNeRONeuralGeometry2023} and IRGS~\cite{guIRGSInterReflectiveGaussian2025}. All methods incorporate designs that handle inter-reflection. NeRO caches spatially varying indirect illumination, and the cache is only supervised end-to-end using groundtruth images. IRGS does not have a cache for indirect illumination, but instead performs raytracing directly on the gaussian volume. %

\noindent\textbf{Co-Located Illumination Comparison.} For co-located light and camera methods, we compare against IRON~\cite{zhangIRONInverseRendering2022a} and WildLight~\cite{chengWildLightInthewildInverse2023}. Neither method handles inter-reflection. %

\subsection{Evaluation of Reflectance \& Geometry}

We evaluate reflectance estimation on a held-out test set, quantitatively on  synthetic data (Fig.~\ref{tab:main_metrics}) and qualitatively on both synthetic and real scenes (Figs.~\ref{fig:qualitative_figure_synthetic},~\ref{fig:qualitative_figure_real}) for albedo and roughness. During re-rendering in validation views, we clip the range of both GT and prediction to \([0, 1]\) to prevent high-intensity pixels dominating the error. We employ scale-invariant loss after correcting white-balance for measuring albedo, similar to prior approaches.

We observe that natural illumination methods often fail in regions with heavy cast shadows and show intensity errors even after scaling (e.g., floors in bedroom, shelf, green cardbox in Fig.~\ref{fig:qualitative_figure_synthetic}). Co-located light–camera methods like IRON and WildLight struggle in multi-object scenes: IRON fails due to poor geometry (see Fig.~\ref{fig:geom_comparison}, while WildLight cannot handle inter-reflections (e.g. side of cabinet in bedroom, green card box in kitchen counter in Fig.~\ref{fig:qualitative_figure_synthetic}). Our method models both near-field lighting and inter-reflections, unlocking the potential of co-located capture.
Similarly, our method significantly outperforms both natural illumination and co-located light-camera capture techniques for roughness map estimation.

We compare the geometry reconstructed by our method against state-of-the-art inverse rendering techniques. Quantitative results on synthetic data are reported in Tab.~\ref{tab:synthetic_geometry_metrics}, and qualitative comparisons on both synthetic and real scenes are shown in Fig.~\ref{fig:geom_comparison}.
IRON~\cite{zhangIRONInverseRendering2022a} does not model near-field illumination and struggles with multi-object scenes. WildLight~\cite{chengWildLightInthewildInverse2023}, which assumes a point light, performs well in open scenes but poorly in concave or complex geometries, such as shelves, counters, or real scenes like tables and window sills.
Natural illumination methods, NeRO~\cite{liuNeRONeuralGeometry2023} and IRGS~\cite{guIRGSInterReflectiveGaussian2025}, perform comparably on geometry, with NeRO slightly better. While our geometry is similar to these methods, our approach significantly outperforms them in material property estimation.

\input{figures/ablation_figure_glow}

\input{figures/ablation_figure_metrics}

\subsection{Ablation studies}

We ablate our key contribution, the dynamic radiance cache, in this section, with additional ablation of our framework in the supplementary material.
In Fig.~\ref{fig:ablation_figure} (kitchen scene), we compare our method with path tracing, direct illumination, and a naive radiance cache using ground-truth geometry. Our results closely match path tracing, while direct illumination shows strong artifacts in albedo and roughness, indicating the need to model complex inter-reflections. The naive cache cannot handle sharp radiance discontinuities from flashlight motion and performs only marginally better than direct illumination.
Tab.~\ref{tab:abl_cache} shows similar trends. Surprisingly, our method slightly outperforms path tracing in albedo and roughness, likely due to lower-variance gradients afforded by the radiance cache, as observed in invNeRAD~\cite{hadadanInverseGlobalIllumination2023}. Path tracing requires 3097 mins, 3.7x more compared to our 821 mins runtime, while our approach is only 1.8x slower than simple direct illumination and adds minimal overhead compared to a naive cache, demonstrating both accuracy and efficiency.

%% file: figures/qualitative_figure_synthetic.tex
\providelength\width
\setlength\width{2.3cm}

\begin{figure*}
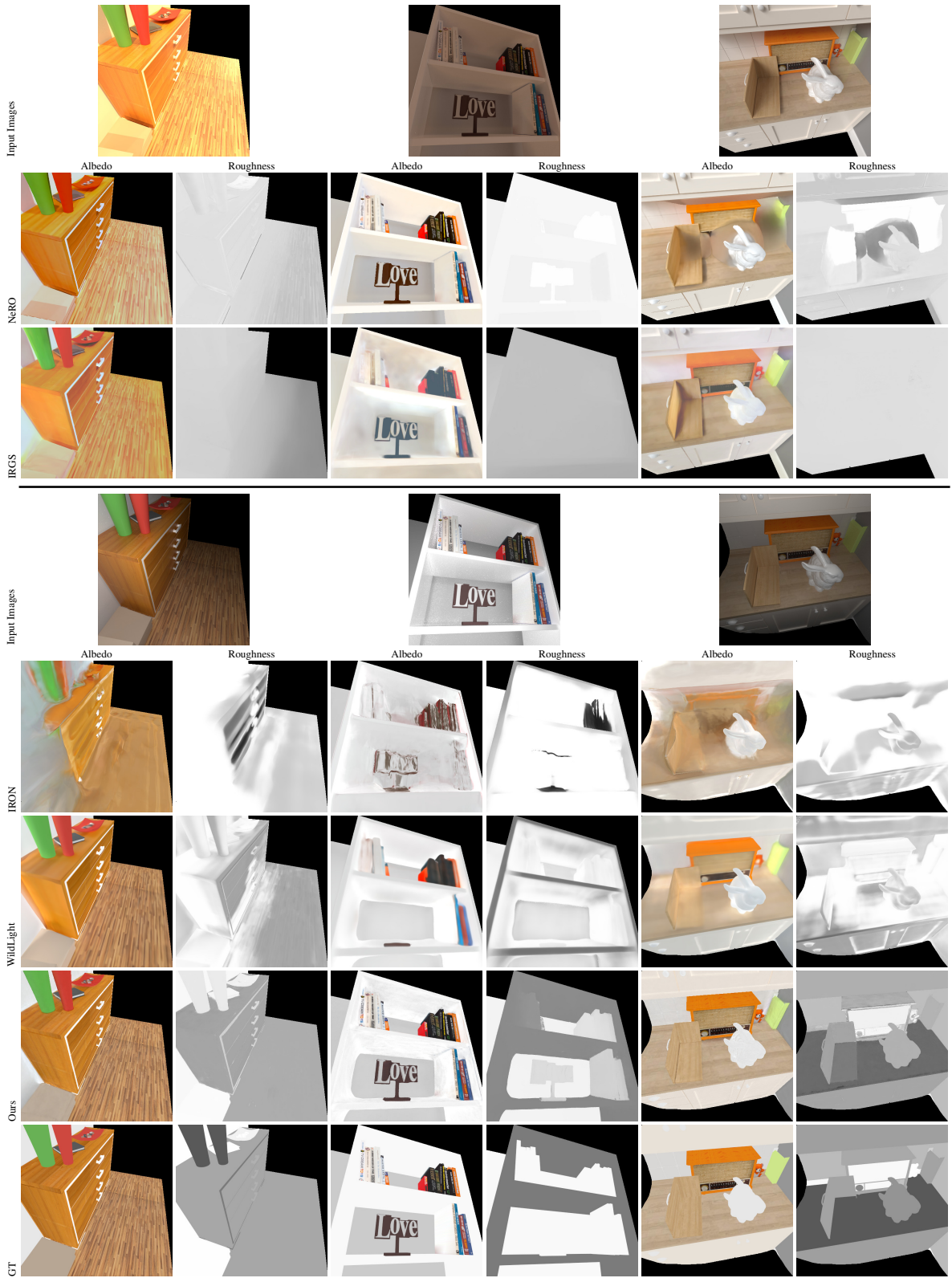

\tiny
\centering

\renewcommand{\tabcolsep}{1pt}

\input generated/qualitative_generated_params_synthetic
\caption{
\textbf{Qualitative comparison of reflectance estimation on synthetic scenes.} We present estimated albedo, and roughness in validation views for the synthetic scenes. Top three rows are natural illumination methods while bottom five rows are co-located methods. Our method produces significantly better albedo and roughness w.r.t. natural illumination methods as co-located capture setup provides additional constraints. Prior co-located light \& camera methods do not model global illumination and perform poorly. (Zoom in for better visualization.) }
\label{fig:qualitative_figure_synthetic}

\end{figure*}

%% file: generated/qualitative_generated_params_synthetic.tex
\begin{tabular}{ccccccc}
& \multicolumn{2}{c}{Bedroom}
& \multicolumn{2}{c}{Shelf}
& \multicolumn{2}{c}{Kitchen}
\\
{\makebox{\rotatebox{90}{Input Images}}}
& \multicolumn{2}{c}{\includegraphics[width=\width]{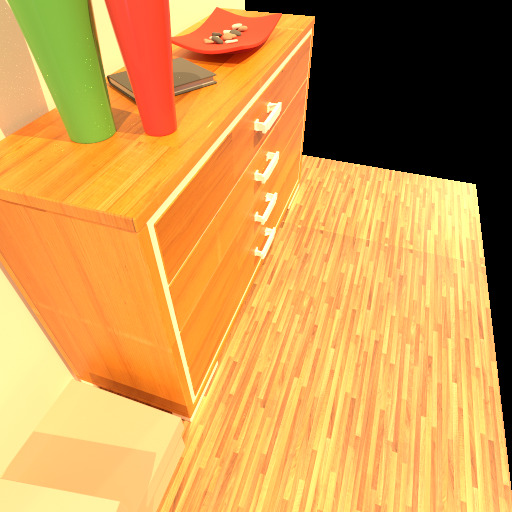}}
& \multicolumn{2}{c}{\includegraphics[width=\width]{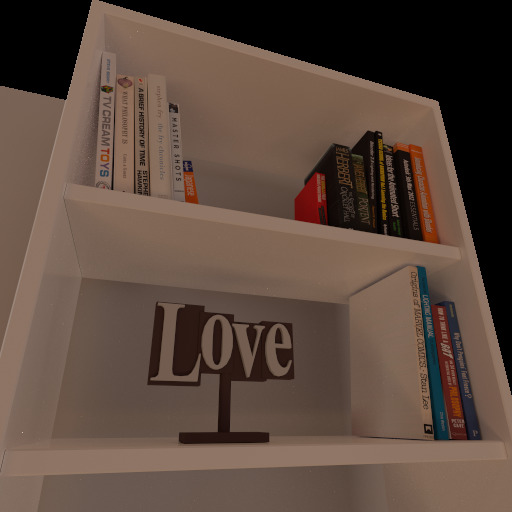}}
& \multicolumn{2}{c}{\includegraphics[width=\width]{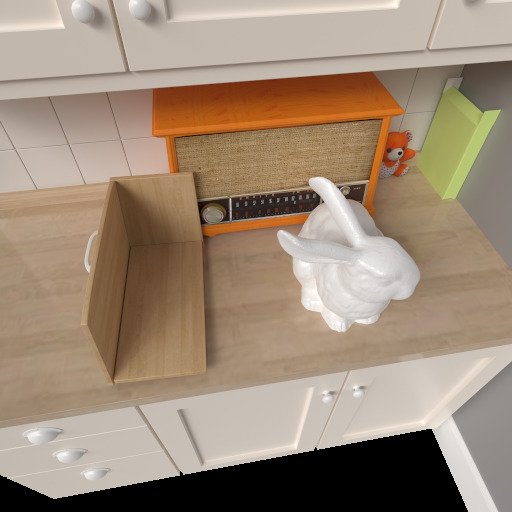}}
\\
&Albedo & Roughness & Albedo & Roughness & Albedo & Roughness  \\
{\makebox{\rotatebox{90}{NeRO}}}
& \includegraphics[width=\width]{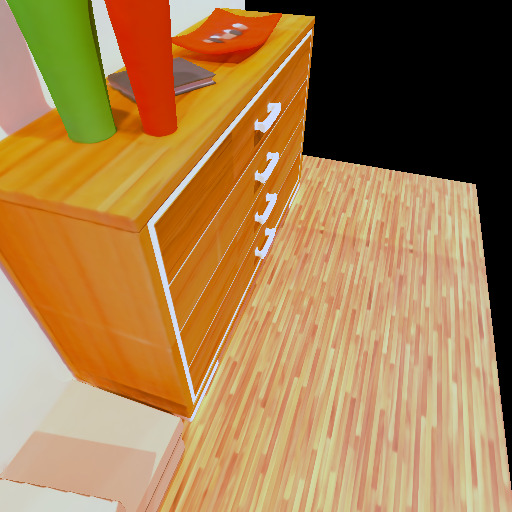}
& \includegraphics[width=\width]{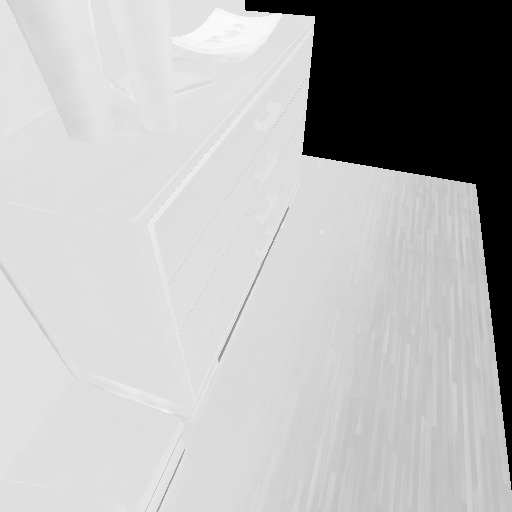}
& \includegraphics[width=\width]{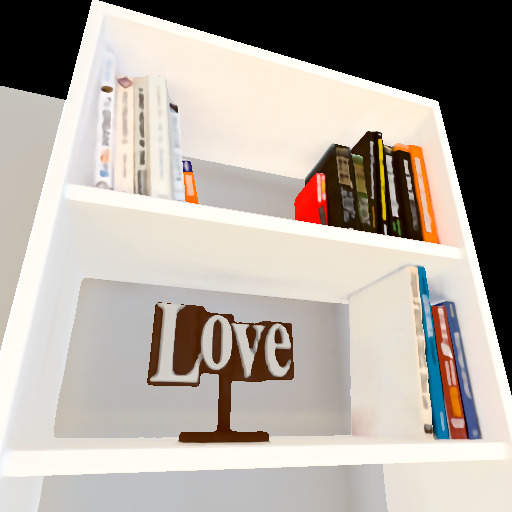}
& \includegraphics[width=\width]{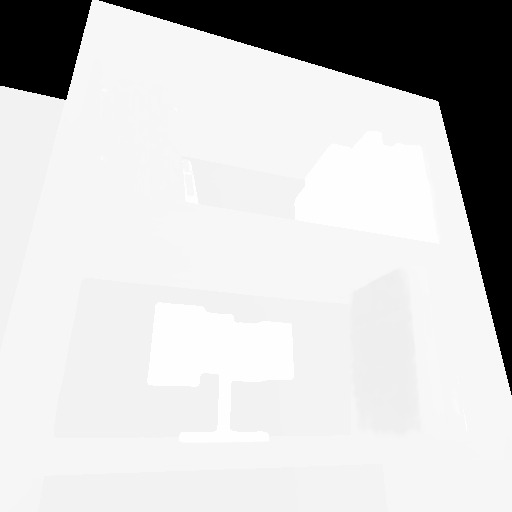}
& \includegraphics[width=\width]{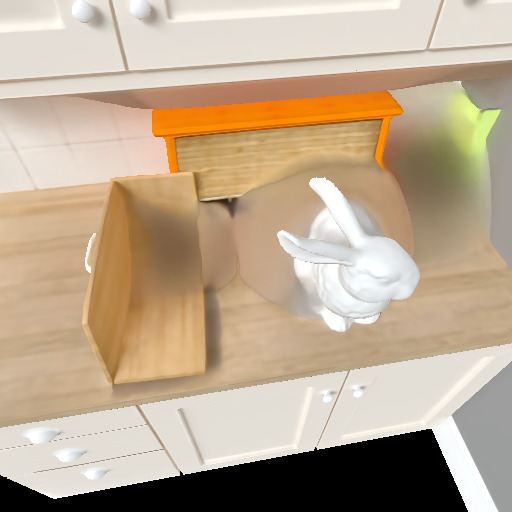}
& \includegraphics[width=\width]{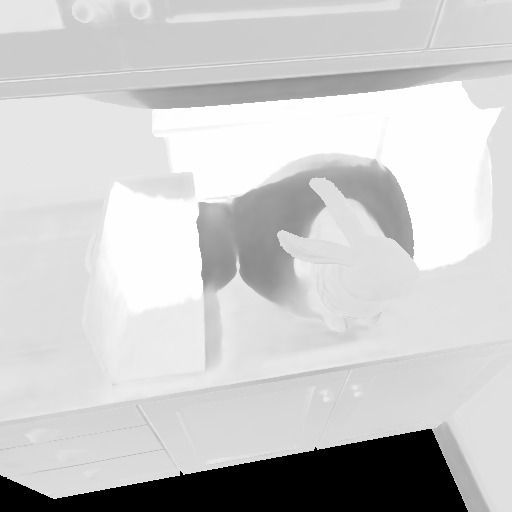}
\\

{\makebox{\rotatebox{90}{IRGS}}}
& \includegraphics[width=\width]{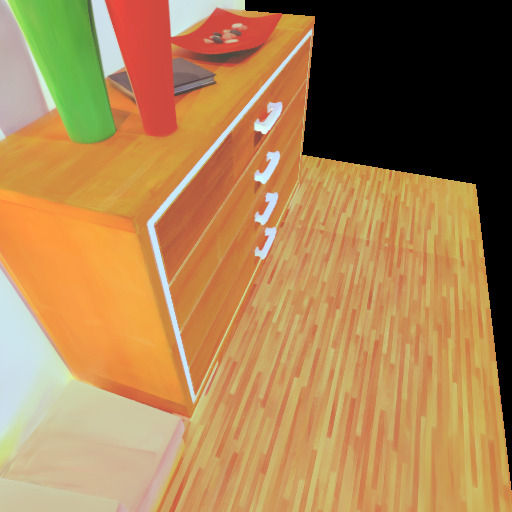}
& \includegraphics[width=\width]{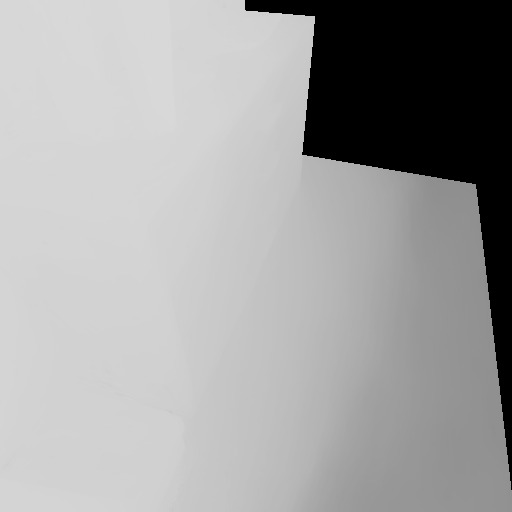}
& \includegraphics[width=\width]{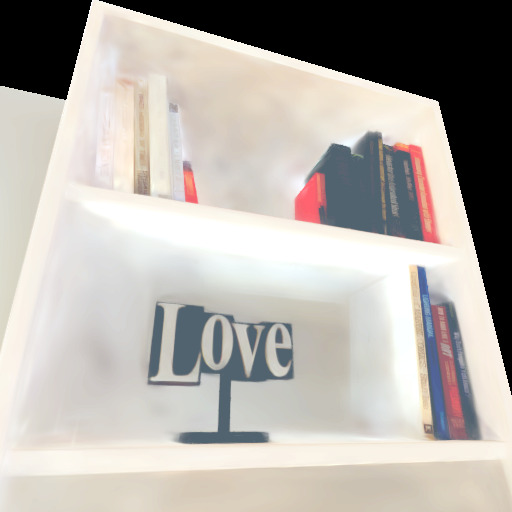}
& \includegraphics[width=\width]{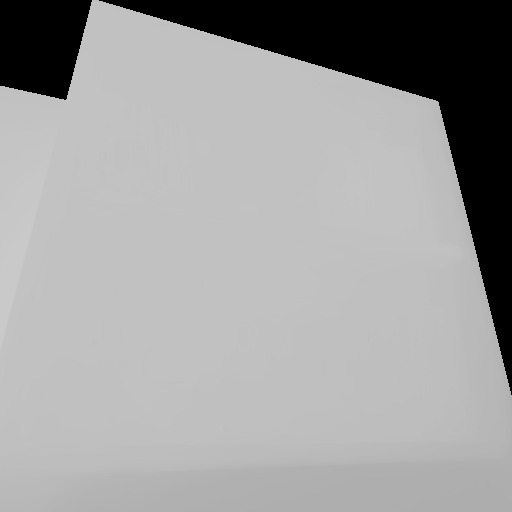}
& \includegraphics[width=\width]{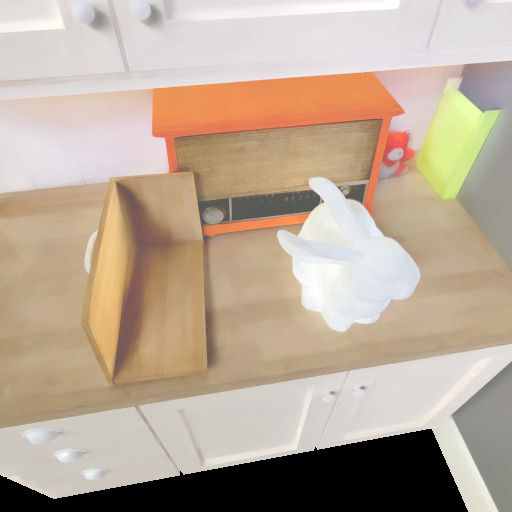}
& \includegraphics[width=\width]{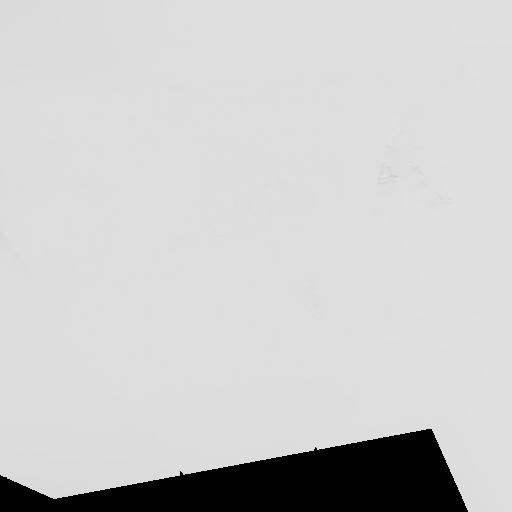}
\\

\cmidrule[1pt]{ 2-7 }
{\makebox{\rotatebox{90}{Input Images}}}
& \multicolumn{2}{c}{\includegraphics[width=\width]{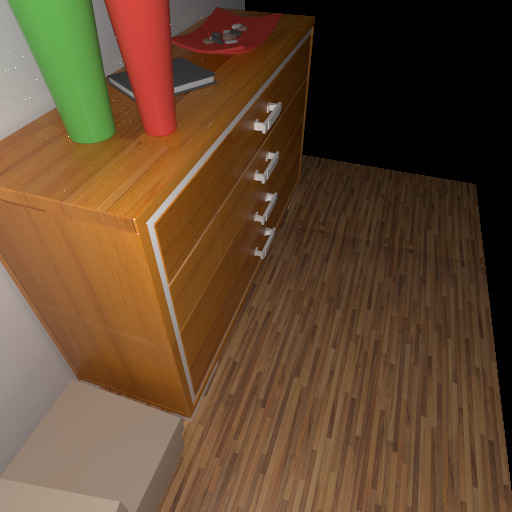}}
& \multicolumn{2}{c}{\includegraphics[width=\width]{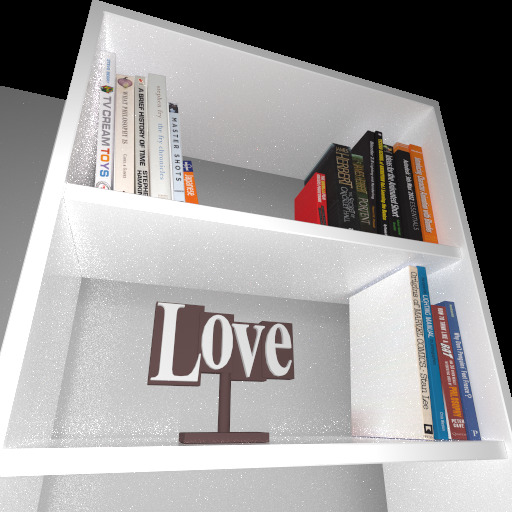}}
& \multicolumn{2}{c}{\includegraphics[width=\width]{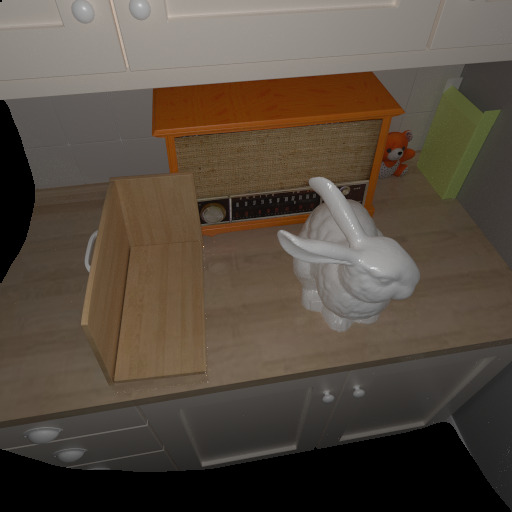}}
\\
&Albedo & Roughness & Albedo & Roughness & Albedo & Roughness  \\
{\makebox{\rotatebox{90}{IRON}}}
& \includegraphics[width=\width]{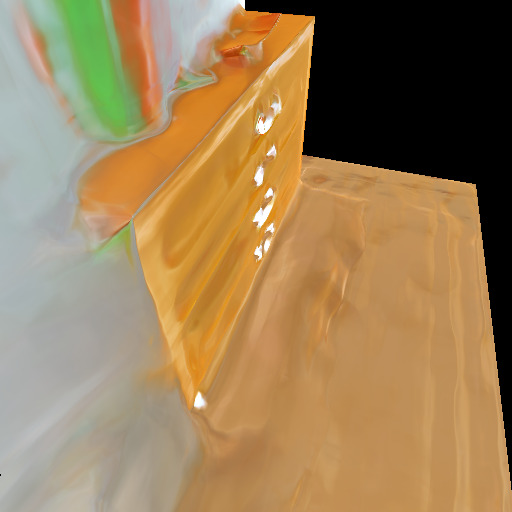}
& \includegraphics[width=\width]{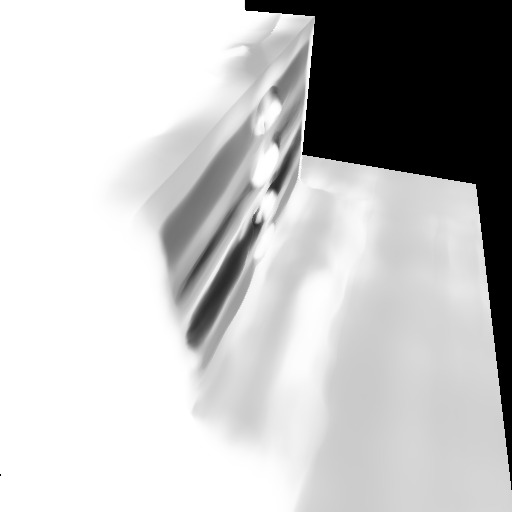}
& \includegraphics[width=\width]{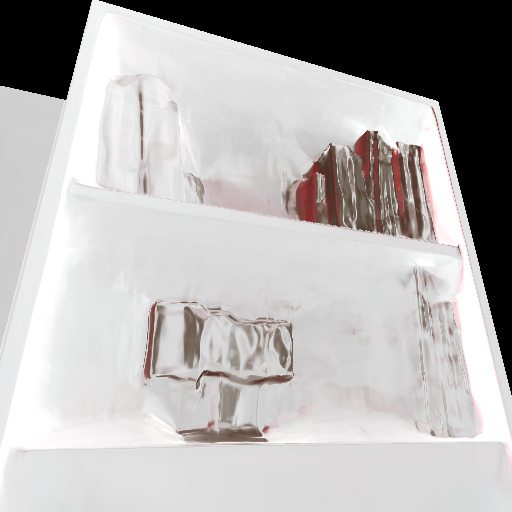}
& \includegraphics[width=\width]{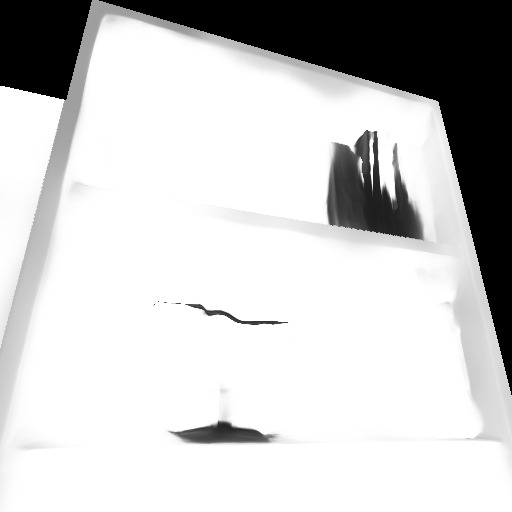}
& \includegraphics[width=\width]{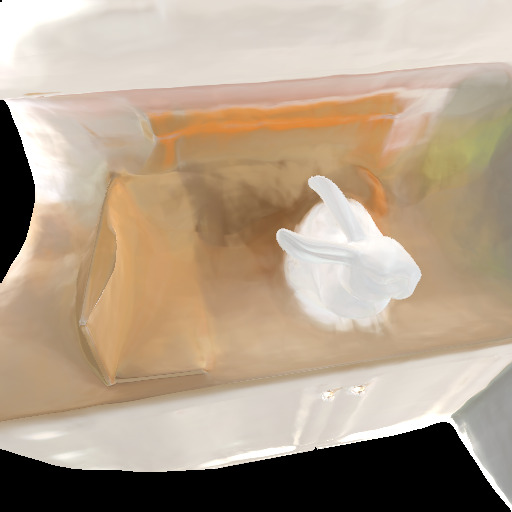}
& \includegraphics[width=\width]{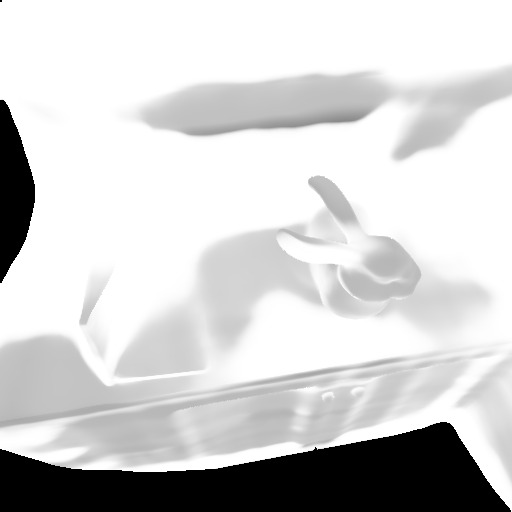}
\\

{\makebox{\rotatebox{90}{WildLight}}}
& \includegraphics[width=\width]{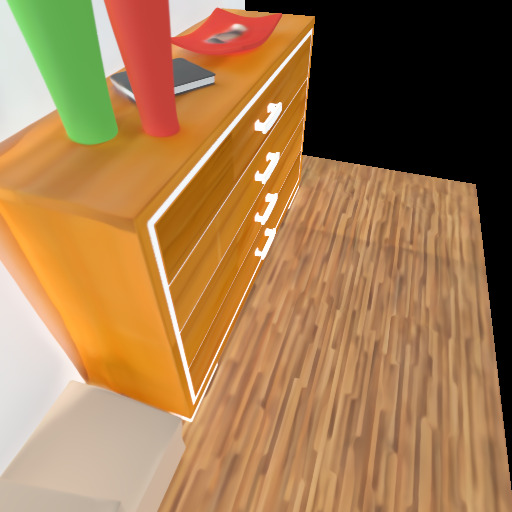}
& \includegraphics[width=\width]{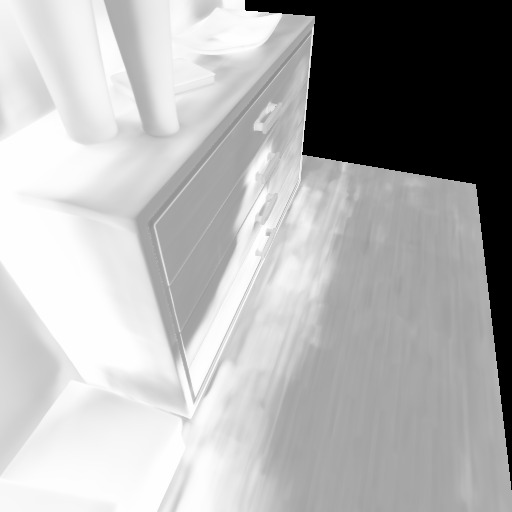}
& \includegraphics[width=\width]{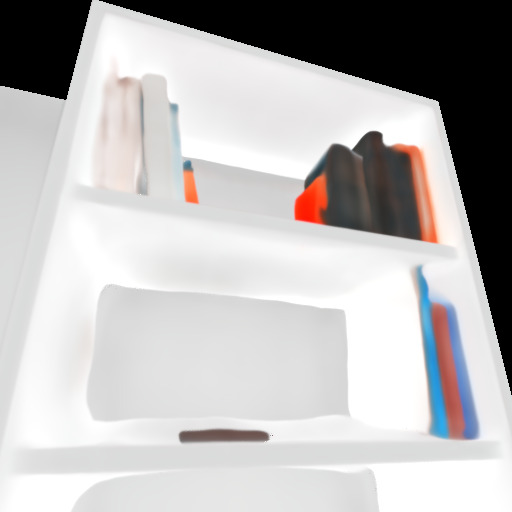}
& \includegraphics[width=\width]{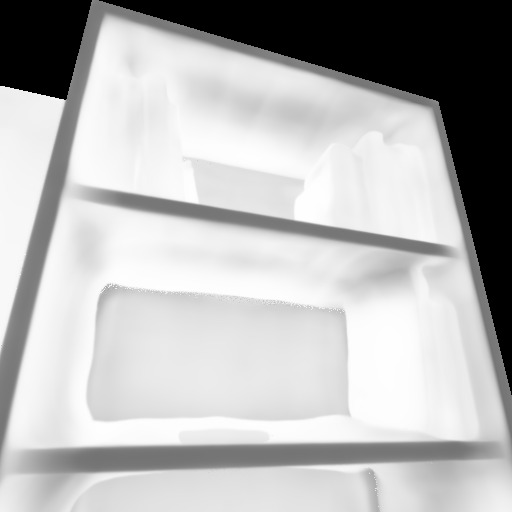}
& \includegraphics[width=\width]{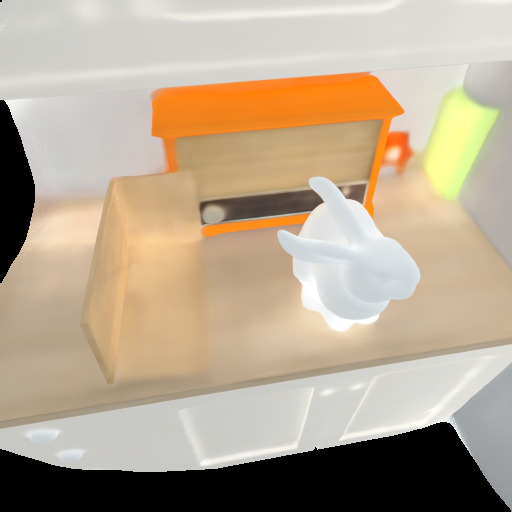}
& \includegraphics[width=\width]{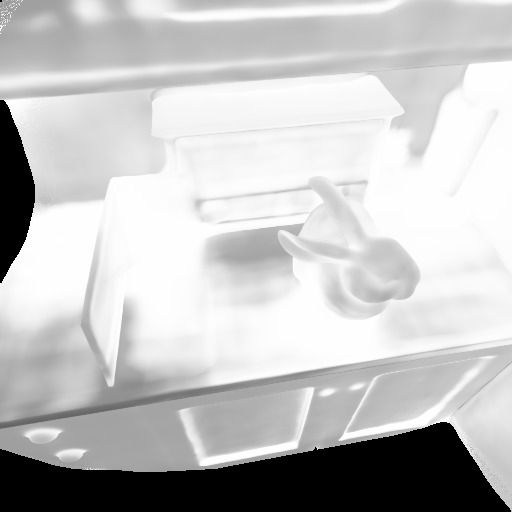}
\\

{\makebox{\rotatebox{90}{Ours}}}
& \includegraphics[width=\width]{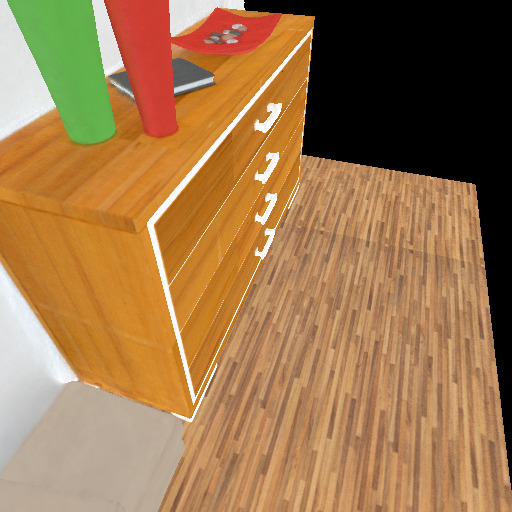}
& \includegraphics[width=\width]{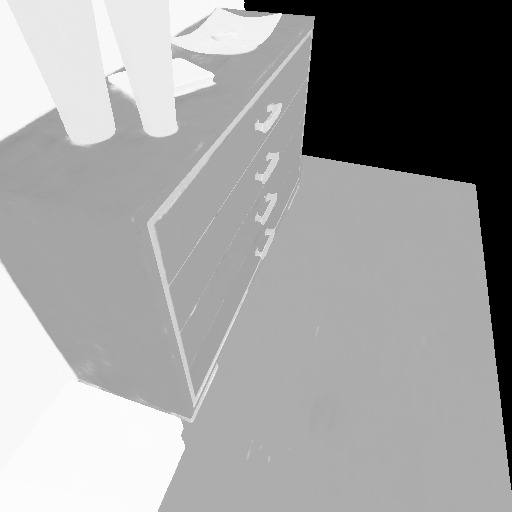}
& \includegraphics[width=\width]{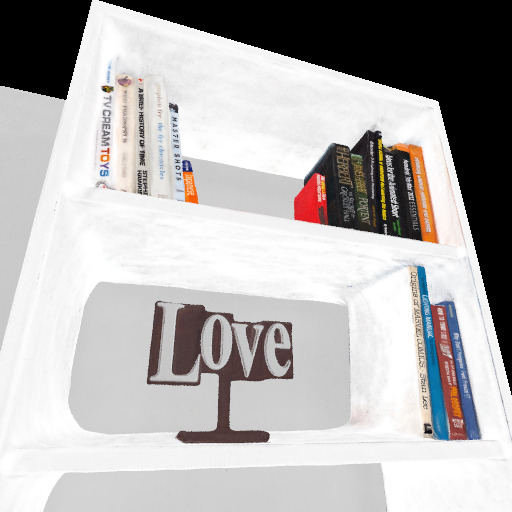}
& \includegraphics[width=\width]{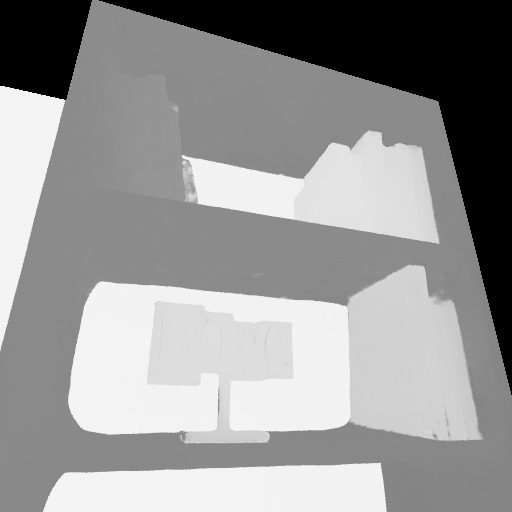}
& \includegraphics[width=\width]{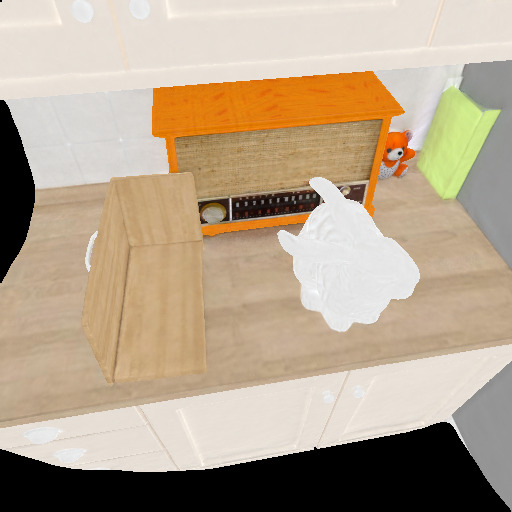}
& \includegraphics[width=\width]{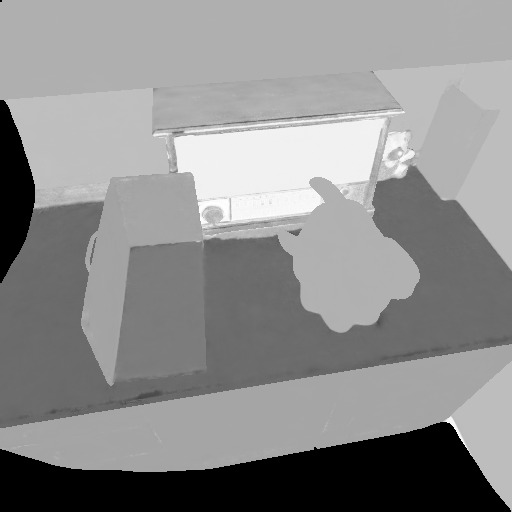}
\\

{\makebox{\rotatebox{90}{GT}}}
& \includegraphics[width=\width]{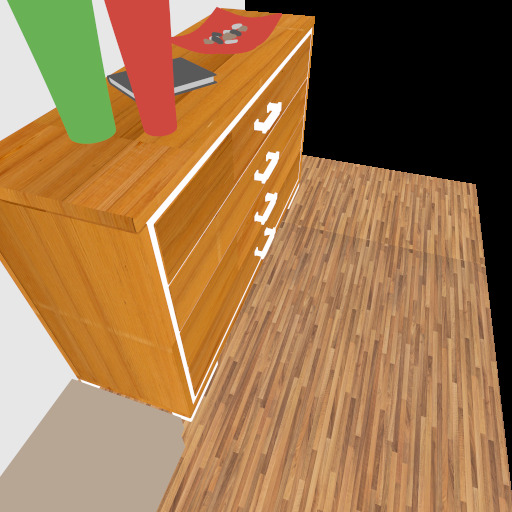}
& \includegraphics[width=\width]{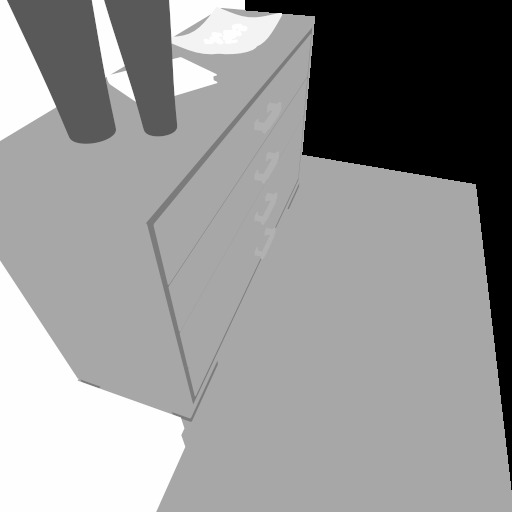}
& \includegraphics[width=\width]{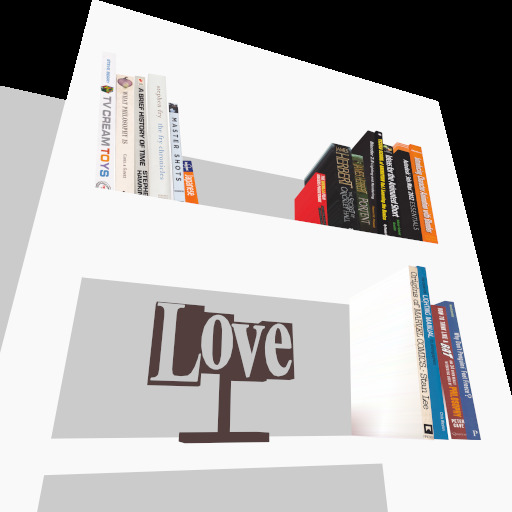}
& \includegraphics[width=\width]{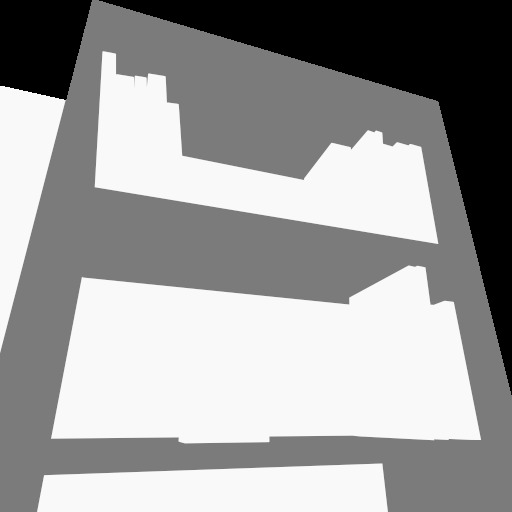}
& \includegraphics[width=\width]{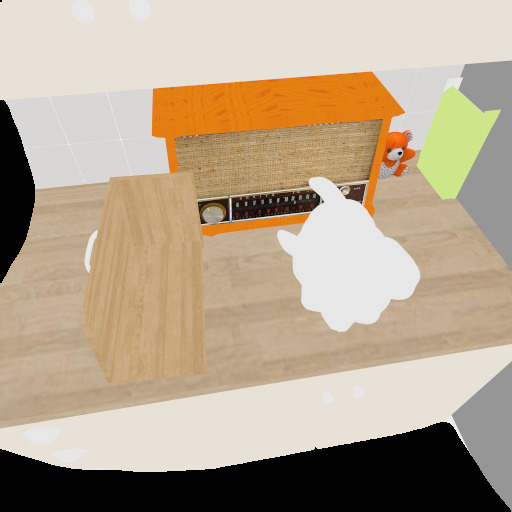}
& \includegraphics[width=\width]{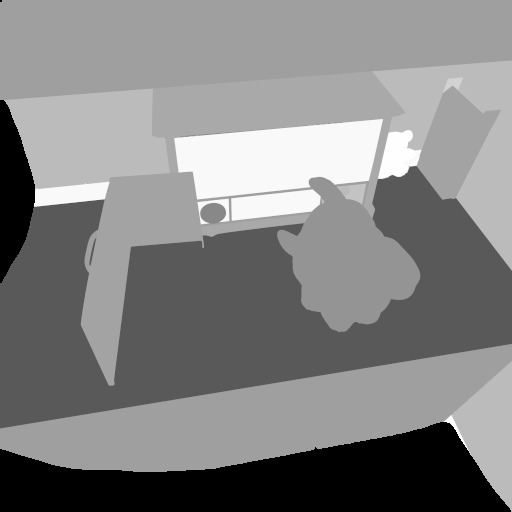}
\\

\end{tabular}

%% file: figures/qualitative_figure_real.tex
\providelength\width
\setlength\width{2.5cm}

\providelength\width
\setlength\width{1.9cm}

\begin{figure*}
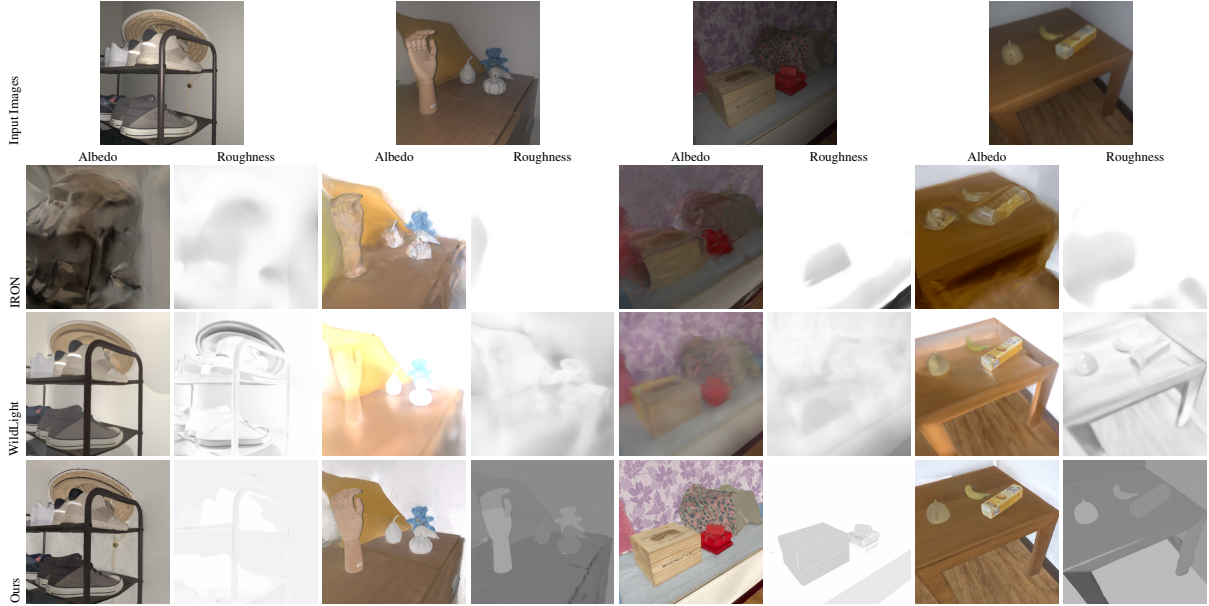

\tiny
\centering

\renewcommand{\tabcolsep}{1pt}

\input generated/qualitative_generated_params_real_col
\caption{
\textbf{Qualitative comparison of reflectance estimation on real scenes captured with co-located light \& camera.} We present estimated albedo, and roughness in validation views for the real scenes. Our method produces significantly better albedo and roughness w.r.t. prior co-located \& camera methods due to our ability to better model global illumination. (Zoom for visualization.) }
\label{fig:qualitative_figure_real}

\end{figure*}

%% file: generated/qualitative_generated_params_real_col.tex
\begin{tabular}{ccccccccc}
& \multicolumn{2}{c}{Shoe Rack}
& \multicolumn{2}{c}{Table}
& \multicolumn{2}{c}{Window Sill}
& \multicolumn{2}{c}{Coffee Table}
\\

{\makebox{\rotatebox{90}{Input Images}}}
& \multicolumn{2}{c}{\includegraphics[width=\width]{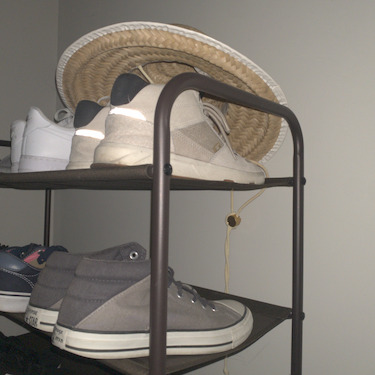}}
& \multicolumn{2}{c}{\includegraphics[width=\width]{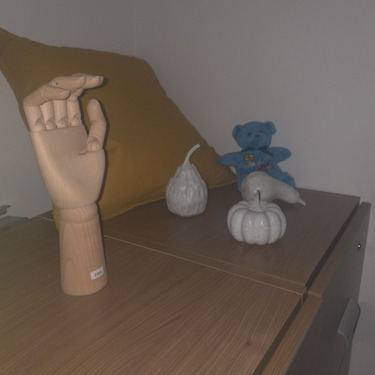}}
& \multicolumn{2}{c}{\includegraphics[width=\width]{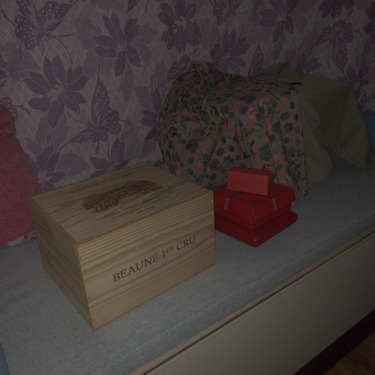}}
& \multicolumn{2}{c}{\includegraphics[width=\width]{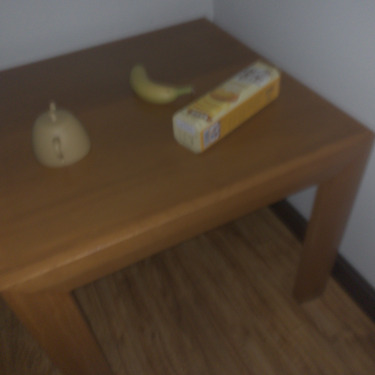}}
\\
&Albedo & Roughness & Albedo & Roughness & Albedo & Roughness & Albedo & Roughness \\
{\makebox{\rotatebox{90}{IRON}}}
& \includegraphics[width=\width]{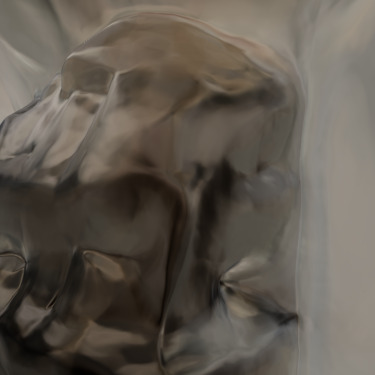}
& \includegraphics[width=\width]{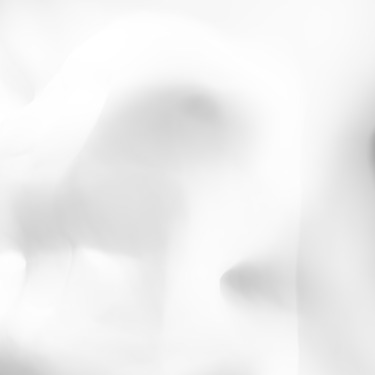}
& \includegraphics[width=\width]{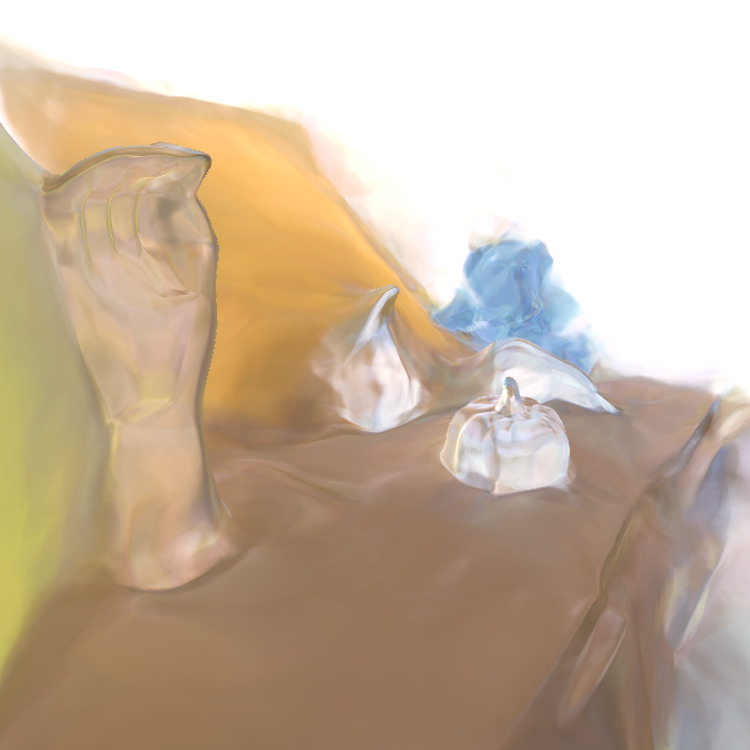}
& \includegraphics[width=\width]{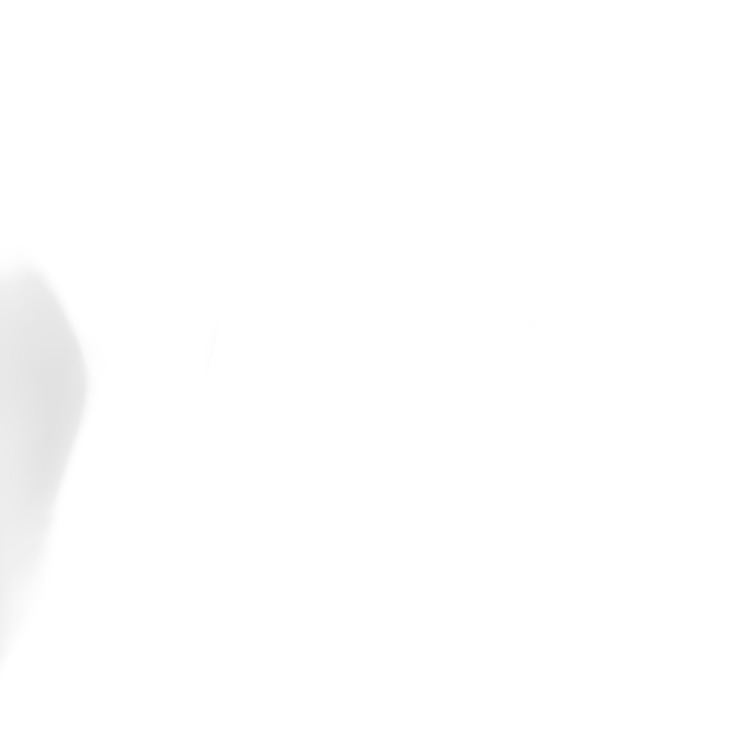}
& \includegraphics[width=\width]{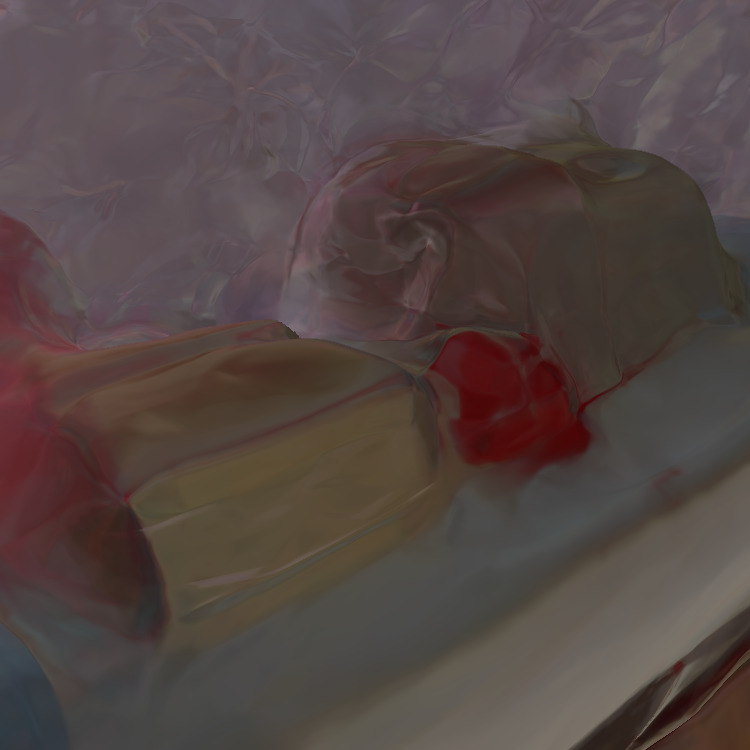}
& \includegraphics[width=\width]{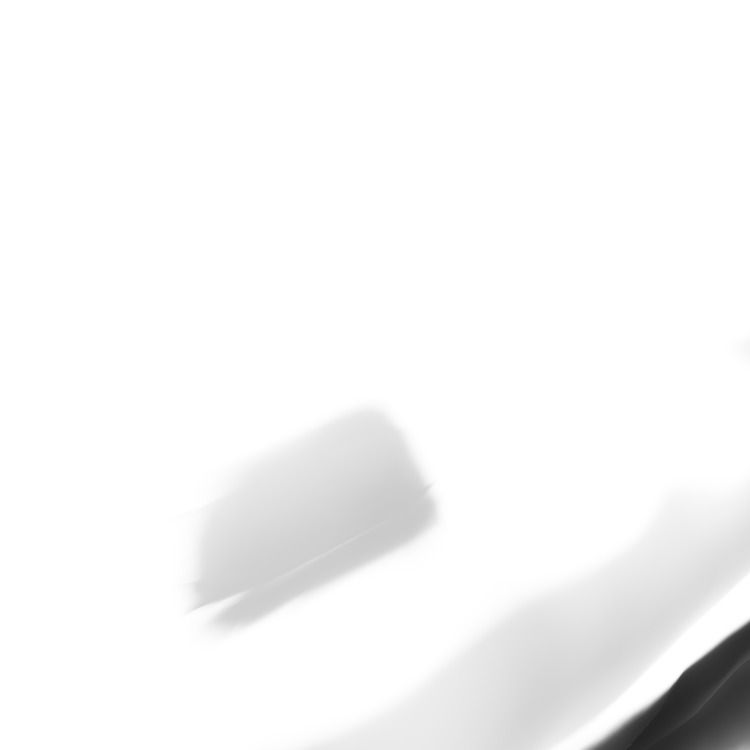}
& \includegraphics[width=\width]{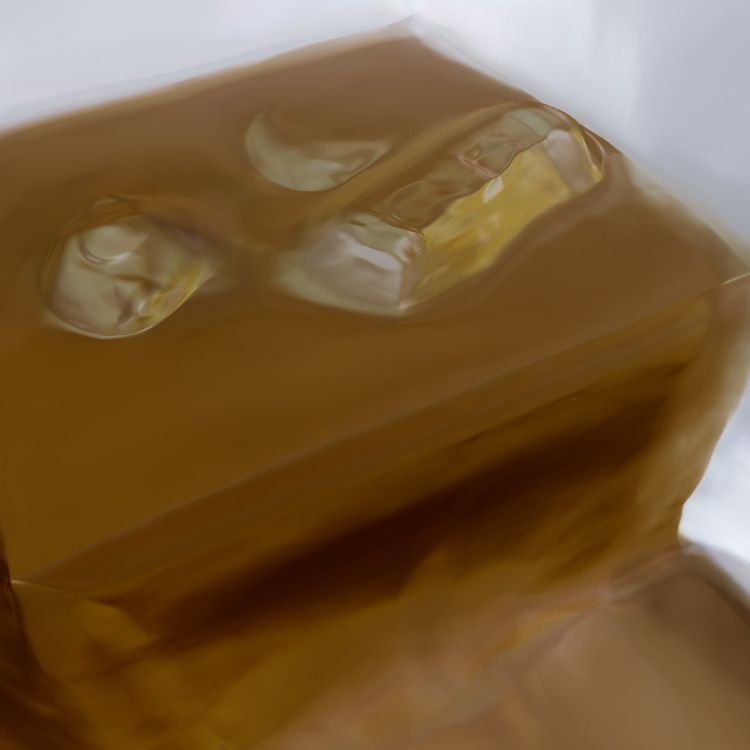}
& \includegraphics[width=\width]{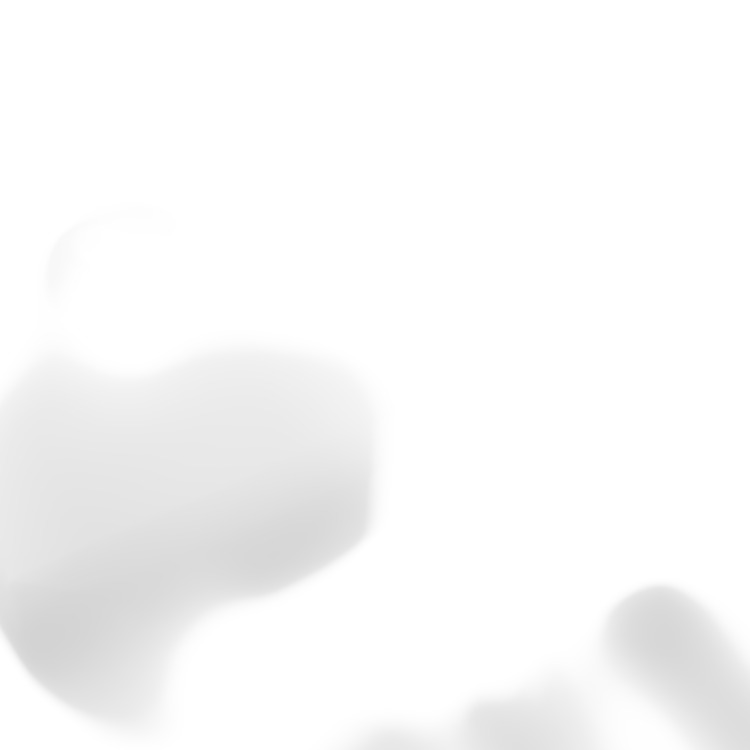}
\\

{\makebox{\rotatebox{90}{WildLight}}}
& \includegraphics[width=\width]{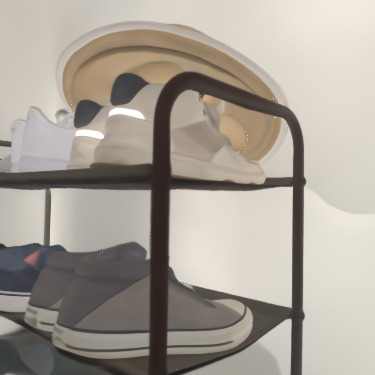}
& \includegraphics[width=\width]{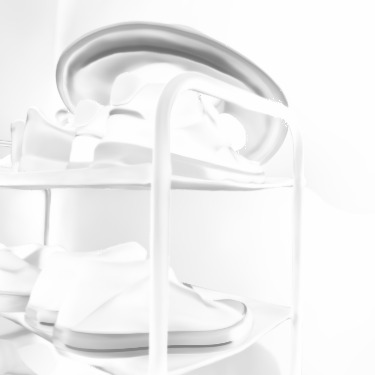}
& \includegraphics[width=\width]{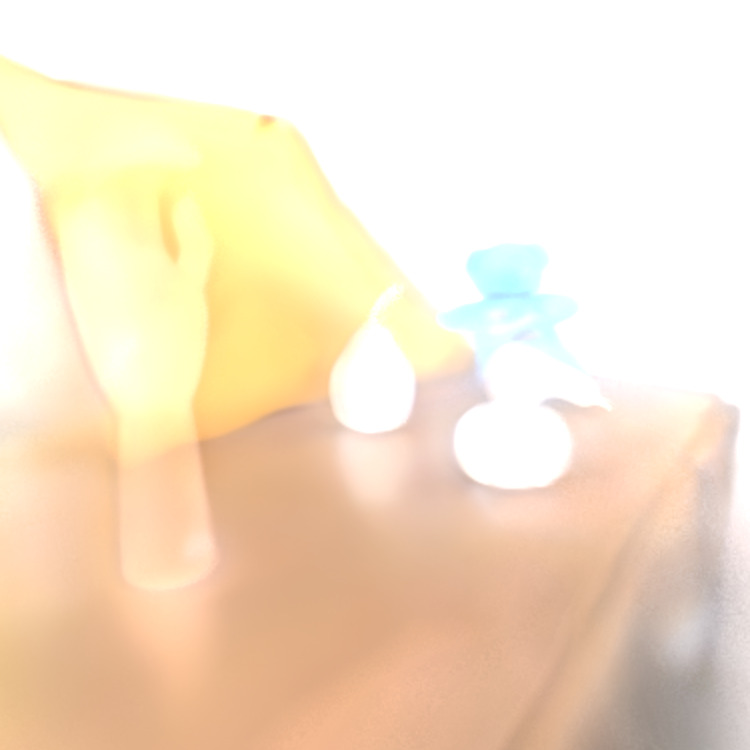}
& \includegraphics[width=\width]{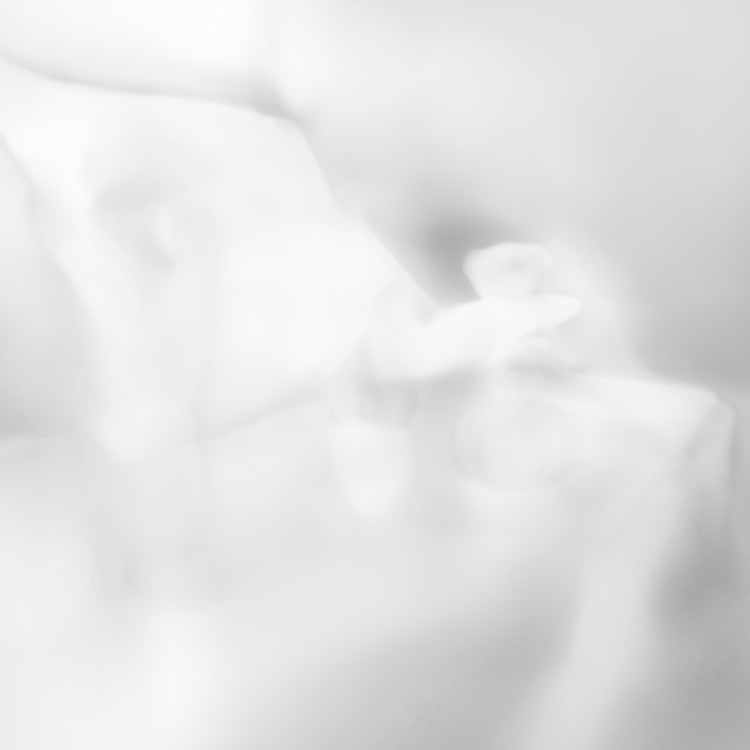}
& \includegraphics[width=\width]{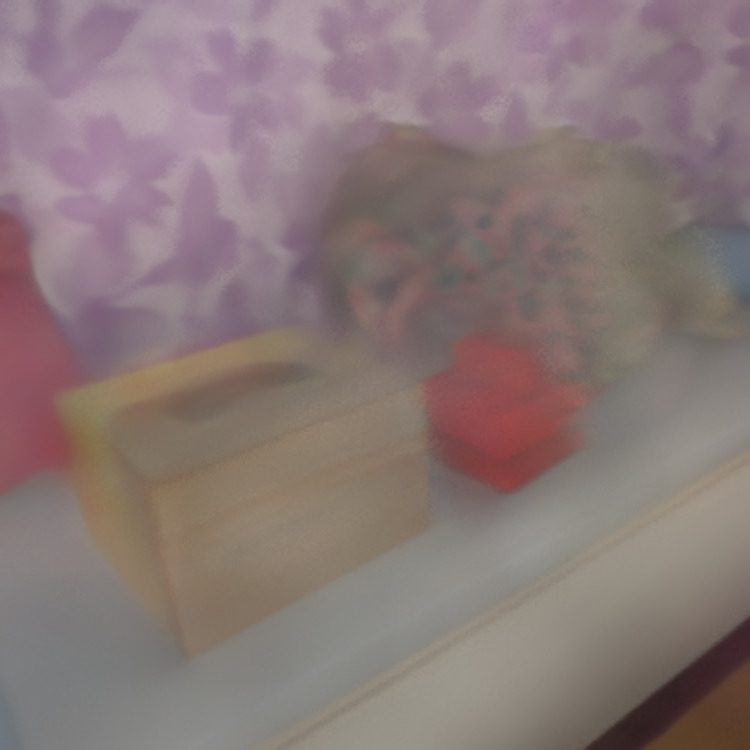}
& \includegraphics[width=\width]{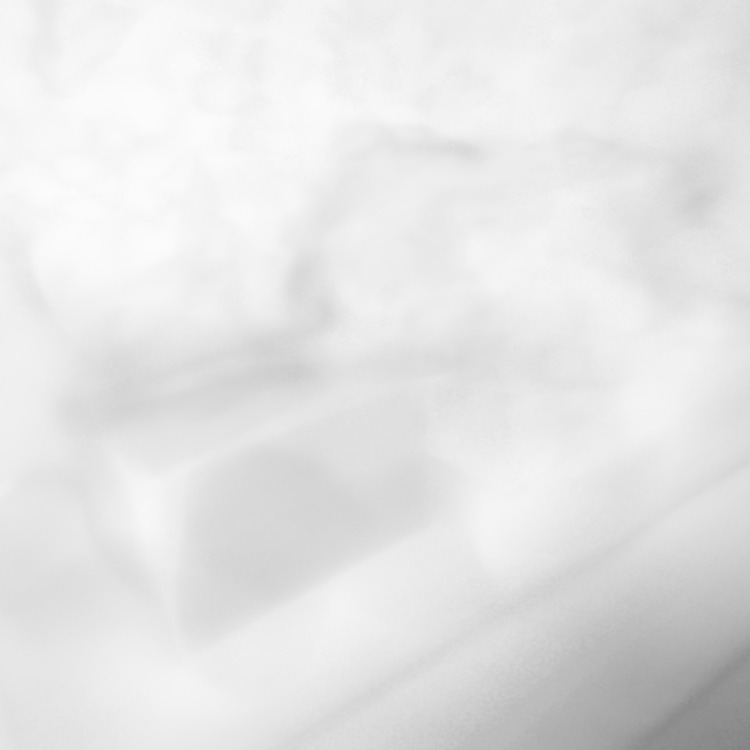}
& \includegraphics[width=\width]{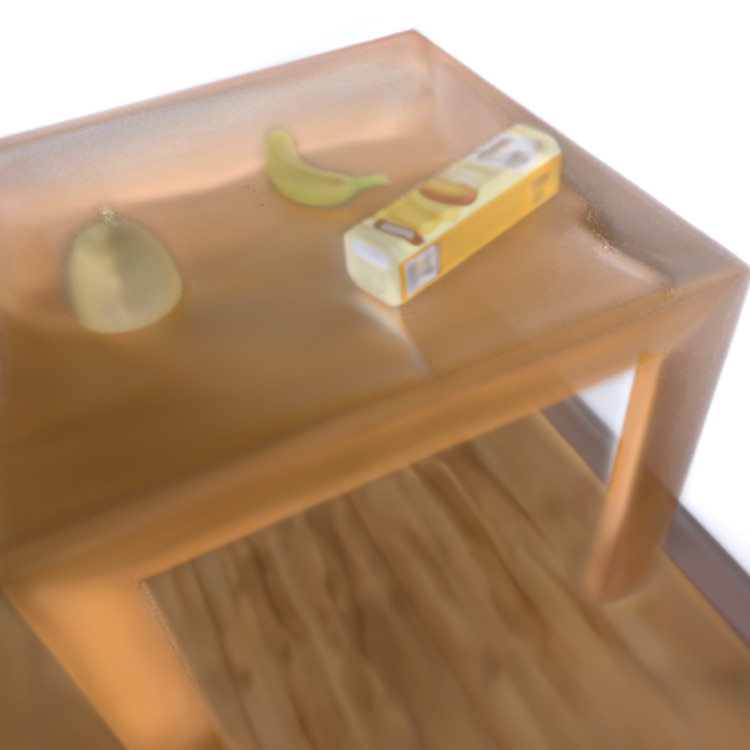}
& \includegraphics[width=\width]{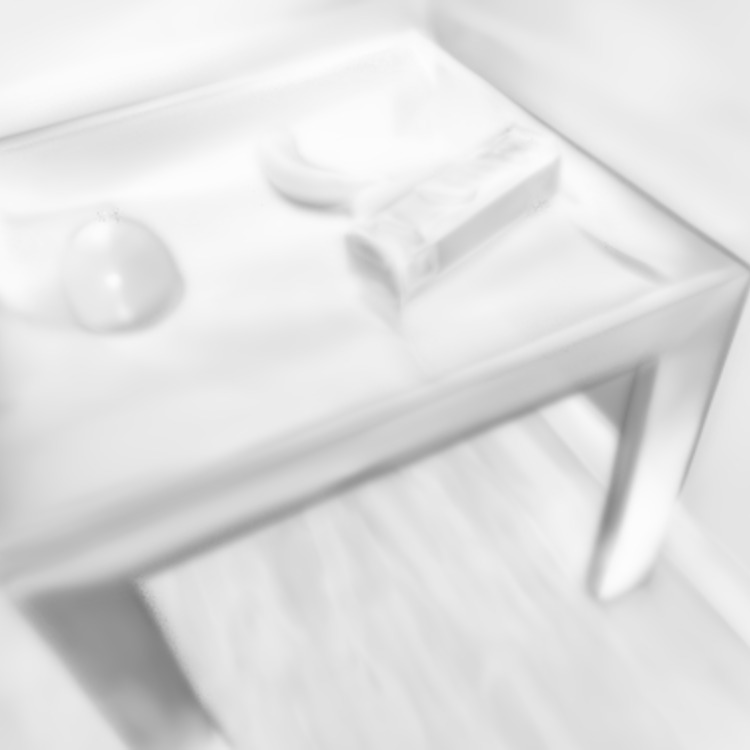}
\\

{\makebox{\rotatebox{90}{Ours}}}
& \includegraphics[width=\width]{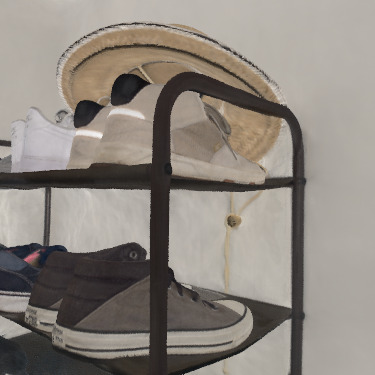}
& \includegraphics[width=\width]{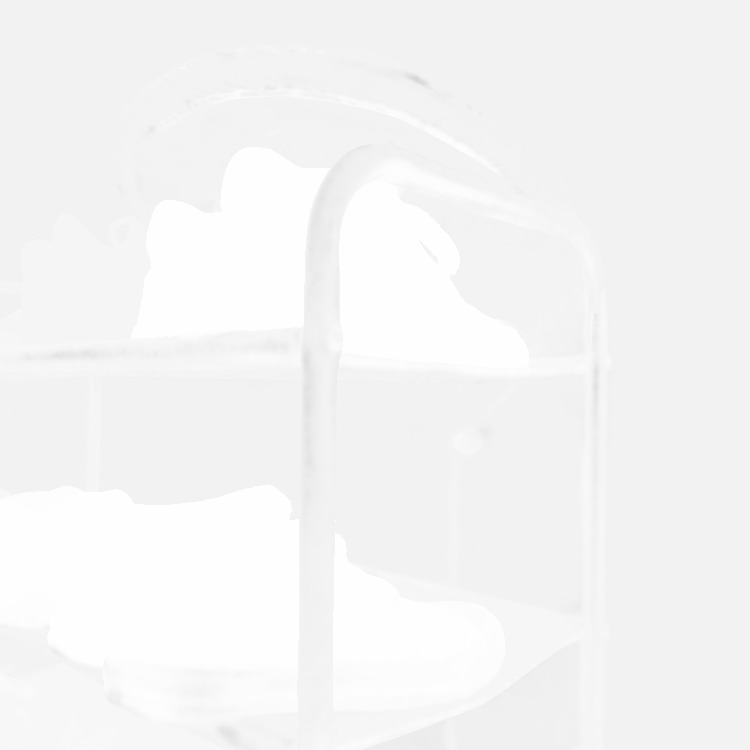}
& \includegraphics[width=\width]{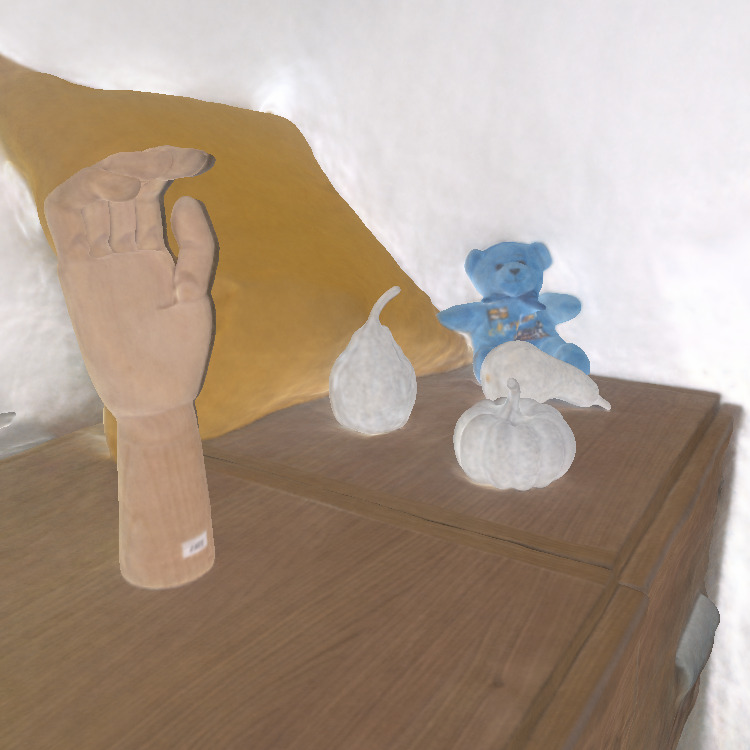}
& \includegraphics[width=\width]{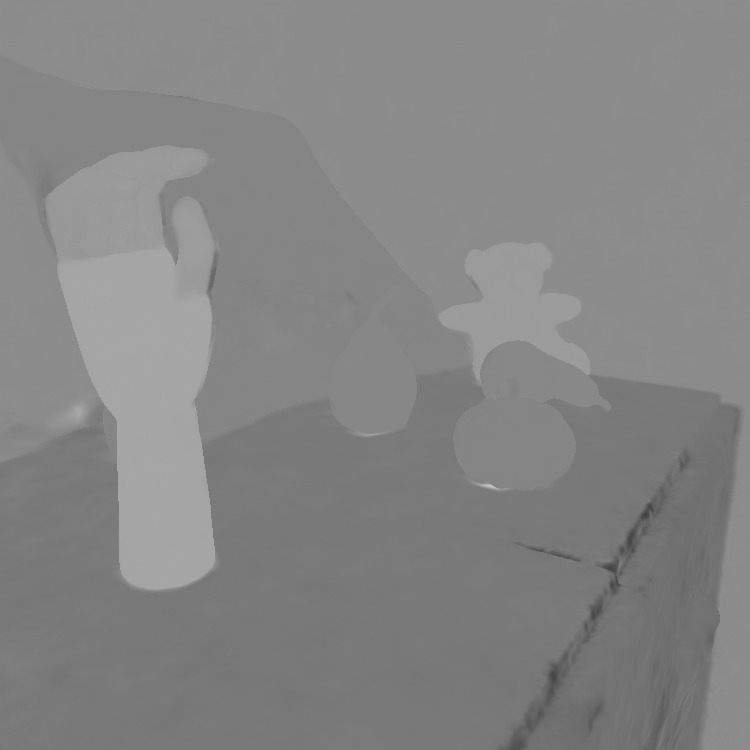}
& \includegraphics[width=\width]{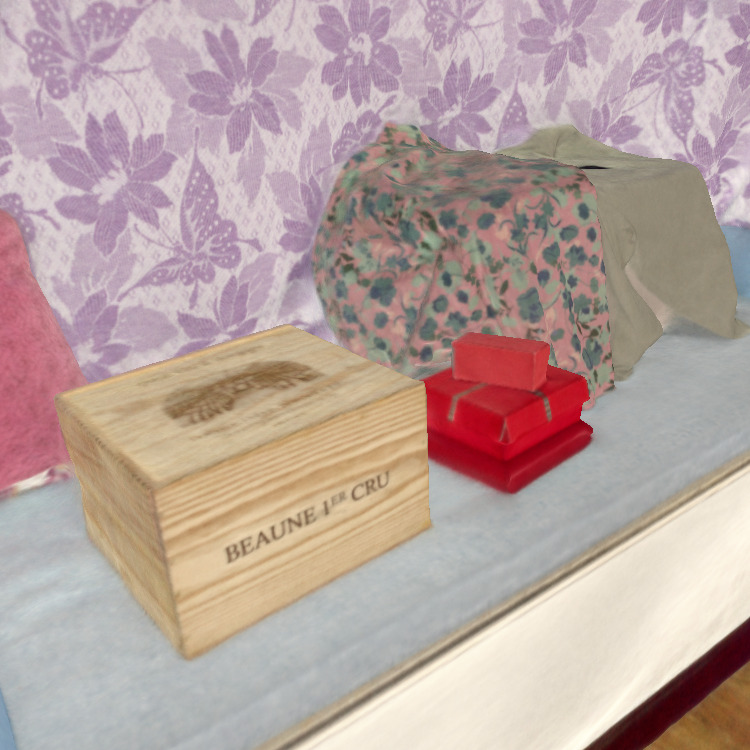}
& \includegraphics[width=\width]{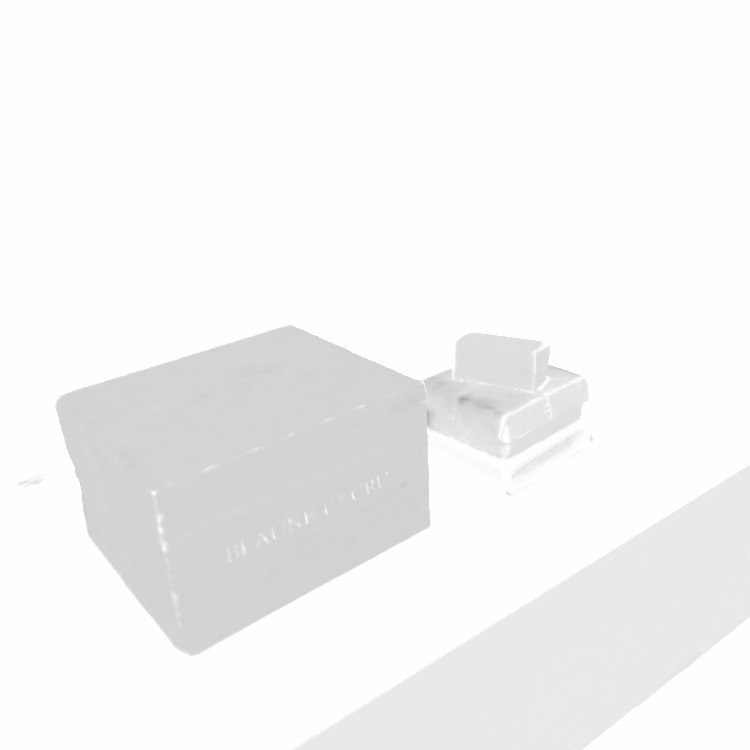}
& \includegraphics[width=\width]{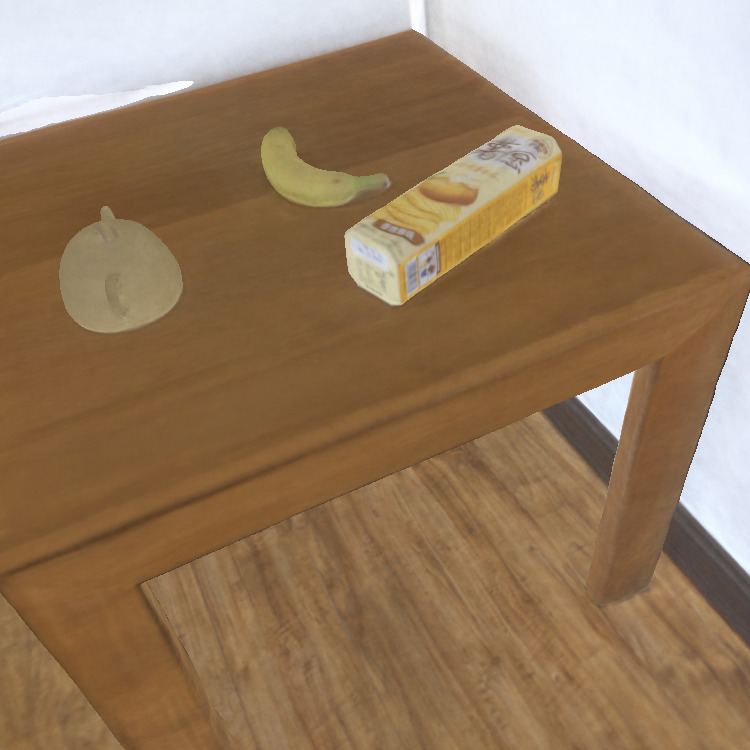}
& \includegraphics[width=\width]{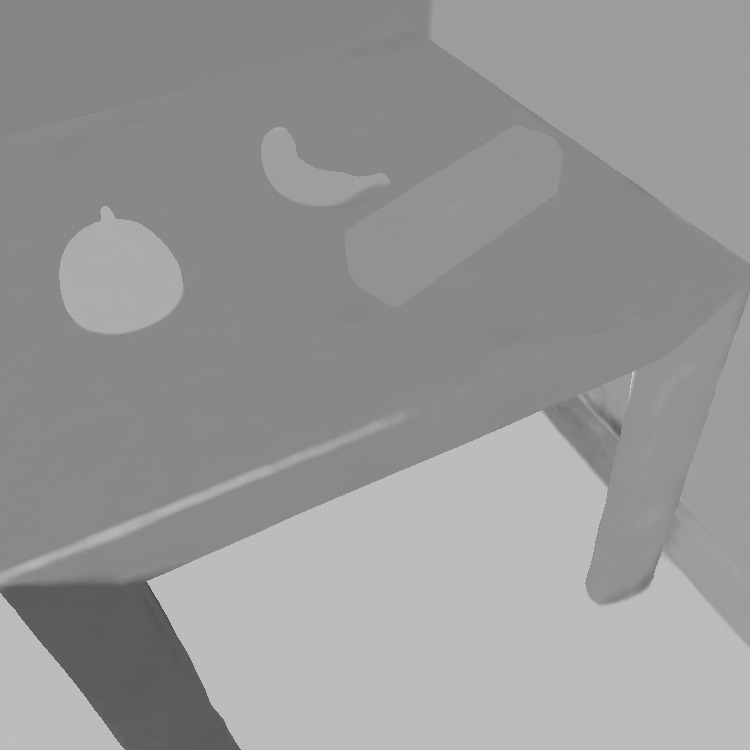}
\\

\end{tabular}

%% file: figures/main_metrics_table.tex
\begin{table}[tbp]
    \centering
    \renewcommand{\tabcolsep}{1pt}
    \resizebox{\columnwidth}{!}{
        \begin{tabular}{crcccccccccccccc}
 & & \multicolumn{4}{c}{Albedo} & & \multicolumn{4}{c}{Roughness} \\
\cmidrule{3-6} \cmidrule{8-11}
  &  & Bedroom  & Shelf & Counter & \textit{Mean} &  & Bedroom  & Shelf & Counter & \textit{Mean} \\
 \cmidrule{3-11}
\input{generated/main_metrics_table_withmean_generated}

    \end{tabular}
  }
      \caption{\textbf{Quantitative evaluation of reflectance on synthetic data}. We present MSE ($\times 10$) of re-rendering in unseen views, albedo, and roughness on the validation set of synthetic data. Best is shown in \textbf{bold} and second places \underline{underlined}. Top two rows are natural illumination methods while bottom three rows are co-located light \& camera methods.} 

    \label{tab:main_metrics}
\end{table}

%% file: generated/main_metrics_table_withmean_generated.tex
NeRO &  & 0.19 & 0.35 & 0.20 & 0.25 &  & 0.97 & 3.03 & 1.81 & 1.94 \\

IRGS &  & 0.16 & 0.54 & 0.14 & 0.28 &  & 0.77 & \underline{1.28} & \underline{1.72} & \underline{1.25} \\

\cmidrule{ 2-11 } 
IRON &  & 0.47 & \underline{0.27} & 0.29 & 0.34 &  & 1.00 & 2.53 & 2.64 & 2.06 \\

WildLight &  & \underline{0.04} & 0.33 & \underline{0.14} & \underline{0.17} &  & \underline{0.64} & 1.34 & 1.93 & 1.30 \\

Ours &  & \textbf{0.03} & \textbf{0.04} & \textbf{0.02} & \textbf{0.03} &  & \textbf{0.23} & \textbf{0.16} & \textbf{0.06} & \textbf{0.15} \\

%% file: figures/synthetic_geometry_metrics.tex
\begin{table}
\centering
\renewcommand{\tabcolsep}{2pt}
\resizebox{\columnwidth}{!}{
\input generated/synthetic_geometry_metrics
}
\caption{\textbf{Quantitative evaluation of geometry on synthetic data.} Our method produces comparable geometry compared to prior state of the arts. Top two rows are natural illumination methods while bottom three rows are co-located light \& camera methods.}
\vspace{-1em}
\label{tab:synthetic_geometry_metrics}
\end{table}

%% file: generated/synthetic_geometry_metrics.tex
\begin{tabular}{ rcccccccccc }
& & \multicolumn{4}{c}{Chamfer ($\times 100$)} & & \multicolumn{4}{c}{Depth Map L1} \\
\cmidrule{3-6} \cmidrule{8-11}
& & {Bedroom}  & {Shelf}  & {Counter} & {Mean}  & & {Bedroom}  & {Shelf}  & {Counter} & {Mean} \\
\cmidrule{3-11}
NeRO & & \underline{0.01} & \underline{0.01} & \underline{0.24} & \underline{0.09} & 
& \underline{0.01} & \textbf{0.00} & \underline{0.03} & \underline{0.01}  \\
IRGS & & 0.15 & 0.45 & 0.62 & 0.41 & 
& 0.03 & 0.09 & 0.04 & 0.05 \\
\cmidrule{3-11}
IRON & & 1.01 & 0.19 & 1.67 & 0.96 & 
& 0.10 & 0.08 & 0.17 & 0.12\\
WildLight & & \textbf{0.00} & 0.01 & 0.44 & 0.15 &
& \textbf{0.00} & 0.03 & 0.09 & 0.04 \\
Ours & & 0.02 & \textbf{0.00} & \textbf{0.04} & \textbf{0.02} & 
& 0.01 & \underline{0.01} & \textbf{0.01} & \textbf{0.01} \\
\end{tabular}

%% file: figures/merged_render_geometry.tex
\providelength\width
\setlength\width{1.0cm}
\begin{figure}
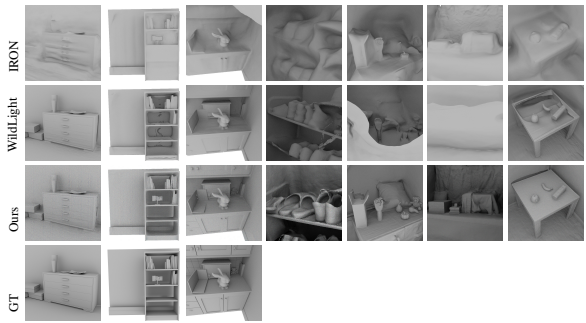

\tiny
\centering
\renewcommand{\tabcolsep}{1pt}
\begin{tabular}{cccccccc}
\input generated/geometry_figure_generated
\end{tabular}
\caption{\textbf{Qualitative comparison of geometry with co-located light \& camera methods.} Our method produces better geometry than prior co-located light \& camera methods. }
\label{fig:geom_comparison}
\end{figure}

%% file: generated/geometry_figure_generated.tex
{\makebox{\rotatebox{90}{\hspace{2pt} IRON}}}
& \includegraphics[width=\width]{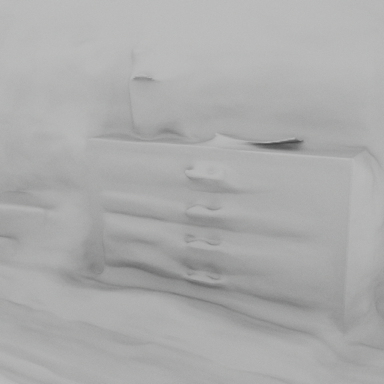}
& \includegraphics[width=\width]{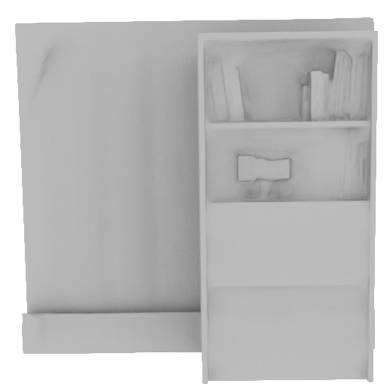}
& \includegraphics[width=\width]{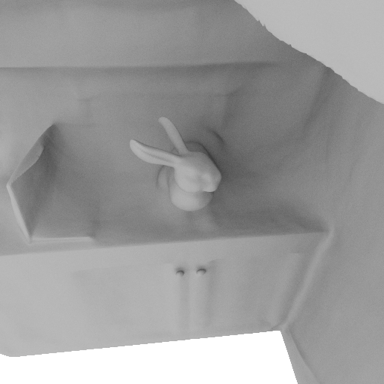}
& \includegraphics[width=\width]{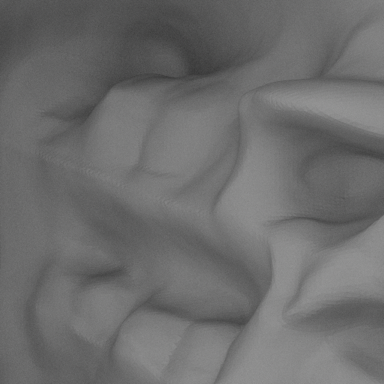}
& \includegraphics[width=\width]{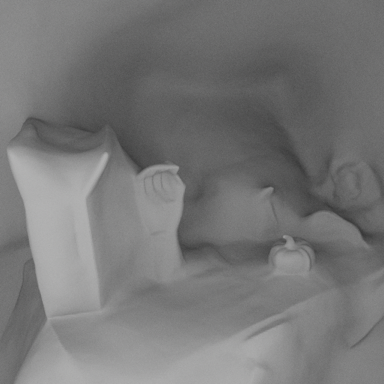}
& \includegraphics[width=\width]{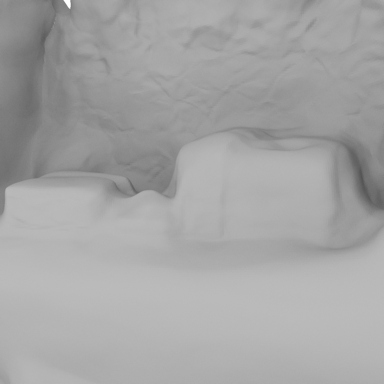}
& \includegraphics[width=\width]{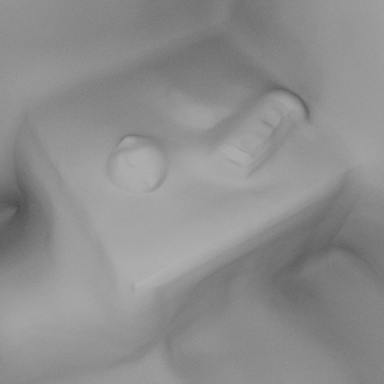}
\\

{\makebox{\rotatebox{90}{\hspace{2pt} WildLight}}}
& \includegraphics[width=\width]{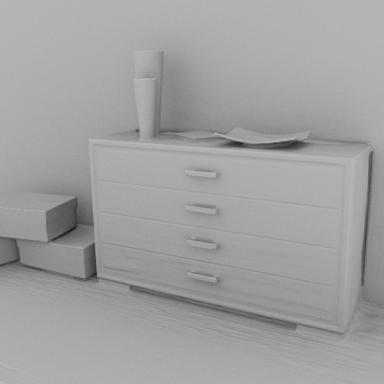}
& \includegraphics[width=\width]{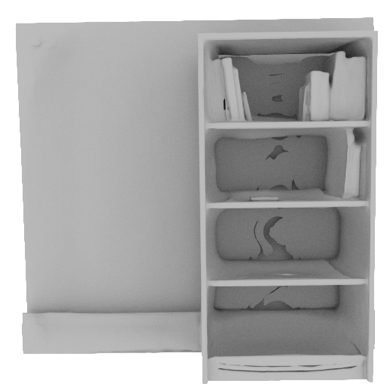}
& \includegraphics[width=\width]{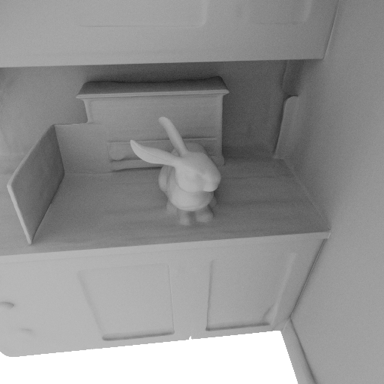}
& \includegraphics[width=\width]{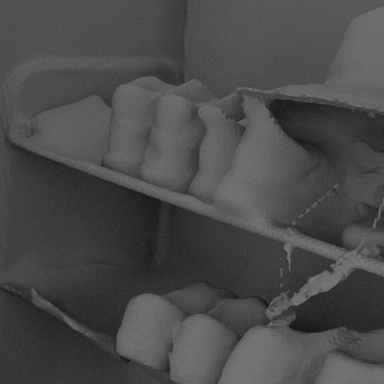}
& \includegraphics[width=\width]{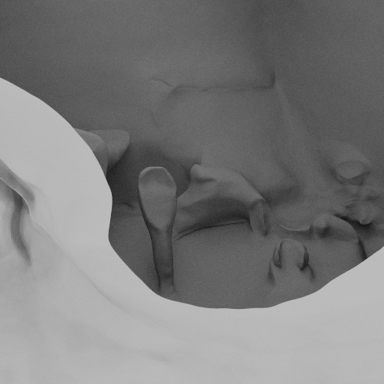}
& \includegraphics[width=\width]{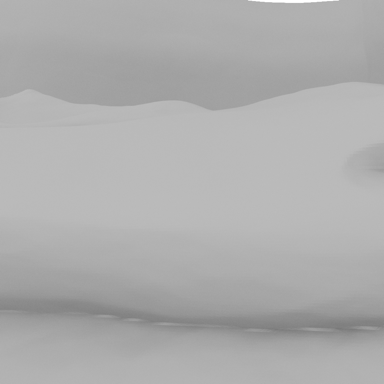}
& \includegraphics[width=\width]{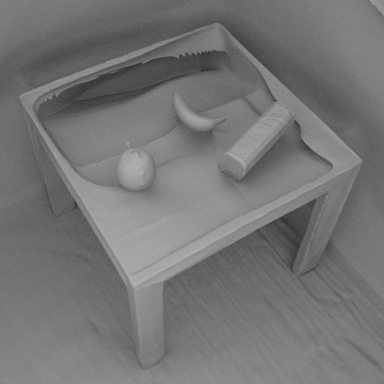}
\\

{\makebox{\rotatebox{90}{\hspace{2pt} Ours}}}
& \includegraphics[width=\width]{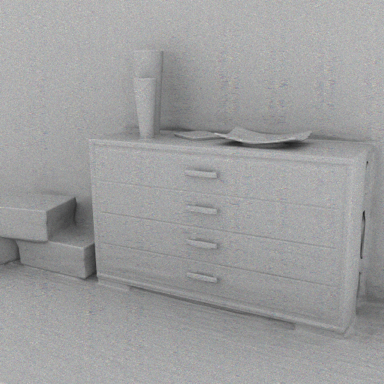}
& \includegraphics[width=\width]{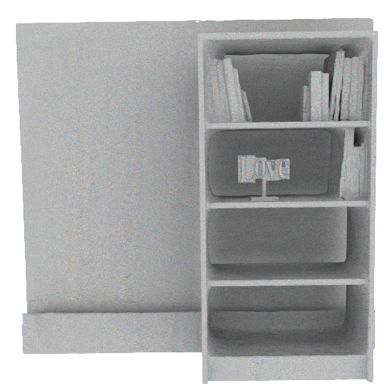}
& \includegraphics[width=\width]{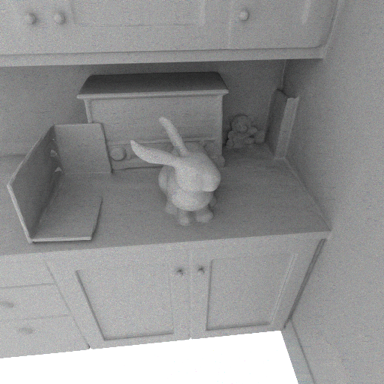}
& \includegraphics[width=\width]{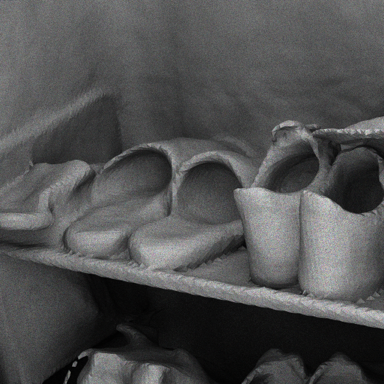}
& \includegraphics[width=\width]{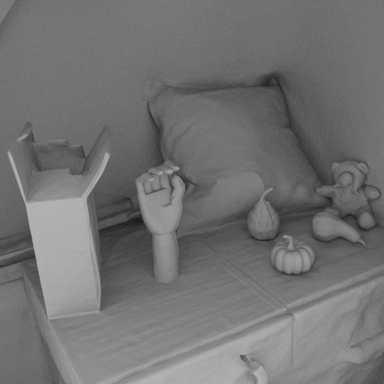}
& \includegraphics[width=\width]{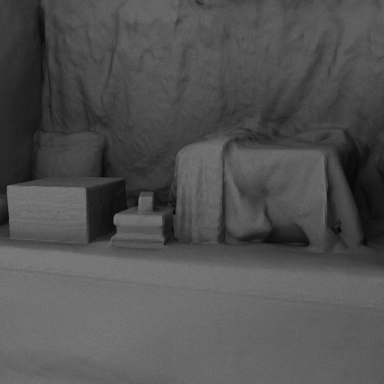}
& \includegraphics[width=\width]{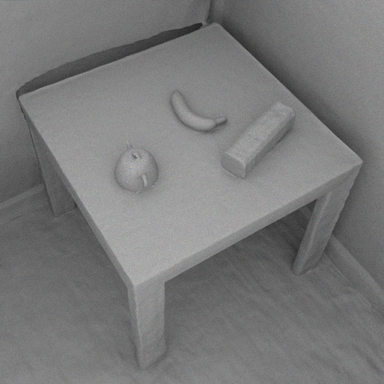}
\\

{\makebox{\rotatebox{90}{\hspace{2pt} GT}}}
& \includegraphics[width=\width]{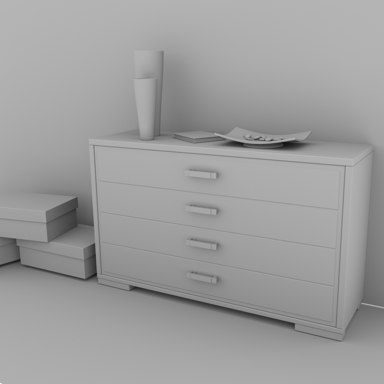}
& \includegraphics[width=\width]{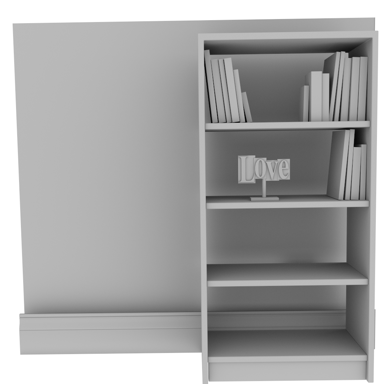}
& \includegraphics[width=\width]{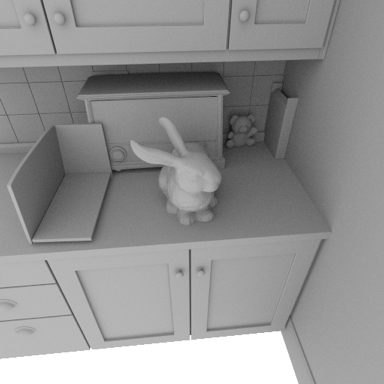}
&
&
&
&
\\

%% file: figures/ablation_figure_glow.tex
\providelength\width
\setlength\width{1.6cm}
\begin{figure}
\tiny
\centering
\renewcommand{\tabcolsep}{1pt}
\begin{tabular}{cccccc}
    &GT & Ours & Path  & Direct & Naive Cache   \\
     \input generated/qualitative_generated_abl
\end{tabular}

\caption{
\textbf{Qualitative results on ablation of radiance cache.} For this ablation, we optimize material properties on ground truth mesh geometry of \textit{kitchen} scene. \textit{Path} is path tracing algorithm for reference. \textit{Direct} is direct illumination renderer. \textit{Naive Cache} is our algorithm but with naive radiance cache instead of dynamic radiance cache.}
\label{fig:ablation_figure}
\end{figure}

%% file: generated/qualitative_generated_abl.tex
\makebox{\rotatebox{90}{Image}}
&\includegraphics[width=\width]{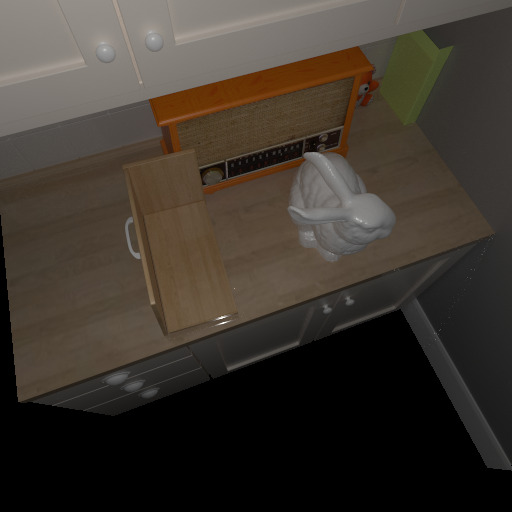}
& \includegraphics[width=\width]{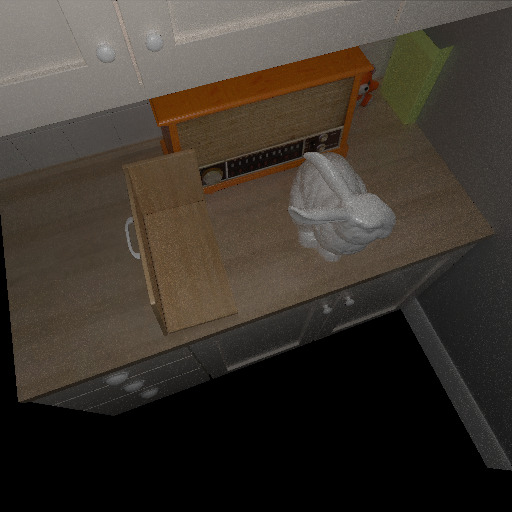}
& \includegraphics[width=\width]{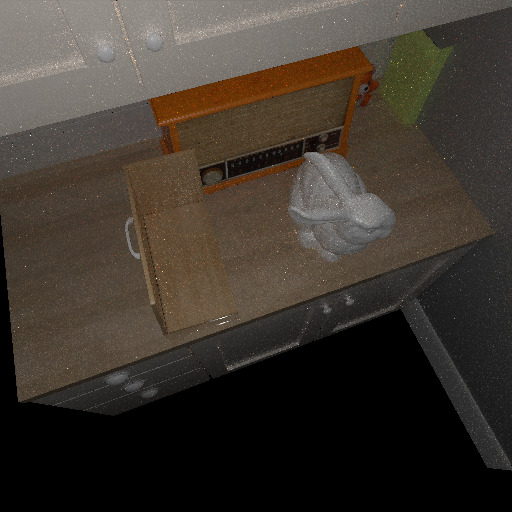}
& \includegraphics[width=\width]{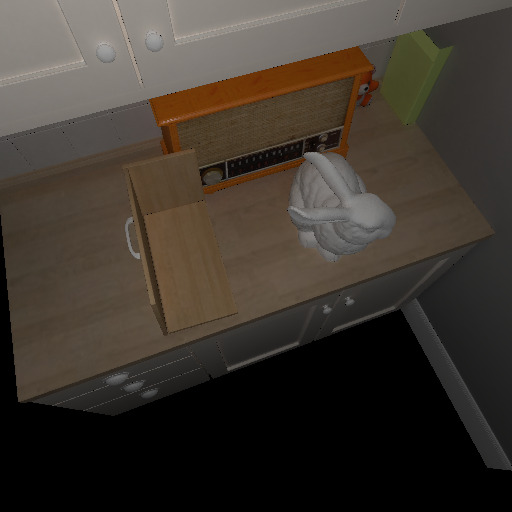}
& \includegraphics[width=\width]{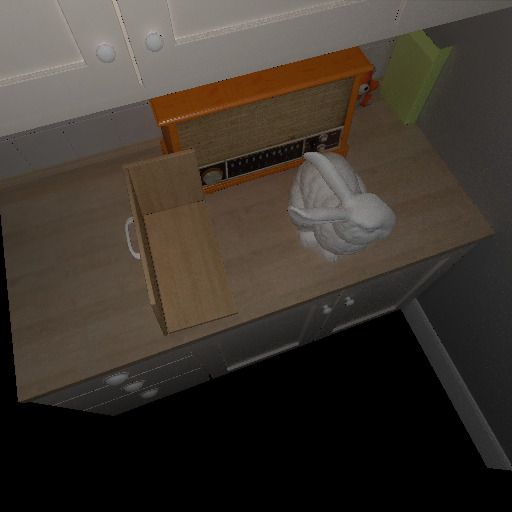}
\\

\makebox{\rotatebox{90}{Albedo}}
&\includegraphics[width=\width]{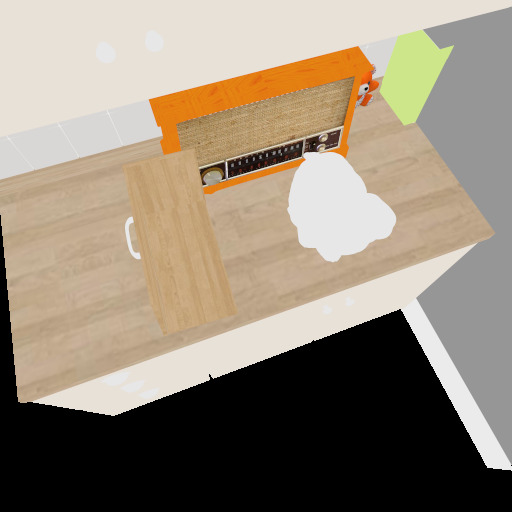}
& \includegraphics[width=\width]{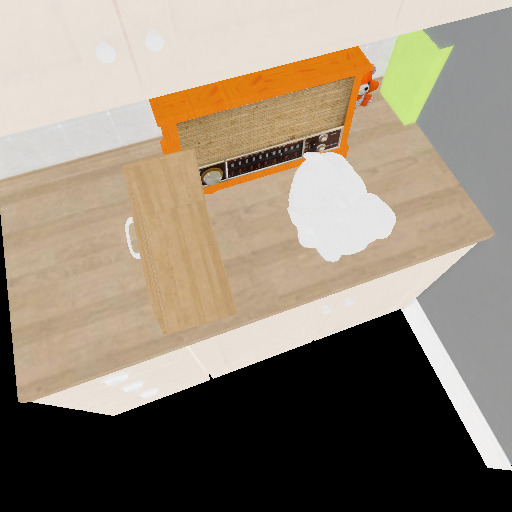}
& \includegraphics[width=\width]{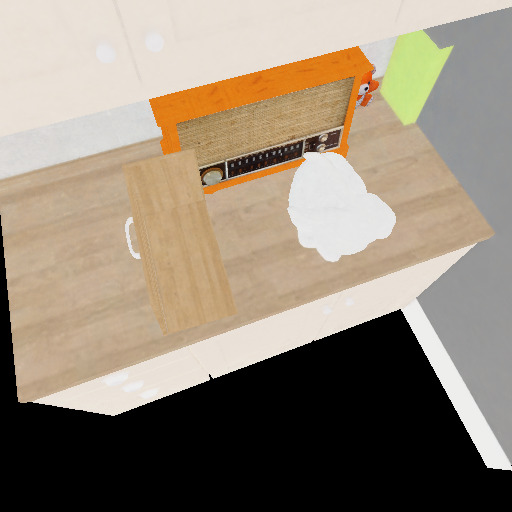}
& \includegraphics[width=\width]{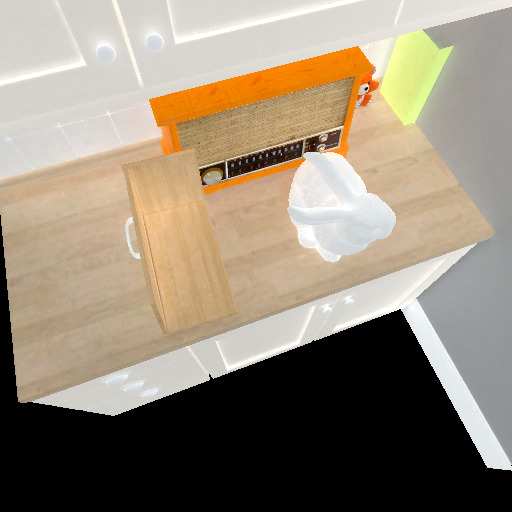}
& \includegraphics[width=\width]{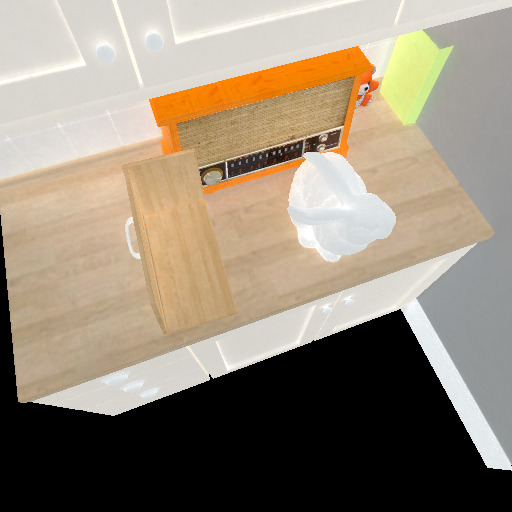}
\\

\makebox{\rotatebox{90}{Roughness}}
&\includegraphics[width=\width]{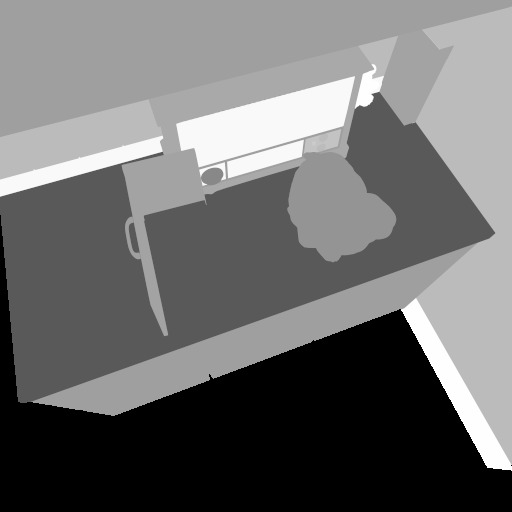}
& \includegraphics[width=\width]{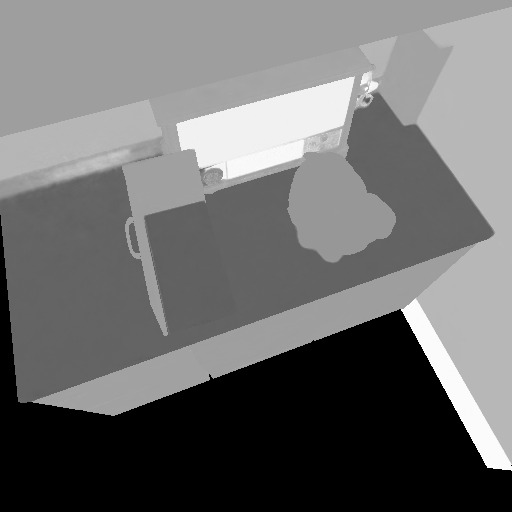}
& \includegraphics[width=\width]{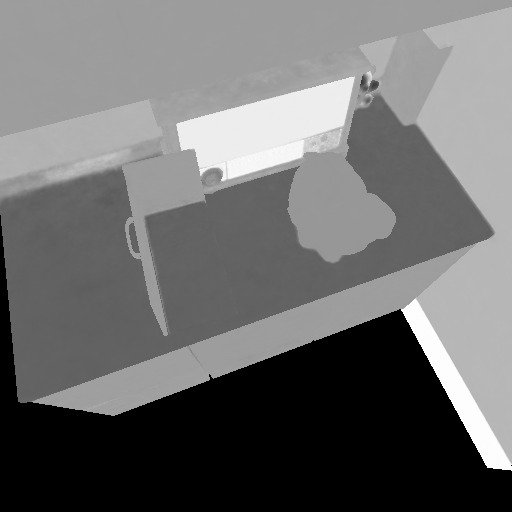}
& \includegraphics[width=\width]{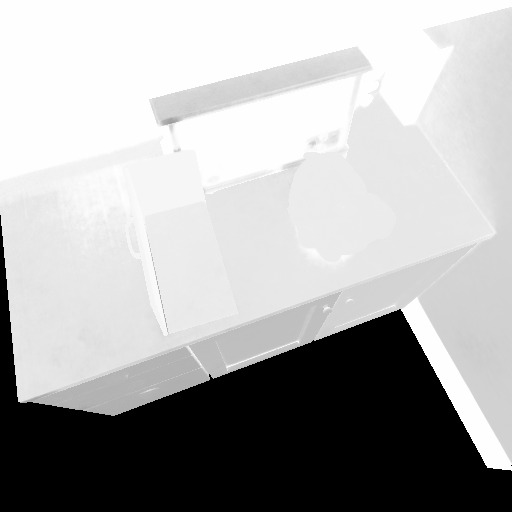}
& \includegraphics[width=\width]{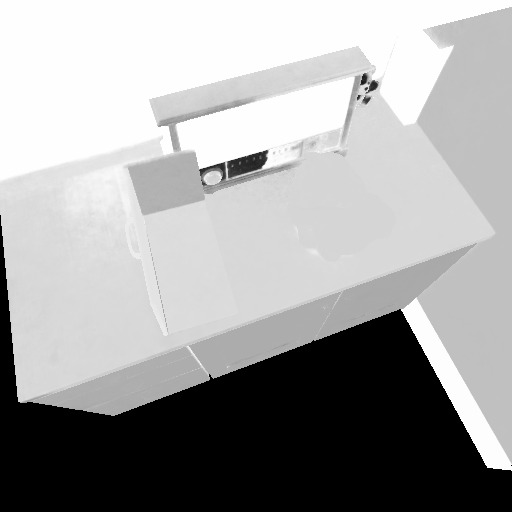}
\\

%% file: figures/ablation_figure_metrics.tex
\begin{table}[tbp]
    \centering
    \renewcommand{\tabcolsep}{5pt}
\begin{tabular}{|c|c|c|c|c|}
\hline
      & Path & Ours & Direct & Naive Cache   \\
\hline
Albedo & 0.38 & \textbf{0.35} & 7.21 & 4.97 \\
\hline
Roughness & 7.46 & \textbf{6.11} & 304.8 & 235.6 \\
\hline
Runtime & 3097 & 821 & \textbf{453} & 794 \\
\hline
\end{tabular}
\caption{
\textbf{Quantitative results on ablation of radiance cache.} For this ablation, we optimize material properties on ground truth mesh geometry of \textit{kitchen} scene. For albedo and roughness, the reported number is MSE (\(x10^3\)), and Runtime is in minutes. \textit{Path} is path tracing algorithm for reference. \textit{Direct} is direct illumination renderer. \textit{Naive Cache} is our algorithm but with naive radiance cache instead of dynamic radiance cache. 
}
\label{tab:abl_cache}
\end{table}

%% file: sections/conclusion.tex
\section{Conclusion}

We introduced GLOW, a Global Illumination-aware inverse rendering framework tailored for multi-object indoor scenes captured with a co-located light–camera setup. By combining neural implicit surface reconstruction with a dynamic radiance cache and a specular-aware loss, GLOW effectively handles near-field lighting, inter-reflections, and moving specular highlights—key challenges that limit existing methods. Extensive experiments on synthetic and real scenes demonstrate that GLOW achieves geometry quality comparable to state-of-the-art natural illumination methods while delivering a substantial improvement in material reflectance estimation. These results highlight the potential of co-located lighting, when paired with proper modeling of global illumination, to significantly advance inverse rendering quality.

\section{Acknowledgement}
This research is based upon work supported by the Office of the Director of National Intelligence (ODNI), Intelligence Advanced Research Projects Activity (IARPA), via IARPA R\&D Contract No. 140D0423C0076. The views and conclusions contained herein are those of the authors and should not be interpreted as necessarily representing the official policies or endorsements, either expressed or implied, of the ODNI, IARPA, or the U.S. Government. The U.S. Government is authorized to reproduce and distribute reprints for Governmental purposes notwithstanding any copyright annotation thereon. Roni Sengupta is supported by the National Institute of Health (NIH) projects \#R21EB035832 and \#R21EB037440 for developing lighting-based 3D reconstruction algorithms. Jiaye Wu and David Jacobs were supported in part by the National Science Foundation under grant no. IIS-1910132 and IIS-2213335.

%% file: sections/supp.tex
\clearpage

\renewcommand{\thesection}{A\arabic{section}}

\section{Overview}
 In this supplementary, we present video demonstrations of the scenes relighted under rich lighting conditions after reconstructed using our method (\ref{sec:suppl_videos}). 

\noindent\textbf{Theories} We provide an background on neural implicit surface volume rendering and physically based rendering (\ref{sec:suppl_background}). We provide additional discussion for non-convexity of co-located photometric stereo during geometry initialization~(\ref{sec:suppl_non_convex}), which motivates us to use SfM points during initialization. We provide a proof (\ref{sec:cont-indir-comp}) that the indirect component of radiance \(\sum_{k=2}^{\infty} \mathcal{T_\phi}^{k}(E) (\mathbf{x}, \omegaout)\) is continuous, which shapes our dynamic radiance cache design.

\noindent\textbf{Experiments} We perform quantitative evaluation on real data following measured albedo in the wild~\ref{sec:maw}. \textbf{All measurements and annotations will be released.} we show ablation of surface angle weighting loss (\ref{sec:saw-supp}). We qualitatively compare with natural illumination methods on material properties (\ref{sec:suppl_qual_natural}), geometry (\ref{sec:suppl_qual_geometry}), re-rendering (\ref{sec:suppl_qual_rerender}). We compare with WildLight~\cite{chengWildLightInthewildInverse2023} on wild-light capture setup (\ref{sec:wildlight_cap}). Finally, we show additional views from our dataset. (\ref{sec:add_views})

\noindent\textbf{Details} We provide additional implementations details on our algorithm (\ref{sec:details}) and real dataset construction (\ref{sec:suppl_data_capture}).

\section{Video Demonstrations} 
\label{sec:suppl_videos}

We render all our scenes under novel lighting conditions with our recovered geometry and material parameters using path tracing renderer mitsuba~\cite{wenzeljakobMitsuba3Renderer}, and denoise with its built-in NVIDIA OptiX denoiser~\cite{parkerOptiXGeneralPurpose2010}. The videos feature ambient lighting with moving point light sources and cameras to demonstrate the practical applications of our method. They are included as separate \texttt{mp4} files.
\section{Background}
\label{sec:suppl_background}
\noindent\textbf{Neural Implicit Surface Volume Rendering} Our method is based on neural implicit surface volume rendering. These techniques~\cite{wangNeuSLearningNeural2021,yarivVolumeRenderingNeural2021} encode geometry with a Signed Distance Field (SDF), which is represented with a neural network called the geometry network \(\operatorname{S_{\Theta_S}}(\mathbf{x})\). They then assume the world is semi-transparent, and develop opacity functions, which map SDF values produced by the geometry network at point \(\mathbf{x}\) to transparency \(w(S_{\Theta_S}(\mathbf{x}))\).

Once transparency is defined, the rest of their system is often similar to NeRF~\cite{mildenhallNeRFRepresentingScenes2021}. Rather than modeling the physics of light, they assume the world directly emits light. The amount of light, or radiance, can be represented with a radiance field, where each scene point is assigned distinct radiance values at every outgoing direction. The radiance field is typically implemented with a neural network called the color network \( \operatorname{C_{\Theta_C}}(\mathbf{x}, \mathbf{v})\rightarrow \mathbf{c} \), where \(\mathbf{c}\) is the radiance and \(\mathbf{v}\) is the outgoing direction.

With \(\mathbf{o}\) as camera position, t as distance along the camera ray, \(\operatorname{p}(t)\) as 3D positions along the camera ray, \(\mathbf{n}\) the surface normal (gradient of SDF \(S_{\Theta_S}(\mathbf{x})\)), \(\mathbf{f}\) the feature vector from the geometry network, we can render the pixels of a view $\operatorname{P}(\mathbf{o}, \mathbf{v})$ with the following equation. More details can be found in NeuS~\cite{wangNeuSLearningNeural2021}.

\begin{equation}
\label{eq:vol_render_main}
    \operatorname{P}(\mathbf{o}, \mathbf{v})=\int_0^{+\infty} w_{\theta_S}(p(t)) \operatorname{C_{\Theta_C}}(\operatorname{p}(t), \mathbf{n}, \mathbf{v}, \mathbf{f}) \text{d}t
\end{equation}

\noindent \textbf{Rendering Equation and Inverse Neural Radiosity.}
InvNeRad~\cite{hadadanInverseGlobalIllumination2023} is an inverse rendering method that leverages a technique for self-supervised training of a radiance cache. InvNeRad is built on top of the rendering equation~\cite{kajiyaRenderingEquation1986}, which defines the outgoing radiance of the scene recursively:
\begin{equation}
\label{eq:rendering_eq}
L(\mathbf{x},\omegaout) = E(\mathbf{x}, \omegaout) +\int_{\mathcal{H}^2} F(\mathbf{x},\omegain, \omegaout) L(r(\mathbf{x},\omegain),-\omegain) d\omegain^{\perp}
\end{equation}
where  \(\mathcal{H}^2\) denotes the hemisphere in the direction of surface normal, and $d\omegain^{\perp}$ is the differential projected solid-angle measure.  \(\mathbf{x}\) is a surface point and \(\omegain\), \(\omegaout\) are the incoming and outgoing directions respectively. \(E\) is the emitter radiance distribution. $L(\mathbf{x},\omegaout)$ is the outgoing radiance. $F(\mathbf{x},\omegain, \omegaout)$ is a bidirectional reflectance distribution function (BRDF). $r(\mathbf{x},\omegain)$ is the ray tracing operator returning the closest surface intersection of ray $(x,\omegain)$.

The equation can be written more compactly in the operator form, where \(\mathcal{T_\phi}\) is the light transport operator based on scene parameters \(\phi\), representing the integral on \(L\). %
\begin{equation}
    \label{eq:rendering_eq_operator}
    L(\mathbf{x}, \omegaout) = E(\mathbf{x}, \omegaout) + \mathcal{T_\phi}(L)(\mathbf{x}, \omegaout)
\end{equation}

InvNeRad introduces radiance cache $L_\theta$ represented as a neural network with parameter set $\theta$, and after bouncing the ray only once, it queries the cache to collect the contribution of the rest of the path. To train the radiance cache, InvNeRad is built on the self-supervision technique introduced by NeRad~\cite{hadadanNeuralRadiosity2021}, which minimizes the radiometric prior loss. The radiometric prior loss encourages the radiance cache \(L_\theta\) to satisfy the rendering equation by substituting actual radiance \(L\) with the radiance cache, and minimizing the difference between left hand side and right hand side of Eq.~\ref{eq:rendering_eq_operator}.

\begin{align}
\label{eq:radiometric_prior}
    \mathcal{L}_{\mathrm{prior}}(\theta) = \| L_{\theta}(\mathbf{x},\omega_o) - (E(\mathbf{x},\omega_o) + \mathcal{T_\phi}(L_{\theta})(\mathbf{x},\omega_o))\|.
\end{align}

\section{Non-Convexity of Co-Located Photometric Stereo} 
\label{sec:suppl_non_convex}

We found that co-located photometric stereo suffers from non convex local minimums. Intuitively, when the reconstructed scene gets larger, there is often significant difference in intensity of each scene point at different scene locations, and such difference in intensity tends to dominate loss, while the error caused by incorrect depth estimates will be relatively small. Therefore, the optimization process will focus on explaining photometric cues, or the effect of shading on surfaces, before moving toward reducing error due to depth estimates. Unlike traditional photometric stereo, photometric cues are fundamentally ambiguous in a co-located light and camera setup, as there is no information about correspondence between pixels in different images.  The optimization process will pick a random geometric explanation that is consistent with all photometric cues, and drift far away from correct geometry. By the time error from depth estimates becomes more significant, the optimization process is stuck in a bad local minimum and unable to escape. 
\section{Continuity of Indirect Component of Radiance} 

\label{sec:cont-indir-comp}
In this section, we will show the continuity of the indirect component: \(\sum_{k=2}^{\infty} \mathcal{T_\phi}^{k}(E) (\mathbf{x}, \omegaout, \mathbf{x_l})\). If \(\sum_{k=2}^{\infty} \mathcal{T_\phi}^{k}(E) (\mathbf{x}, \omegaout, \mathbf{x_l})\) is continuous, then the term can be approximated with a sufficiently large neural network by the universal approximation theorem.

Our strategy focuses on using mathematical induction to prove every single bounce \(\mathcal{T_\phi}^{k}(E) (\mathbf{x}, \omegaout, \mathbf{x_l})\) is continuous w.r.t. flashlight position for \(k \ge 2\).

\begin{figure}
  \centering

  \includegraphics[width=\linewidth]{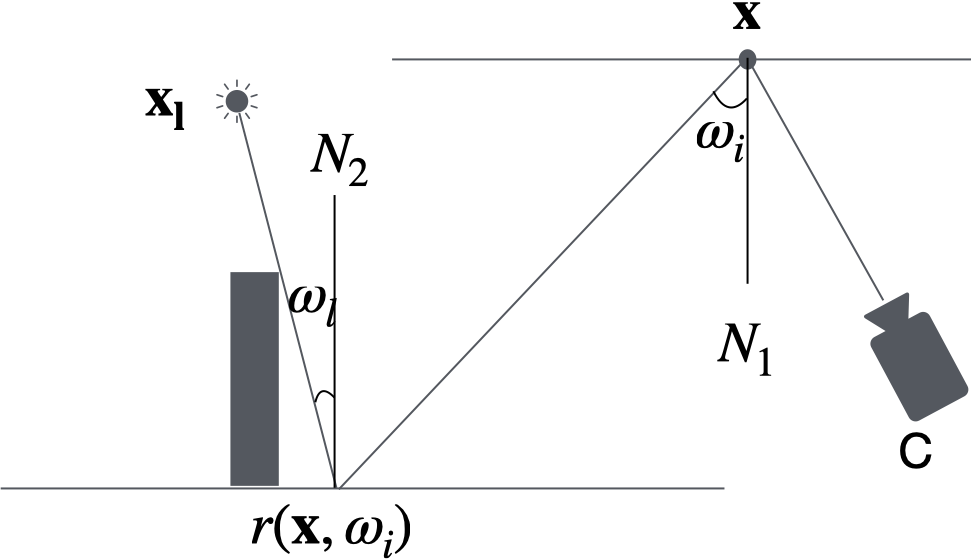}
  \caption{A simple scene for demostrating continuity and differentiability of the indirect component.}
  \label{fig:cont_scene}
\end{figure}

Before getting started, we make a couple of assumptions. The scene is illuminated by single isotropic point light source. We assume scene geometry \(\mathbb{S}\) consists of a finite number of objects, which are 2-manifold smoothly embedded in \(\mathbb{R}^3\). The physical dimension of the entire scene is finite. We also assume the flashlight energy is finite. We assume flashlight position \(\mathbf{x_l}\) is not in the closure of any object. We differentiate between cast shadows versus attached shadows, which are surfaces oriented away from the light. We model attached shadows as an effect of BRDF \(F(\mathbf{x}, \omegain, \omegaout) = 0\) when \(\omegain \cdot \omegaout < 0\) for non-transparent objects. \(\mathbbm{V}(\mathbf{x_l}\leftrightarrow\mathbf{x})\) is the binary visibility function of light \(\mathbf{x_l}\) from scene point \(\mathbf{x}\), which is 0 if \(\mathbf{x_l}\) is not visibile from \(\mathbf{x}\) and 1 otherwise. Recall we assume SV-BRDF \(F(\mathbf{x}, \omegain, \omegaout)\) is a continuous function. Therefore, illumination change caused by attached shadows is smooth relative to flashlight positions. When a surface is both occluded and oriented away from light, we model the occluded part as a cast shadow.

The base case is that the second bounce \(\mathcal{T_\phi}^{2}(E) (\mathbf{x}, \omegaout, \mathbf{x_l})\) is continuous and differentiable wrt. flashlight position. For illustrative purposes, we present a simple scene in Fig.~\ref{fig:cont_scene}.

We write down the second bounce only rendering equation at point \(\mathbf{x}\), where \(L^1\) is the radiance of the scene under only direct illumination, \(\mathbf{x_l}\) is the emitter location, and \(F(\mathbf{x},\omegain, \omegaout)\) is the BRDF function which we assume to be a smooth function. \(r(\mathbf{x}, \omegain)\) is the ray casting operator which finds the first intersection of a ray starting at point \(\mathbf{x}\) toward direction \(\omegain\). We denote the direction at \(r(\mathbf{x}, \omegain)\) to light location as \(\omega_l\). We denote the surface normal at \(\mathbf{x}\) as \(N_1\), and the surface normal 
at \(r(\mathbf{x}, \omegain)\) as \(N_2\)
\begin{multline}
\label{eq:radiance_cont_b2}
\mathcal{T_\phi}^{2}(E)(\mathbf{x},\omegaout, \mathbf{x_l}) = \\
\int_{\mathcal{H}^2} F(\mathbf{x}, \omegaout, \omegain)  (\omega_i \cdot N_1)  \\
\frac{ \mathbbm{V}{( r(\mathbf{x}, \omegain) \leftrightarrow \mathbf{x_l})} F(r(\mathbf{x}, \omegain), \omegain, \omega_l) (\omega_l \cdot N_2)}{\|\mathbf{x_l} - r(\mathbf{x}, \omegain)\|^2 } d\omegain^{\perp} 
\end{multline}

To make the notation more compact, we define \(f(\mathbf{x}, \omegaout, \omegain, \mathbf{x_l})\) as the following.
\begin{multline}
  f(\mathbf{x}, \omegaout, \omegain, \mathbf{x_l}) = F(\mathbf{x}, \omegaout, \omegain)  (\omegain \cdot N_1)  \\
\frac{  F(r(\mathbf{x}, \omegain), \omegain, \omega_l) (\omega_l \cdot N_2)}{\|\mathbf{x_l} - r(\mathbf{x}, \omegain)\|^2 } 
\end{multline}

\begin{multline}
\label{eq:radiance_cont_compact}
\mathcal{T_\phi}^{2}(E)(\mathbf{x},\omegaout, \mathbf{x_l}) = 
\int_{\mathcal{H}^2}\mathbbm{V}{( r(\mathbf{x}, \omegain) \leftrightarrow \mathbf{x_l})} \\
  f(\mathbf{x}, \omegaout, \omegain, \mathbf{x_l}) d\omegain^{\perp} 
\end{multline}

To show the continuity of \(\mathcal{T_\phi}^{2}(E)(\mathbf{x}, \omegaout, \mathbf{x_l})\) in light position, we want to show that for any sequence \(\mathbf{x_{ln}} \rightarrow \mathbf{x_l}\), we have the following.
\begin{multline}
    \int_{\mathcal{H}^2}{\mathbbm{V}{(r(\mathbf{x}, \omegain) \leftrightarrow\mathbf{x_{ln}}  )f(\mathbf{x}, \omegaout, \mathbf{x_{ln}})}} d\omegain^{\perp} \rightarrow \\
      \int_{\mathcal{H}^2}{\mathbbm{V}{( r(\mathbf{x}, \omegain) \leftrightarrow \mathbf{x_l} )f(\mathbf{x}, \omegaout, \mathbf{x_l})} d\omegain^{\perp} } 
\end{multline}

However, since \(f(\mathbf{x}, \omegaout, \mathbf{x_l})\) is continuous, we will be focusing on the behavior of \(\mathbbm{V}{(\mathbf{x_l}\leftrightarrow \mathbf{x}^\prime)}\) under the sequence. Finally we will examine the behavior of the integral.

Our overall goal is to show the set of points where the visibility function can change is only the cast shadow boundary as \(\mathbf{x_{ln}} \rightarrow \mathbf{x_l}\).

Here we introduce a couple of definitions that will be used throughout the proof. Consider a point \(\mathbf{x}^\prime = r(\mathbf{x}, \omegain)\) reachable from secondary bounce location \(\mathbf{x}\) in direction \(\omegain\). Consider the points along the ray segment \(\mathbf{x_l}-\mathbf{x}^\prime\):
\begin{equation}
    p(\mathbf{x_l}, \mathbf{x}^\prime, t) = \mathbf{x_l} + t (\mathbf{x}^\prime - \mathbf{x_l})
\end{equation}

\begin{lemma}
\label{lem:unoccluded-visibility-const}
    Assume \(\mathbf{x}^\prime\) is not on the shadow boundary of any object in the scene with respect to the light source \(\mathbf{x_l}\). Additionally suppose that \(\mathbf{x}^\prime\) is unoccluded. We wish to prove that there exists a neighborhood around \(\mathbf{x_l}\) such that if \(\mathbf{x_l}\) is perturbed to a position in that neighborhood, \(\mathbf{x}^\prime\) is still unoccluded from its perspective.
\end{lemma}

\begin{figure}
    \centering
    \includegraphics[width=0.7\linewidth]{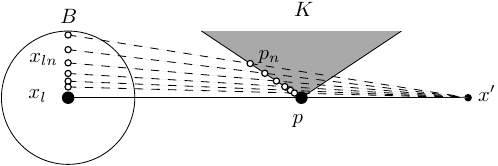}
    \caption{In unoccluded case, the light visibility does not change except the boundary.}
    \label{fig:proof-unoccluded-case}
\end{figure}

\begin{proof}

Suppose to the contrary. Then there exists a sequence of positions \(\mathbf{x_{ln}}\) which converge to \(\mathbf{x_l}\) such that the ray from each \(\mathbf{x_{ln}}\) to \(\mathbf{x}^\prime\) always passes through some object in the scene. Since \(\mathbf{x_l}\) is not in the closure of any object, there exists a ball \(B\) around \(\mathbf{x_l}\) of positive radius that does not intersect the geometry of the scene. Hence an infinite amount of these \(\mathbf{x_{ln}}\) lie in that ball \(B\). Consider only the \(\mathbf{x_{ln}}\) in this ball. Since the scene only has a finite amount of objects, an infinite amount of the occluders defined for the \(\mathbf{x_{ln}}\) must be the same object. That is to say, there exists an object \(K\) such that for an infinite amount of the \(\mathbf{x_{ln}}\) the ray from \(\mathbf{x_{ln}}\) to \(\mathbf{x}^\prime\) always hits this object. Note this object must be strictly between \(\mathbf{x_l}\) and \(\mathbf{x}^\prime\) since we are only considering the \(\mathbf{x_{ln}}\) in \(B\). Now consider only the \(\mathbf{x_{ln}}\) whose ray to \(\mathbf{x}^\prime\) hits \(K\). Let the sequence of points they first hit on \(K\) be \(p_n\). 

Since \(p_n\) is on a compact manifold smoothly embeded in \(R^3\),

Since the \(p_n\) are on a compact space and get infinitely close to the segment from \(\mathbf{x_l}\) to \(\mathbf{x}^\prime\), by the compactness of \(K\) there must be some point \(p\) on \(K\) along this line. Hence \(p\) intersects the line segment from \(\mathbf{x_l}\) to \(\mathbf{x}^\prime\), and so \(\mathbf{x}^\prime\) must be on the shadow boundary. This contradicts what we assumed about \(\mathbf{x}^\prime\). See Figure \ref{fig:proof-unoccluded-case} for an illustration. \end{proof}

\begin{lemma}
\label{lem:occluded-visibility-const}
Assume \(\mathbf{x}^\prime\) is not on the shadow boundary of any object in the scene with respect to the light source \(\mathbf{x_l}\). Additionally suppose that \(\mathbf{x}^\prime\) is occluded. We wish to prove that there exists a neighborhood around \(\mathbf{x_l}\) such that if \(\mathbf{x_l}\) is perturbed to a position in that neighborhood, \(\mathbf{x}^\prime\) is still occluded from its perspective.
\end{lemma}

\begin{figure}
    \centering
    \includegraphics[width=0.7\linewidth]{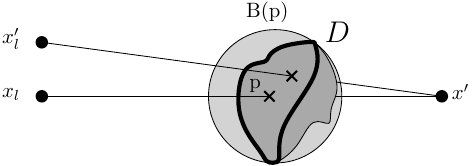}
    \caption{In occluded case, the light visibility does not change except the boundary.}
    \label{fig:proof-occluded-case}
\end{figure}

\begin{proof}

Since \(\mathbf{x}^\prime\) is occluded but not on the shadow boundary of any object in the scene with respect to the light source \(\mathbf{x_l}\), then the line segment from \(\mathbf{x_l}\) to \(\mathbf{x}^\prime\) intersects the relative interior of some surface \(K\) at a point, \(p\), strictly between \(\mathbf{x}^\prime\) and \(\mathbf{x_l}\).

Since \(p\) is in the relative interior of \(K\) then there exists a closed ball around \(p\), \(B(p)\), whose radius is smaller than \(\frac{1}{2}\min(d(\mathbf{x_l},p),d(p,\mathbf{x}^\prime))\), such that \(D=B(p)\cap K\) is in the relative interior of \(K\). Since the topological circle \(\partial D\) (the relative boundary of \(D\)) is compact, let \(\theta\) be the minimum angle \(\angle \mathbf{x_l}\mathbf{x}^\prime y\) makes for any \(y\) on \(\partial D\). Note that \(\theta>0\) because \(p\) is the unique point for which it is equal zero and \(p\) is not on \(\partial D\).

Choose \(B(\mathbf{x_l})\) to be a ball around \(\mathbf{x_l}\) such that it has small enough radius so as not to intersect \(D\) and such that for any \(z\) in \(B(\mathbf{x_l})\), \(\angle \mathbf{x_l}\mathbf{x}^\prime x\) makes a smaller angle than \(\theta\). Since for all \(\mathbf{x_l}^\prime \in B(\mathbf{x_l})\), \(\angle \mathbf{x_l}\mathbf{x}^\prime\mathbf{x_l}^\prime < \theta\), then the segment from \(\mathbf{x_l}^\prime\) to \(\mathbf{x}^\prime\) passes through the hole of the topological circle \(\partial D\) and hence intersects \(D\) before it reaches \(\mathbf{x}^\prime\), leaving \(\mathbf{x}^\prime\) occluded. And so if \(\mathbf{x_l}\) is perturbed in that ball \(B(\mathbf{x_l})\), \(\mathbf{x}^\prime\) is still occluded from it by \(D\). See Figure \ref{fig:proof-occluded-case} for an illustration.\end{proof}

With Lemma~\ref{lem:unoccluded-visibility-const} and Lemma~\ref{lem:occluded-visibility-const}, we can show the second bounce radiance is indeed continuous. Denote \(S\) as the shadow boundary. We conclude for any sequence \(\mathbf{x_{ln}} \rightarrow \mathbf{x_l}\), we have the following.
\begin{equation}
    \forall \mathbf{x^\prime} \notin S \, \mathbbm{V}{(\mathbf{x_{ln}}\leftrightarrow \mathbf{x}^\prime)}\rightarrow 
     \mathbbm{V}{(\mathbf{x_l}\leftrightarrow \mathbf{x}^\prime) }
\end{equation}

First we note $f(\mathbf{x}, \omegaout, \omegain, \mathbf{x_{ln}})$ is continuous. Moreover, since \(\mathbbm{V}{( r(\mathbf{x}, \omegain) \leftrightarrow \mathbf{x_l})}
  f(\mathbf{x}, \omegaout, \omegain, \mathbf{x_l})\) is bounded (all its components are bounded) by a constant, say by \(M\), we can define \(g(\mathbf{x},\omegaout, \omegain) = M\), and \(|f(\mathbf{x}, \omegaout, \omegain, \mathbf{x_l})| <g(\mathbf{x},\omegaout, \omegain)  \). By the dominanted convergence theorem, as \(\mathbf{x_{ln}} \rightarrow \mathbf{x_l}\), we have the following.
\begin{align*}
    & \mathcal{T_\phi}^{2}(E)(\mathbf{x},\omegaout, \mathbf{x_{ln}}) \\
    & = \int_{\mathcal{H}^2}{\mathbbm{V}{(\mathbf{x_{ln}}\leftrightarrow \mathbf{x}^\prime)f(\mathbf{x}, \omegaout, \omegain, \mathbf{x_{ln}})}} d\omegain^{\perp}  \\
    & = \int_{\mathcal{H}^2 \setminus S}{\mathbbm{V}{(\mathbf{x_{ln}}\leftrightarrow \mathbf{x}^\prime)f(\mathbf{x}, \omegaout, \omegain, \mathbf{x_{ln}})}} d\omegain^{\perp}  \\
    & \rightarrow \int_{\mathcal{H}^2 \setminus S}{\mathbbm{V}{(\mathbf{x_l}\leftrightarrow \mathbf{x}^\prime)f(\mathbf{x}, \omegaout, \omegain, \mathbf{x_l})} d\omegain^{\perp} } \\
    & = \int_{\mathcal{H}^2}{\mathbbm{V}{(\mathbf{x_l}\leftrightarrow \mathbf{x}^\prime)f(\mathbf{x}, \omegaout, \omegain, \mathbf{x_l})} d\omegain^{\perp} } \\
    & = \mathcal{T_\phi}^{2}(E)(\mathbf{x},\omegaout, \mathbf{x_l}) 
\end{align*}
We conclude \(\mathcal{T_\phi}^{2}(E)(\mathbf{x},\omegaout, \mathbf{x_l})\) is continuous with respect to flashlight position $\mathbf{x_l}$, which is our base case for our proof by mathematical induction.

For the inductive case on the \(N\)-th bounce, we can assume the \((N-1)\)-th bounce is continuous. The radiance is the following.

\begin{multline}
\label{eq:radiance_cont_bn}
\mathcal{T}^{N}_{\phi}(E)(\mathbf{x},\omegaout, \mathbf{x_l})  = \\
\int_{\mathcal{H}^2} F(\mathbf{x},\omegain, \omegaout) \mathcal{T}^{N-1}_{\phi}(E)(\mathbf{x},\omegaout, \mathbf{x_l})  d\omegain^{\perp}
\end{multline}

It is easy to notice \(\mathcal{T}^{N}_{\phi}(E)(\mathbf{x},\omegaout, \mathbf{x_l})\) is  the integration of a continuous brdf \(F\) and continuous function \(\mathcal{T}^{N-1}_{\phi}(E)(\mathbf{x},\omegaout, \mathbf{x_l})\). Therefore, we conclude \(\mathcal{T}^{N}_{\phi}(E)(\mathbf{x},\omegaout, \mathbf{x_l})\) is continuous.

\section{Additional Comparisons}
\label{sec:suppl_add_comp}
In this section, we provide additional comparisons against state of the art methods.

\subsection{Measured Albedo in the Wild 2.0}
\label{sec:maw}
\begin{table}[ht]
\centering
\resizebox{\linewidth}{!}{%
\begin{tabular}{lcccccc}
    \toprule
    \multirow{2.5}{*}{Method} &
    \multicolumn{3}{c}{Albedo Intensity (MSE \(\times10^2\))} &
    \multicolumn{3}{c}{Albedo Chromaticity (Delta-E)} \\
    \cmidrule(lr){2-4} \cmidrule(lr){5-7}
    & Coffee table & Shoe rack & Window still &
      Coffee table & Shoe rack & Window still \\
    \midrule
    IRGS        & \underline{0.09} & \underline{0.18} & 1.56 & 9.1714 & 3.2322 & 9.0345 \\
    NeRO        & 4.86 & 0.31 & 1.47 & 4.7499 & 7.5724 & 4.0790 \\
    IRON        & 0.14 & 3.75 & \underline{0.54} & 4.9089 & 5.1342 & 4.0915 \\
    WildLight  & 0.85 & 0.26 & 1.01 & \textbf{3.7362} & \underline{1.3104} & \underline{3.8298} \\
    Ours       & \textbf{0.07} & \textbf{0.07} & \textbf{0.10} & 6.1491 & \textbf{1.0236} & \textbf{2.6620} \\
    \bottomrule
\end{tabular}%
}
\caption{Quantitative evaluation of albedo on real data following measured albedo in the wild~\cite{wuMeasuredAlbedoWild2023}. We report albedo intensity, which measures intensity difference with ground-truth albedo in the grayscale component, and albedo chromaticity, which measure chromaticity difference with ground-truth albedo.}
\label{tab:suppl_maw}
\end{table}
To quantitatively compare the quality of albedo on real data, we follow measured albedo in the wild (MAW)~\cite{wuMeasuredAlbedoWild2023} to collect ground-truth measured albedo labels on our real datasets \textit{Coffee Table}, \textit{Shock Rack} and \textit{Window Sill}.

 MAW contains few images per scene and is better suited to evaluate single image inverse rendering techniques. On the other hand, our dataset contains densely captured images for each scene and is suitable for evaluating multi-view inverse rendering methods. The measurements and annotations will be released together with our dataset upon acceptance. 

Following MAW, we evaluate algorithms with albedo intensity in MSE and albedo chromaticity in Delta-E, and report the performance in Tab.~\ref{tab:suppl_maw}. We found that similar to the results on synthetic data, our method significantly outperforms natural illumination baselines in albedo estimation, highlighting the importance of co-located light \& camera setup in contraining the ambiguous inverse rendering problem.

\subsection{Surface Angle Weighting loss}
\label{sec:saw-supp}
\input{figures/loss_abl_2}

In Fig.~\ref{fig:loss_abl}, we compare our surface angle weighting loss to  state-of-the-art loss weighting schemes, which includes Ref-NeuS Loss~\cite{geRefNeuSAmbiguityReducedNeural2023} and Adaptive Huber loss from neural-pbir~\cite{sunNeuralPBIRReconstructionShape2023}.

Ref-NeuS~\cite{geRefNeuSAmbiguityReducedNeural2023} is based on variation between corresponding pixels across images, and is unsuitable to use on co-located light and camera capture setup, as pixel intensity changes due to different light position. Therefore, we compare Ref-NeuS on natural illumination against our geometry initialization stage (stage 1) on co-located light and camera setup. We found Ref-NeuS tend to reconstruct concave regions poorly, potentially due to down-weighting of errors in such regions.

We also compare our full stage 1 with a variant without such surface angle weighting loss, we found the variant without such surface angle weighting loss produce artifacts in regions with specular inter-reflections, while our surface angle weighting loss prevents such artifacts. We also included Adaptive Huber from Neural PBIR~\cite{sunNeuralPBIRReconstructionShape2023} for comparison, which we found is unable to prevent these artifacts.

\subsection{Natural Illumination Methods Material Comparisons on Real Rata}

\label{sec:suppl_qual_natural}
\input{figures/qualitative_figure_real_natural}
In Fig.~\ref{fig:qualitative_figure_real_natural}, we show comparisons against state of art natural illumination methods NeRO~\cite{liuNeRONeuralGeometry2023} and IRGS~\cite{guIRGSInterReflectiveGaussian2025} on real data.  We found that our method significantly outperforms the natural illumination baselines, especially in material properties.

\subsection{Natural Illumination Methods Geometry Comparisons}
\label{sec:suppl_qual_geometry}

\input{figures/merged_render_geometry_natural}
We compare the geometry of our method with natural illumination methods in Fig.~\ref{fig:qualitative_figure_real_natural}. We do not claim we achieve better than state-of-the-art natural illumination geometry, but only comparable.

\subsection{Qualitative Comparisons on re-rendering}
\label{sec:suppl_qual_rerender}
\input{figures/qualitative_figure_rerender_new}

In Fig.~\ref{fig:qualitative_figure_rerender}, we visualize the re-rendering performance of colocated and natural illumination methods. We found all methods perform well in re-rendering as along as geometry is reasonably reconstructed. Nevertheless, natural illumination methods often are unable to recover accurate material properties due to inherent ambiguity in inverse rendering. 
\subsection{WildLight Capture Setup}
\label{sec:wildlight_cap}
\input{figures/suppl_wildlight}
In the main paper, we show WildLight results as a darkroom co-located light and camera method. In Figure~\ref{fig:suppl_wildlight}, we show qualitative results of WildLight on WildLight style capture setup (co-located light and camera under ambient natural illumination). We found WildLight \cite{chengWildLightInthewildInverse2023} still fail to converge under prominent inter-reflections.
\subsection{Additional Views in Dataset}
\label{sec:add_views}
We show additional views from our dataset on re-rendering, albedo and roughness in Fig.~\ref{fig:suppl_qualitative_figure_synthetic_1}, Fig.~\ref{fig:suppl_qualitative_figure_synthetic_2}, Fig.~\ref{fig:suppl_qualitative_figure_synthetic_3}, Fig.~\ref{fig:suppl_qualitative_figure_real_1}, Fig.~\ref{fig:suppl_qualitative_figure_real_2}, Fig.~\ref{fig:suppl_qualitative_figure_real_3}, Fig.~\ref{fig:suppl_qualitative_figure_real_4}.

\section{Implementation Details} %
 Here we described additional details of implementation. 

\label{sec:details}

\subsection{Stage 1 Geometry Initialization}
With our synthetic dataset, the images can contain ``background pixels'' where the primary ray from the camera does not intersect with any scene geometry during rendering. Since we are  using environment map for our scenes, the values of these pixels are undefined, and we use a per image binary mask to ignore these pixels during training. Consequently, we do not put any supervision in the background region. To prevent the network from producing arbitrary values for the background, we adopt mask loss commonly used by prior works~\cite{wangNeuSLearningNeural2021,yarivMultiviewNeuralSurface2020}. NeuS~\cite{wangNeuSLearningNeural2021} defines unbiased weights \(w_{k,i}\) along the k-th camera ray based on the underlying signed distance field. Denote \(\hat{O}_k=\sum_{i=1}^nw_i\) as the the sum of weights along the k-th camera ray, \(M_k = \{0, 1\}\) as the value of the binary mask on the k-th pixel, and \(\operatorname{BCE}\) as the binary cross entropy loss, we have the following equation.

\begin{equation}
    L_\text{mask}=\operatorname{BCE}(M_k, \hat{O}_k)
\end{equation}

Such mask loss is only used for synthetic data, and not used for real data.

Stage 1 is trained with ADAM optimizer, batch size of 512 rays, learning rate of \(5\times10^{-4}\). %

\subsection{Stage 2 Physically Based Rendering}
To integrate the outer integral, we need to sample \(N\) 3d points \(p(t)\) along camera ray, which is importance sampled similar to NeuS\cite{wangNeuSLearningNeural2021}. For every sampled points, we evaluate the light transport operator \(\mathcal{T_\phi}\) by sampling \(M\) incoming light directions \(w_i\) through bsdf sampling~\cite{pharrPhysicallyBasedRendering2023}. Finally, the radiance cache \(L_\theta\) is queried at intersection points \(L_\theta(r(p(t), \omegain))\). In our experiments, we set \(N\) to be 128 as NeuS, and \(M\) to be 1.

As noted by Dejan Azinovic~\cite{azinovicInversePathTracing2019a}, naive automatic differentiation gives incorrect result, and it is important to draw independent samples for Monte Carlo estimator of gradient of reconstruction loss against output \(\frac{\partial \mathcal{L}_\text{recons}}{\partial \hat{y}}\) and gradient of output against parameters \(\frac{\partial \hat{y}}{\partial \phi}\), which we follow.

When we use principled BRDF~\cite{burleyPhysicallyBasedShading}. We set all fixed parameters to zeros, except for ``specular'' (which sometimes called ``specular albedo''), which we set to 0.6 for synthetic scenes and 0.5 for real scenes. 

Stage 2 is trained with ADAM optimizer at batch size of 512 rays, learning rate of \(5\times10^{-5}\). We downscale the gradient against SDF network \(S_{\theta_S}\) by \(10^{-1}\) as we found it improves stability. 

Since at the beginning of stage 2, only the geometry is initialized, but not the material properties and radiance cache. To initialize the material properties and radiance cache, we freeze the geometry at the beginning of stage 2 until the material properties and radiance cache become initialized, then we train all parameters jointly.

\noindent \textbf{Sampling Radiometric Prior Loss} One important design decision of the radiometric prior loss in the inverse rendering system is how to sample points for evaluating the radiometric prior loss. We evaluate on all points \(p(t)\) sampled along the camera ray during volume rendering. 

We also include an extra bounce prior similar to InvNeRad. To adapt such loss for volume geometry while keeping computational cost manageable, we only take 1 point out of points sampled along primary camera ray during physically based volume rendering, and sample \(N=128\) points along its income light direction, we we apply extra bounce prior loss on these sampled points.

To ensure thorough coverage of the radiance cache at all location under every light location, when calculating the extra bounce prior radiosity loss, we evaluate both the radiance cache and direct/indirect illumination term under a randomly sampled light location from training data.

\subsection{Stage 3 Material Optimization}
Here we provide additional details regarding the final stage of our system. We optimize for BRDF on fixed mesh geometry extracted from geometry network from stage 2. The algorithm is similar to invNeRAD~\cite{hadadanInverseGlobalIllumination2023}, where we intersect rays from camera with mesh geometry of the scene, and render under radiance cache global illumination. However, we use our dynamic radiance cache instead of naive radiance cache, and the dynamic radiance cache is trained same as stage 2, except instead of evaluating the loss on volume points, we can only evaluate on surface points.

We train our final with ADAM optimizer at learning rate of \(5\times10^{-4}\)  with batch size of 512. 
\section{Real Data Capture Setup}
\label{sec:suppl_data_capture}

We capture all of our real data using an iPhone XS Max and an iPhone 11 Pro. We capture all the image with ProCamera app on iOS as raw dng file. During capture, we keep manual and fixed white balance, focus and exposure. For co-located capture, we also keep flashlight constantly on through the capture session. We process the raw files with RawPy~\cite{riechertLetmaikRawpy2025}, which is a python interface around libraw~\cite{llcLibRawLibRaw2025}. We perform structure-from-motion reconstruction using pixel perfect sfm~\cite{lindenbergerPixelPerfectStructureFromMotionFeaturemetric}. We apply camera undistortion parameters estimated by pixel perfect sfm to our captured images. We found that both iPhone XS Max and iPhone 11 Pro experience significant vignetting. To calibrate for vignetting, we use a piece of white paper on a sunny day under direct sunlight as the calibration target. We model the vignetting as 6-th degree even order polynomial~\cite{goldmanVignetteExposureCalibration2005}, and apply vignetting correction accordingly. We store all final processed images as 16-bit unsigned png images with linear response curve without any gamma curve applied, which are used for all following experiments.

\input{figures/suppl_qualitative_figure}

\clearpage

%% file: figures/loss_abl_2.tex
\providelength\width
\setlength\width{3.0cm}
\begin{figure*}
\tiny
\centering
\renewcommand{\tabcolsep}{1pt}
\begin{tabular}{cc|ccc}
Ref-Neus & Stage 1 Full & Stage 1 w/o Surface Angle Weighting & Adaptive Huber & Stage 1 Full \\
\includegraphics[width=\width]{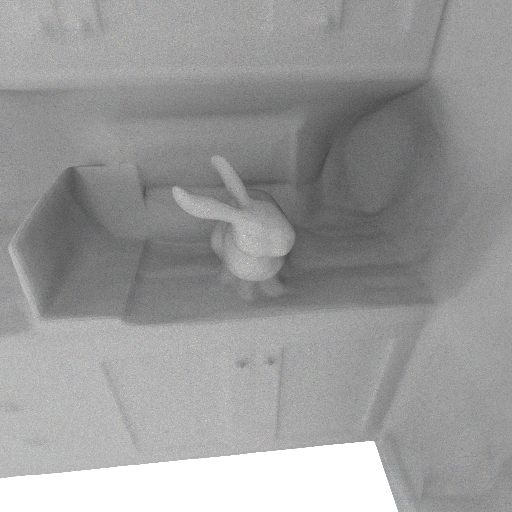} &
\includegraphics[width=\width]{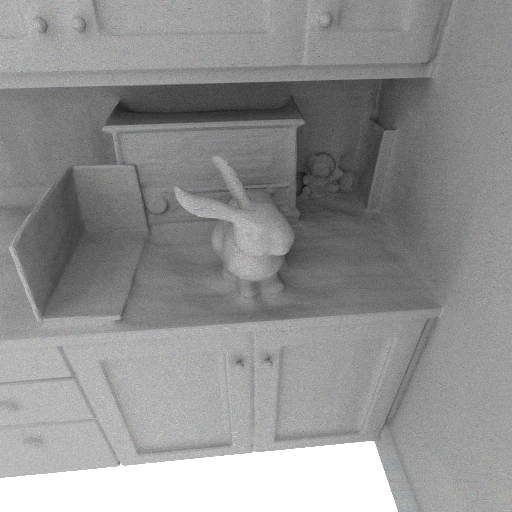} &

\includegraphics[width=\width]{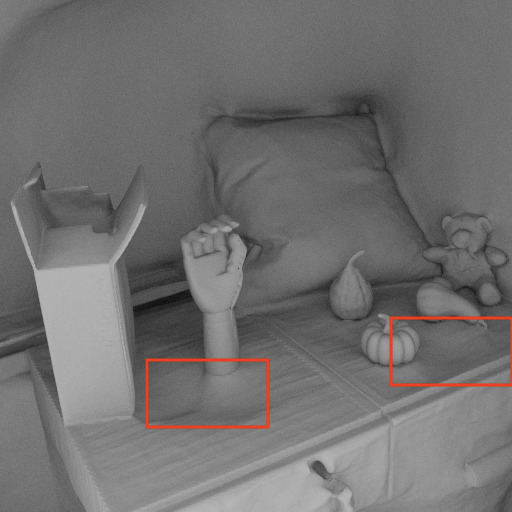} &
\includegraphics[width=\width]{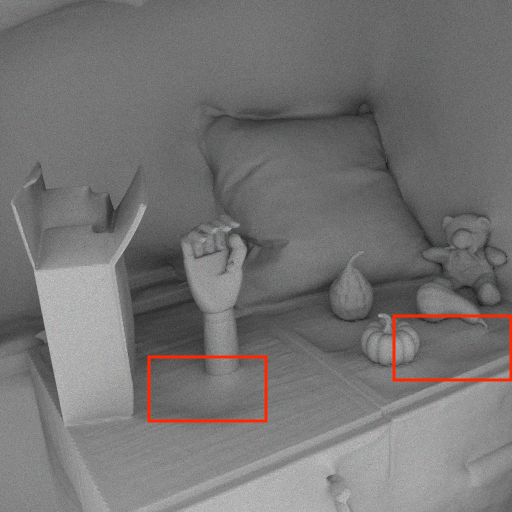} &
\includegraphics[width=\width]{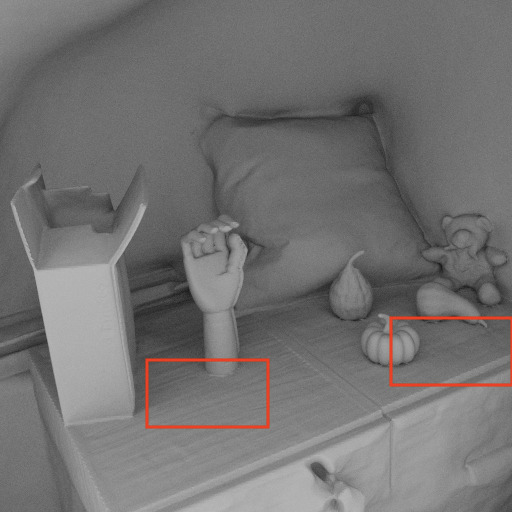}
\\

\end{tabular}
\captionof{figure}{
\textbf{Ablation of surface angle weighting.} Left: we compare Ref-NeuS with our geometry initialization stage. Ref-NeuS loss is unsuitable for co-located light \& camera setup, so Ref-NeuS is trained on natural illumination. Ref-NeuS tends to produce artifacts in concave regions, potentially due to downweighting in these regions. Right: Red box highlights areas where specular inter-reflection causes artifacts in geometry without surface angle weighting. Such errors become significantly less pronounced in our full geometry initialization stage with surface angle weighting loss. Adaptive Huber from neural-pbir~\cite{sunNeuralPBIRReconstructionShape2023} does not prevent such artifacts.}
\label{fig:loss_abl}
\end{figure*}

%% file: figures/qualitative_figure_real_natural.tex
\providelength\width
\setlength\width{2.5cm}

\begin{figure*}
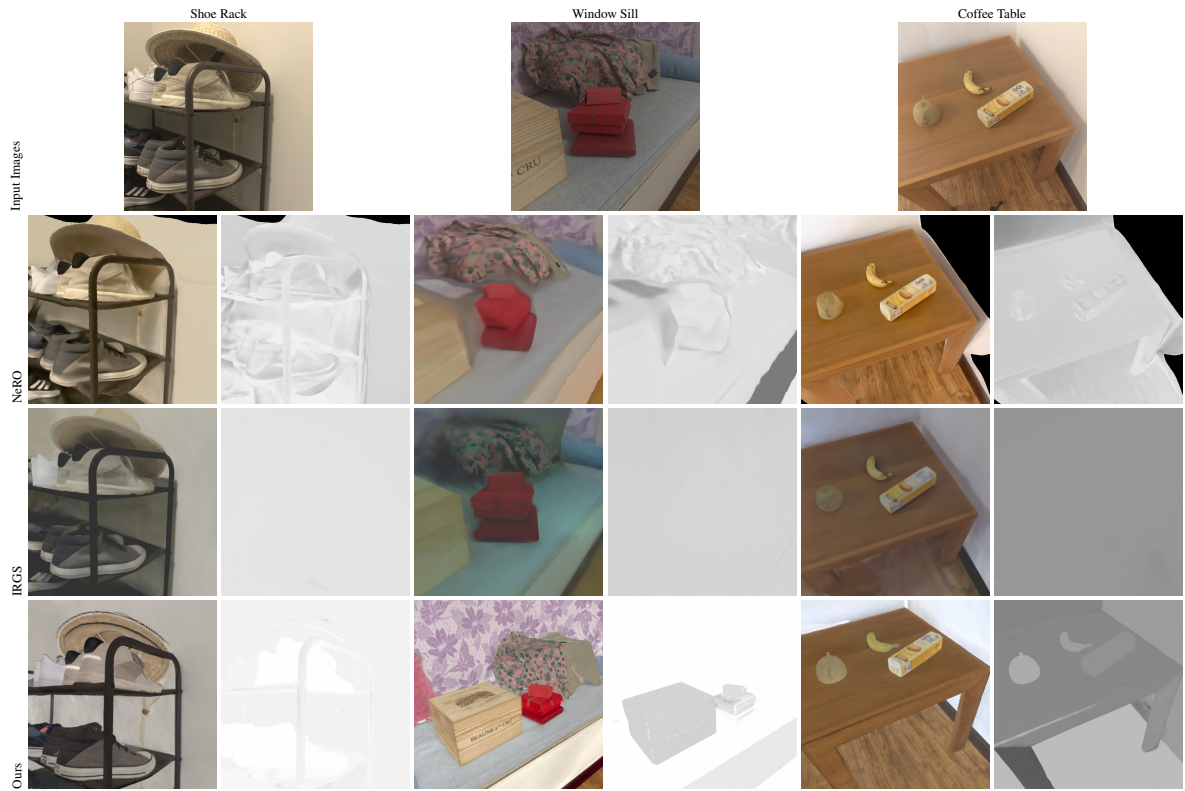

\tiny
\centering

\renewcommand{\tabcolsep}{1pt}

\input generated/qualitative_generated_params_real_natural

\caption{
\textbf{Qualitative comparison of reflectance estimation on real scenes.} We present estimated albedo, and roughness in validation views for the real scenes. Our method produces significantly better albedo, roughness w.r.t. natural illumination methods as co-located capture setup provides
additional constraints. (Zoom in for better visualization.) }
\label{fig:qualitative_figure_real_natural}

\end{figure*}

%% file: generated/qualitative_generated_params_real_natural.tex
\begin{tabular}{ccccccc}
& \multicolumn{2}{c}{Shoe Rack}
& \multicolumn{2}{c}{Window Sill}
& \multicolumn{2}{c}{Coffee Table}
\\
{\makebox{\rotatebox{90}{Input Images}}}
& \multicolumn{2}{c}{\includegraphics[width=\width]{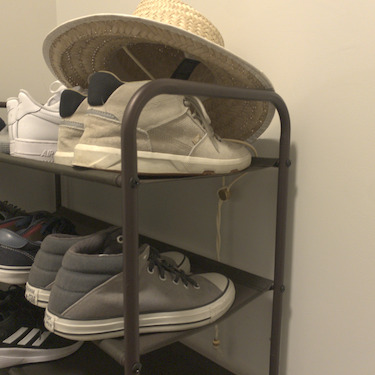}}
& \multicolumn{2}{c}{\includegraphics[width=\width]{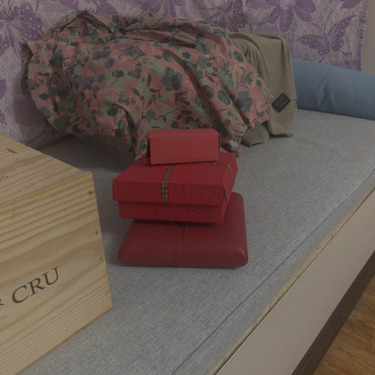}}
& \multicolumn{2}{c}{\includegraphics[width=\width]{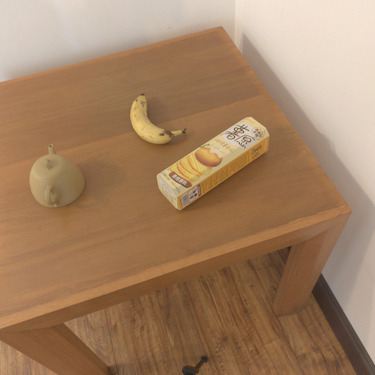}}
\\
{\makebox{\rotatebox{90}{NeRO}}}
& \includegraphics[width=\width]{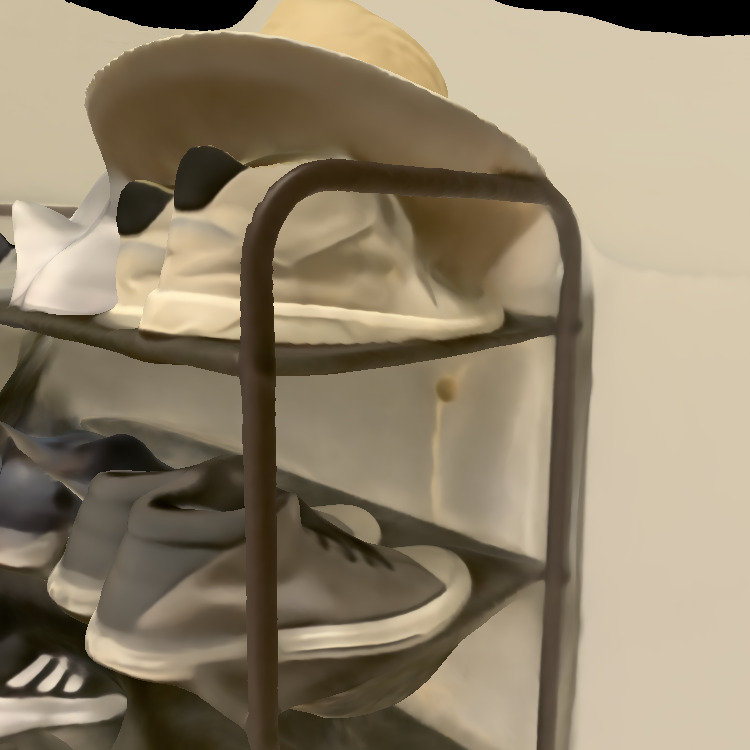}
& \includegraphics[width=\width]{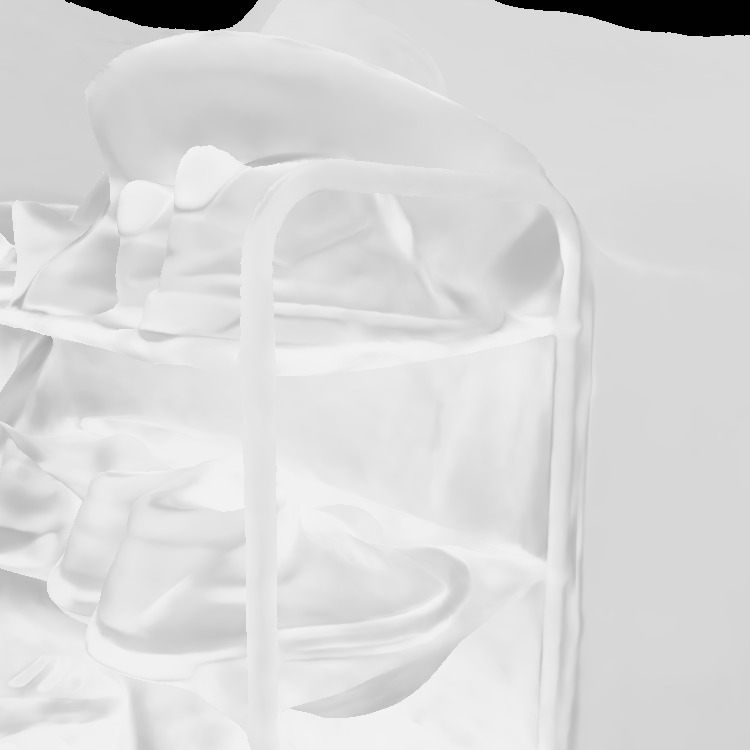}
& \includegraphics[width=\width]{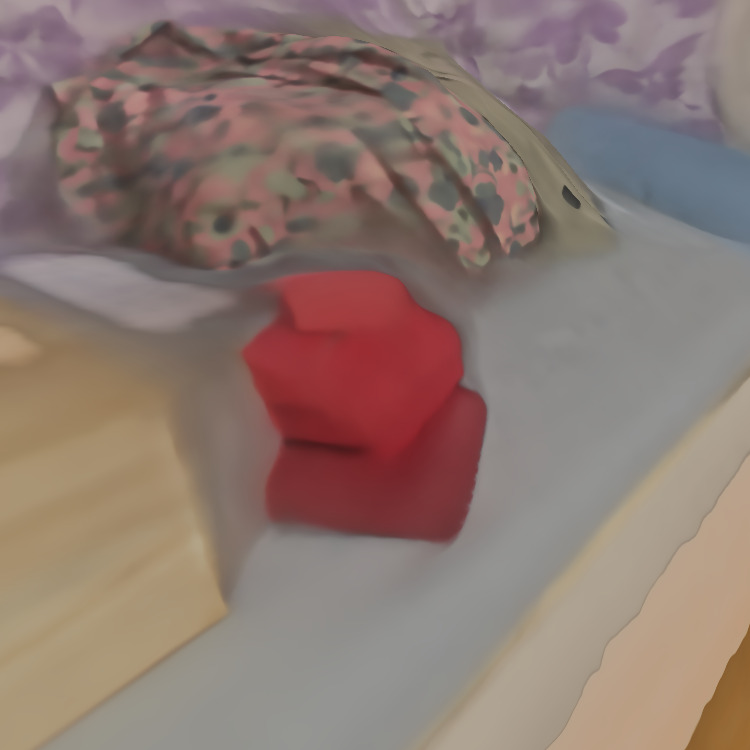}
& \includegraphics[width=\width]{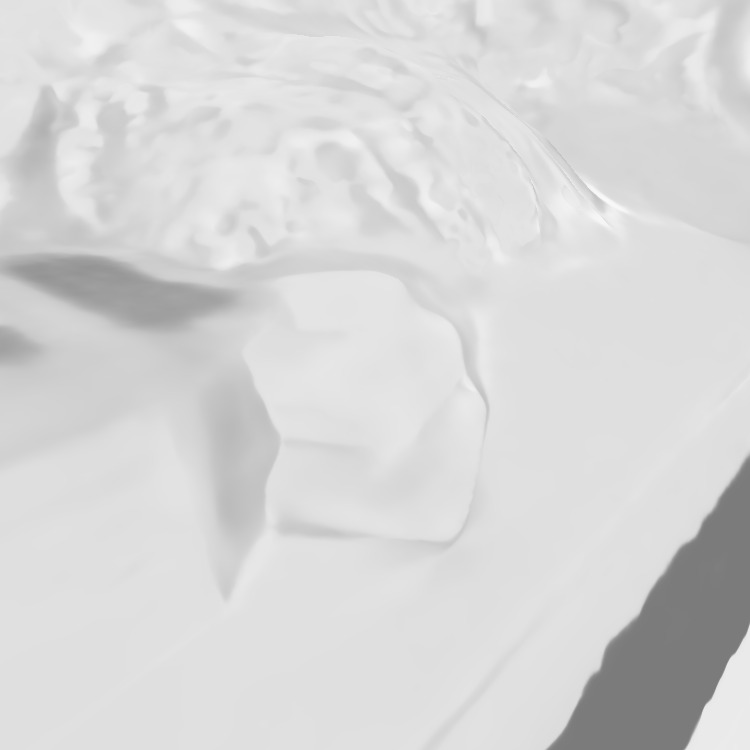}
& \includegraphics[width=\width]{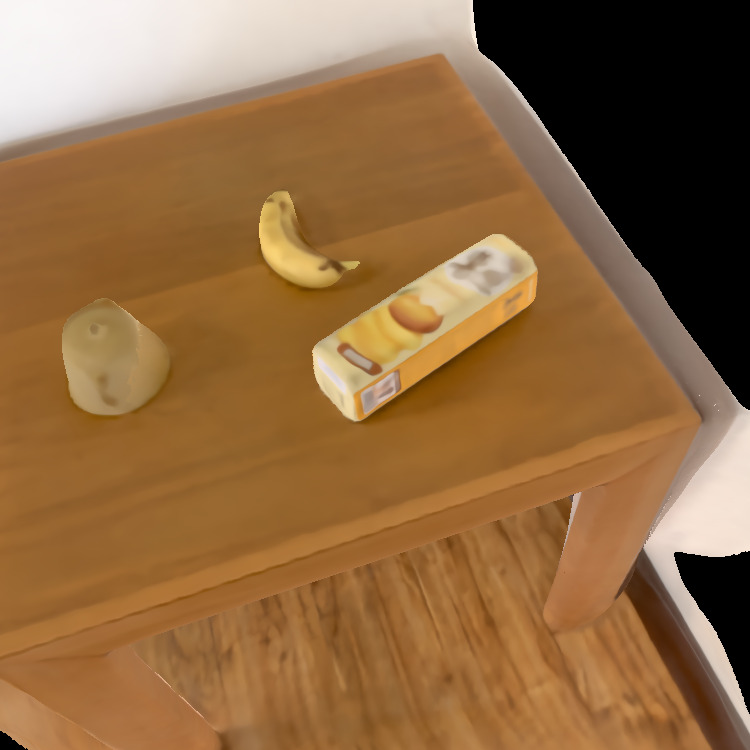}
& \includegraphics[width=\width]{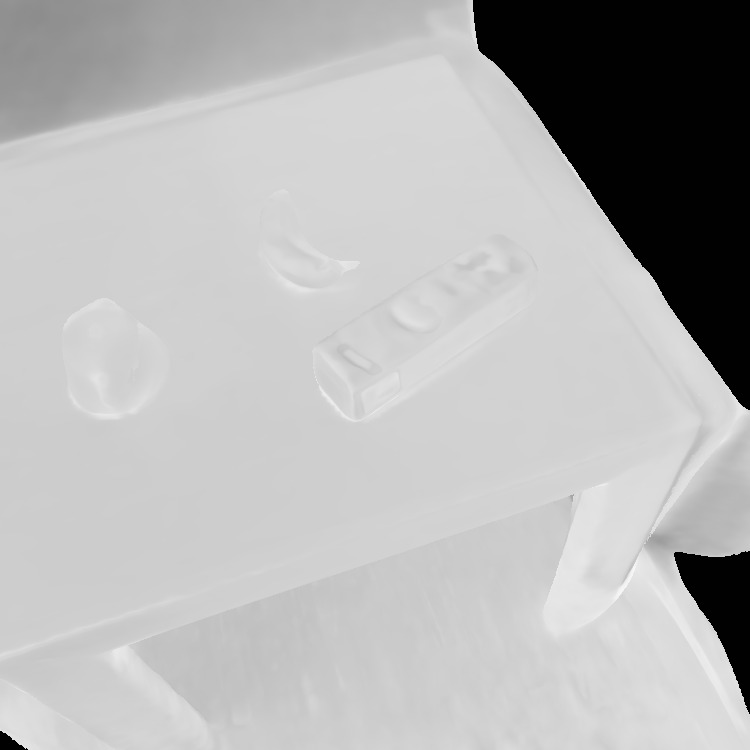}
\\

{\makebox{\rotatebox{90}{IRGS}}}
& \includegraphics[width=\width]{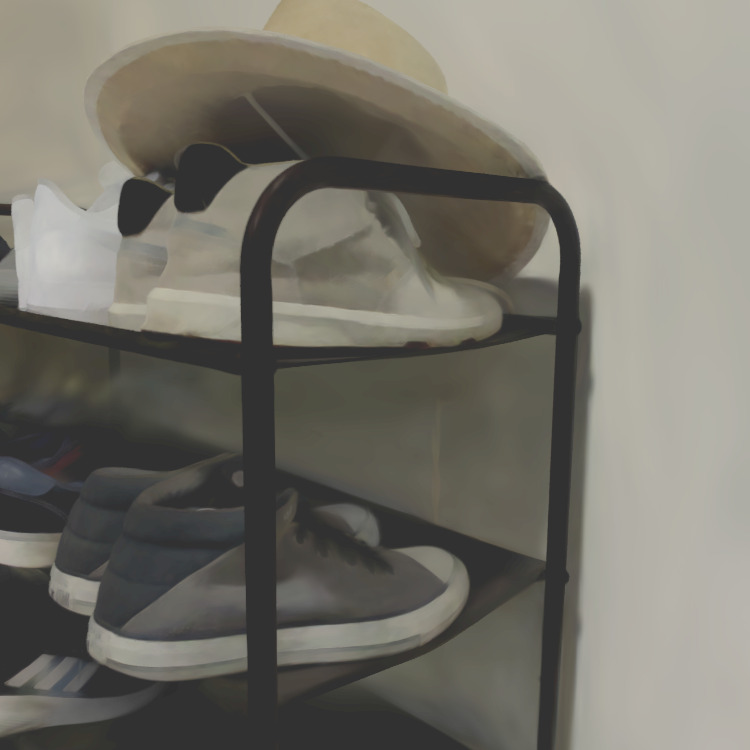}
& \includegraphics[width=\width]{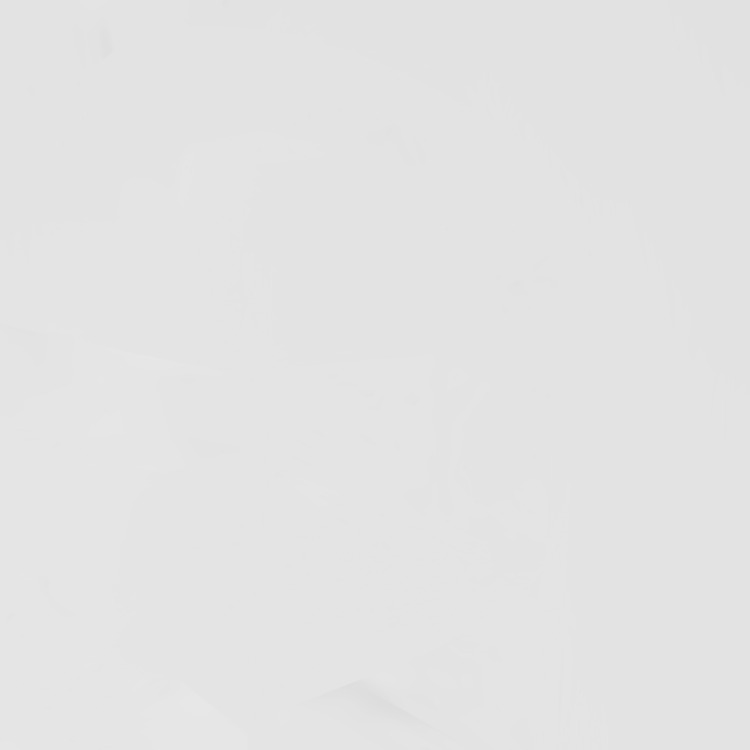}
& \includegraphics[width=\width]{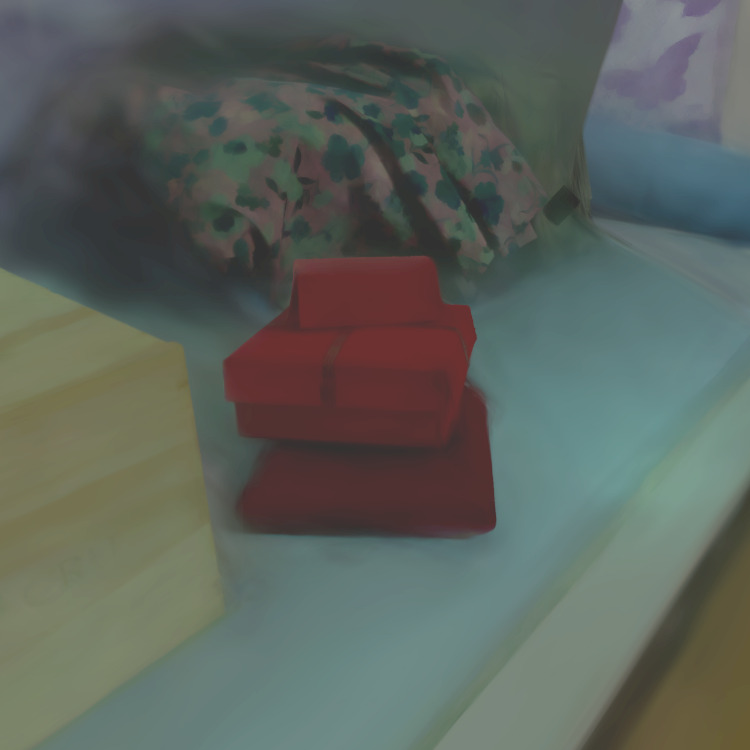}
& \includegraphics[width=\width]{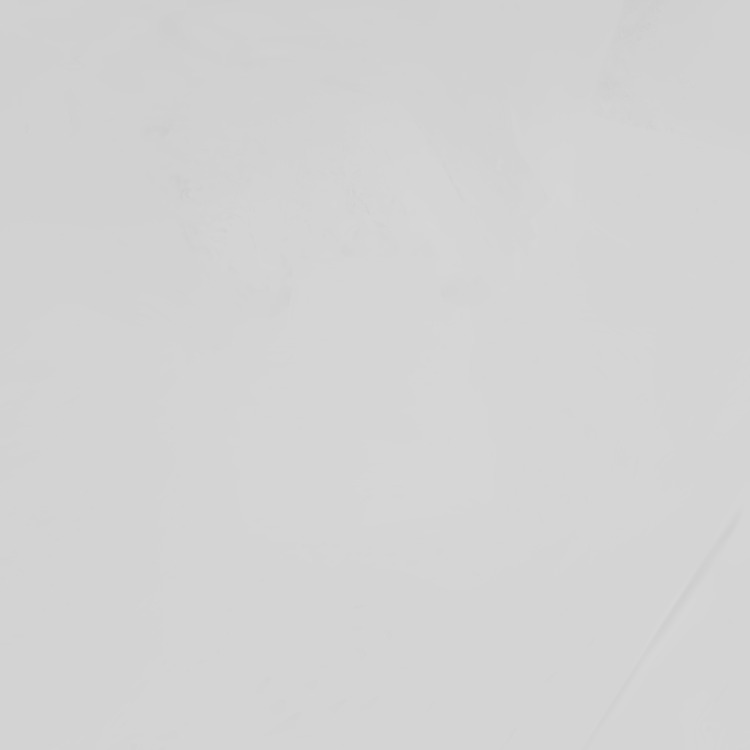}
& \includegraphics[width=\width]{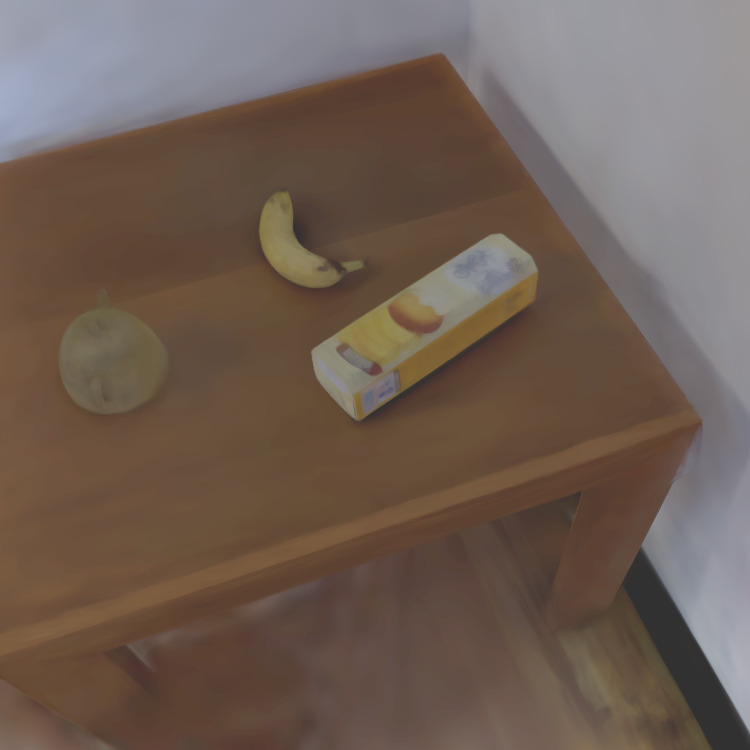}
& \includegraphics[width=\width]{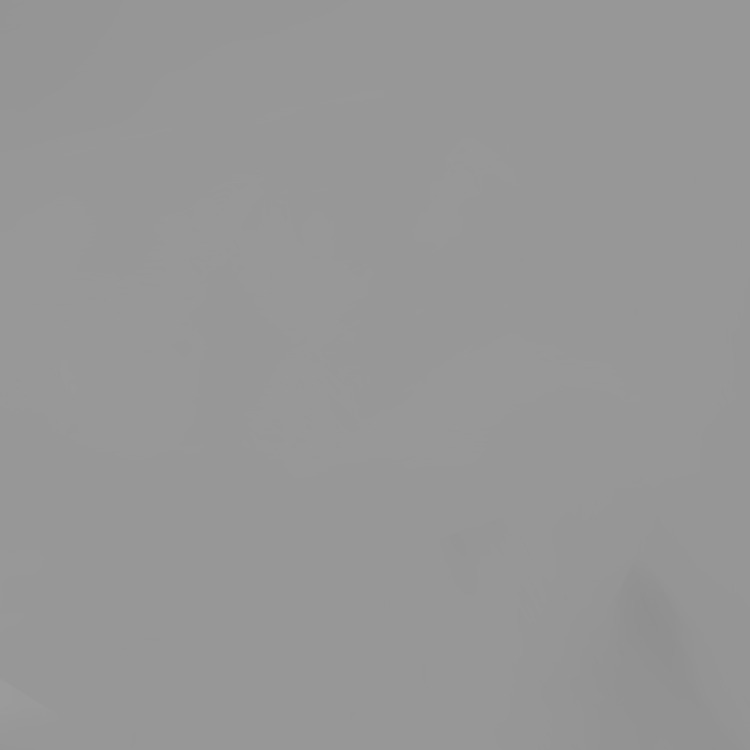}
\\

{\makebox{\rotatebox{90}{Ours}}}
& \includegraphics[width=\width]{images/main_fig/shoe_rack_01_08_collocated/albedo/4_nerad.jpg}
& \includegraphics[width=\width]{images/main_fig/shoe_rack_01_08_collocated/roughness/4_nerad.jpg}
& \includegraphics[width=\width]{images/main_fig/home_staged_window/albedo/9_nerad.jpg}
& \includegraphics[width=\width]{images/main_fig/home_staged_window/roughness/9_nerad.jpg}
& \includegraphics[width=\width]{images/main_fig/home_coffe_table_3/albedo/9_nerad.jpg}
& \includegraphics[width=\width]{images/main_fig/home_coffe_table_3/roughness/9_nerad.jpg}
\\

\end{tabular}

%% file: figures/merged_render_geometry_natural.tex
\providelength\width
\setlength\width{2.0cm}
\begin{figure*}
\tiny
\centering
\renewcommand{\tabcolsep}{1pt}
\begin{tabular}{ccccccc}
& Bedroom & Shelf & Kitchen & Shoe Rack & Coffee Table & Window Sill \\
{\makebox{\rotatebox{90}{\hspace{2pt} NeRO}}}
& \includegraphics[width=\width]{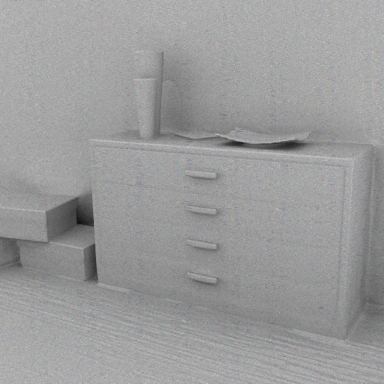}
& \includegraphics[width=\width]{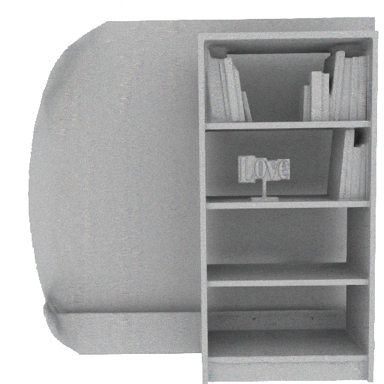}
& \includegraphics[width=\width]{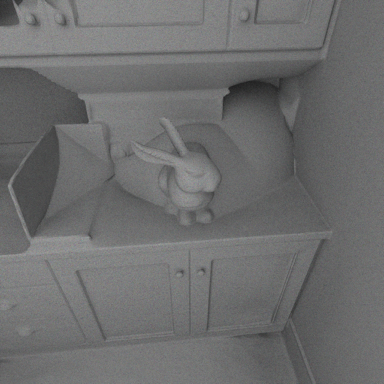}
& \includegraphics[width=\width]{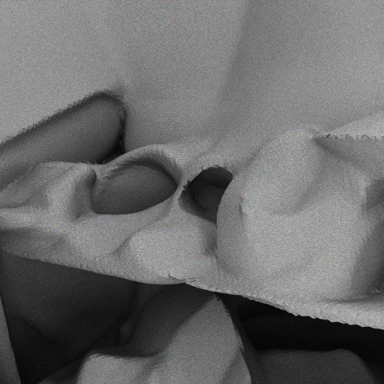}
& \includegraphics[width=\width]{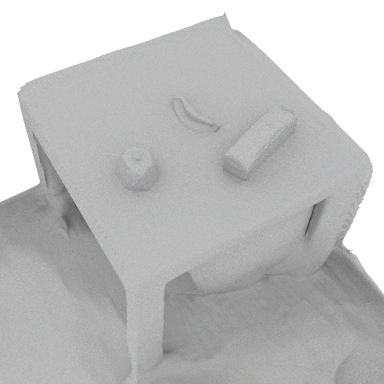}
& \includegraphics[width=\width]{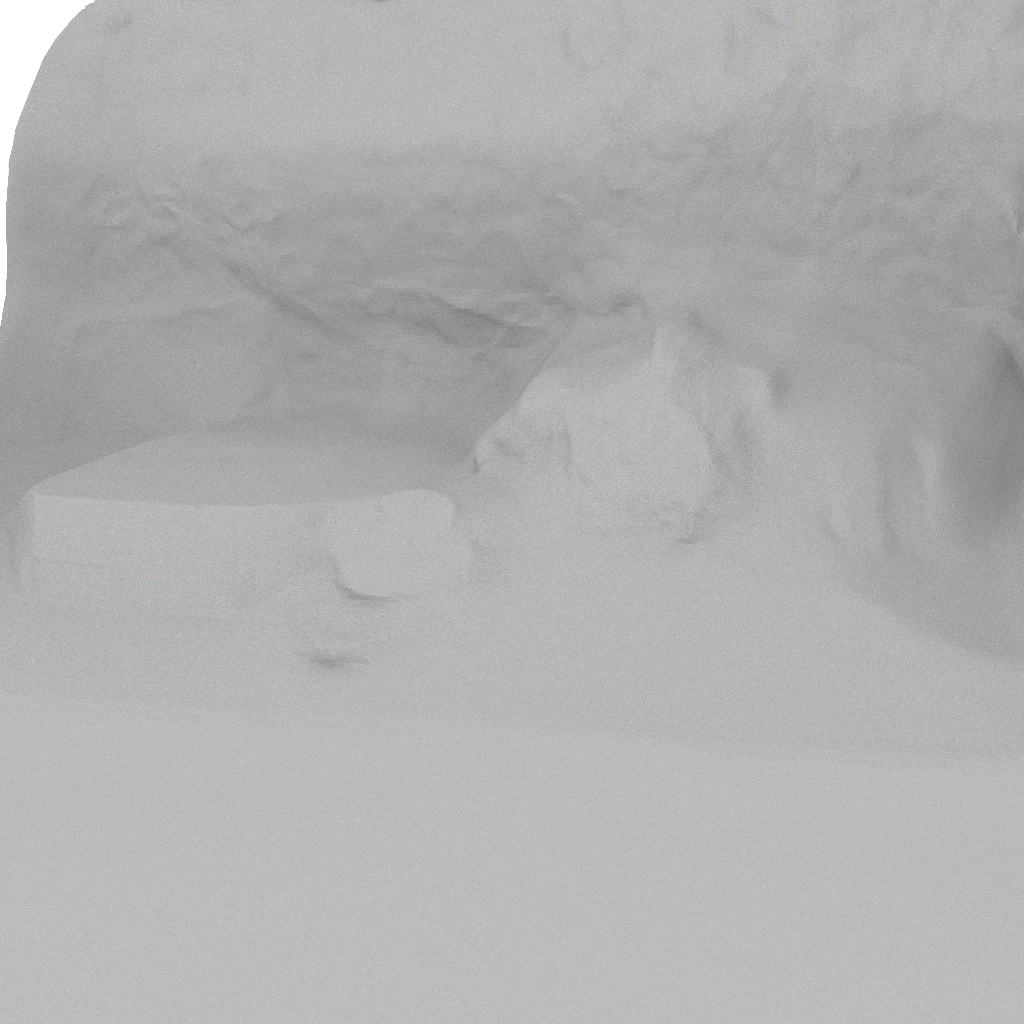}
\\

{\makebox{\rotatebox{90}{\hspace{2pt} IRGS}}}
& \includegraphics[width=\width]{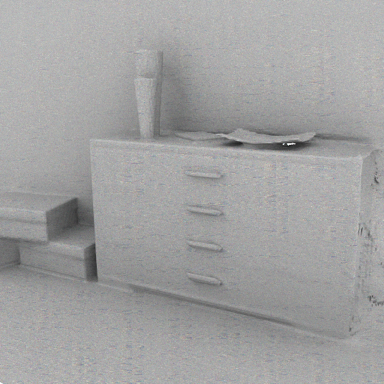}
& \includegraphics[width=\width]{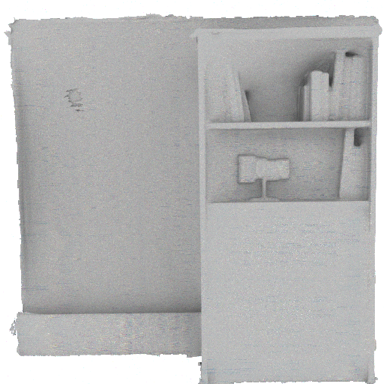}
& \includegraphics[width=\width]{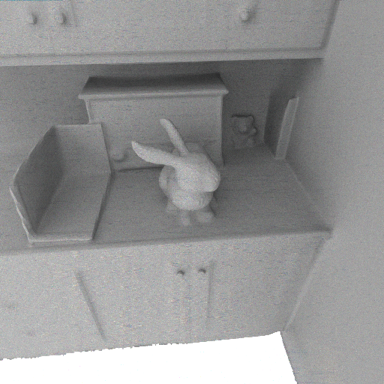}
& \includegraphics[width=\width]{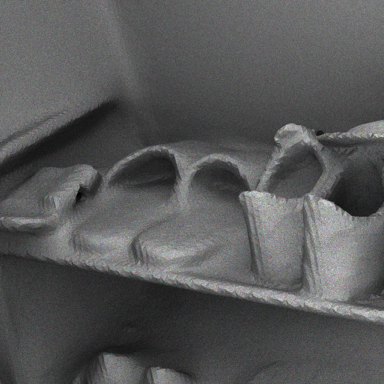}
& \includegraphics[width=\width]{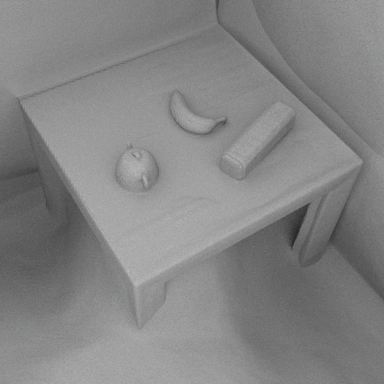}
& \includegraphics[width=\width]{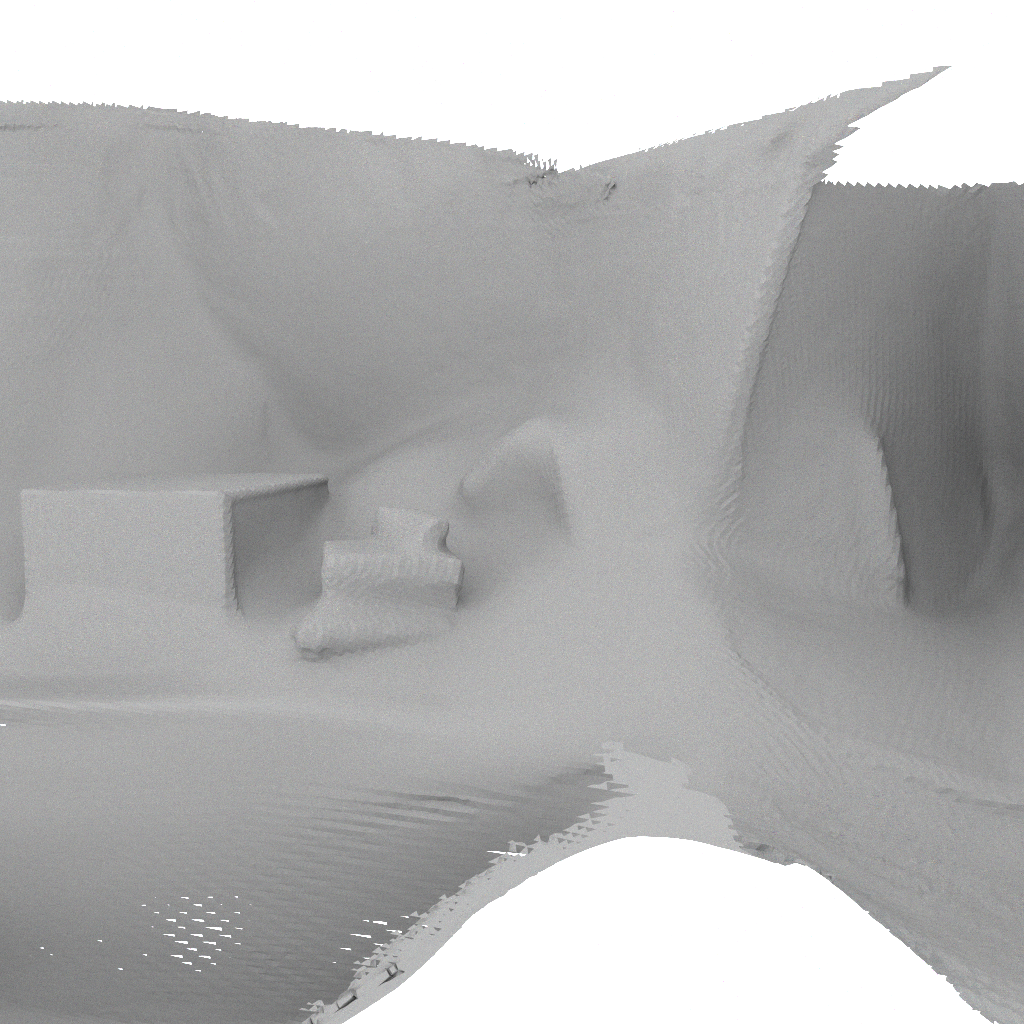}
\\
{\makebox{\rotatebox{90}{\hspace{2pt} Ours}}}
& \includegraphics[width=\width]{images/vis_geom/colloc-bedroom-1/Ours/ambient/wildlight_geometry_0.png}
& \includegraphics[width=\width]{images/vis_geom/colloc-living-room-1/Ours/ambient/wildlight_geometry_0.png}
& \includegraphics[width=\width]{images/vis_geom/custom_kitchen_ver9_principled_path/Ours/ambient/wildlight_geometry_0.png}
& \includegraphics[width=\width]{images/vis_geom/shoe_shelf_vignetting/Ours/ambient/wildlight_geometry_0.png}
& \includegraphics[width=\width]{images/vis_geom/home_coffe_table_3/Ours/ambient/wildlight_geometry_0.png}
& \includegraphics[width=\width]{images/vis_geom/home_staged_window/Ours/ambient/wildlight_geometry_0.png}
\\

{\makebox{\rotatebox{90}{\hspace{2pt} GT}}}
& \includegraphics[width=\width]{images/vis_geom/colloc-bedroom-1/GT/ambient/gt_geometry_0.png}
& \includegraphics[width=\width]{images/vis_geom/colloc-living-room-1/GT/ambient/gt_geometry_0.png}
& \includegraphics[width=\width]{images/vis_geom/custom_kitchen_ver9_principled_path/GT/ambient/gt_geometry_0.png}
&
&
&
\end{tabular}
\caption{\textbf{Qualitative comparison of geometry with co-located light \& camera methods.} The geometry of our method is comparable or better than natural illumination methods. }
\label{fig:geom_comparison_natural}
\end{figure*}

%% file: figures/qualitative_figure_rerender_new.tex
\providelength\width
\setlength\width{2.0cm}
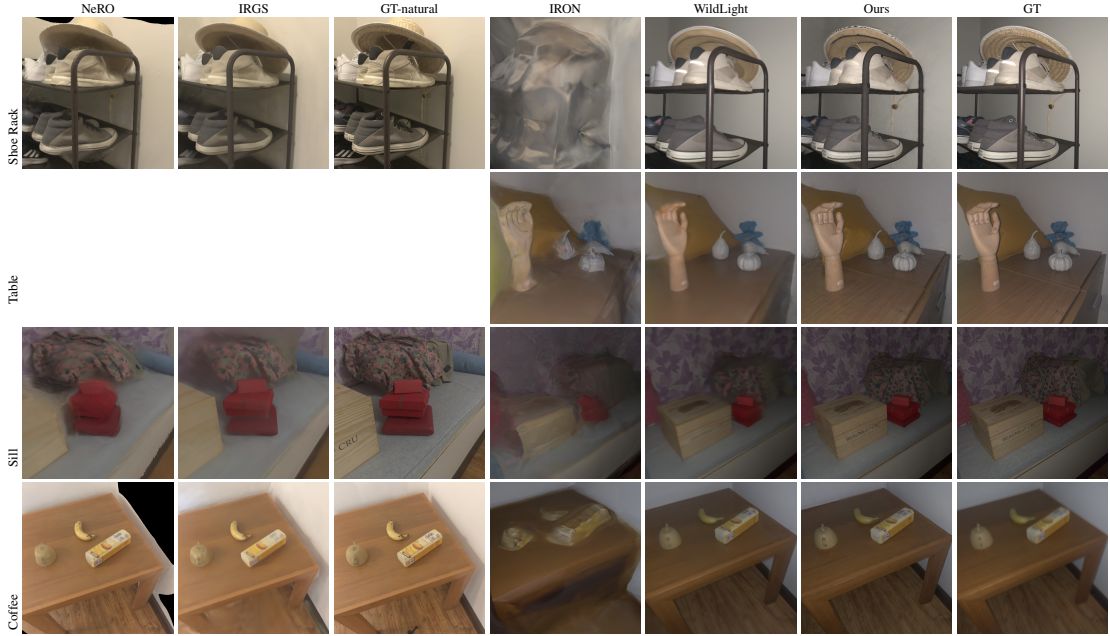
\begin{figure*}
\tiny
\centering
\renewcommand{\tabcolsep}{1pt}
\begin{tabular}{cccccccc}
  & NeRO & IRGS & GT-natural & IRON & WildLight & Ours & GT \\

  \input{generated/qualitative_generated_real_val}
  
\end{tabular}
\caption{\textbf{Qualitative comparison of re-rendering on real scenes.} We present re-rendering in validation views for the real scenes. Our method produces better re-rendering w.r.t. IRON~\cite{zhangIRONInverseRendering2022a}  and WildLight~\cite{chengWildLightInthewildInverse2023} as we are able to model global illumination. Our re-rendering is comparable to natural illumination methods, but co-located setup allow us to recover better material properties.}

\label{fig:qualitative_figure_rerender}

\end{figure*}

%% file: generated/qualitative_generated_real_val.tex
{\makebox{\rotatebox{90}{\hspace{0pt} Shoe Rack}}}
& \includegraphics[width=\width]{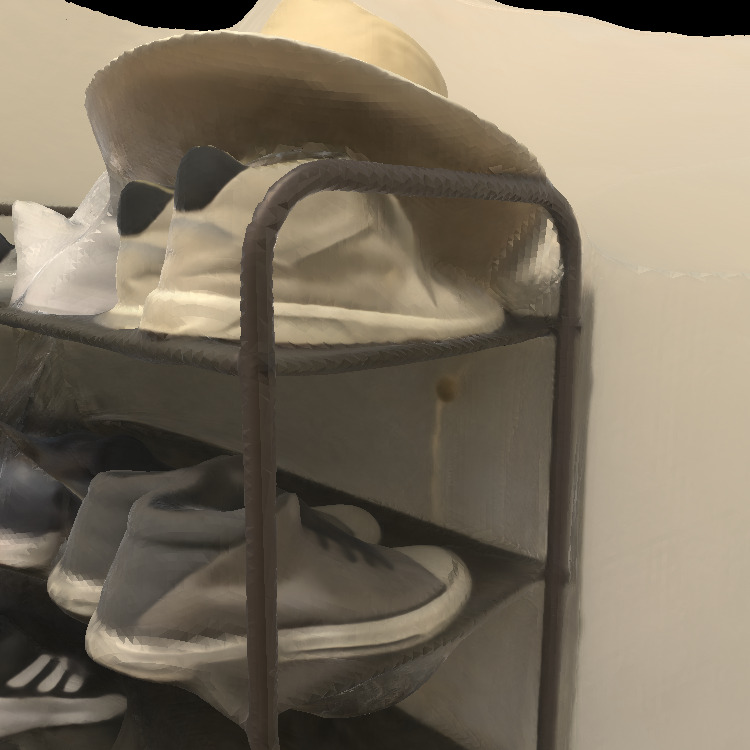}
& \includegraphics[width=\width]{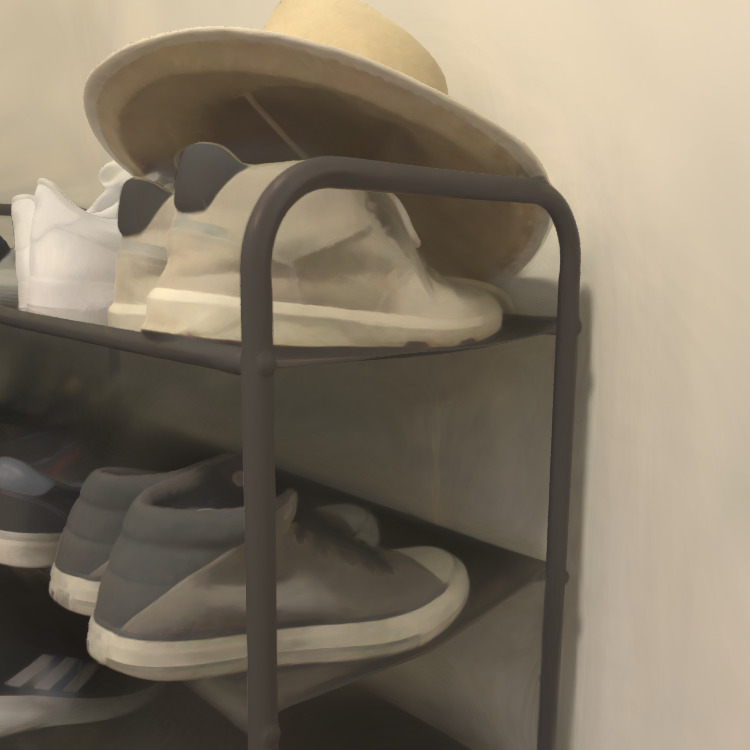}
& \includegraphics[width=\width]{images/main_fig/shoe_rack_01_08_collocated-natural/rerender/9_gt.jpg}
& \includegraphics[width=\width]{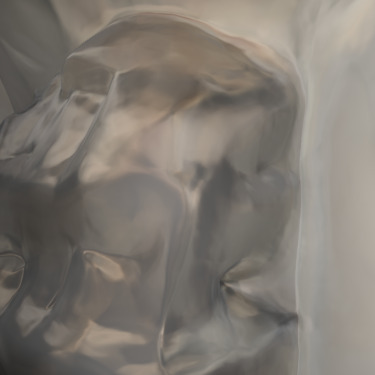}
& \includegraphics[width=\width]{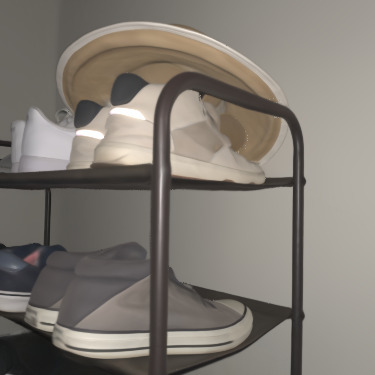}
& \includegraphics[width=\width]{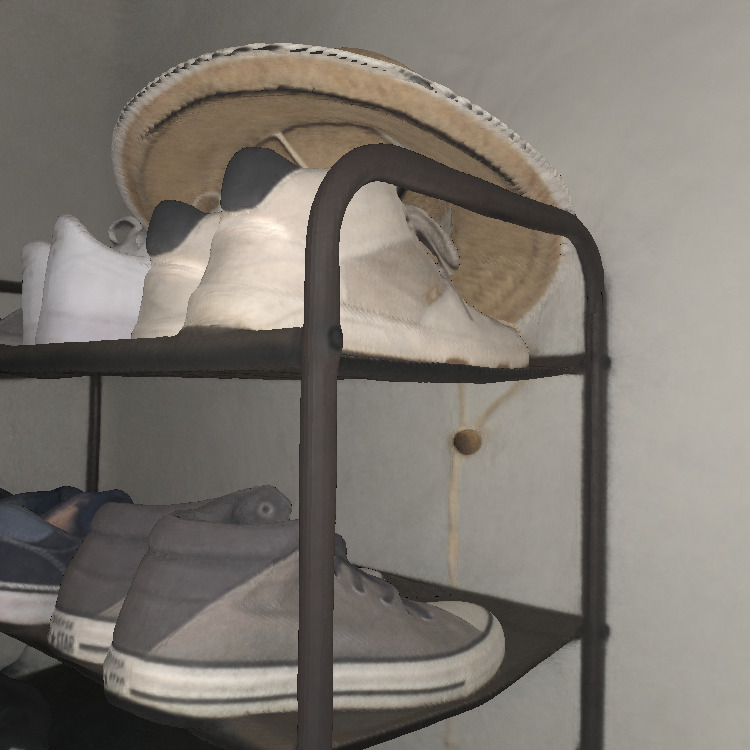}
& \includegraphics[width=\width]{images/main_fig/shoe_rack_01_08_collocated/rerender/4_gt.jpg}
\\

{\makebox{\rotatebox{90}{\hspace{6pt} Table}}}
&
&
&
& \includegraphics[width=\width]{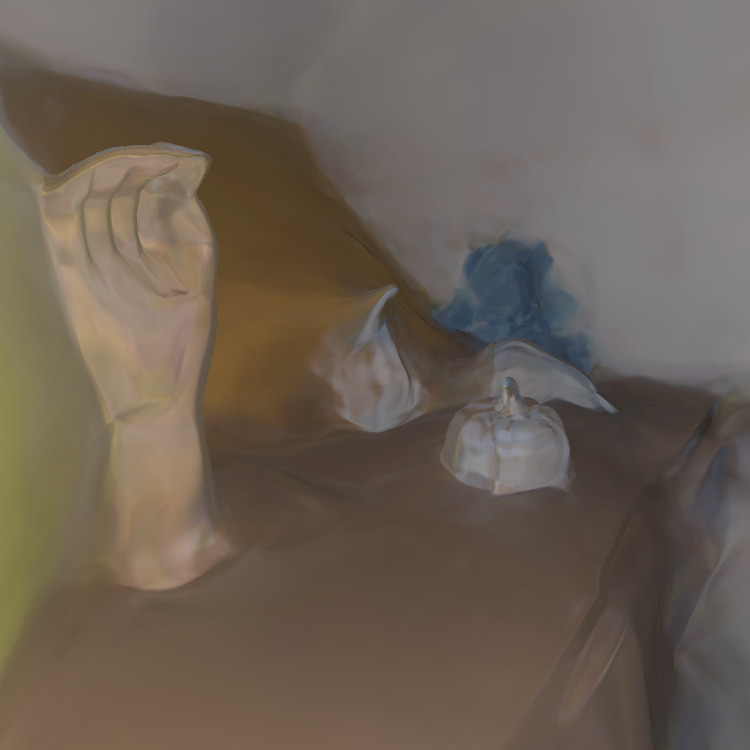}
& \includegraphics[width=\width]{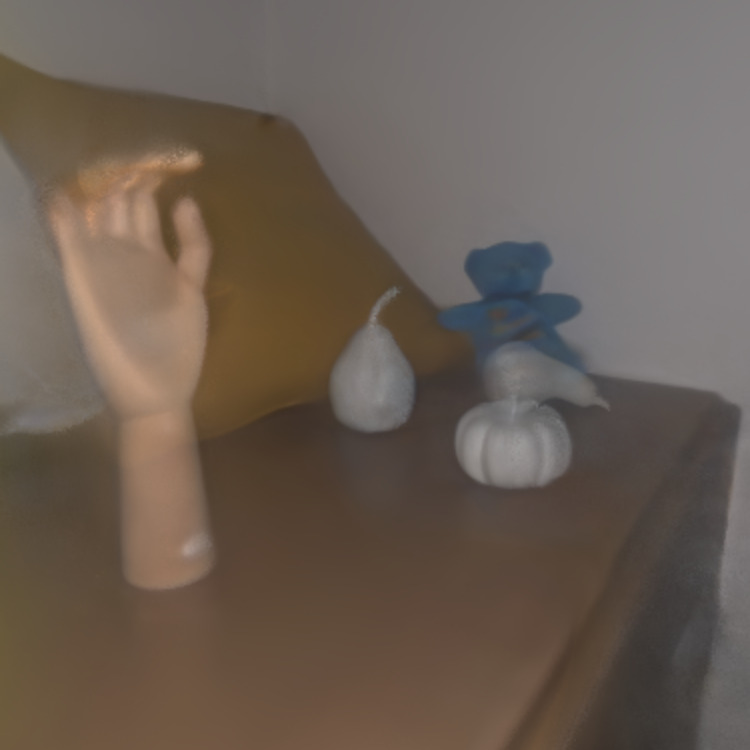}
& \includegraphics[width=\width]{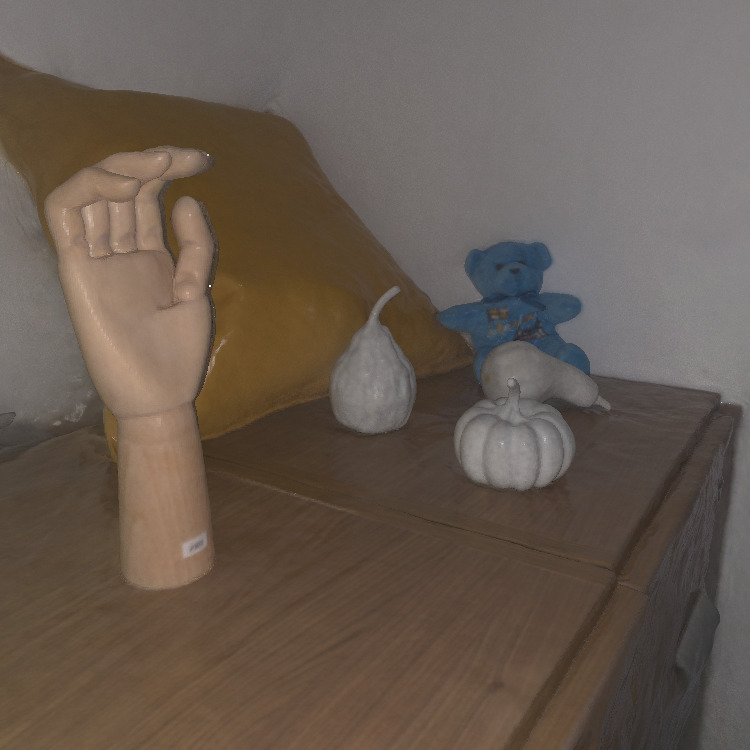}
& \includegraphics[width=\width]{images/main_fig/irb_4th_printer_scene_vignetting_real/rerender/15_gt.jpg}
\\

{\makebox{\rotatebox{90}{\hspace{3pt} Sill}}}
& \includegraphics[width=\width]{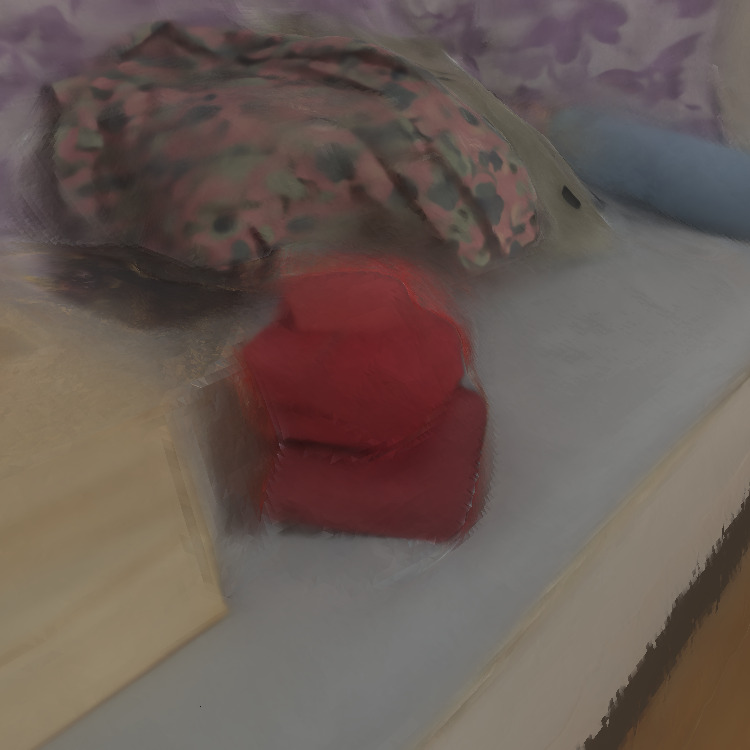}
& \includegraphics[width=\width]{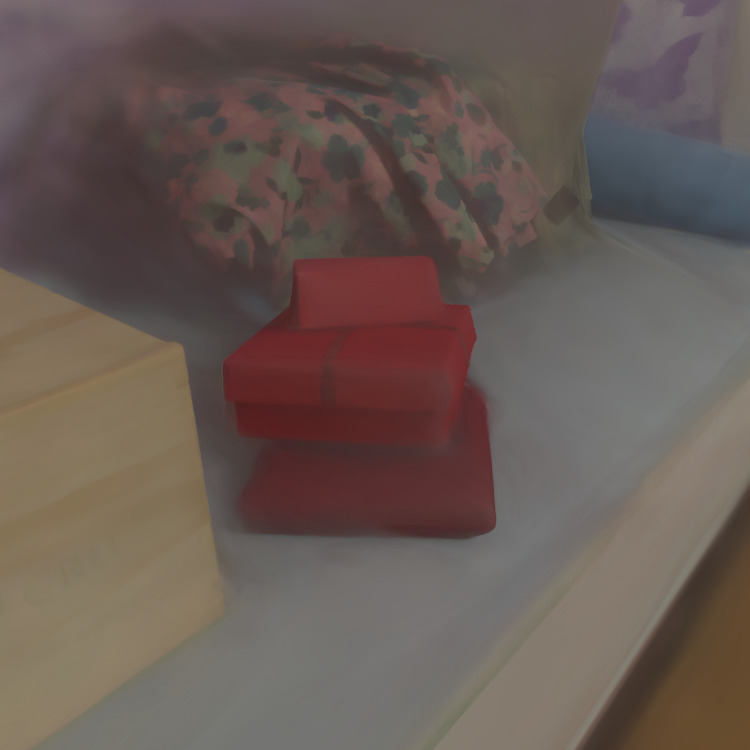}
& \includegraphics[width=\width]{images/main_fig/home_staged_window-natural/rerender/14_gt.jpg}
& \includegraphics[width=\width]{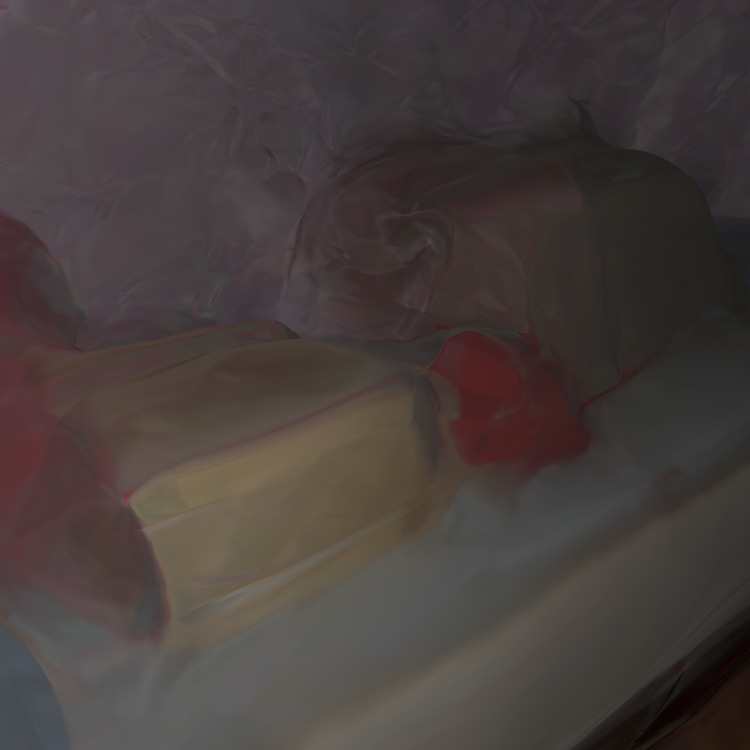}
& \includegraphics[width=\width]{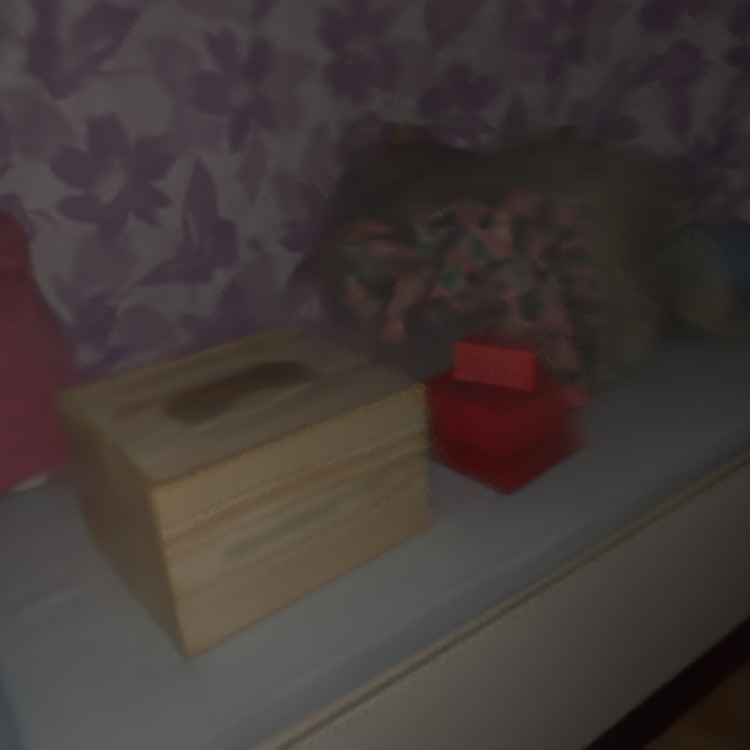}
& \includegraphics[width=\width]{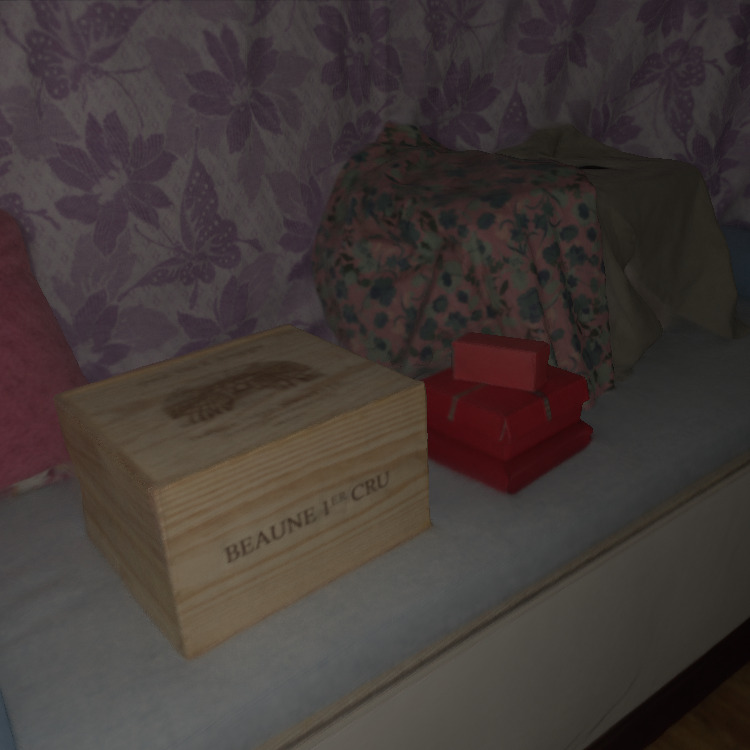}
& \includegraphics[width=\width]{images/main_fig/home_staged_window/rerender/9_gt.jpg}
\\

{\makebox{\rotatebox{90}{\hspace{0pt} Coffee}}}
& \includegraphics[width=\width]{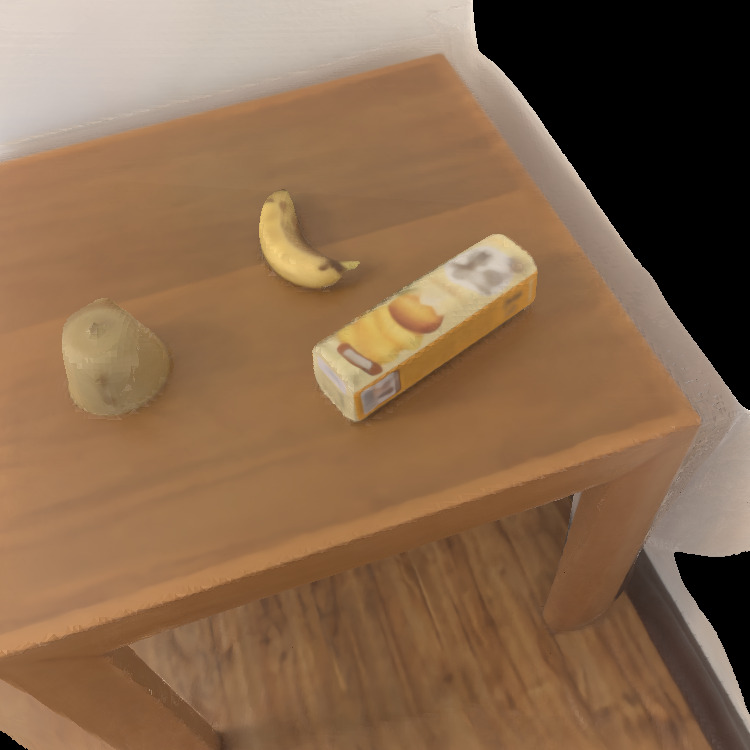}
& \includegraphics[width=\width]{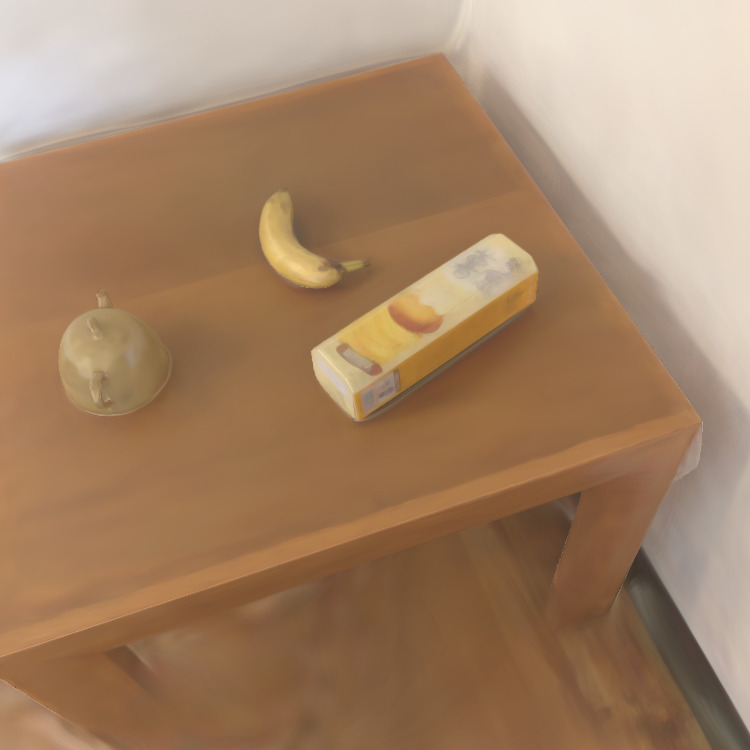}
& \includegraphics[width=\width]{images/main_fig/home_coffe_table_3-natural/rerender/30_gt.jpg}
& \includegraphics[width=\width]{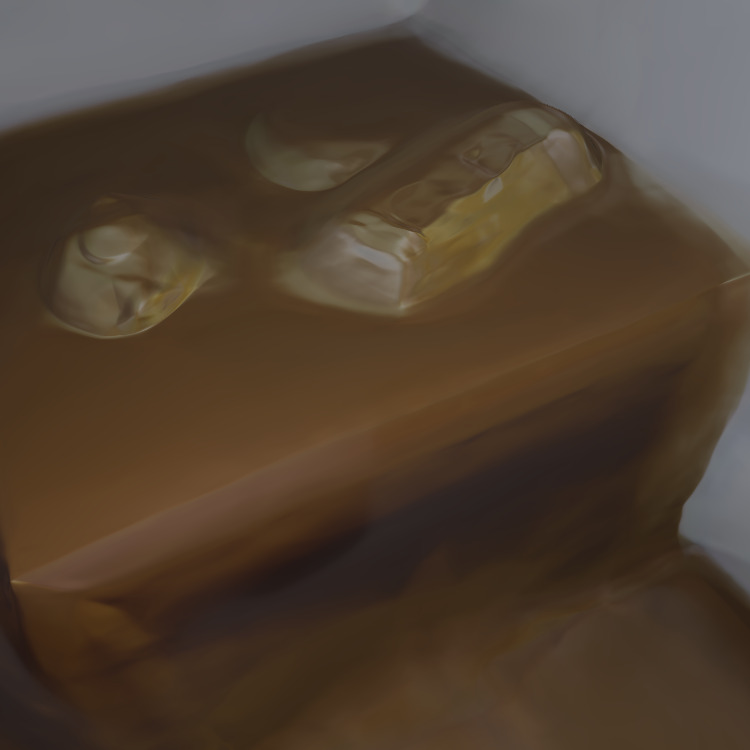}
& \includegraphics[width=\width]{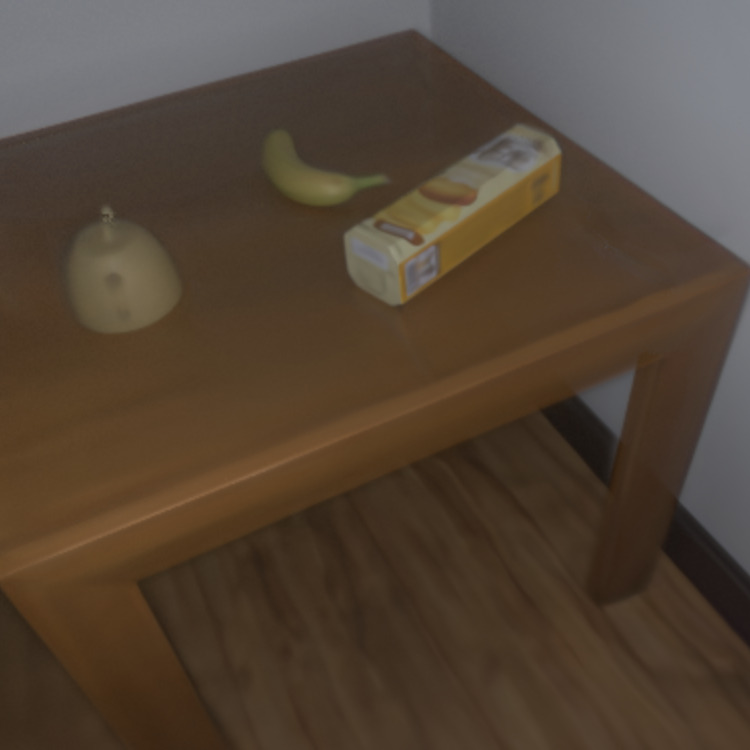}
& \includegraphics[width=\width]{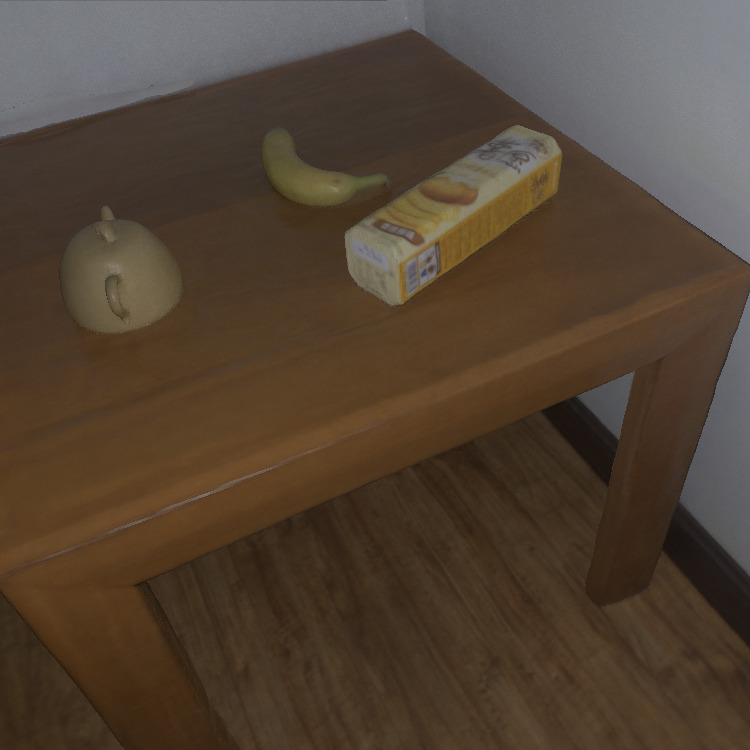}
& \includegraphics[width=\width]{images/main_fig/home_coffe_table_3/rerender/9_gt.jpg}
\\

%% file: figures/suppl_wildlight.tex
\providelength\width
\setlength\width{2.3cm}
\begin{figure*}
\tiny
\centering

\renewcommand{\tabcolsep}{1pt}
\begin{tabular}{cccccc}

& Image & Re-Render & Albedo & Roughness & Geometry  \\

\input{generated/suppl_wildlight}

\end{tabular}
\vspace{-10pt}
\caption{
Qualitative results of WildLight on Co-located Light and Camera under ambient natural illumination.
}
\label{fig:suppl_wildlight}
\end{figure*}
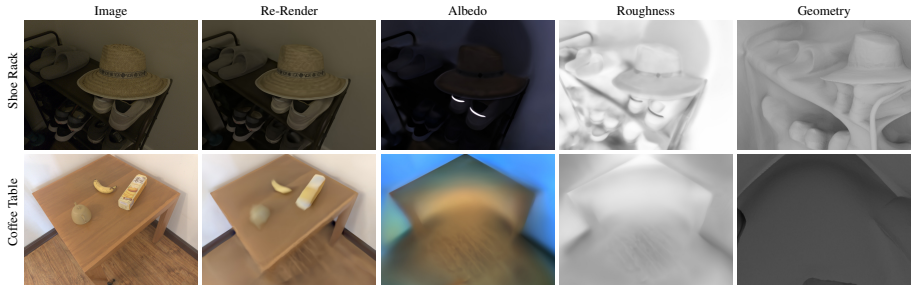

%% file: generated/suppl_wildlight.tex
{\makebox[5pt]{\rotatebox{90}{\hspace{5mm} \tiny Shoe Rack}}}
& \includegraphics[width=\width]{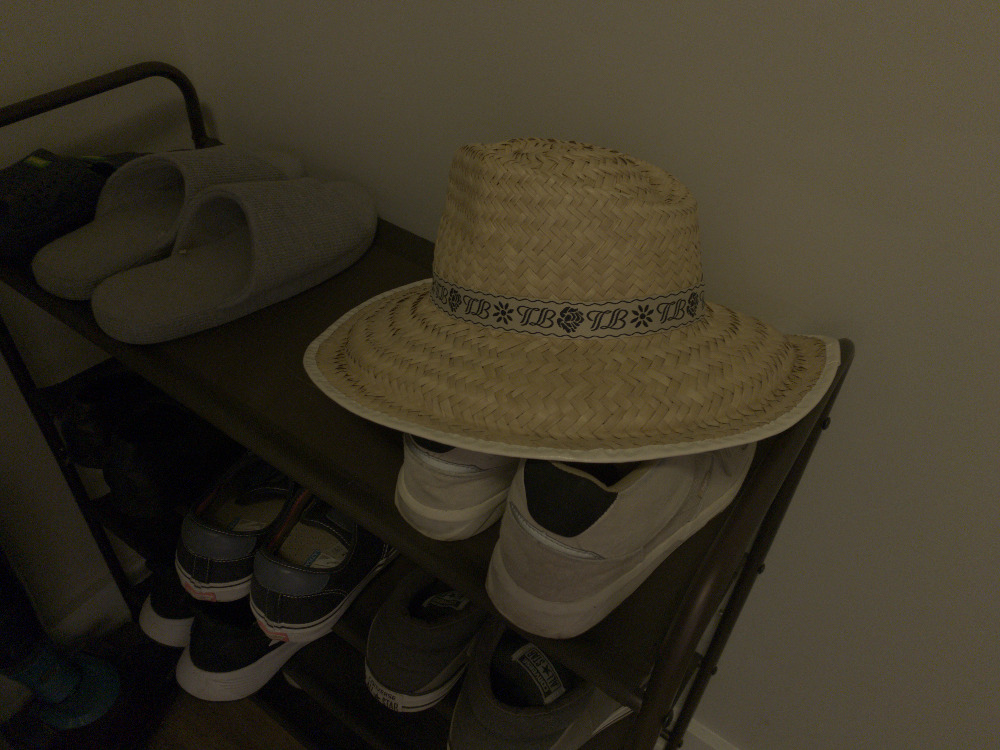}
& \includegraphics[width=\width]{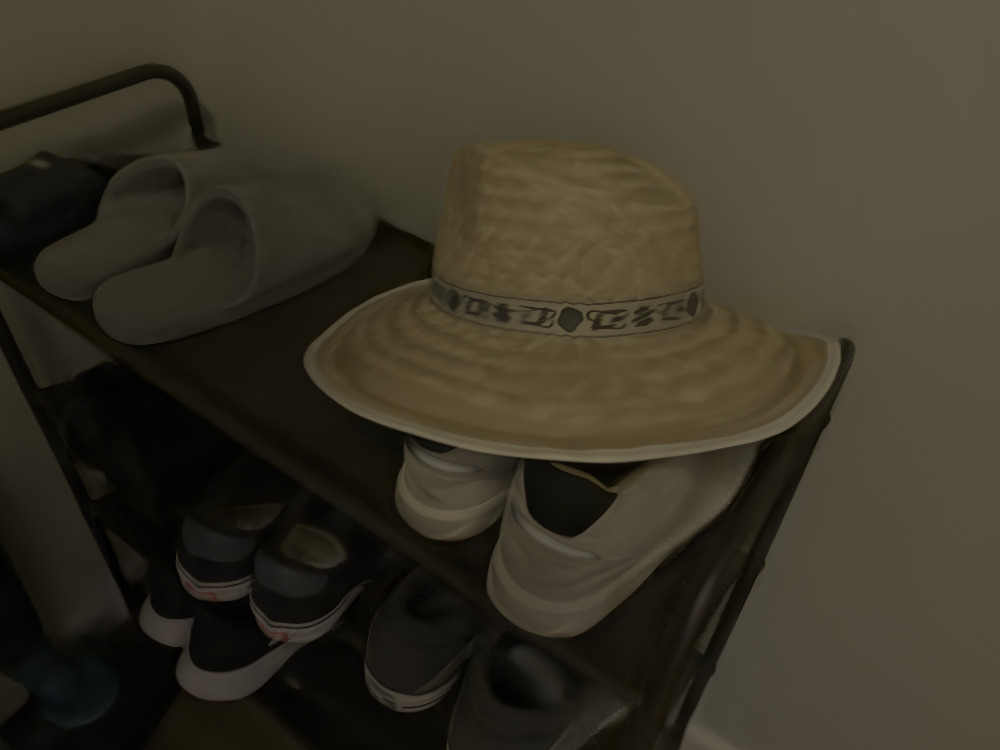}
& \includegraphics[width=\width]{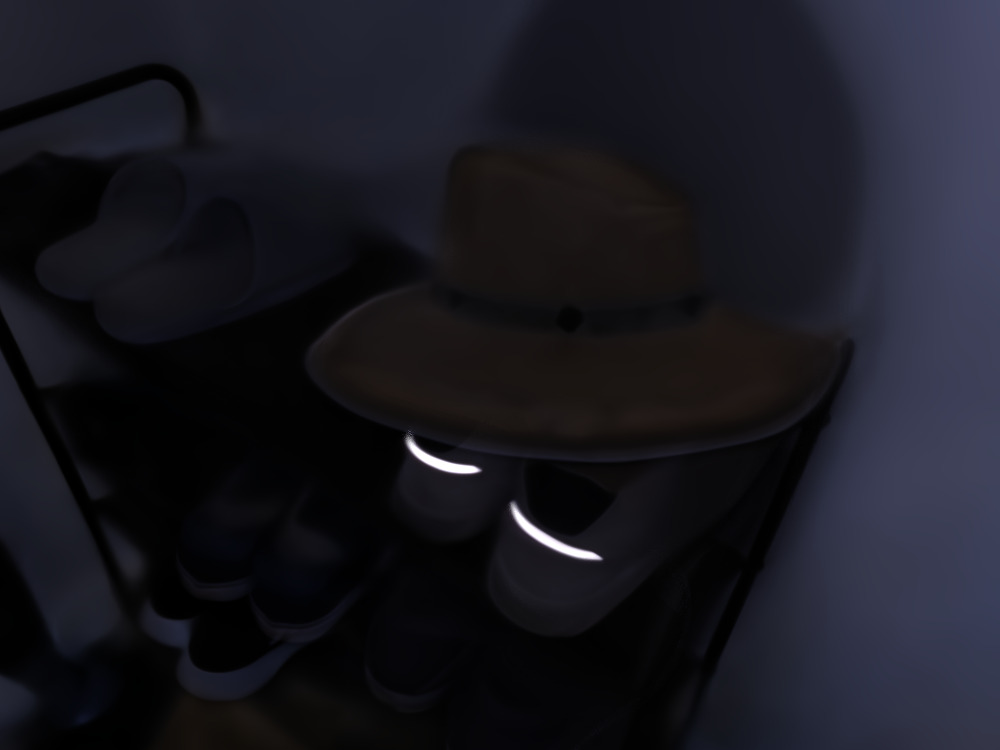}
& \includegraphics[width=\width]{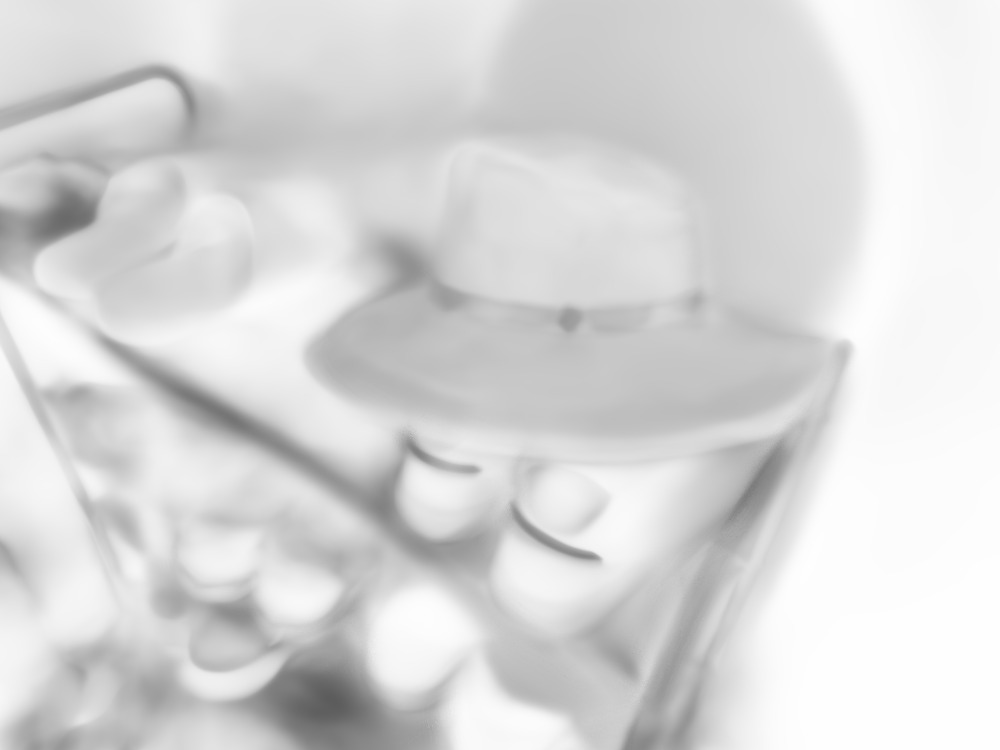}
& \includegraphics[width=\width]{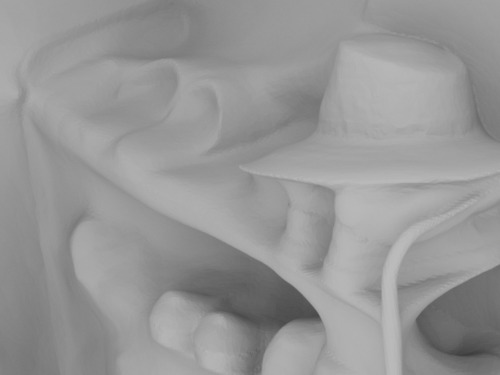}
\\

{\makebox[5pt]{\rotatebox{90}{\hspace{5mm}\tiny Coffee Table}}}
& \includegraphics[width=\width]{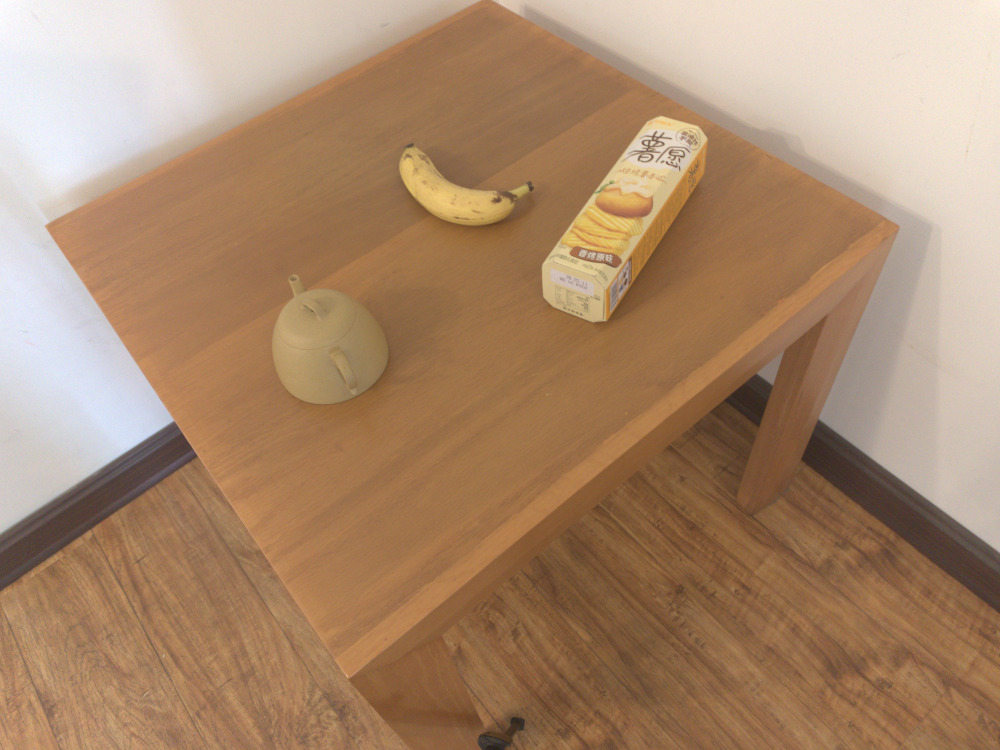}
& \includegraphics[width=\width]{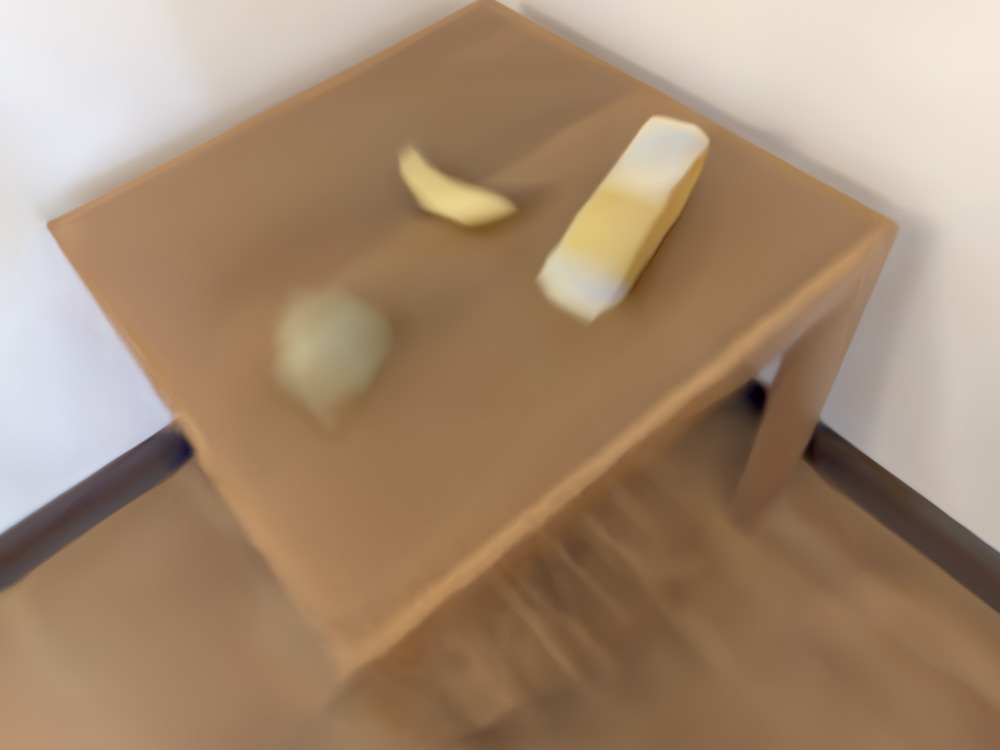}
& \includegraphics[width=\width]{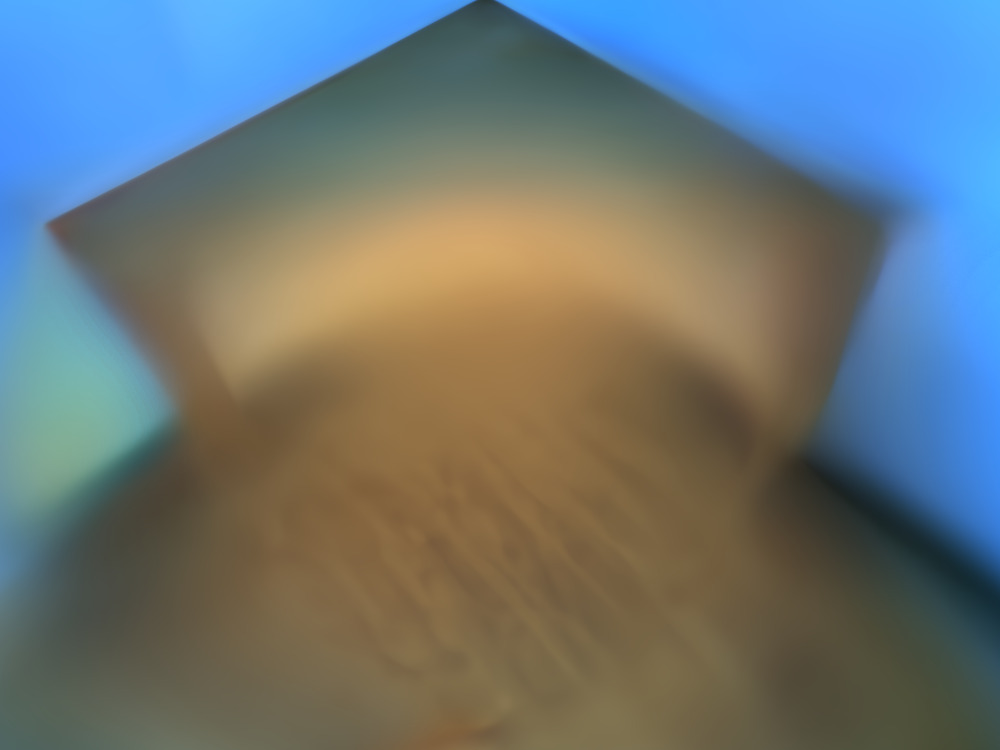}
& \includegraphics[width=\width]{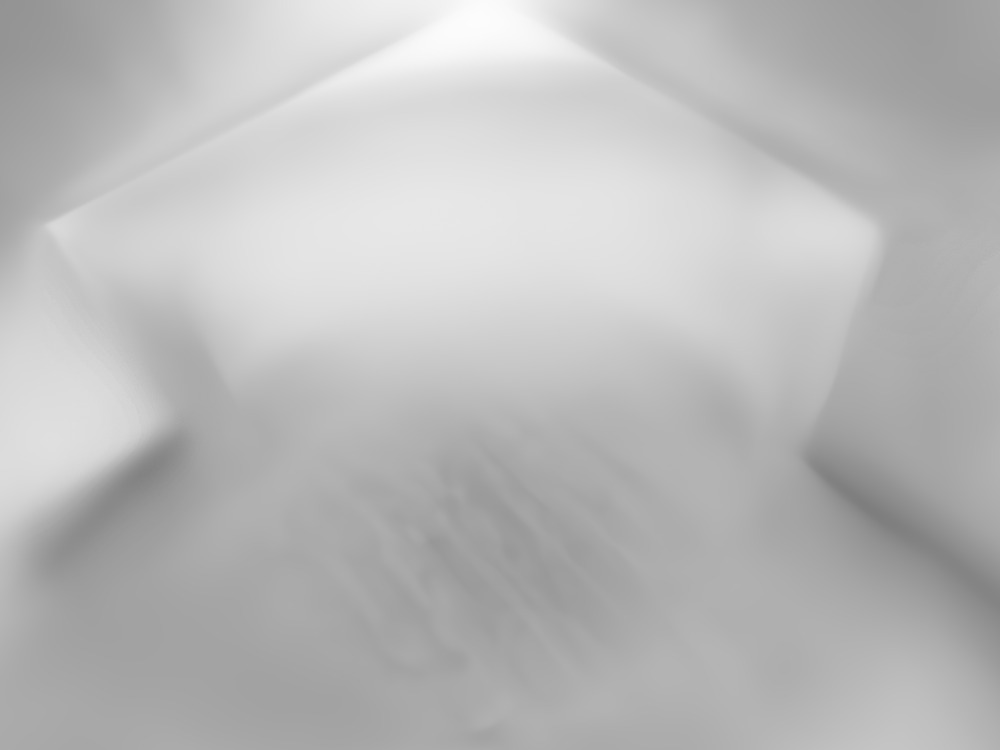}
& \includegraphics[width=\width]{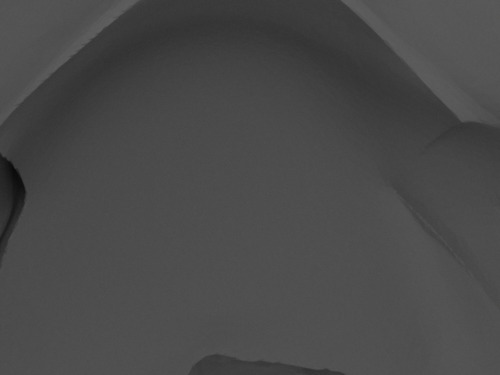}
\\

%% file: figures/suppl_qualitative_figure.tex
\providelength\width
\setlength\width{2.3cm}
\renewcommand{\tabcolsep}{1pt}

\begin{figure*}
\tiny
\centering
\begin{tabular}{ccccccc}
    & NeRO & IRGS & IRON & WildLight & Ours & gt \\
     \input{generated/suppl_qualitative_generated_synthetic_0}
\end{tabular}

\vspace{-10pt}
\caption{
Qualitative comparison on synthetic scene \textit{bedroom}. 
}
\label{fig:suppl_qualitative_figure_synthetic_1}
\end{figure*}
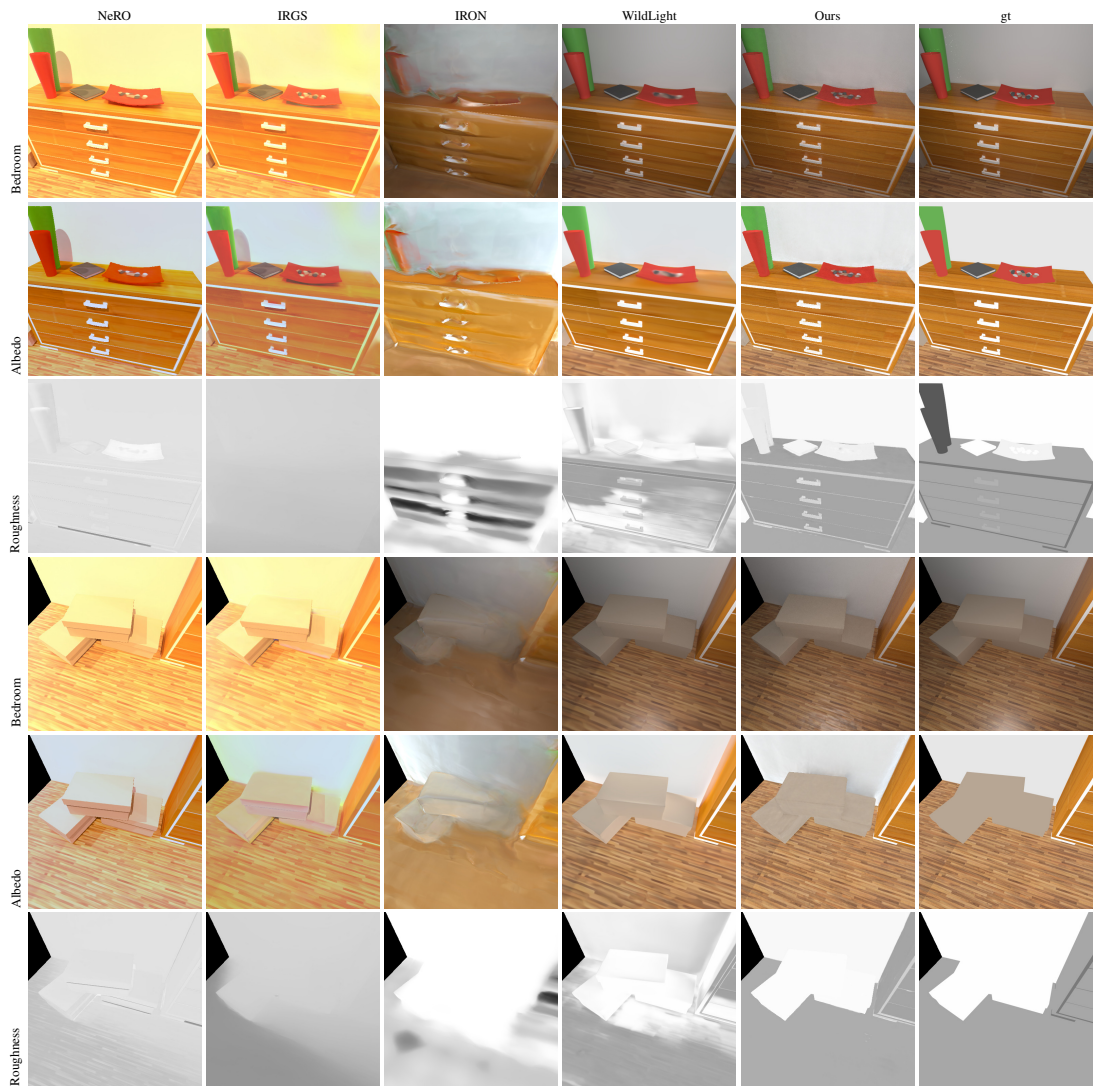

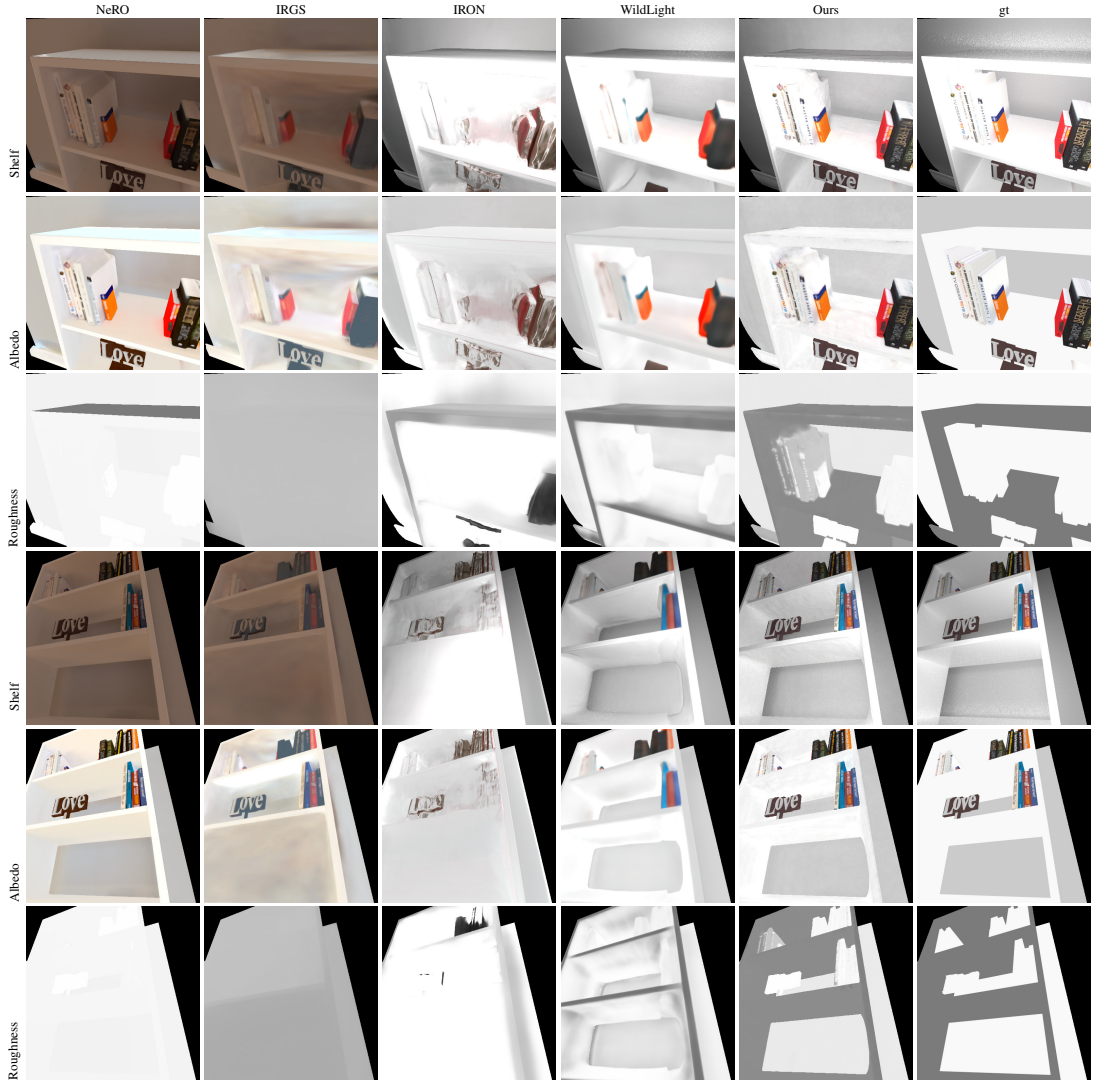
\begin{figure*}
\tiny
\centering
\begin{tabular}{ccccccc}
    & NeRO & IRGS & IRON & WildLight & Ours & gt \\
\input{generated/suppl_qualitative_generated_synthetic_1}
\end{tabular}
\vspace{-10pt}
\caption{
Qualitative comparison on synthetic scene \textit{shelf}. 
}
\label{fig:suppl_qualitative_figure_synthetic_2}
\end{figure*}

\begin{figure*}
\tiny
\centering
\begin{tabular}{ccccccc}
    & NeRO & IRGS & IRON & WildLight & Ours & gt \\
\input{generated/suppl_qualitative_generated_synthetic_2}
\end{tabular}
\vspace{-10pt}
\caption{
Qualitative comparison on synthetic scene \textit{kitchen counter}.
}
\label{fig:suppl_qualitative_figure_synthetic_3}
\end{figure*}
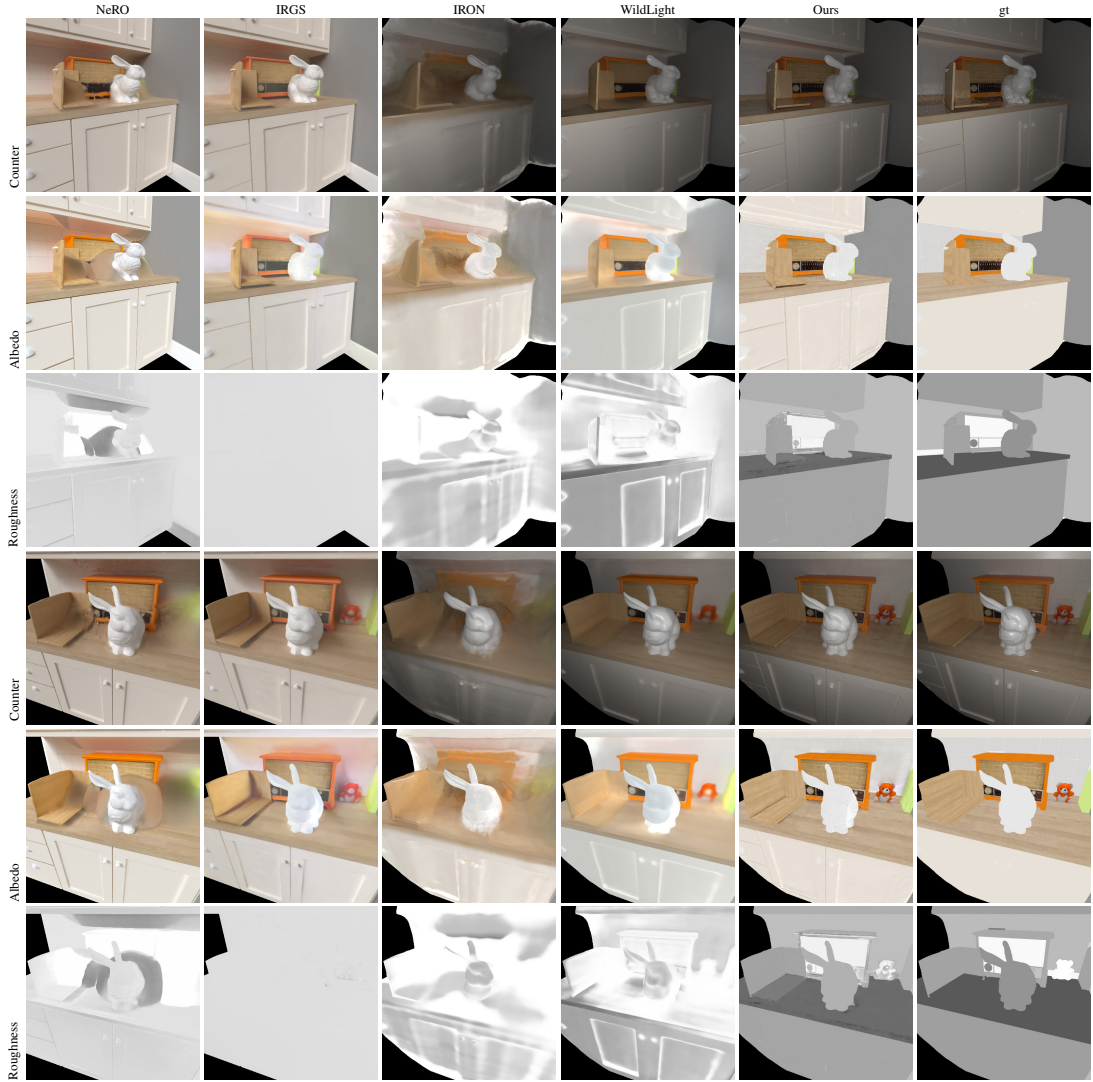

\begin{figure*}
\tiny
\centering
\begin{tabular}{ccccccc}
    & NeRO & IRGS & IRON & WildLight & Ours & gt \\
\input{generated/suppl_qualitative_generated_real_0}
\end{tabular}
\vspace{-10pt}
\caption{
Qualitative comparison on real scene \textit{shoe rack}.
}
\label{fig:suppl_qualitative_figure_real_1}
\end{figure*}
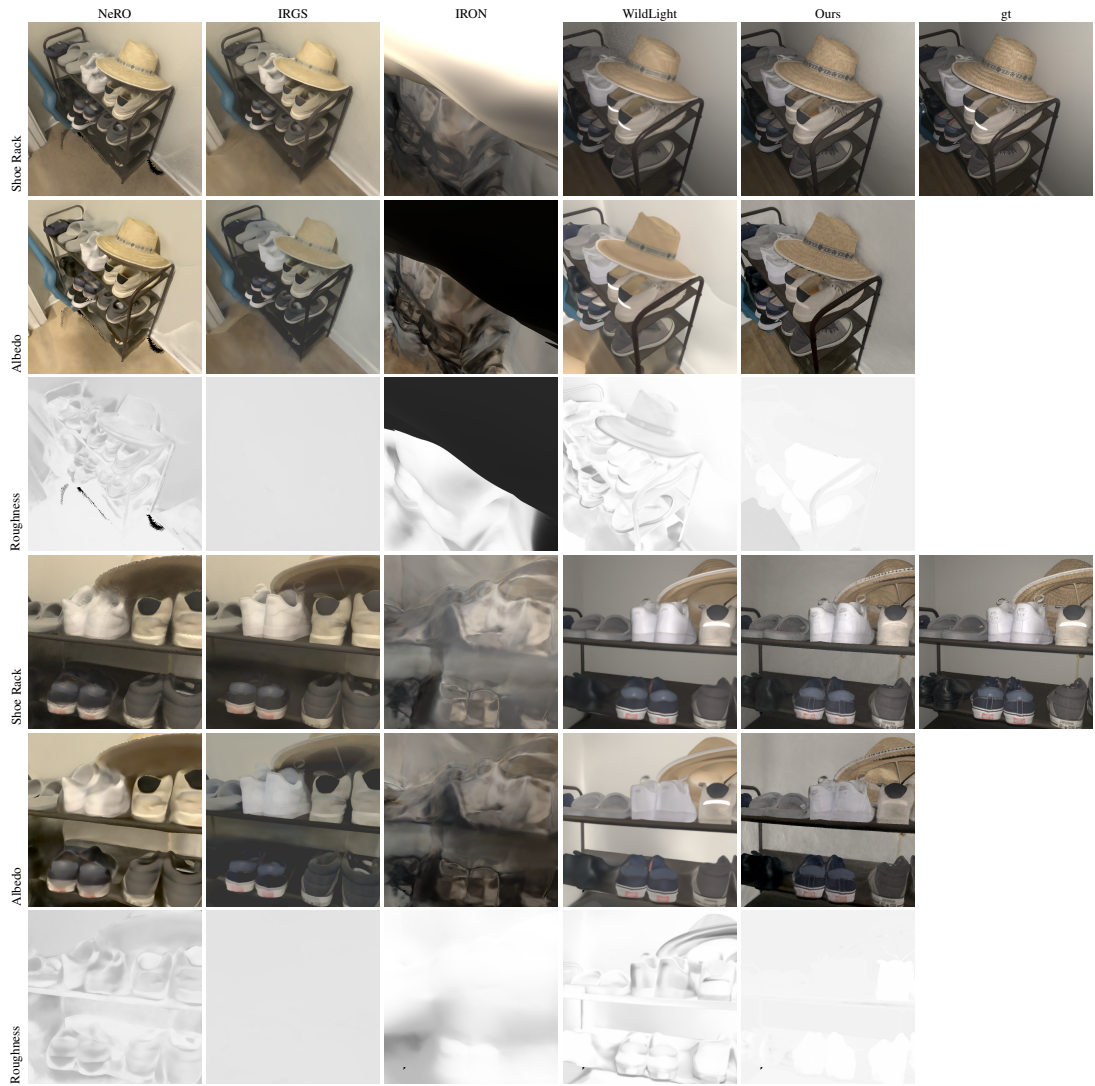

\begin{figure*}
\tiny
\centering
\begin{tabular}{ccccccc}
   & NeRO & IRGS & IRON & WildLight & Ours & gt \\
\input{generated/suppl_qualitative_generated_real_1}
\end{tabular}
\vspace{-10pt}
\caption{
Qualitative comparison on real scene \textit{table}.
}
\label{fig:suppl_qualitative_figure_real_2}
\end{figure*}

\begin{figure*}
\tiny
\centering
\begin{tabular}{ccccccc}
    & NeRO & IRGS & IRON & WildLight & Ours & gt \\
\input{generated/suppl_qualitative_generated_real_2}
\end{tabular}
\vspace{-10pt}
\caption{
Qualitative comparison on real scene \textit{window sill}.
}
\label{fig:suppl_qualitative_figure_real_3}
\end{figure*}
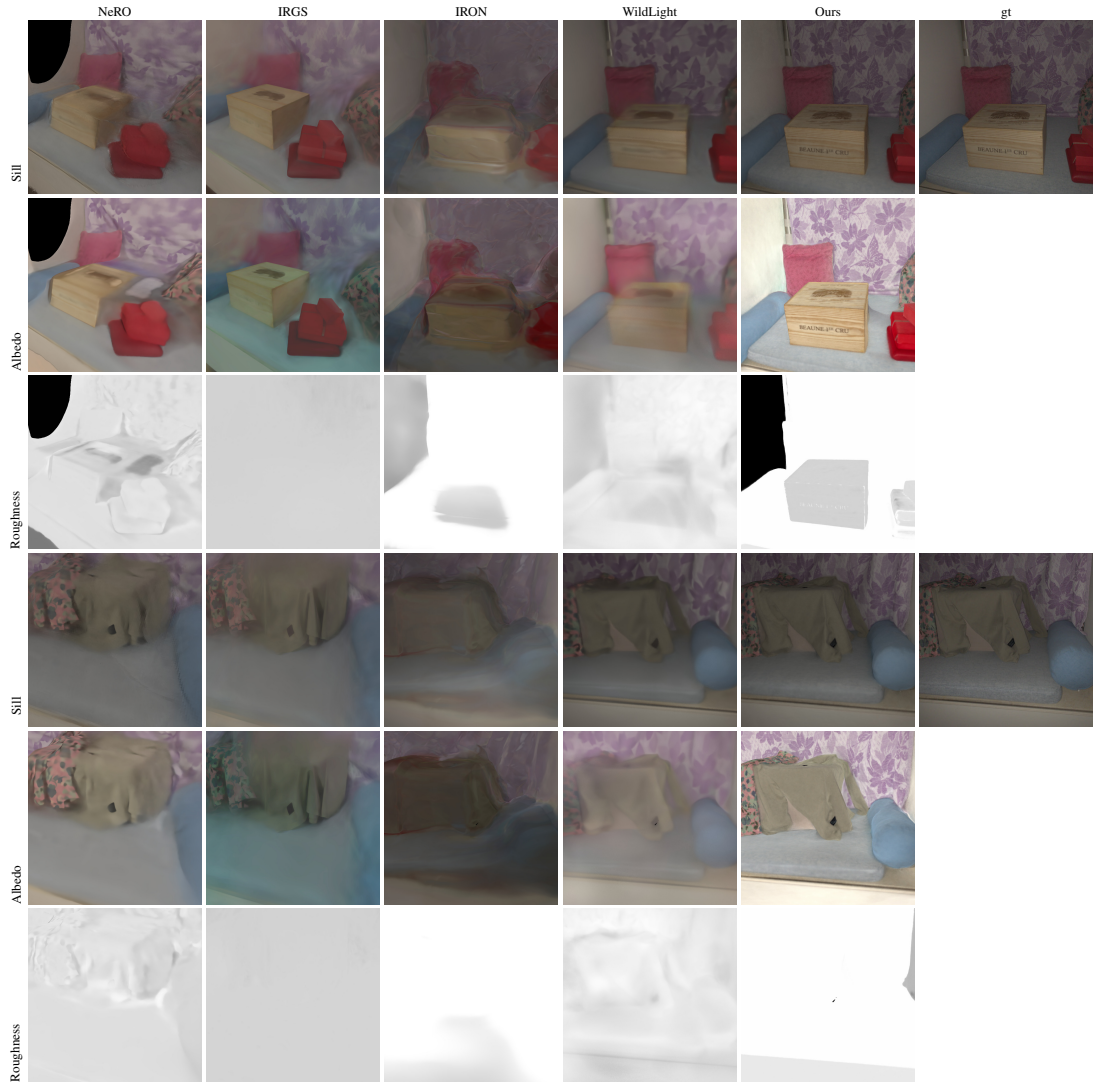

\begin{figure*}
\footnotesize
\centering
\begin{tabular}{ccccccc}
    & NeRO & IRGS & IRON & WildLight & Ours & gt \\
\input{generated/suppl_qualitative_generated_real_3}
\end{tabular}
\vspace{-10pt}
\caption{
Qualitative comparison on real scene \textit{coffee table}.
}
\label{fig:suppl_qualitative_figure_real_4}
\end{figure*}

%% file: generated/suppl_qualitative_generated_synthetic_0.tex
{\makebox[5pt]{\rotatebox{90}{\tiny \hspace{0pt} Bedroom}}} &
\includegraphics[width=\width]{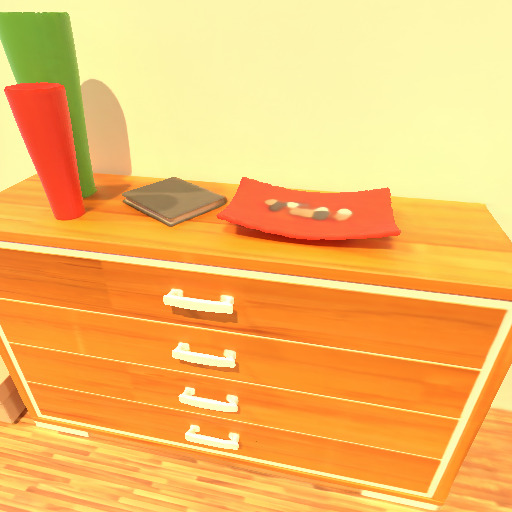} &
\includegraphics[width=\width]{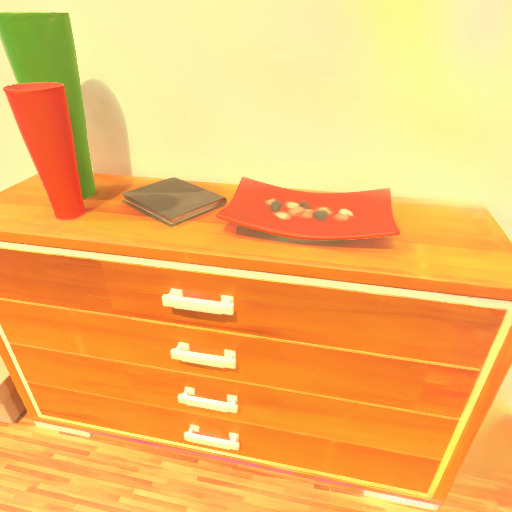} &
\includegraphics[width=\width]{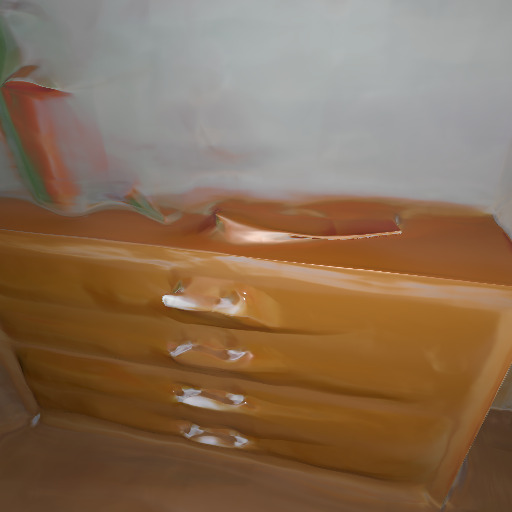} &
\includegraphics[width=\width]{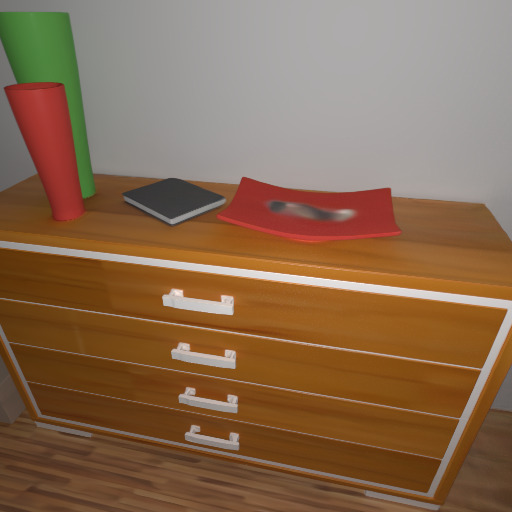} &
\includegraphics[width=\width]{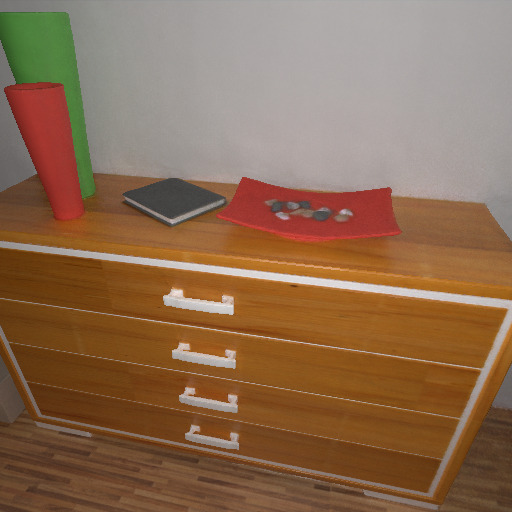} &
\includegraphics[width=\width]{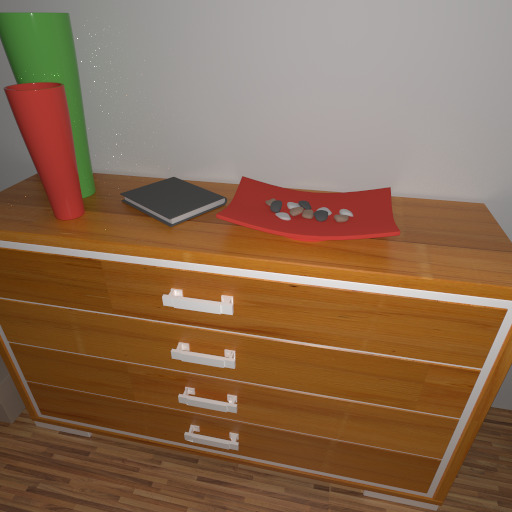} \\
{\makebox[5pt]{\rotatebox{90}{\tiny Albedo}}} &
\includegraphics[width=\width]{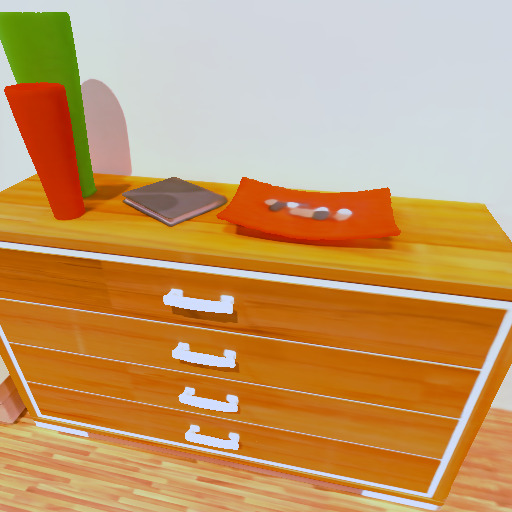} &
\includegraphics[width=\width]{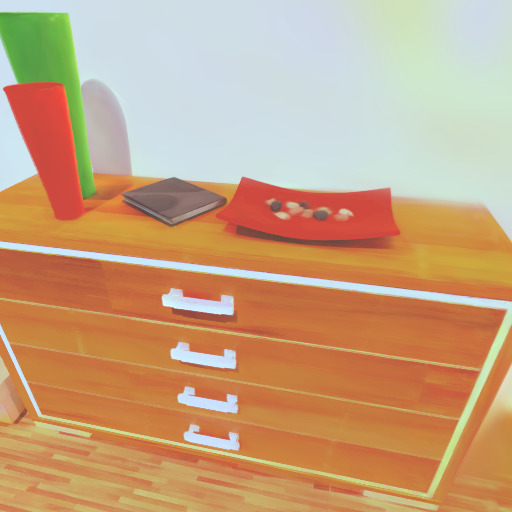} &
\includegraphics[width=\width]{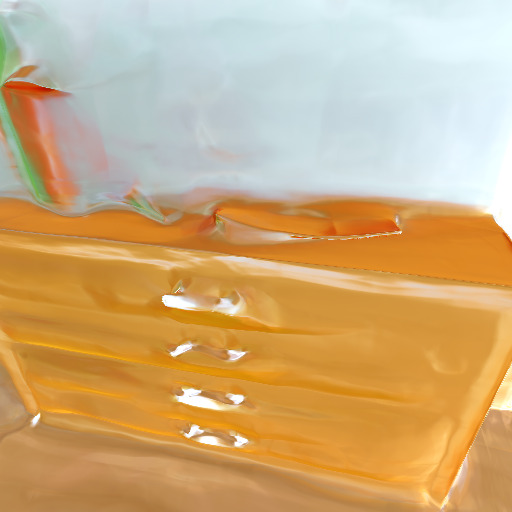} &
\includegraphics[width=\width]{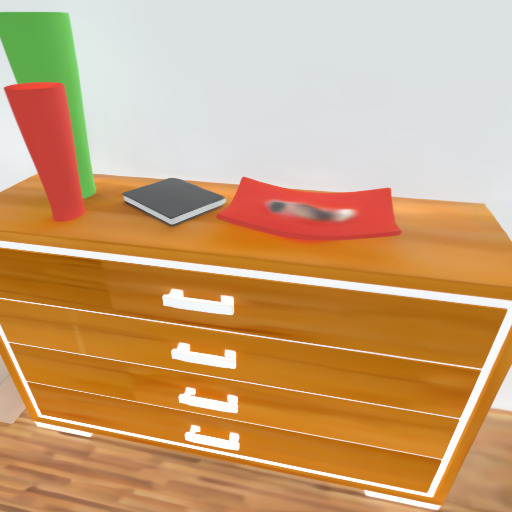} &
\includegraphics[width=\width]{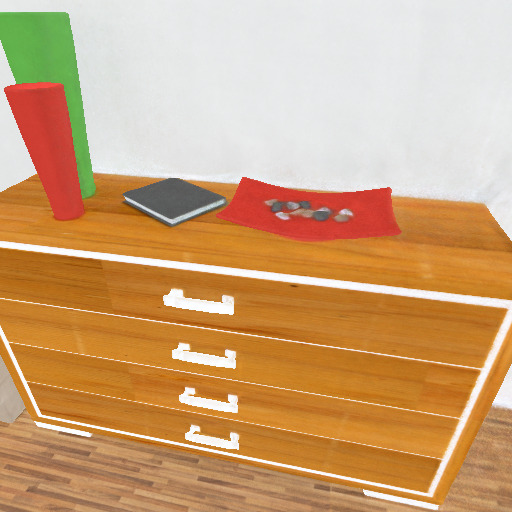} &
\includegraphics[width=\width]{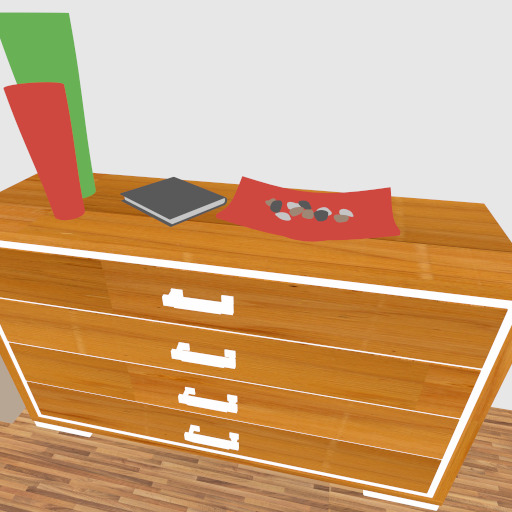} \\
{\makebox[5pt]{\rotatebox{90}{\tiny Roughness}}} &
\includegraphics[width=\width]{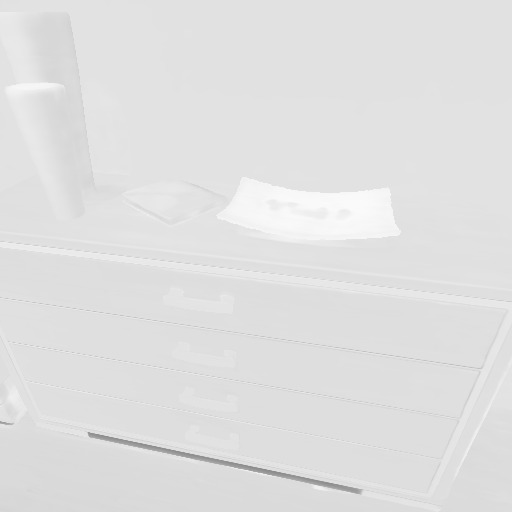} &
\includegraphics[width=\width]{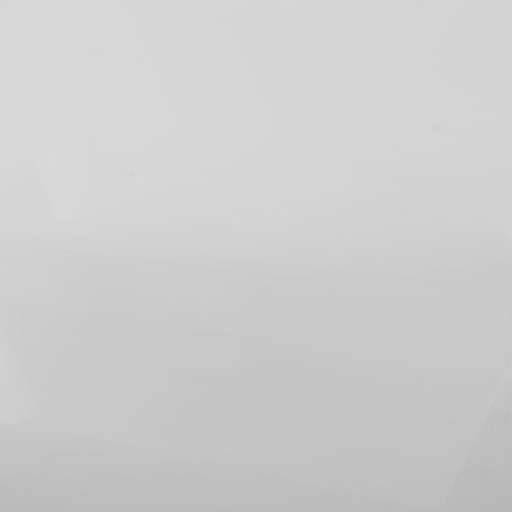} &
\includegraphics[width=\width]{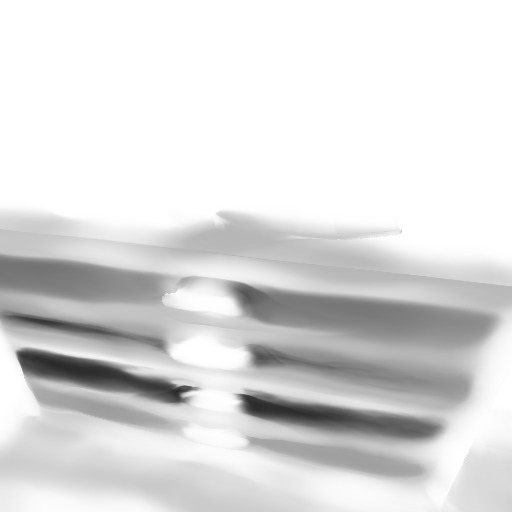} &
\includegraphics[width=\width]{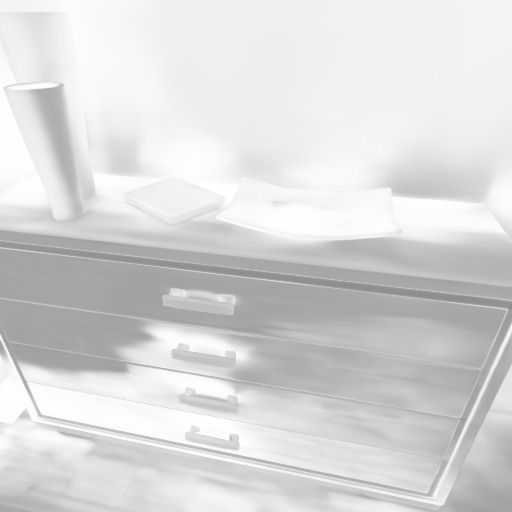} &
\includegraphics[width=\width]{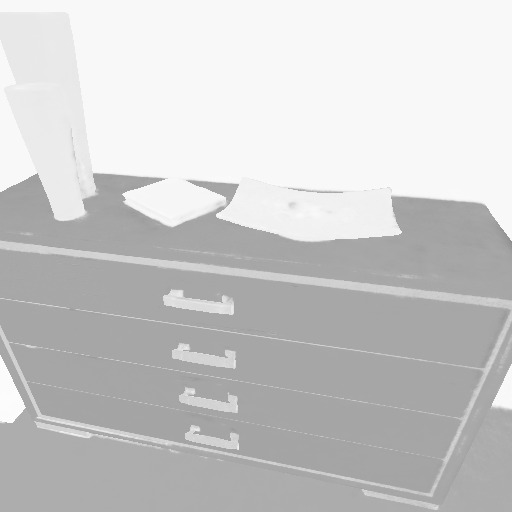} &
\includegraphics[width=\width]{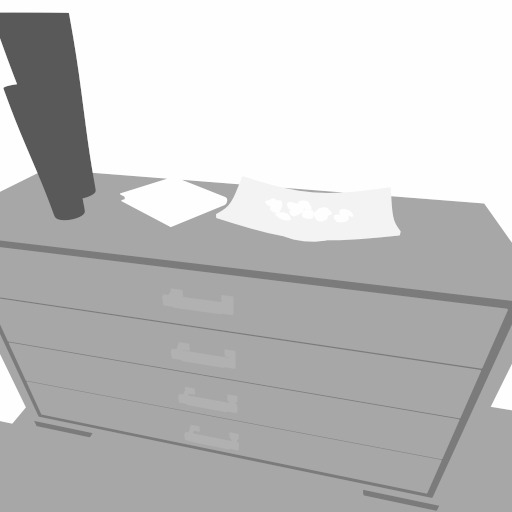} \\
{\makebox[5pt]{\rotatebox{90}{\tiny \hspace{0pt} Bedroom}}} &
\includegraphics[width=\width]{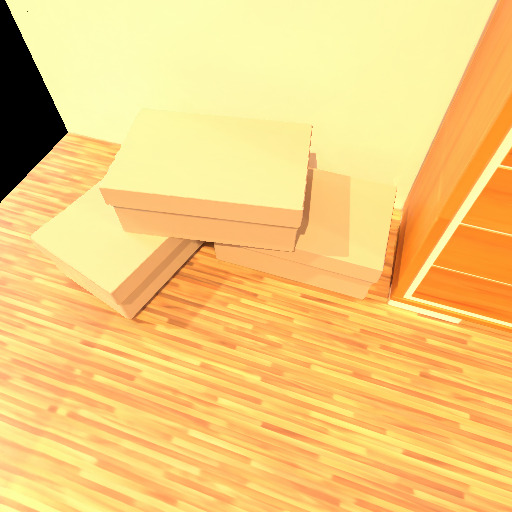} &
\includegraphics[width=\width]{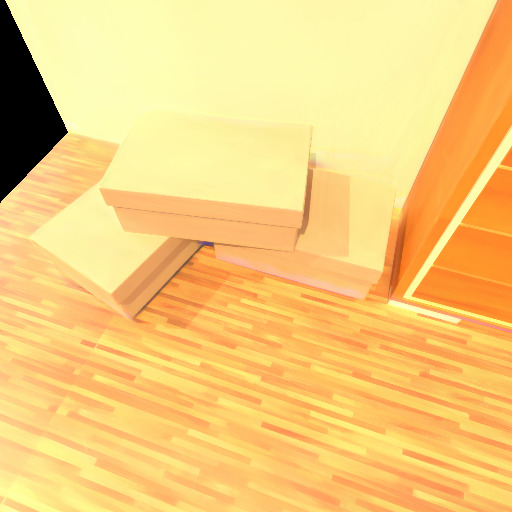} &
\includegraphics[width=\width]{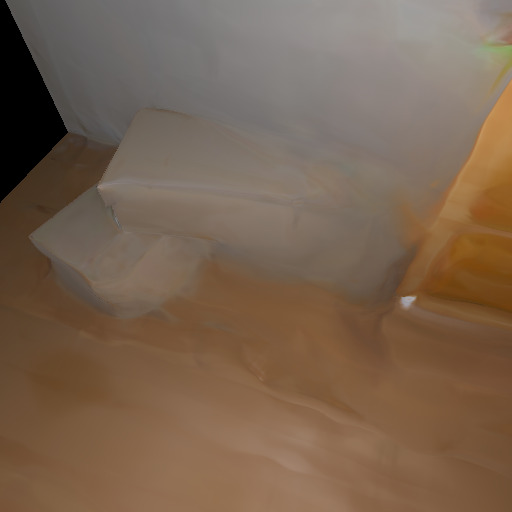} &
\includegraphics[width=\width]{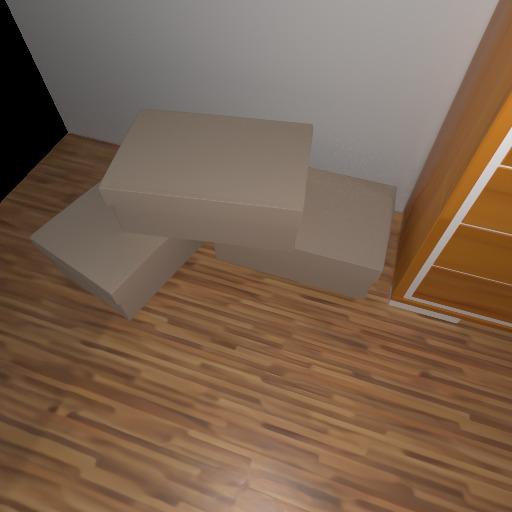} &
\includegraphics[width=\width]{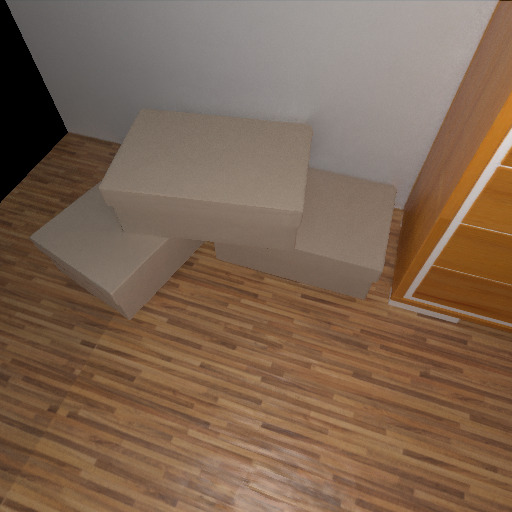} &
\includegraphics[width=\width]{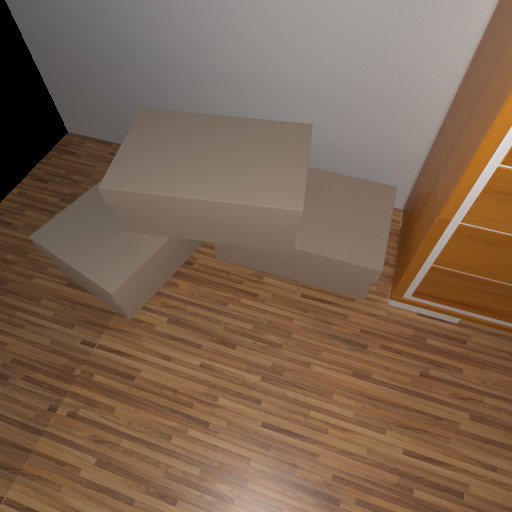} \\
{\makebox[5pt]{\rotatebox{90}{\tiny Albedo}}} &
\includegraphics[width=\width]{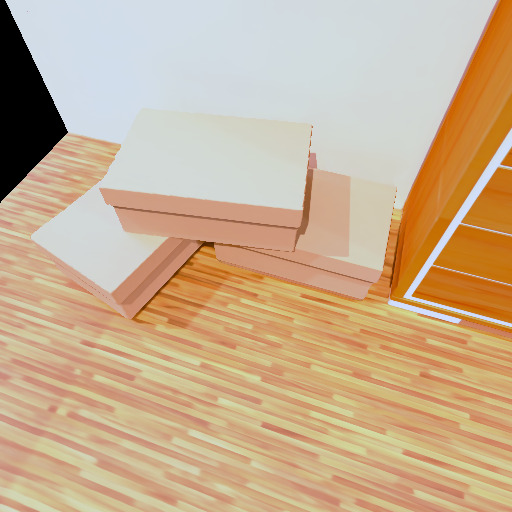} &
\includegraphics[width=\width]{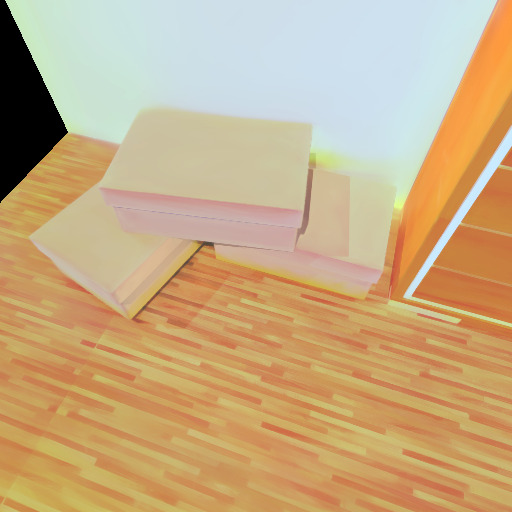} &
\includegraphics[width=\width]{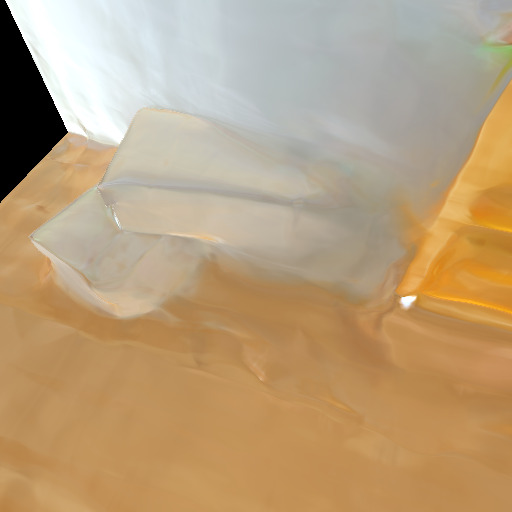} &
\includegraphics[width=\width]{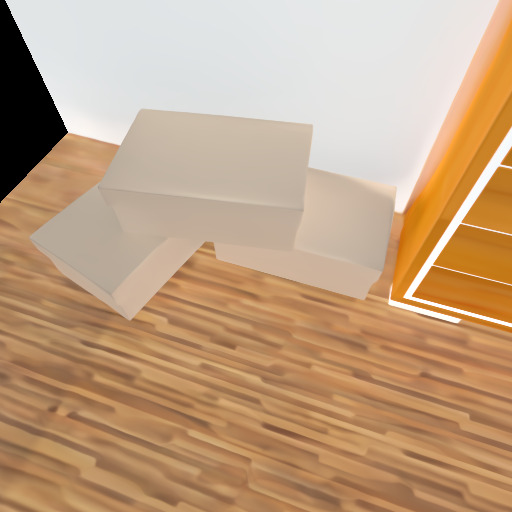} &
\includegraphics[width=\width]{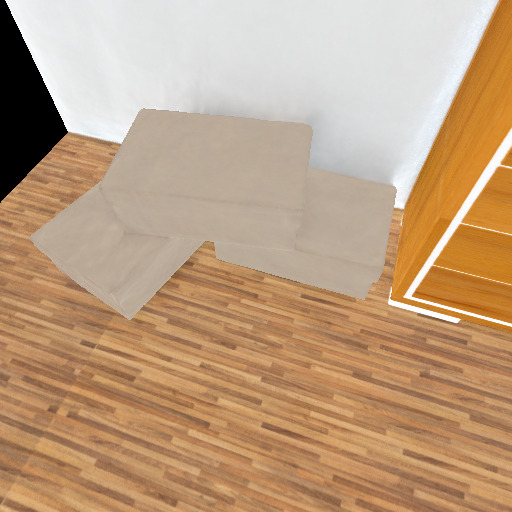} &
\includegraphics[width=\width]{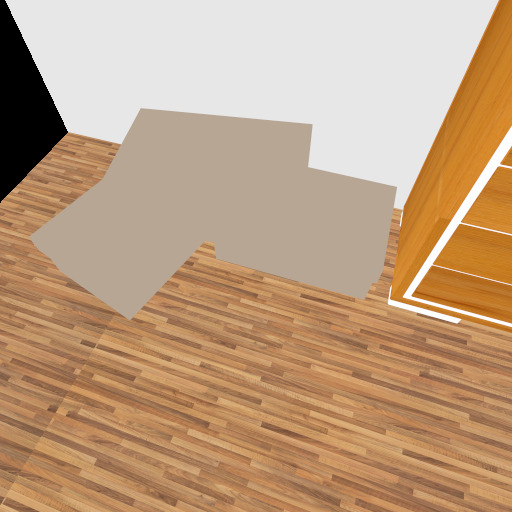} \\
{\makebox[5pt]{\rotatebox{90}{\tiny Roughness}}} &
\includegraphics[width=\width]{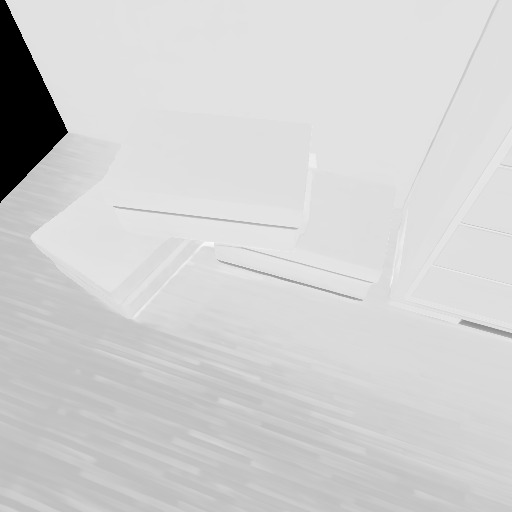} &
\includegraphics[width=\width]{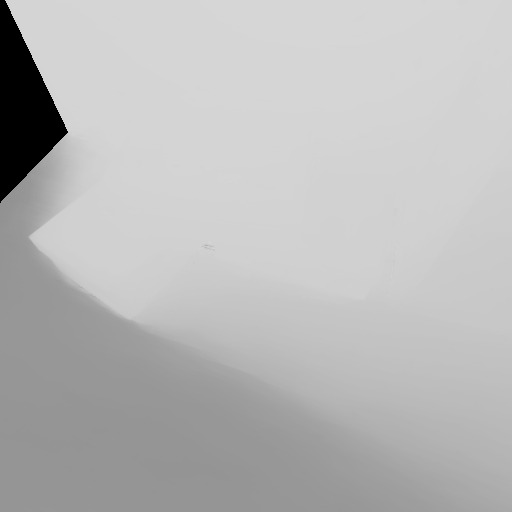} &
\includegraphics[width=\width]{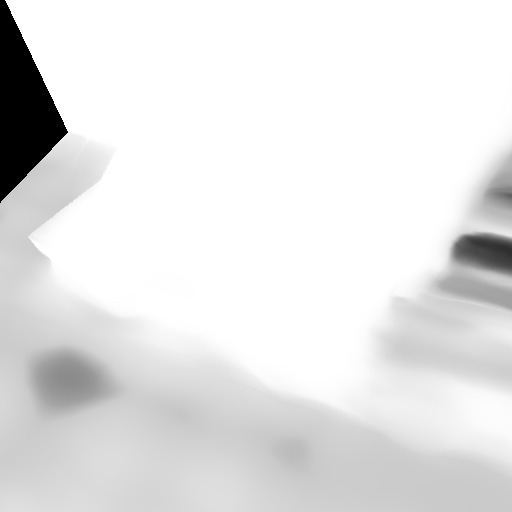} &
\includegraphics[width=\width]{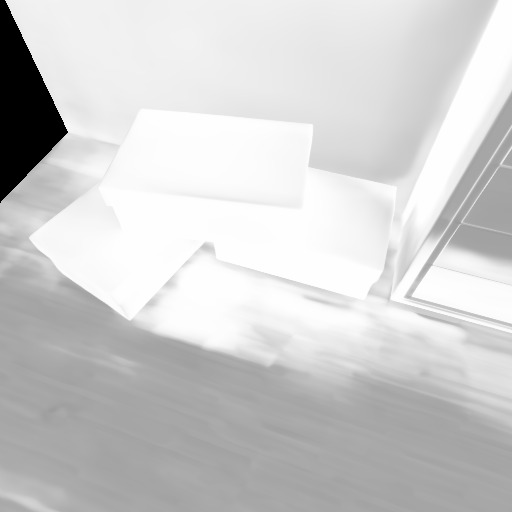} &
\includegraphics[width=\width]{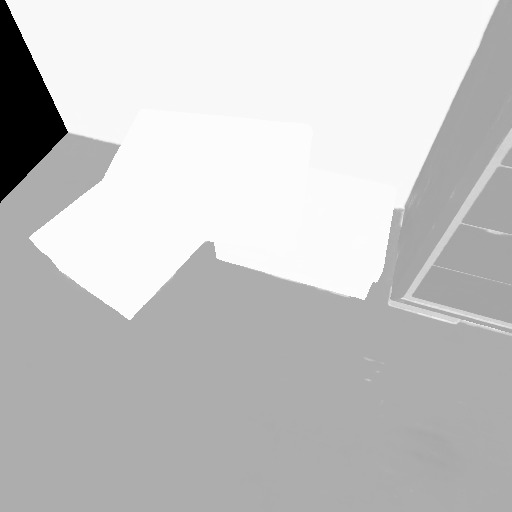} &
\includegraphics[width=\width]{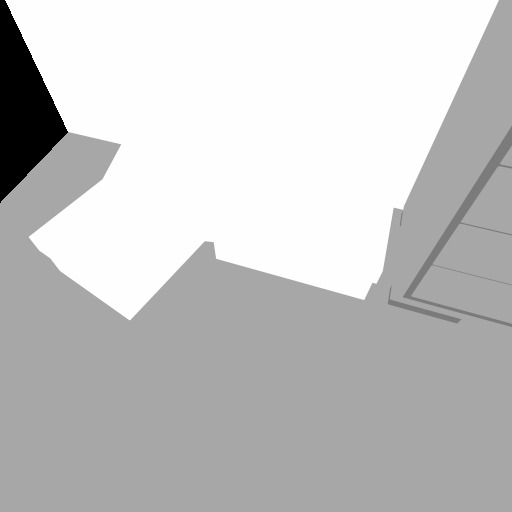} \\

%% file: generated/suppl_qualitative_generated_synthetic_1.tex
{\makebox[5pt]{\rotatebox{90}{\tiny \hspace{4pt} Shelf}}} &
\includegraphics[width=\width]{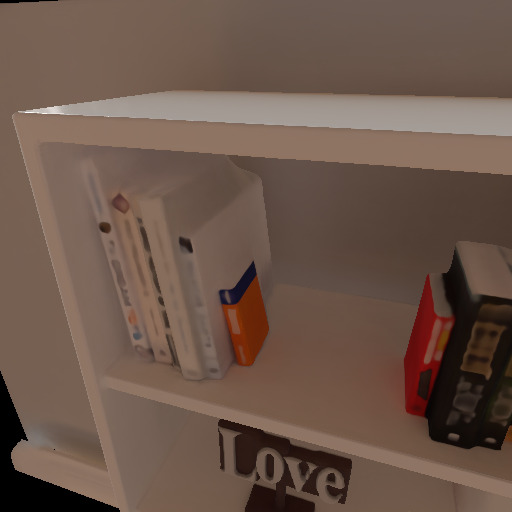} &
\includegraphics[width=\width]{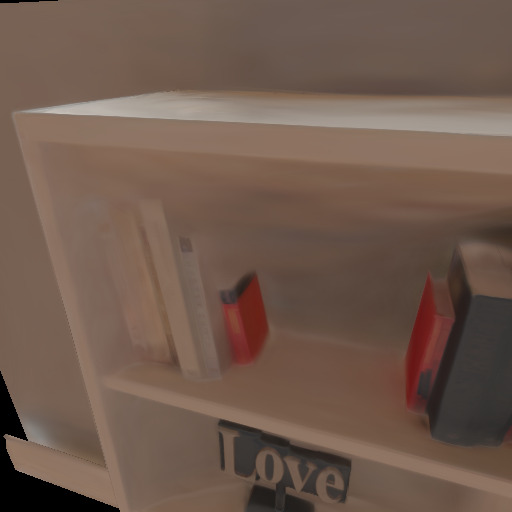} &
\includegraphics[width=\width]{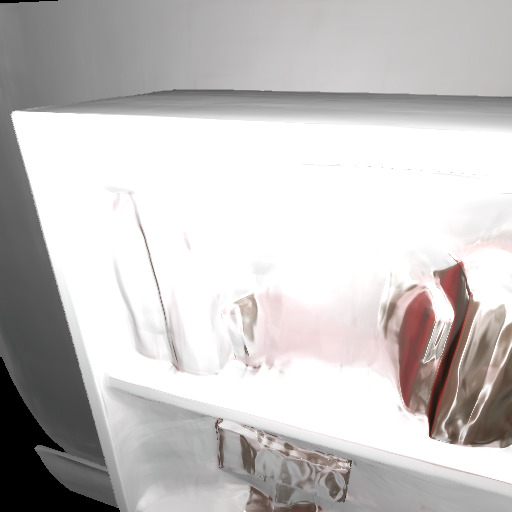} &
\includegraphics[width=\width]{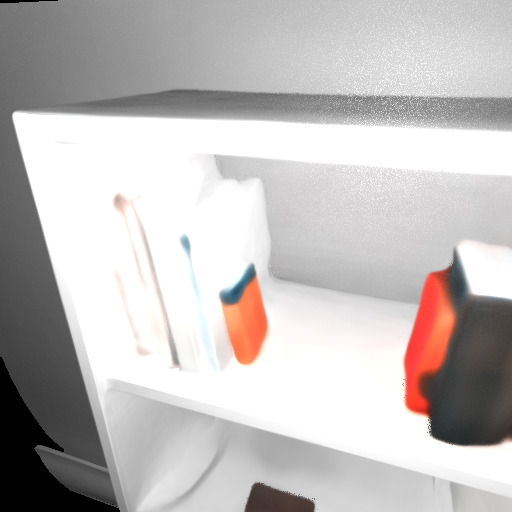} &
\includegraphics[width=\width]{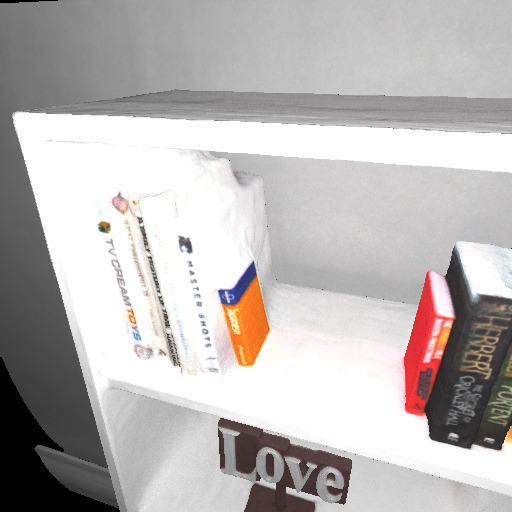} &
\includegraphics[width=\width]{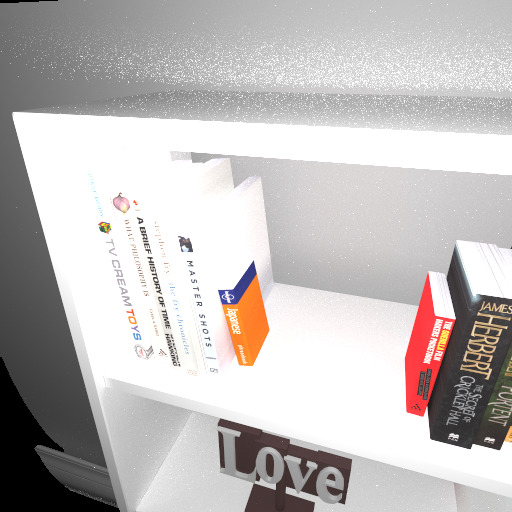} \\
{\makebox[5pt]{\rotatebox{90}{\tiny Albedo}}} &
\includegraphics[width=\width]{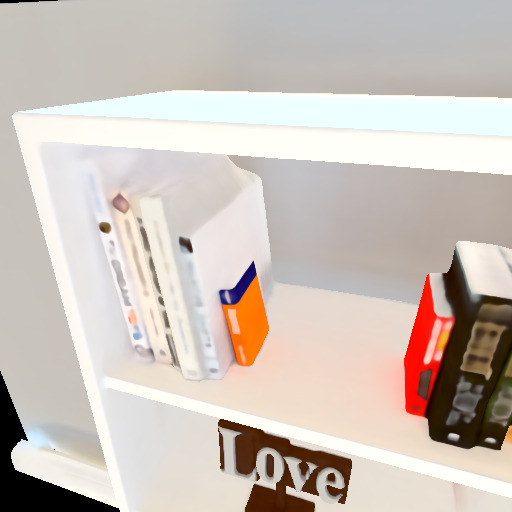} &
\includegraphics[width=\width]{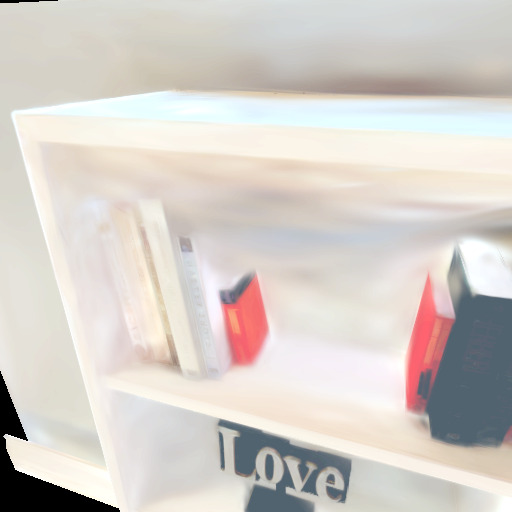} &
\includegraphics[width=\width]{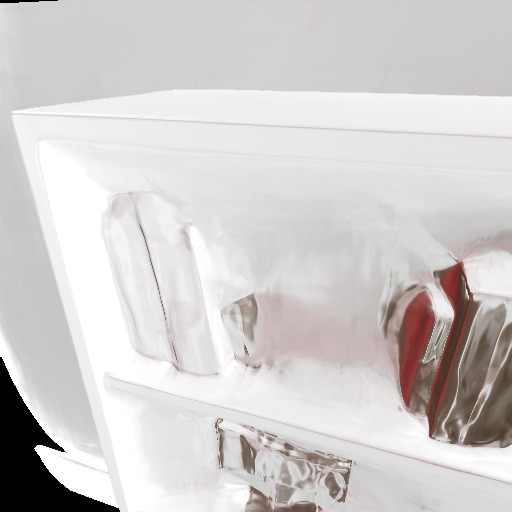} &
\includegraphics[width=\width]{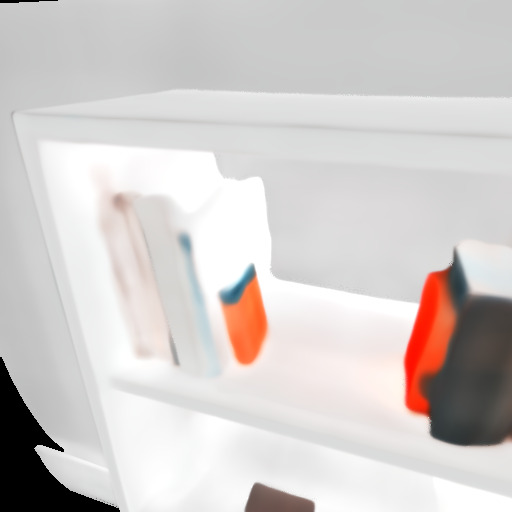} &
\includegraphics[width=\width]{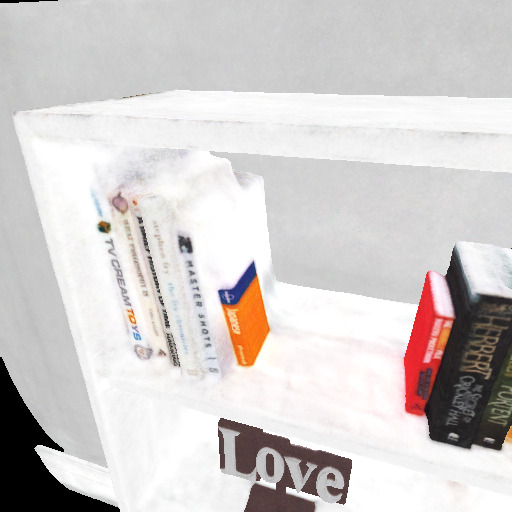} &
\includegraphics[width=\width]{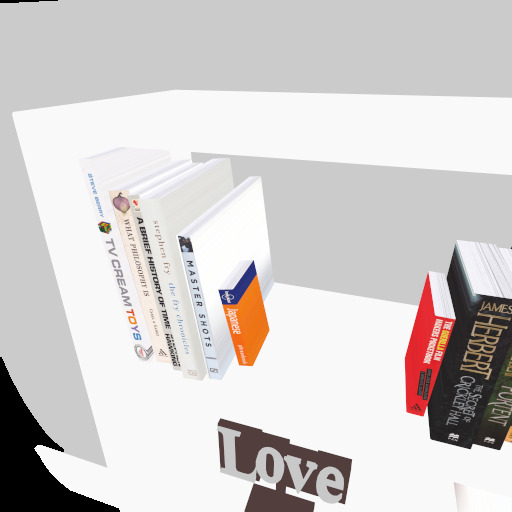} \\
{\makebox[5pt]{\rotatebox{90}{\tiny Roughness}}} &
\includegraphics[width=\width]{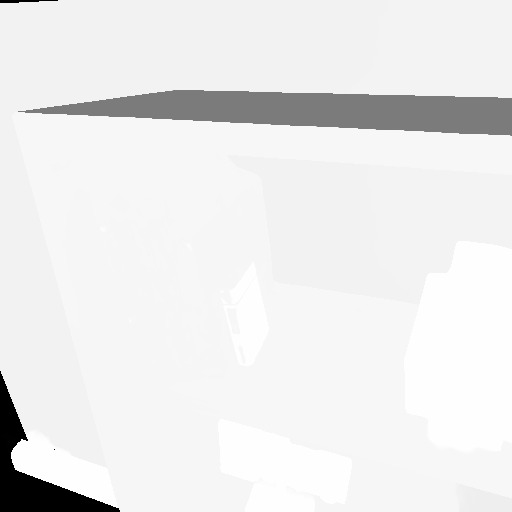} &
\includegraphics[width=\width]{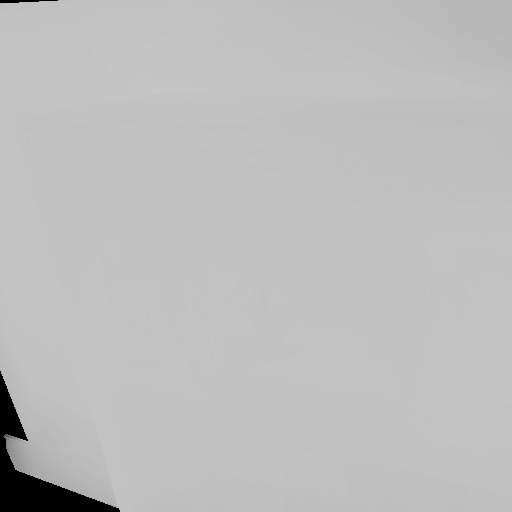} &
\includegraphics[width=\width]{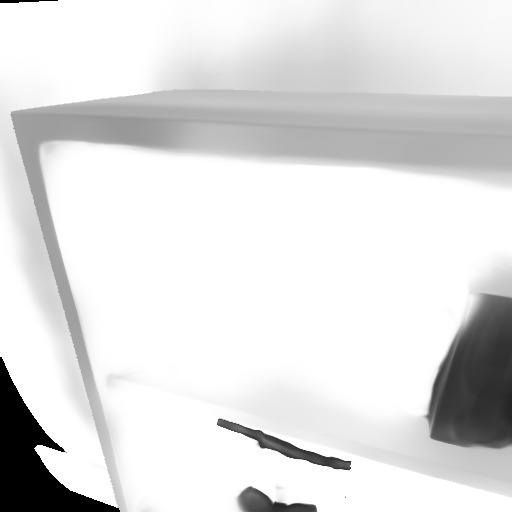} &
\includegraphics[width=\width]{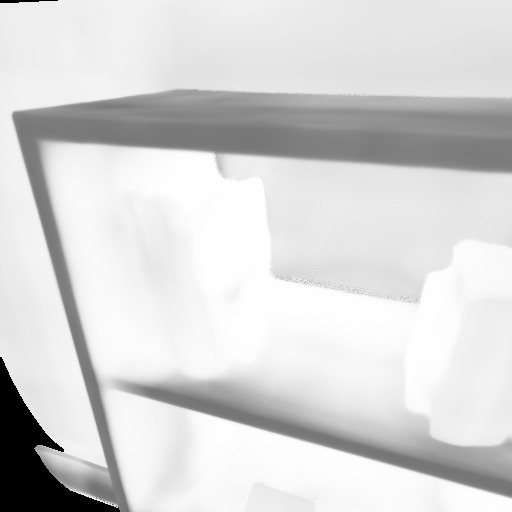} &
\includegraphics[width=\width]{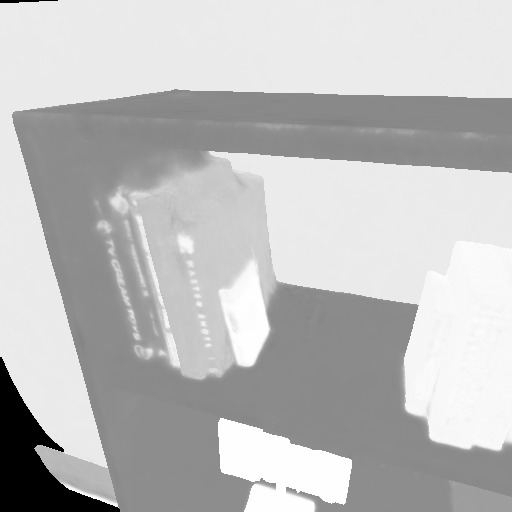} &
\includegraphics[width=\width]{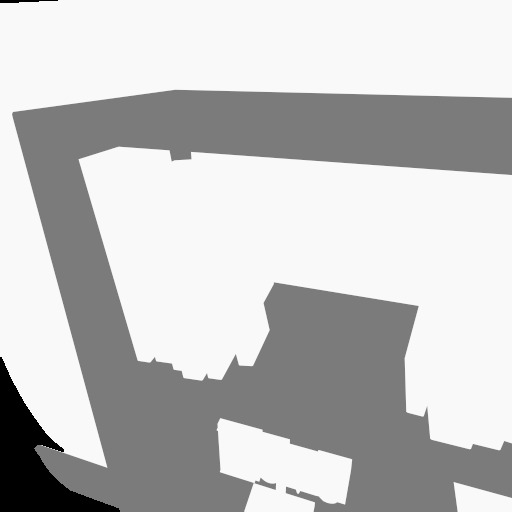} \\
{\makebox[5pt]{\rotatebox{90}{\tiny \hspace{4pt} Shelf}}} &
\includegraphics[width=\width]{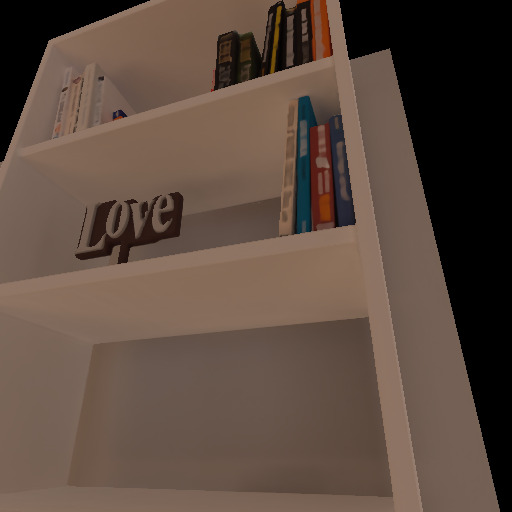} &
\includegraphics[width=\width]{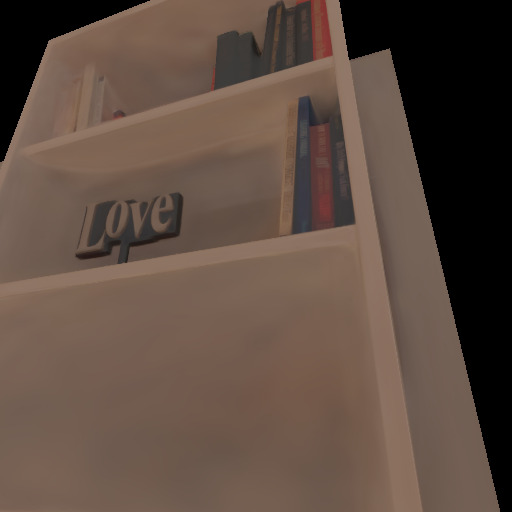} &
\includegraphics[width=\width]{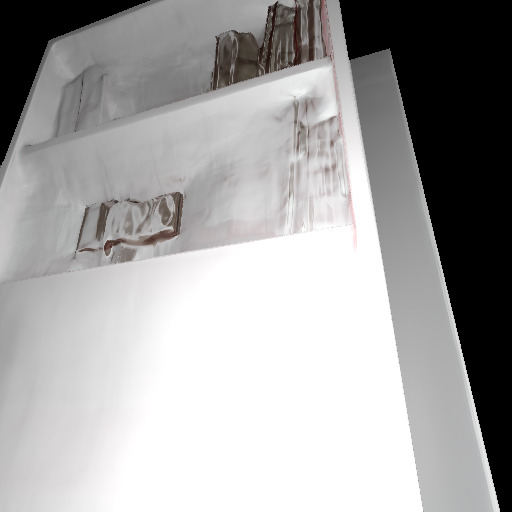} &
\includegraphics[width=\width]{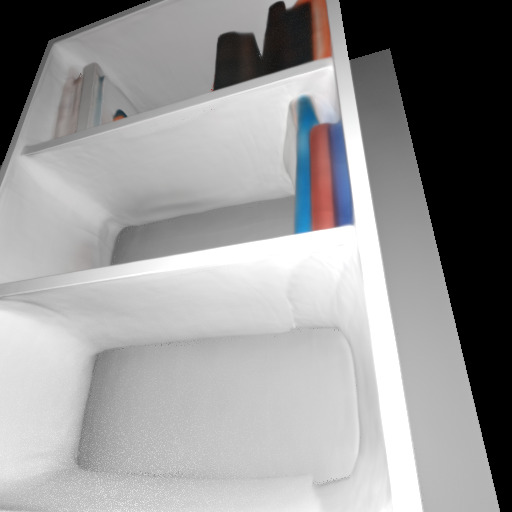} &
\includegraphics[width=\width]{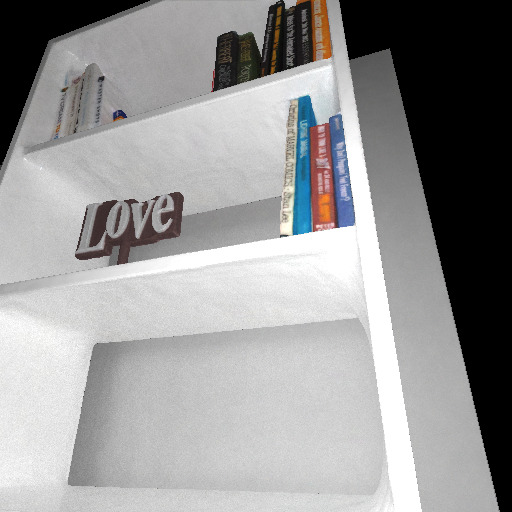} &
\includegraphics[width=\width]{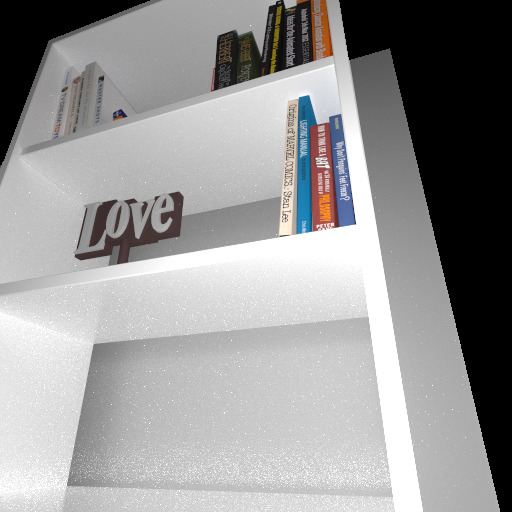} \\
{\makebox[5pt]{\rotatebox{90}{\tiny Albedo}}} &
\includegraphics[width=\width]{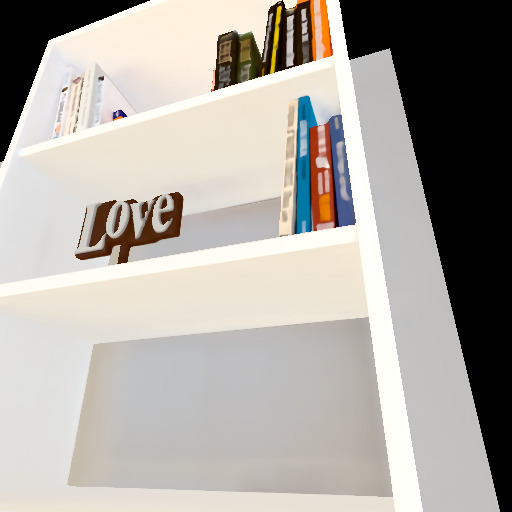} &
\includegraphics[width=\width]{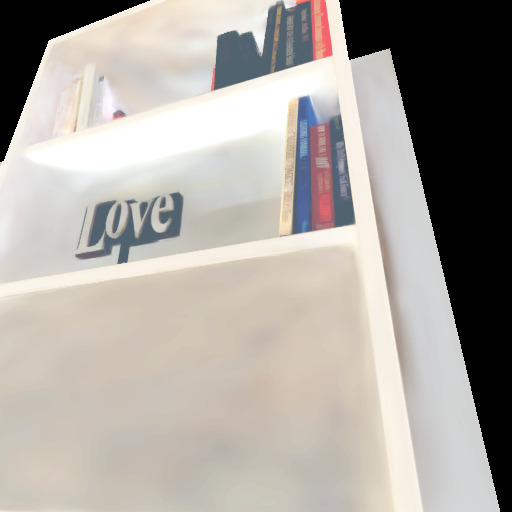} &
\includegraphics[width=\width]{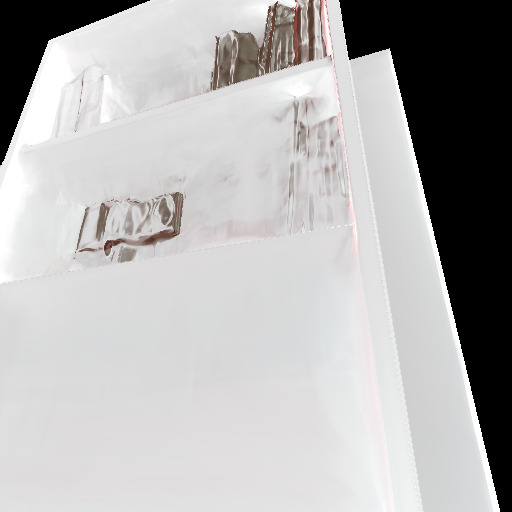} &
\includegraphics[width=\width]{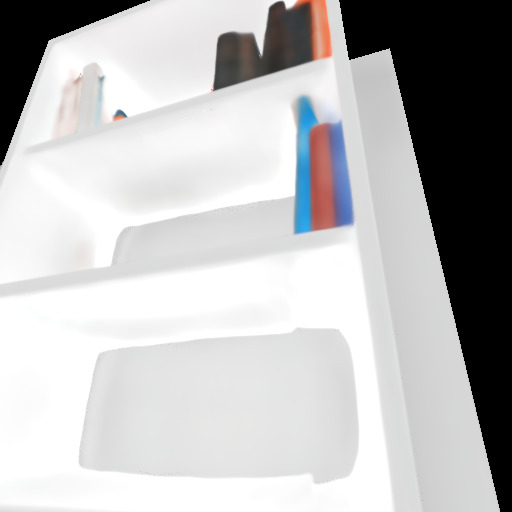} &
\includegraphics[width=\width]{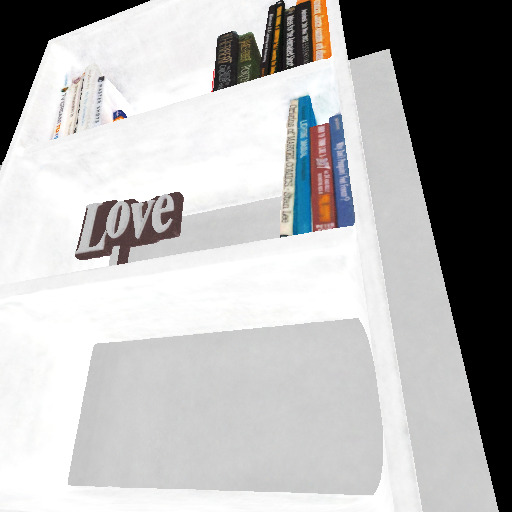} &
\includegraphics[width=\width]{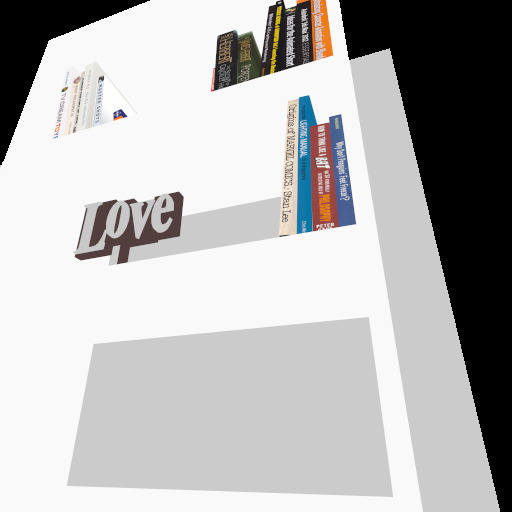} \\
{\makebox[5pt]{\rotatebox{90}{\tiny Roughness}}} &
\includegraphics[width=\width]{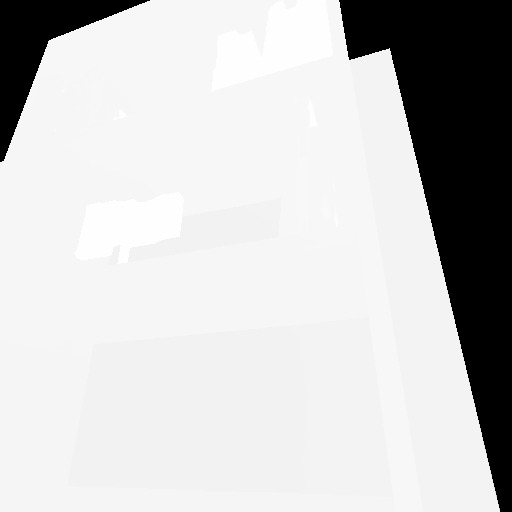} &
\includegraphics[width=\width]{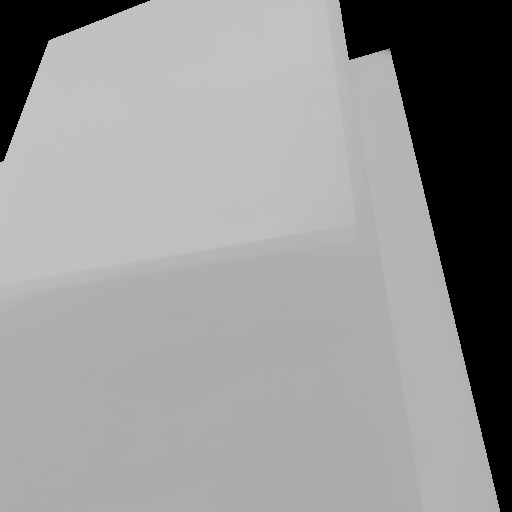} &
\includegraphics[width=\width]{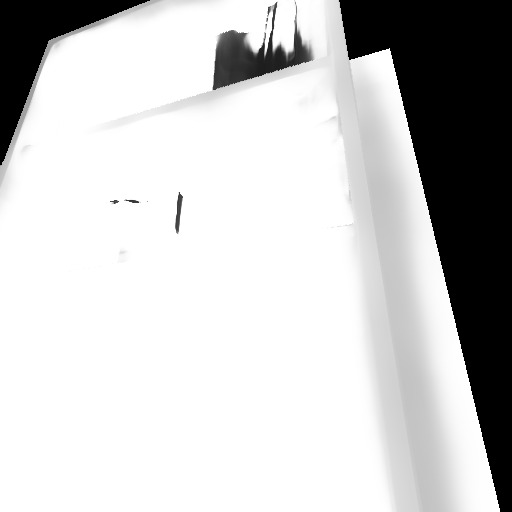} &
\includegraphics[width=\width]{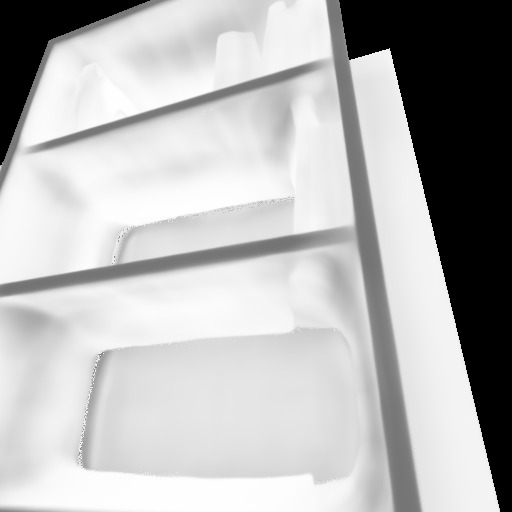} &
\includegraphics[width=\width]{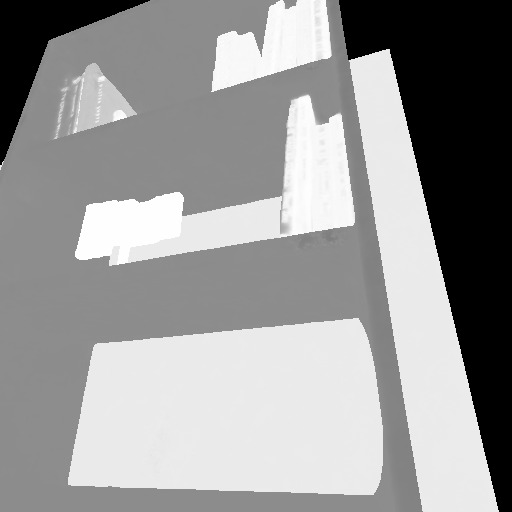} &
\includegraphics[width=\width]{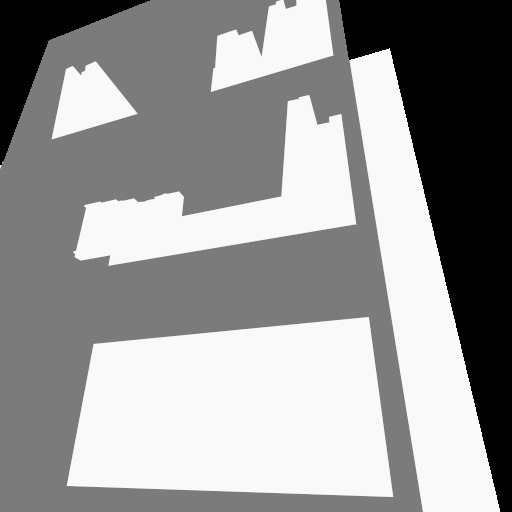} \\

%% file: generated/suppl_qualitative_generated_synthetic_2.tex
{\makebox[5pt]{\rotatebox{90}{\tiny \hspace{0pt} Counter}}} &
\includegraphics[width=\width]{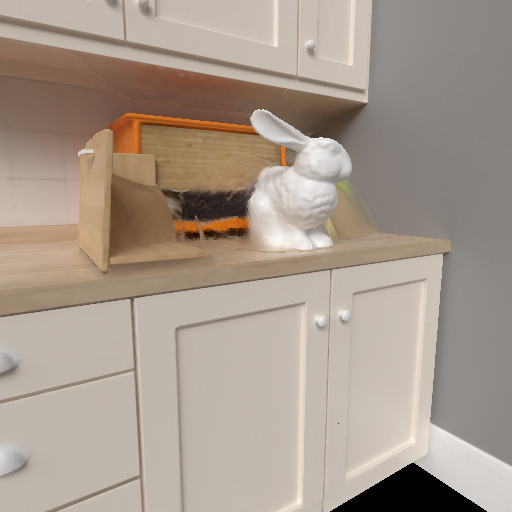} &
\includegraphics[width=\width]{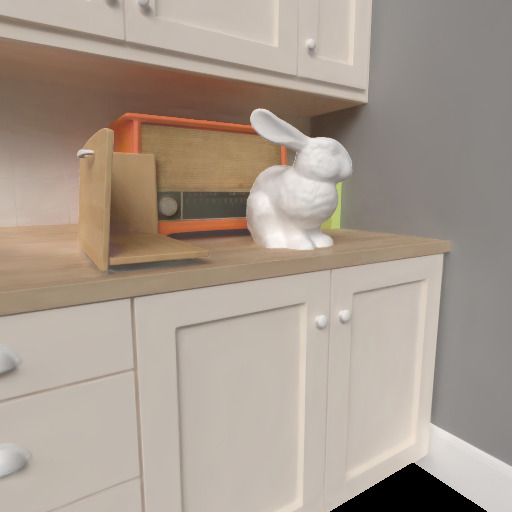} &
\includegraphics[width=\width]{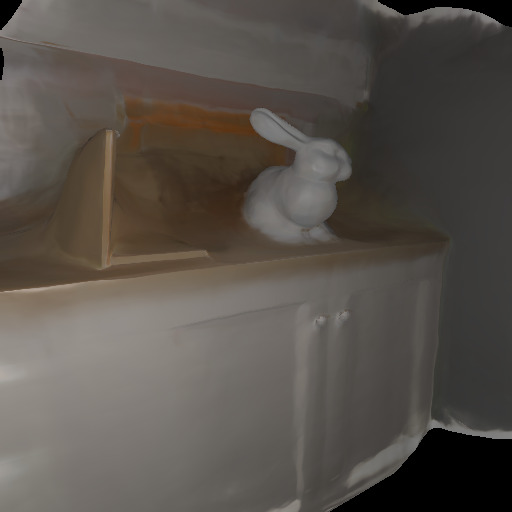} &
\includegraphics[width=\width]{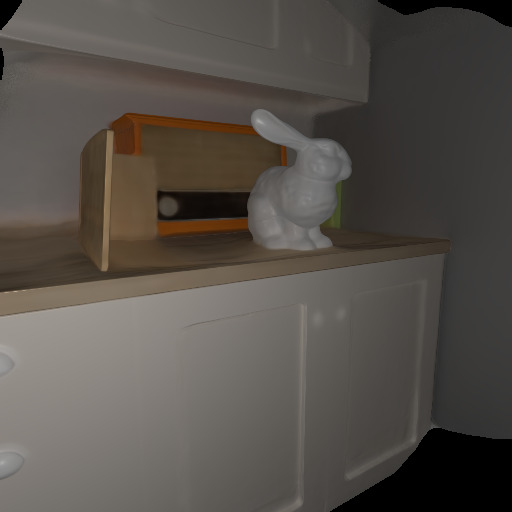} &
\includegraphics[width=\width]{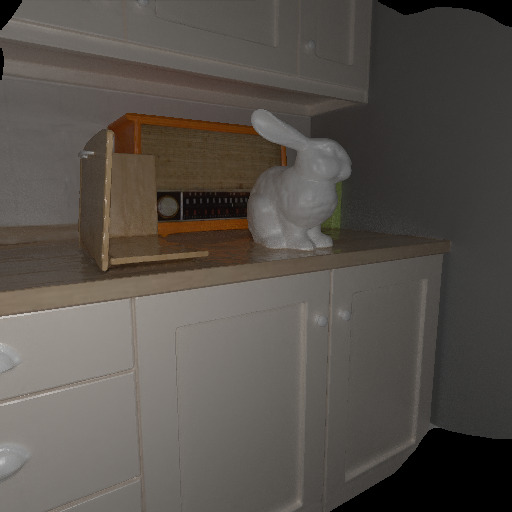} &
\includegraphics[width=\width]{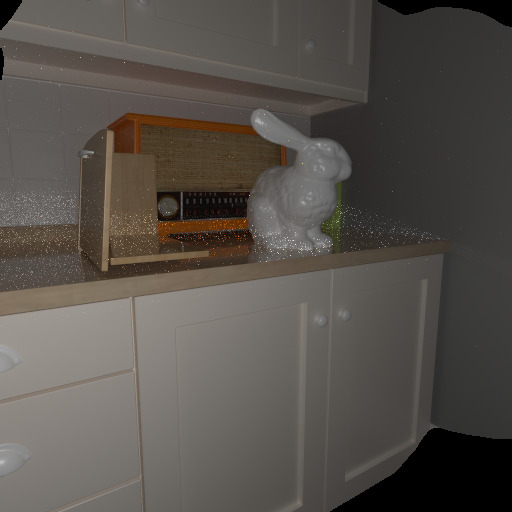} \\
{\makebox[5pt]{\rotatebox{90}{\tiny Albedo}}} &
\includegraphics[width=\width]{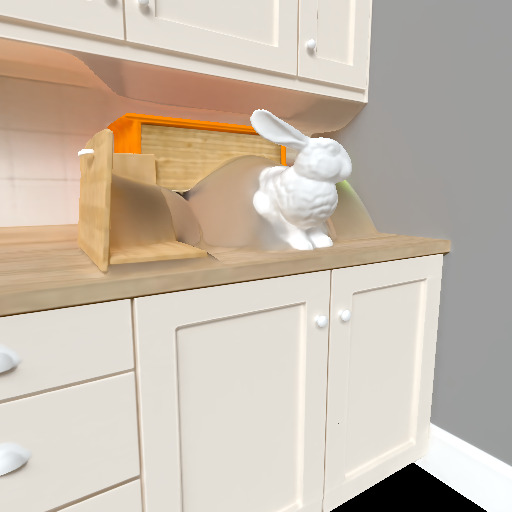} &
\includegraphics[width=\width]{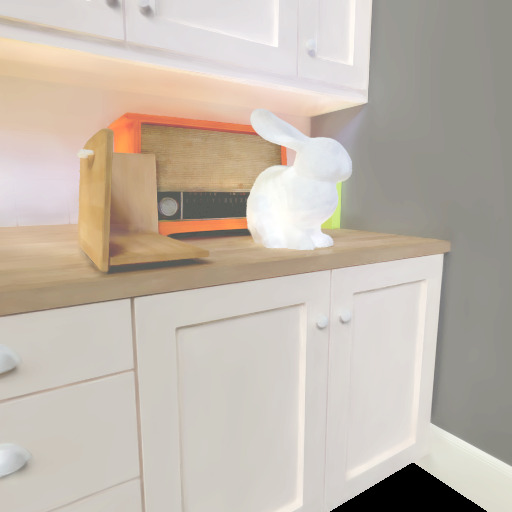} &
\includegraphics[width=\width]{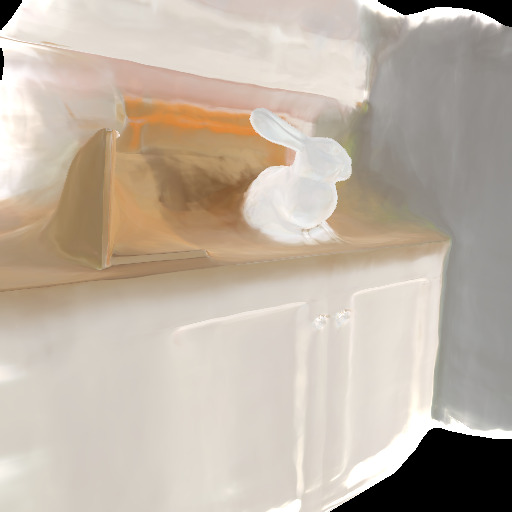} &
\includegraphics[width=\width]{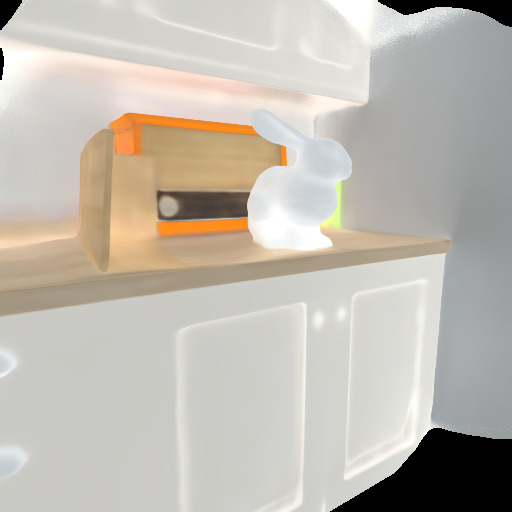} &
\includegraphics[width=\width]{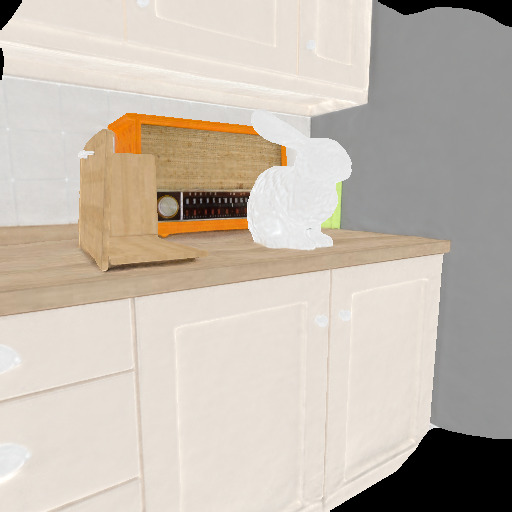} &
\includegraphics[width=\width]{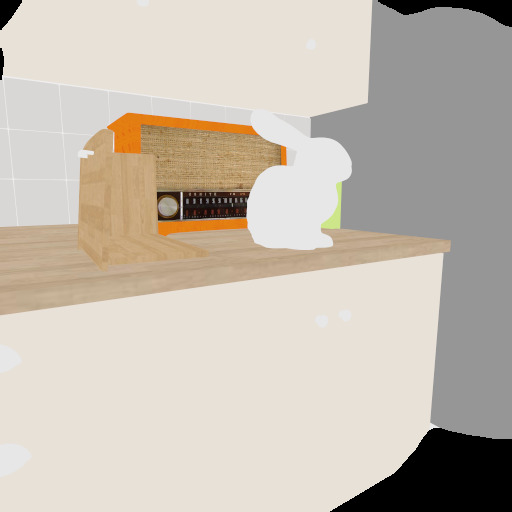} \\
{\makebox[5pt]{\rotatebox{90}{\tiny Roughness}}} &
\includegraphics[width=\width]{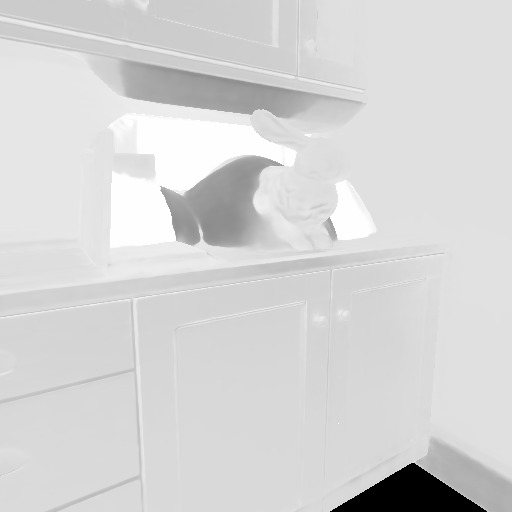} &
\includegraphics[width=\width]{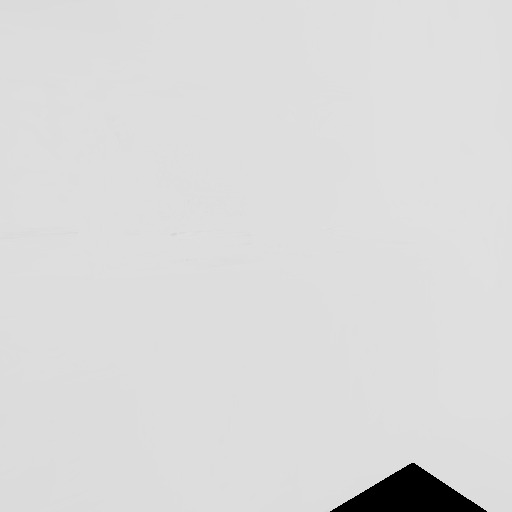} &
\includegraphics[width=\width]{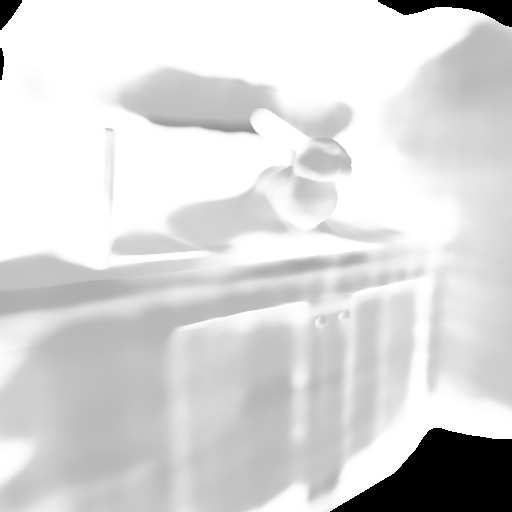} &
\includegraphics[width=\width]{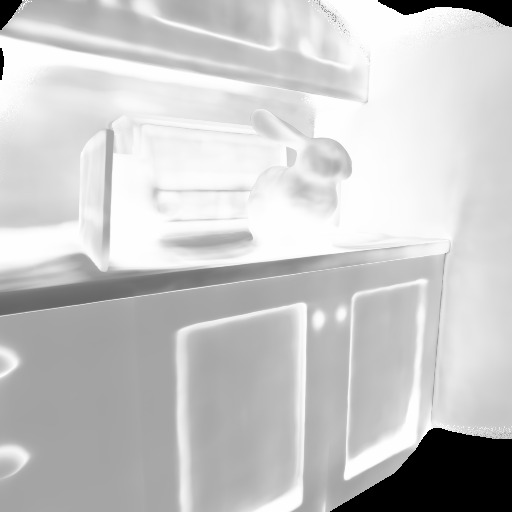} &
\includegraphics[width=\width]{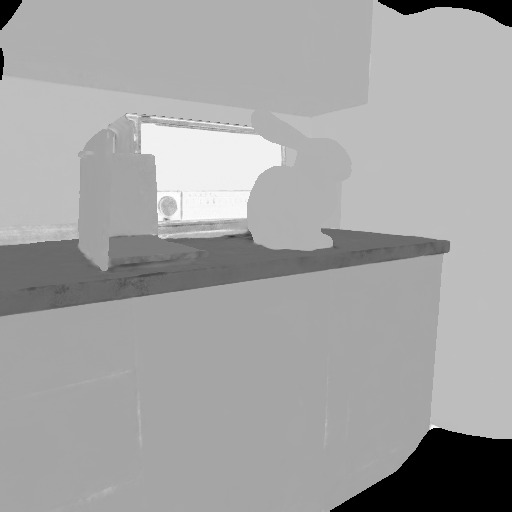} &
\includegraphics[width=\width]{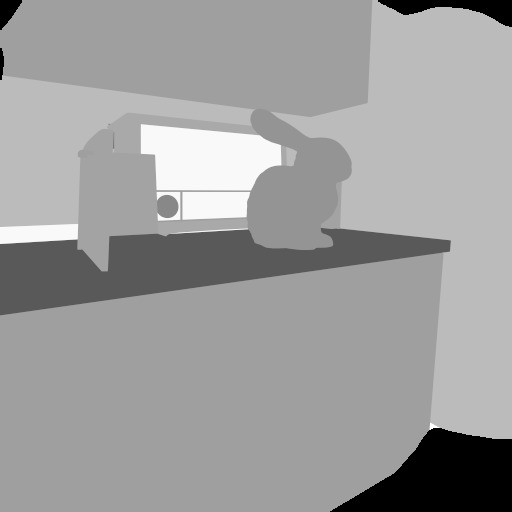} \\
{\makebox[5pt]{\rotatebox{90}{\tiny \hspace{0pt} Counter}}} &
\includegraphics[width=\width]{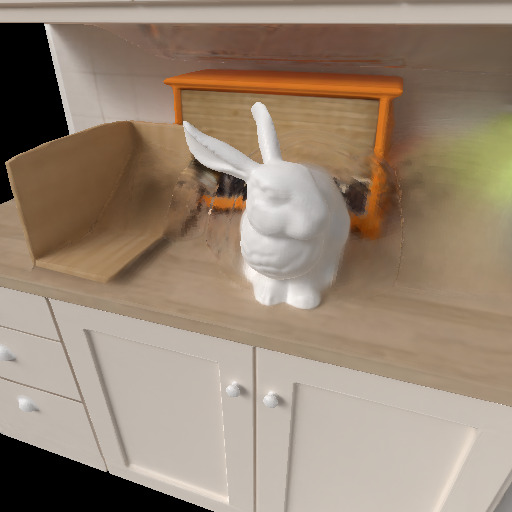} &
\includegraphics[width=\width]{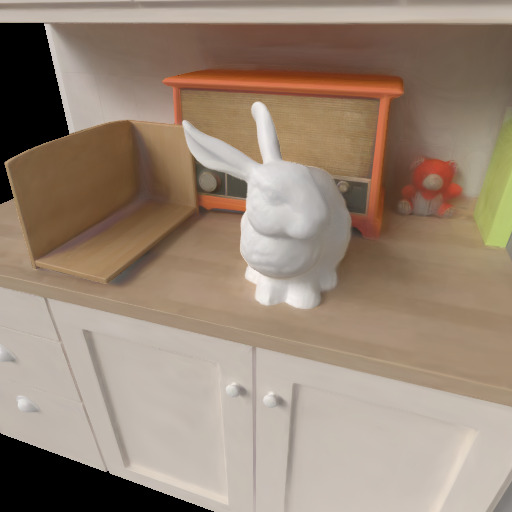} &
\includegraphics[width=\width]{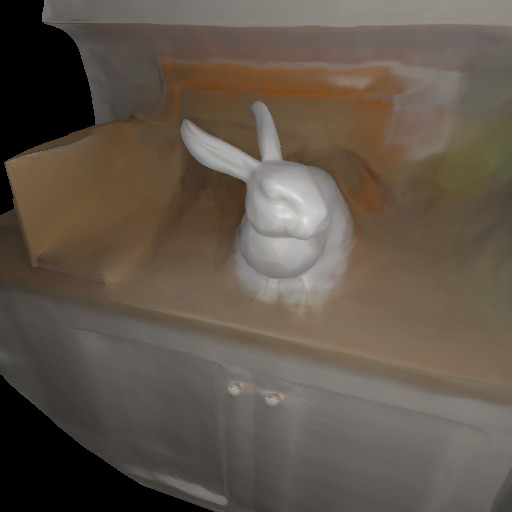} &
\includegraphics[width=\width]{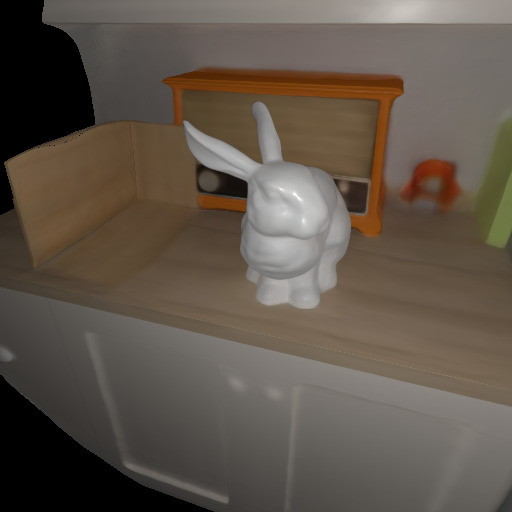} &
\includegraphics[width=\width]{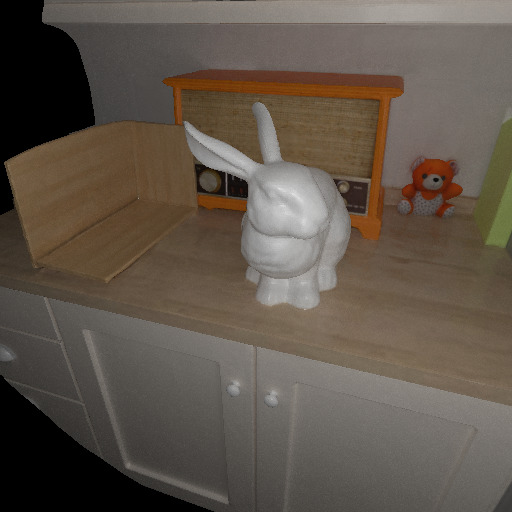} &
\includegraphics[width=\width]{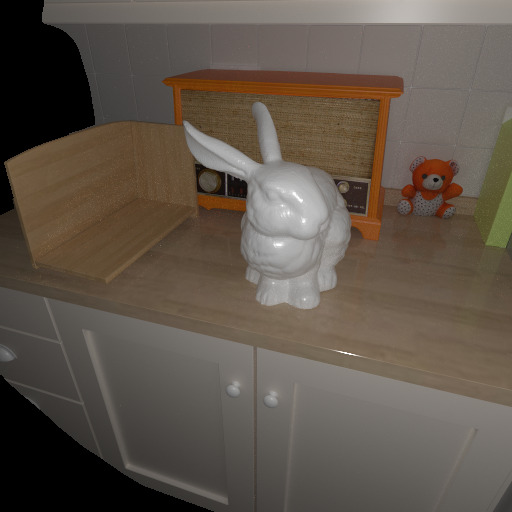} \\
{\makebox[5pt]{\rotatebox{90}{\tiny Albedo}}} &
\includegraphics[width=\width]{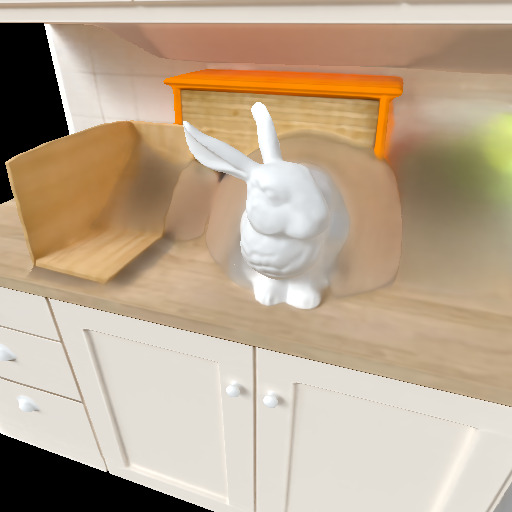} &
\includegraphics[width=\width]{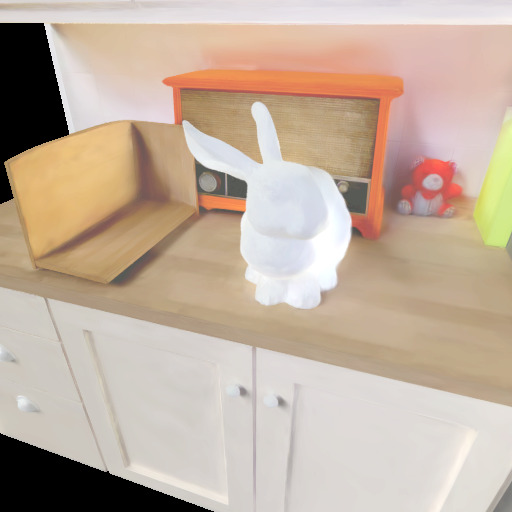} &
\includegraphics[width=\width]{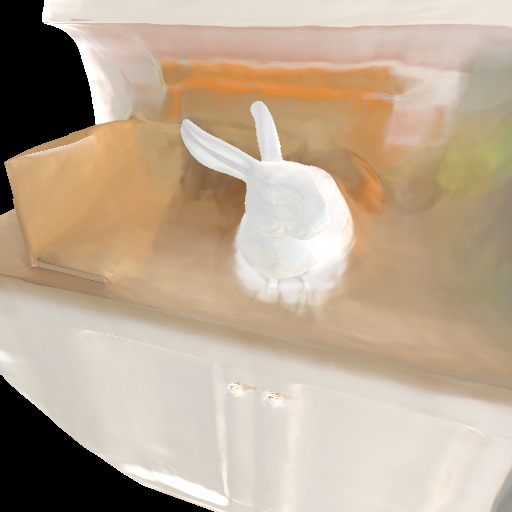} &
\includegraphics[width=\width]{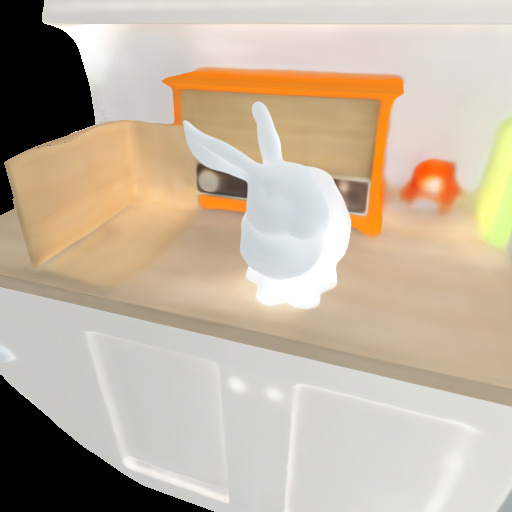} &
\includegraphics[width=\width]{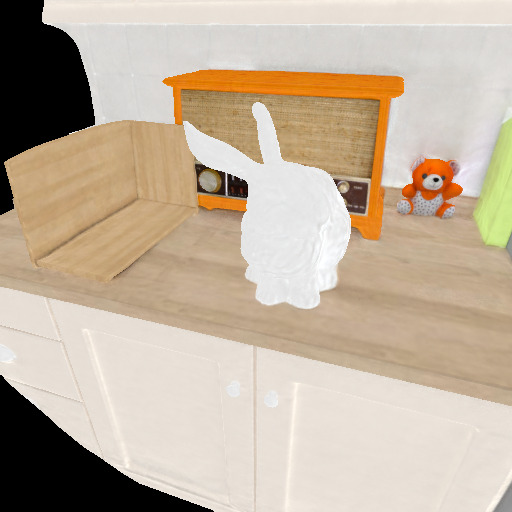} &
\includegraphics[width=\width]{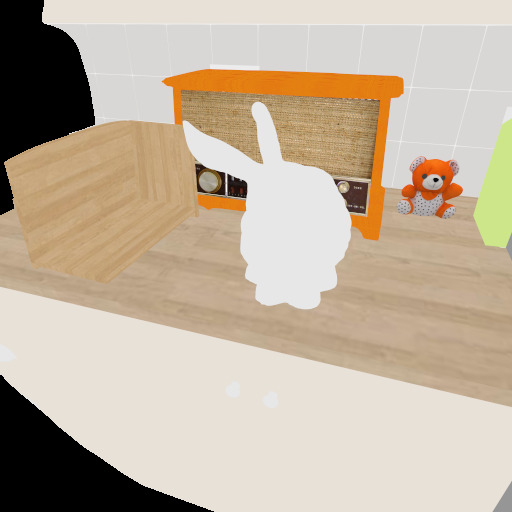} \\
{\makebox[5pt]{\rotatebox{90}{\tiny Roughness}}} &
\includegraphics[width=\width]{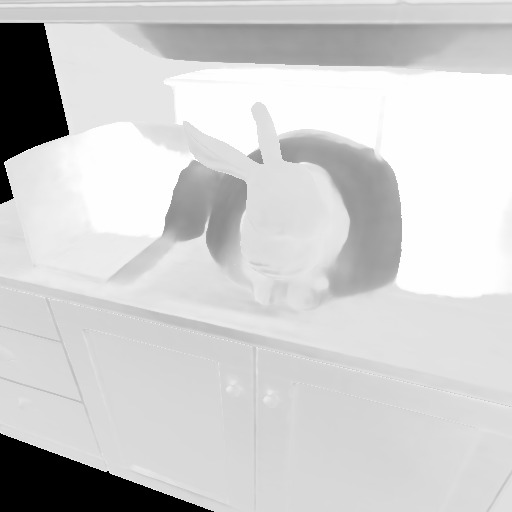} &
\includegraphics[width=\width]{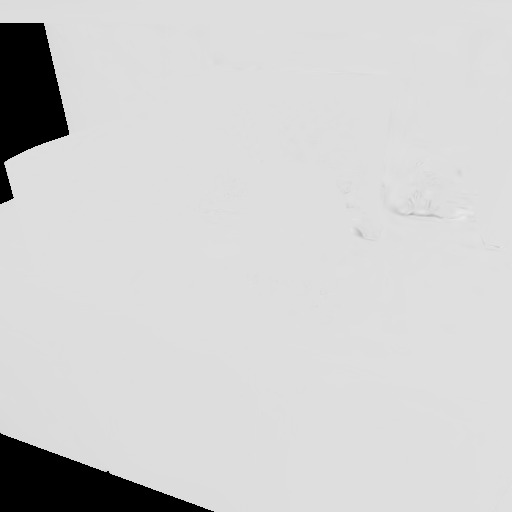} &
\includegraphics[width=\width]{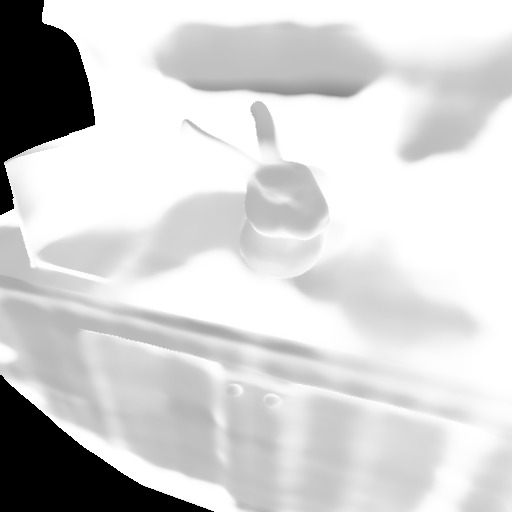} &
\includegraphics[width=\width]{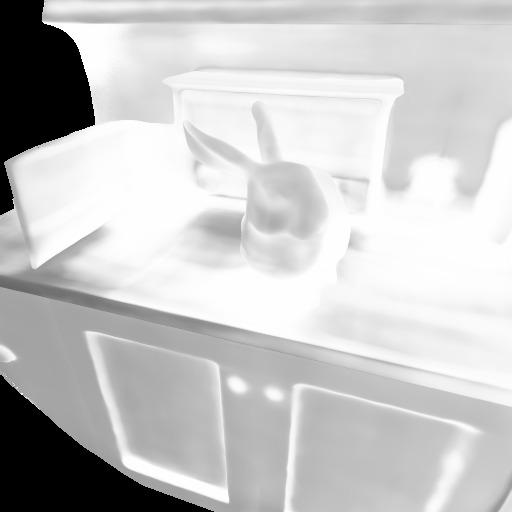} &
\includegraphics[width=\width]{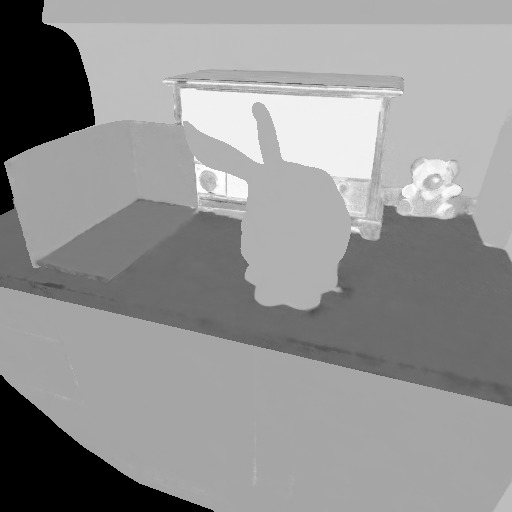} &
\includegraphics[width=\width]{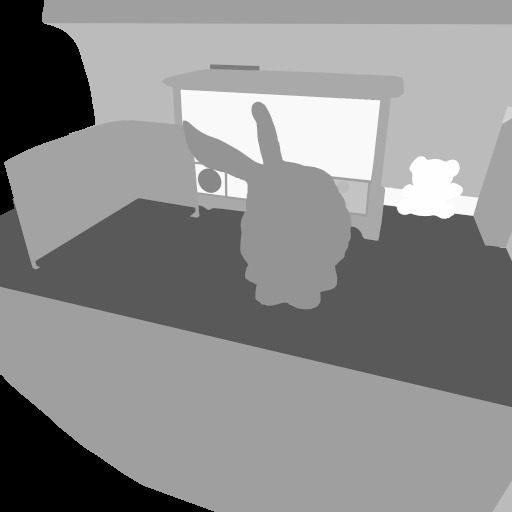} \\

%% file: generated/suppl_qualitative_generated_real_0.tex
{\makebox[5pt]{\rotatebox{90}{\tiny \hspace{0pt} Shoe Rack}}} &
\includegraphics[width=\width]{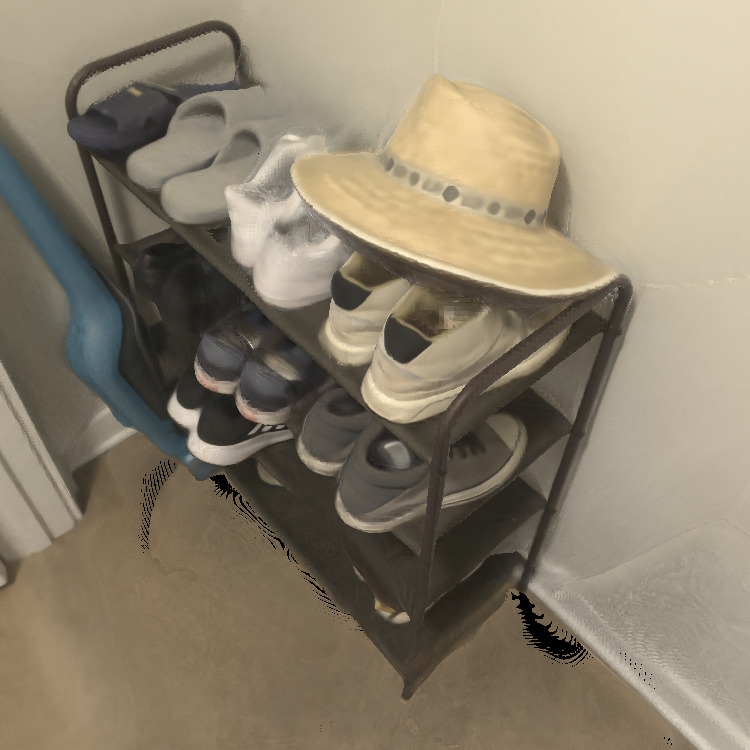} &
\includegraphics[width=\width]{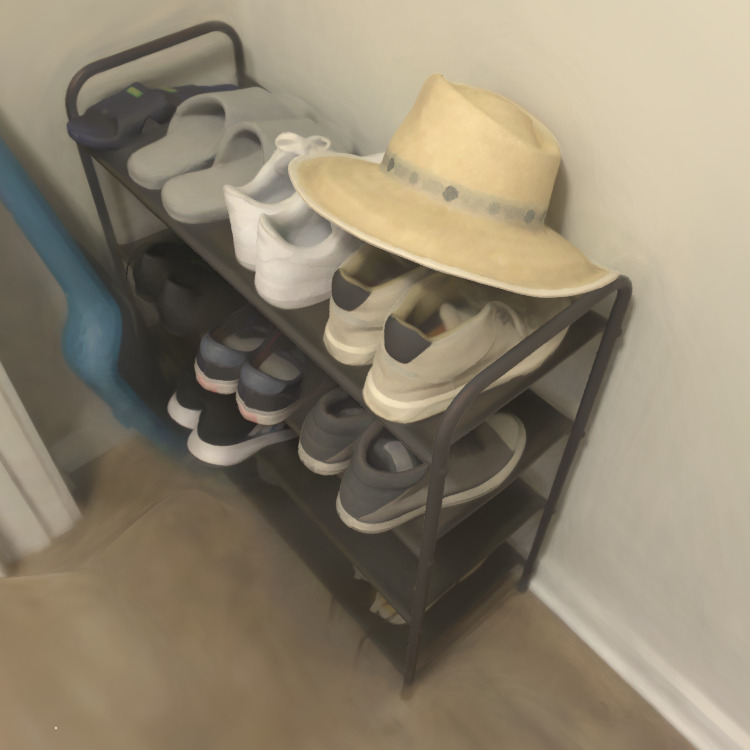} &
\includegraphics[width=\width]{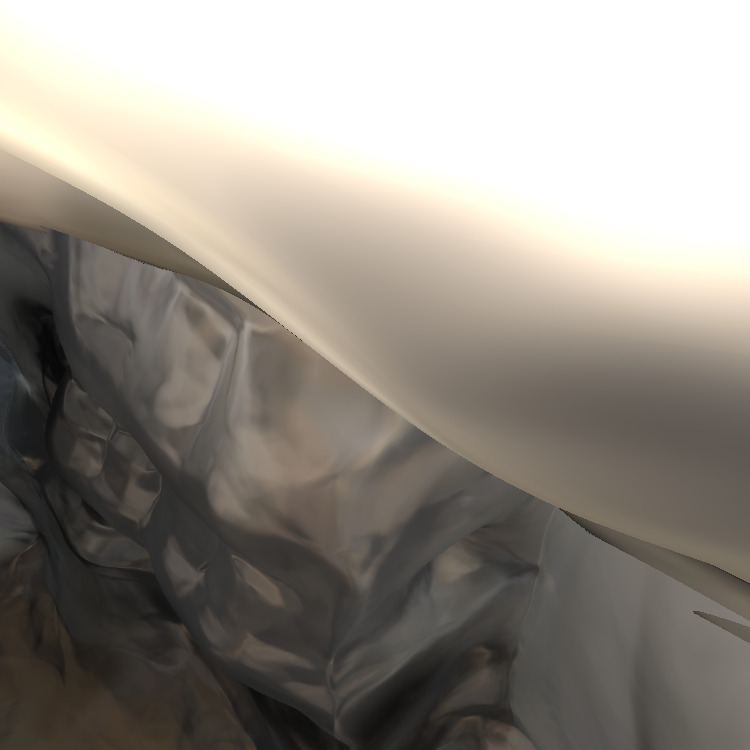} &
\includegraphics[width=\width]{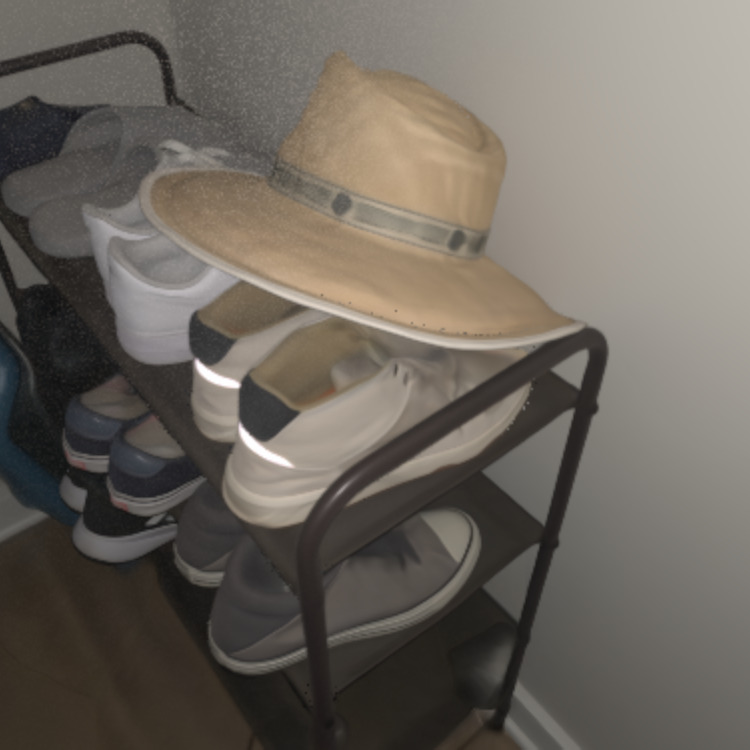} &
\includegraphics[width=\width]{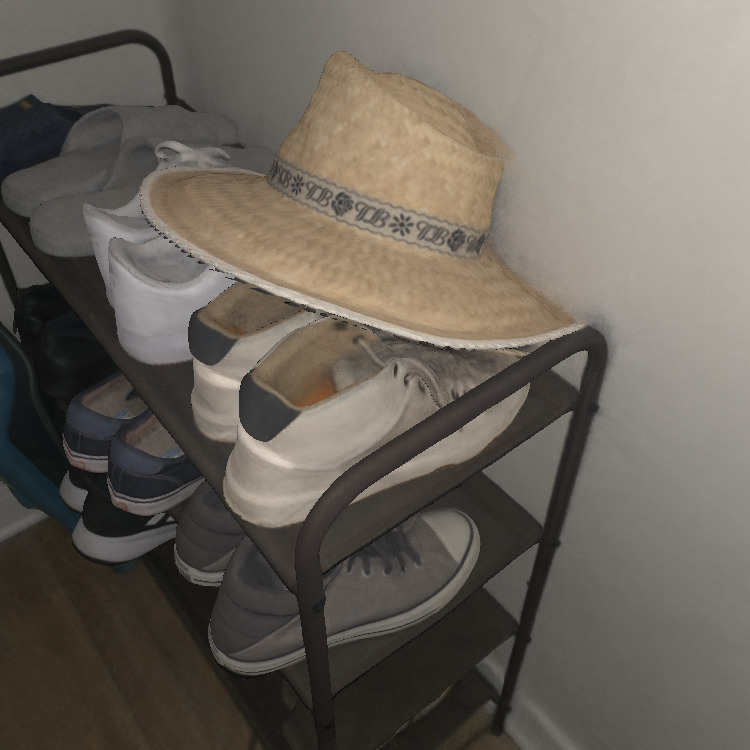} &
\includegraphics[width=\width]{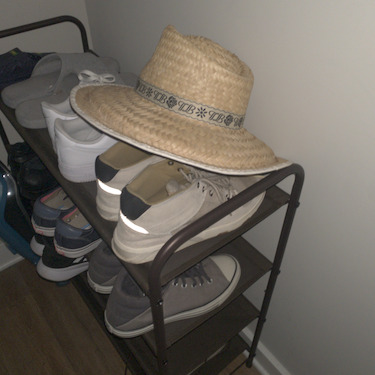} \\
{\makebox[5pt]{\rotatebox{90}{\tiny Albedo}}} &
\includegraphics[width=\width]{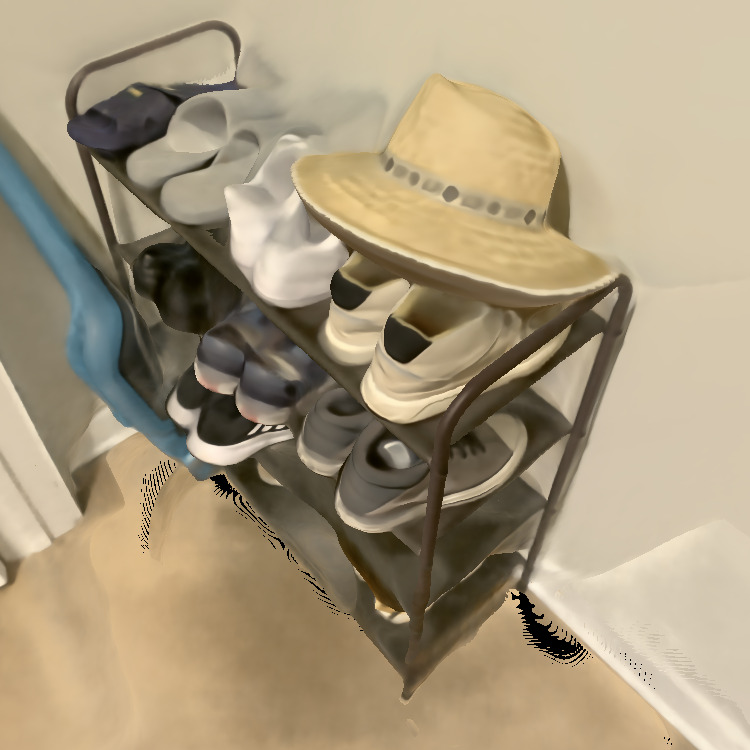} &
\includegraphics[width=\width]{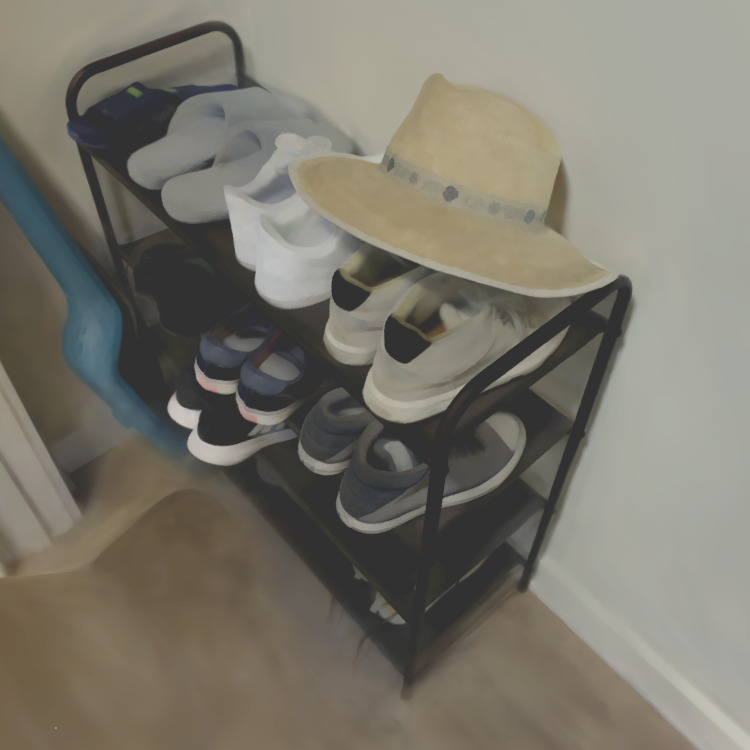} &
\includegraphics[width=\width]{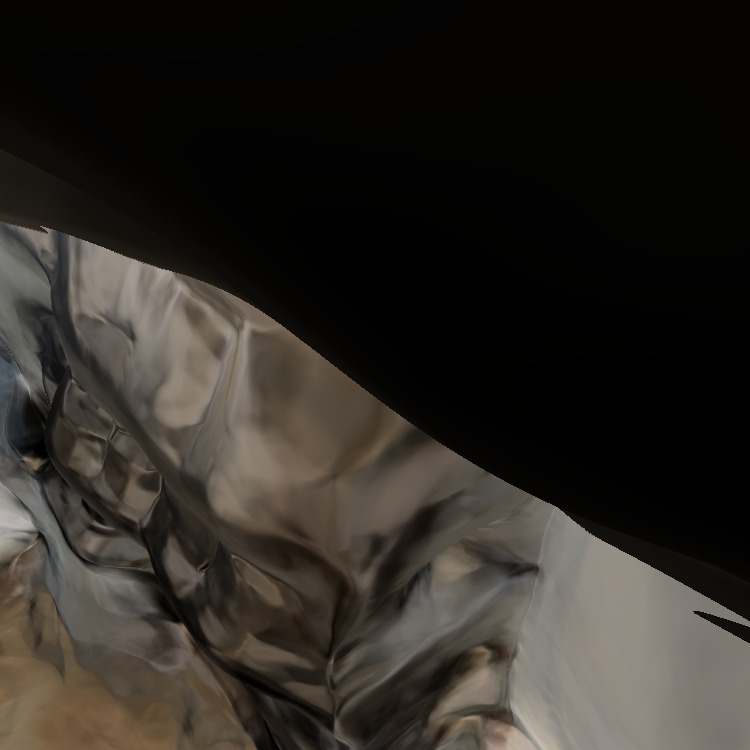} &
\includegraphics[width=\width]{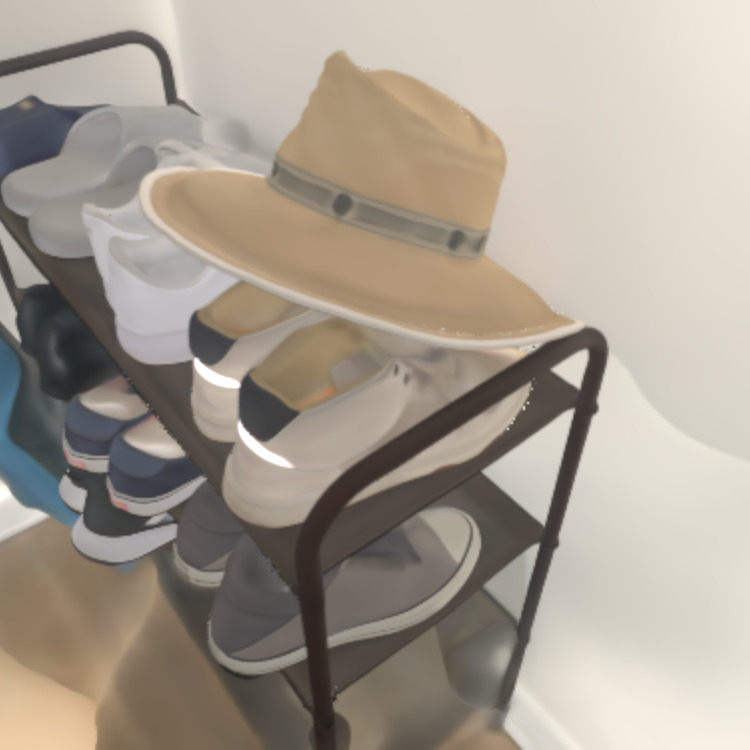} &
\includegraphics[width=\width]{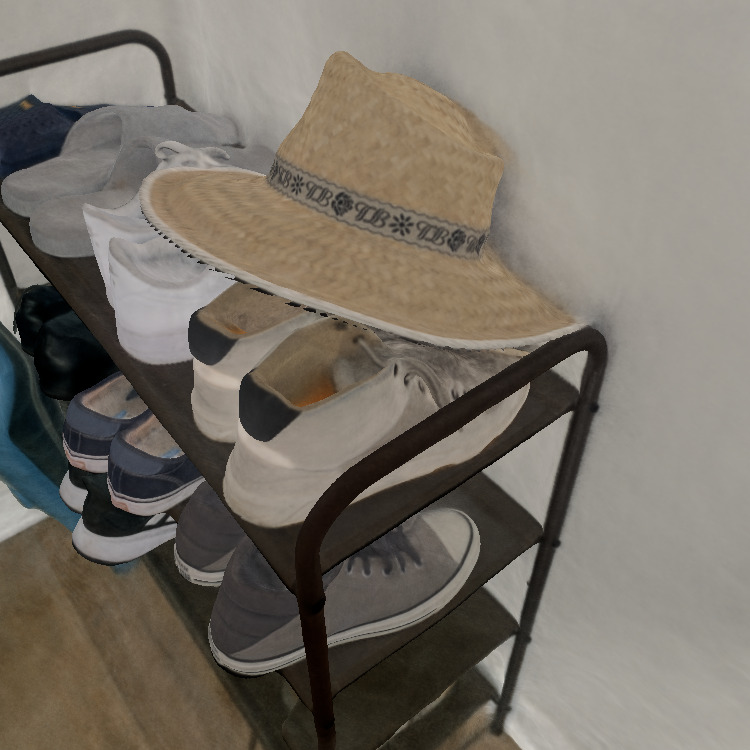} &
 \\
{\makebox[5pt]{\rotatebox{90}{\tiny Roughness}}} &
\includegraphics[width=\width]{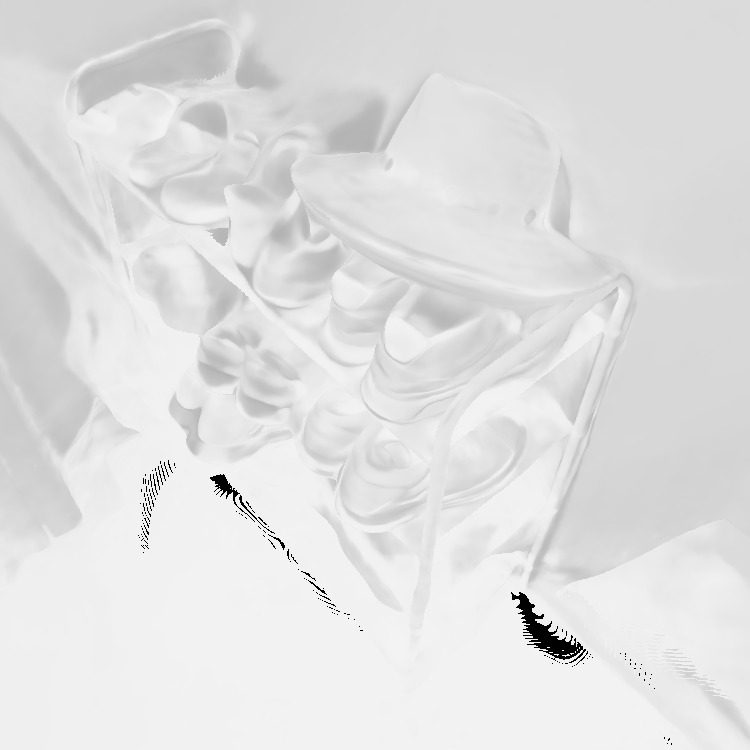} &
\includegraphics[width=\width]{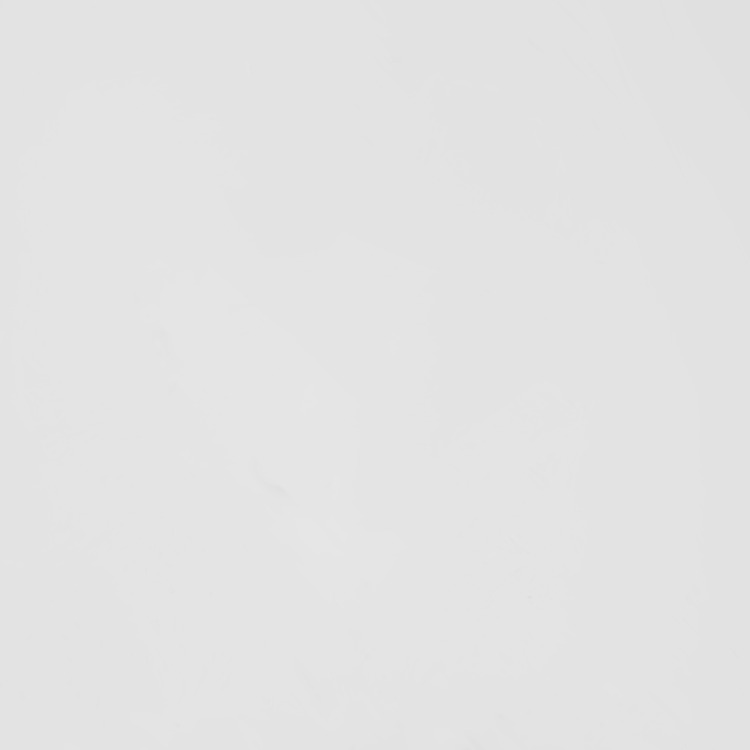} &
\includegraphics[width=\width]{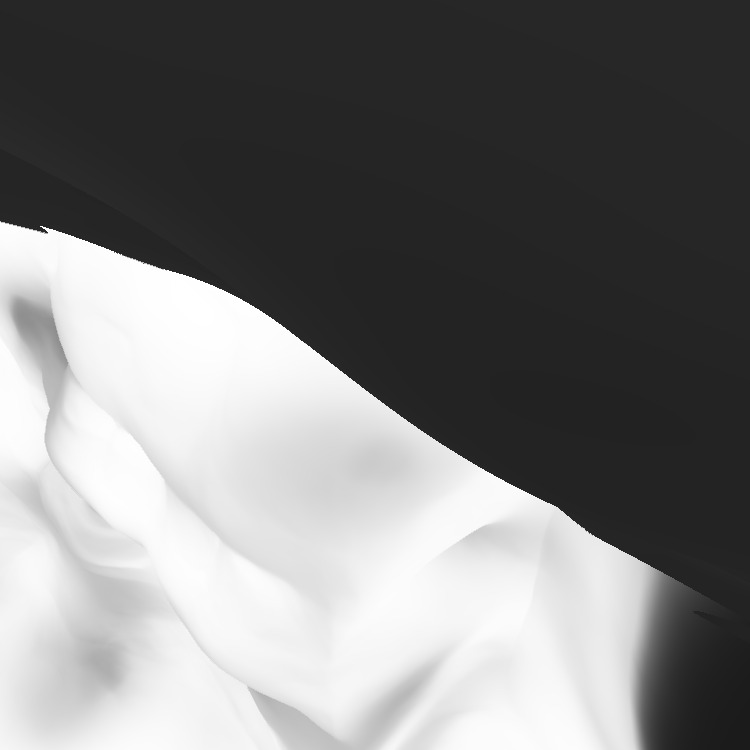} &
\includegraphics[width=\width]{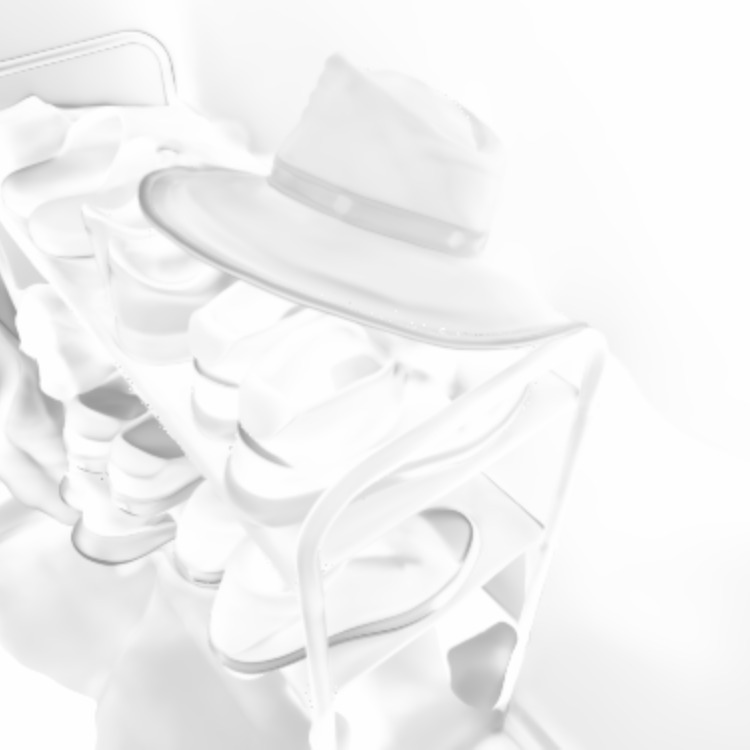} &
\includegraphics[width=\width]{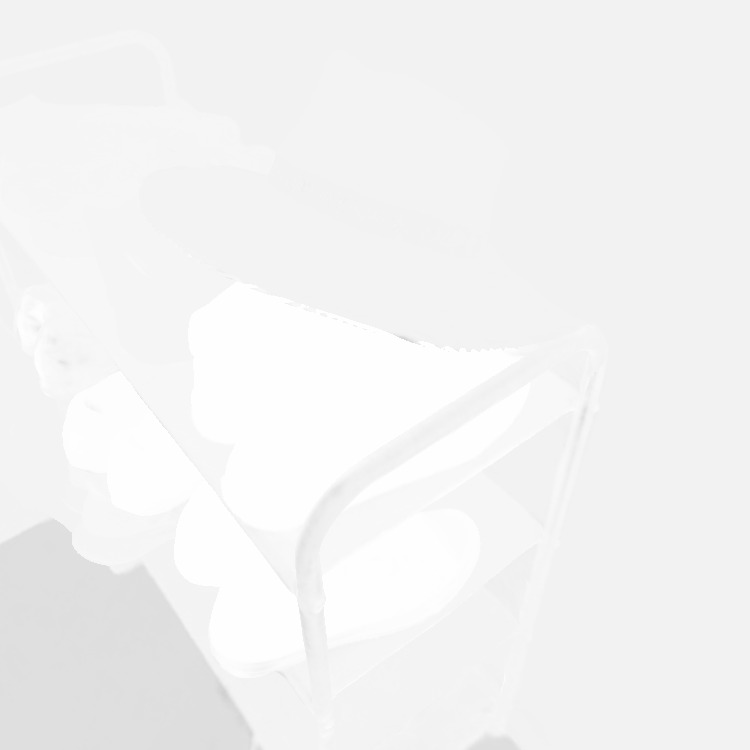} &
 \\
{\makebox[5pt]{\rotatebox{90}{\tiny \hspace{0pt} Shoe Rack}}} &
\includegraphics[width=\width]{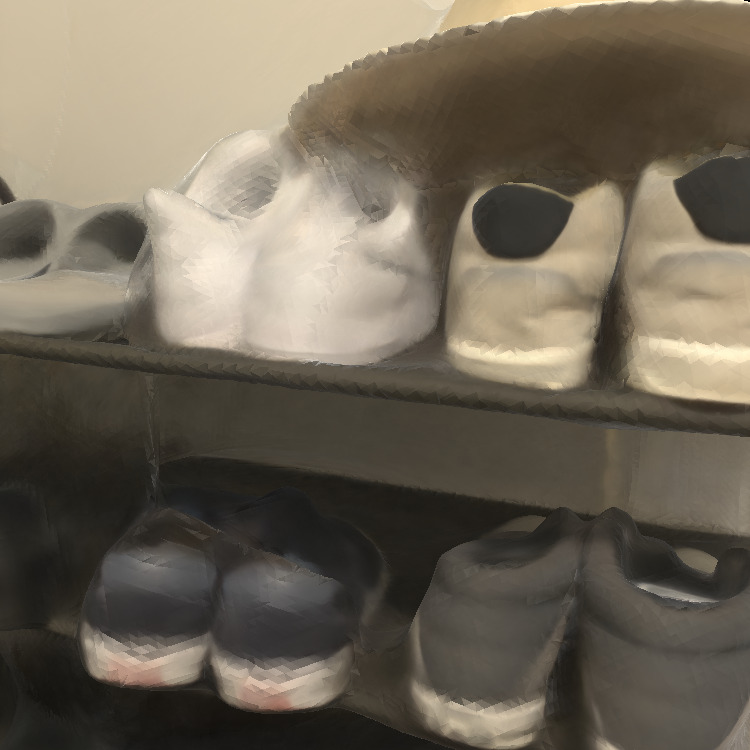} &
\includegraphics[width=\width]{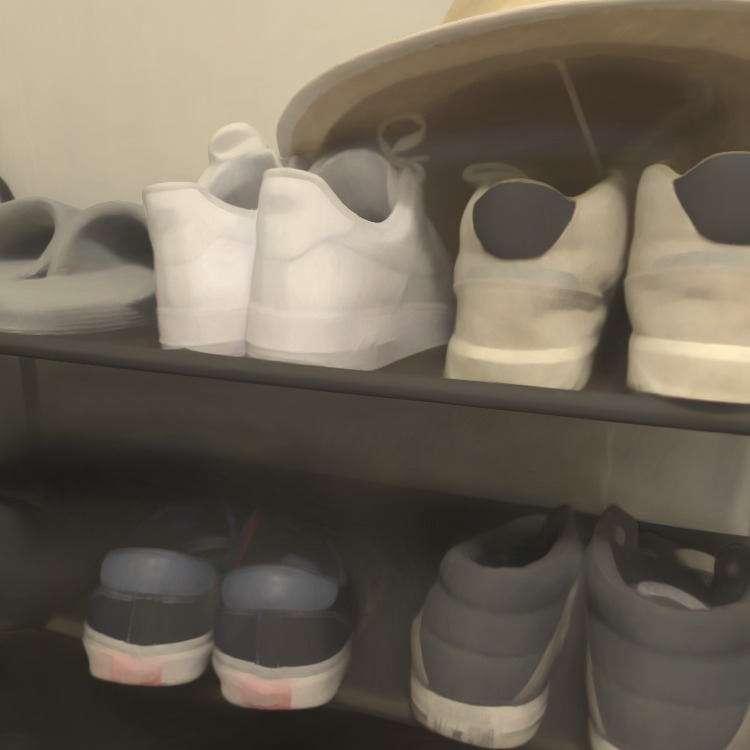} &
\includegraphics[width=\width]{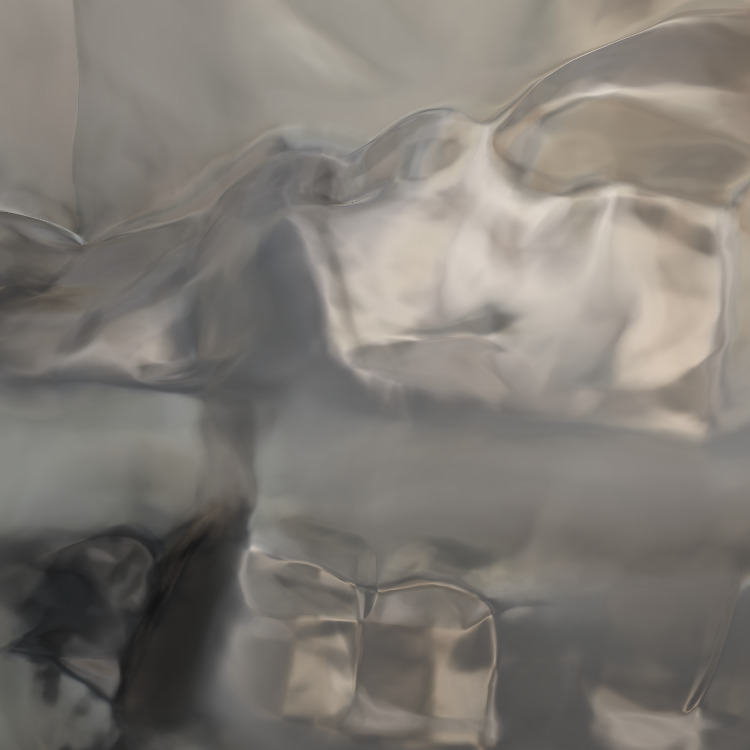} &
\includegraphics[width=\width]{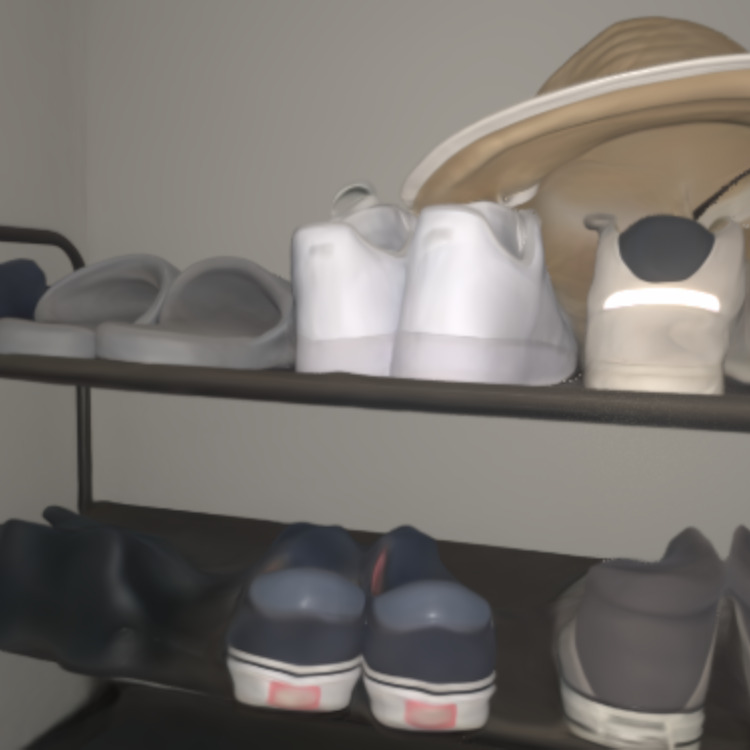} &
\includegraphics[width=\width]{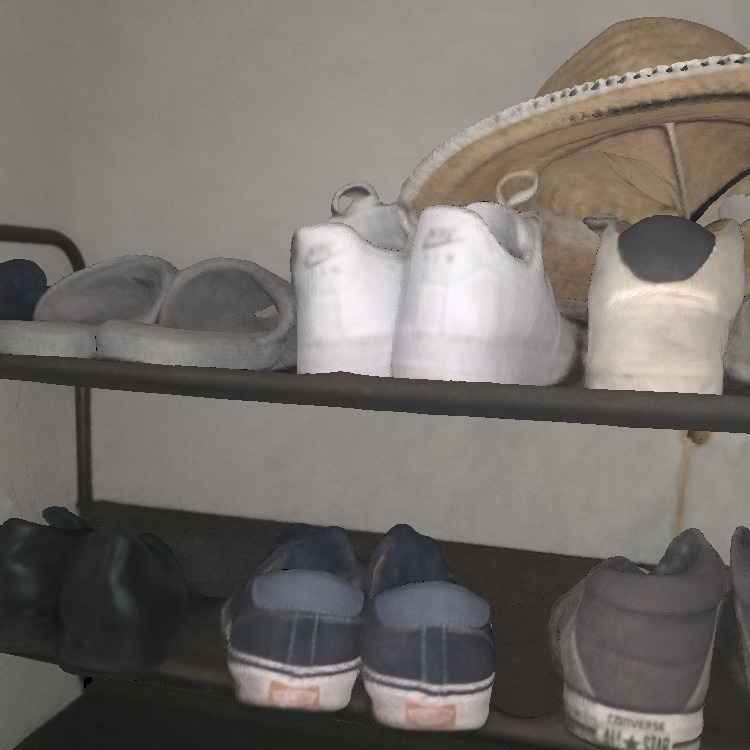} &
\includegraphics[width=\width]{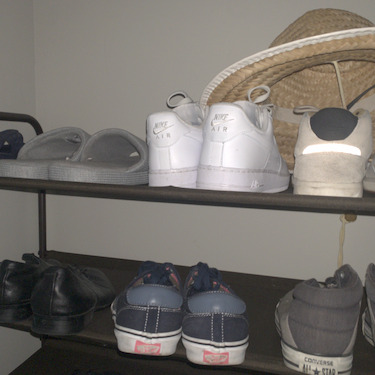} \\
{\makebox[5pt]{\rotatebox{90}{\tiny Albedo}}} &
\includegraphics[width=\width]{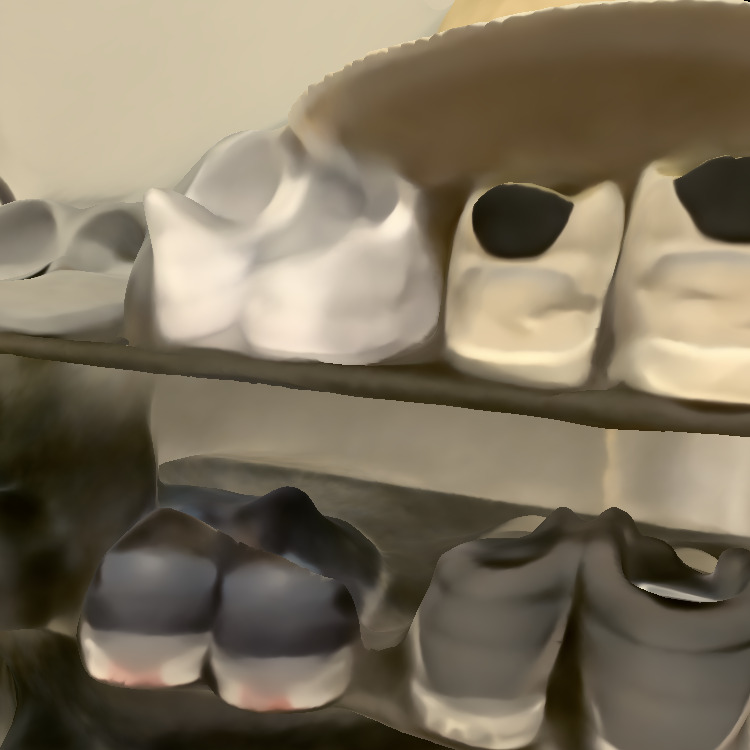} &
\includegraphics[width=\width]{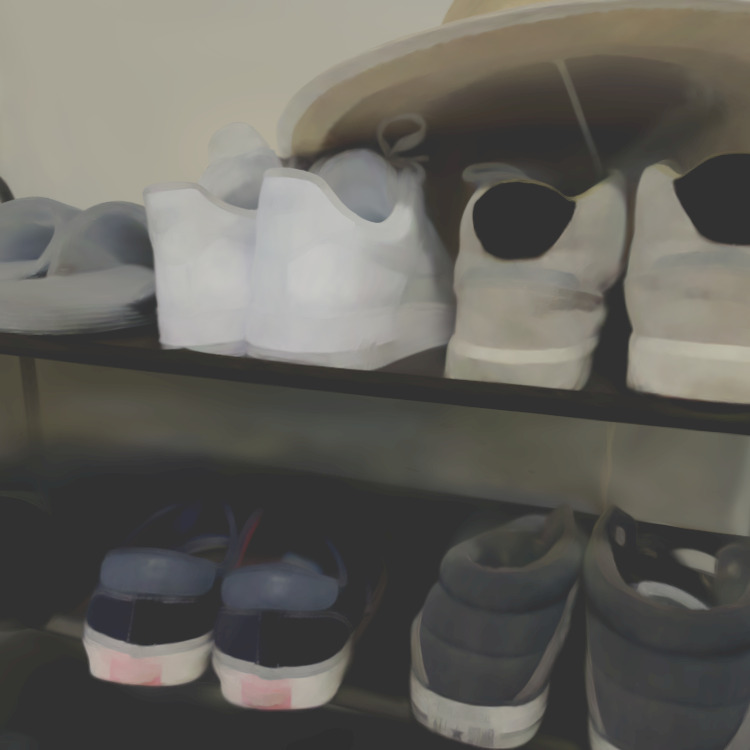} &
\includegraphics[width=\width]{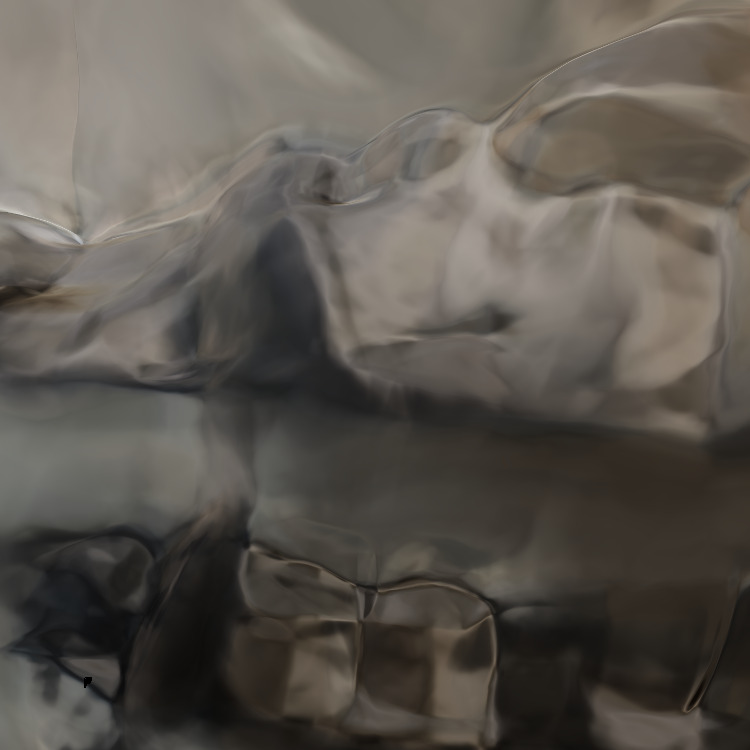} &
\includegraphics[width=\width]{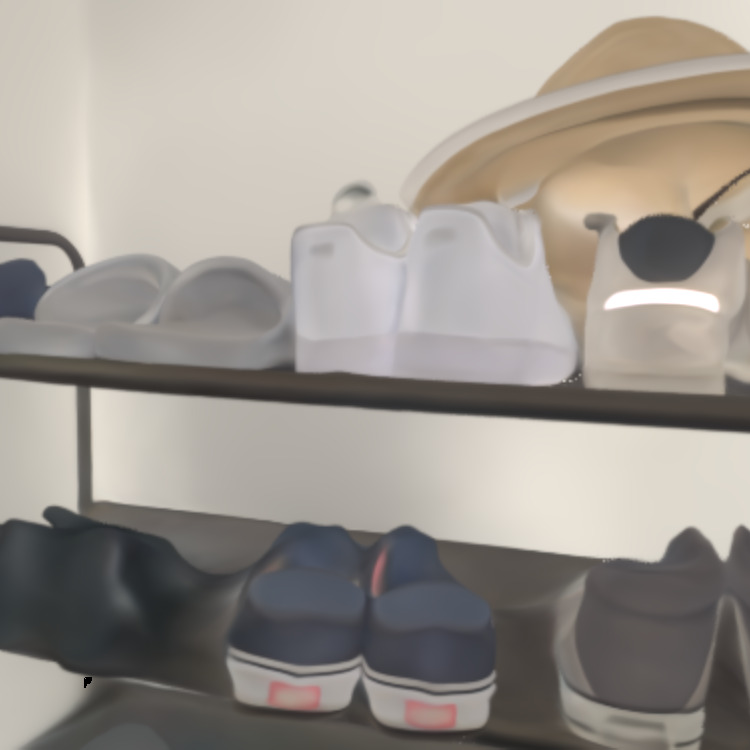} &
\includegraphics[width=\width]{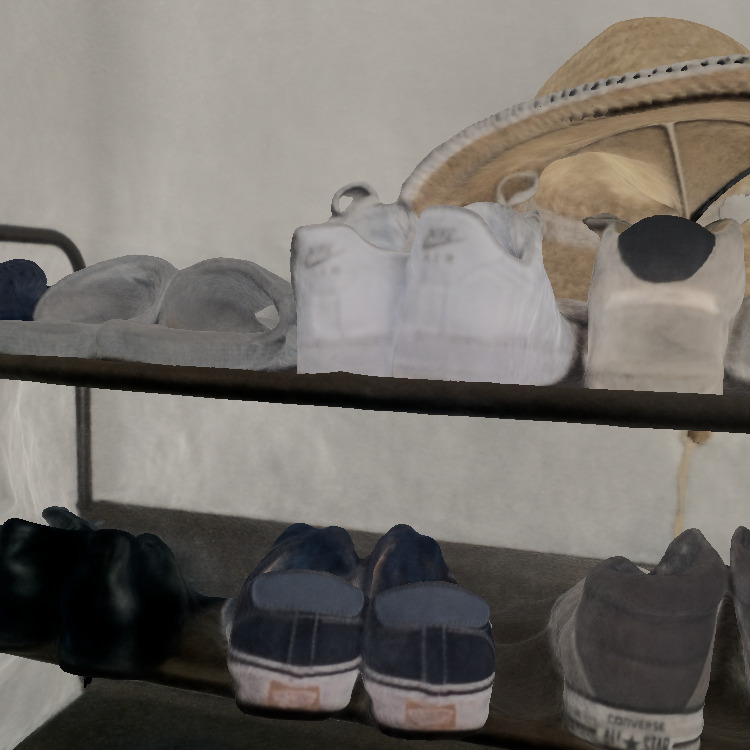} &
 \\
{\makebox[5pt]{\rotatebox{90}{\tiny Roughness}}} &
\includegraphics[width=\width]{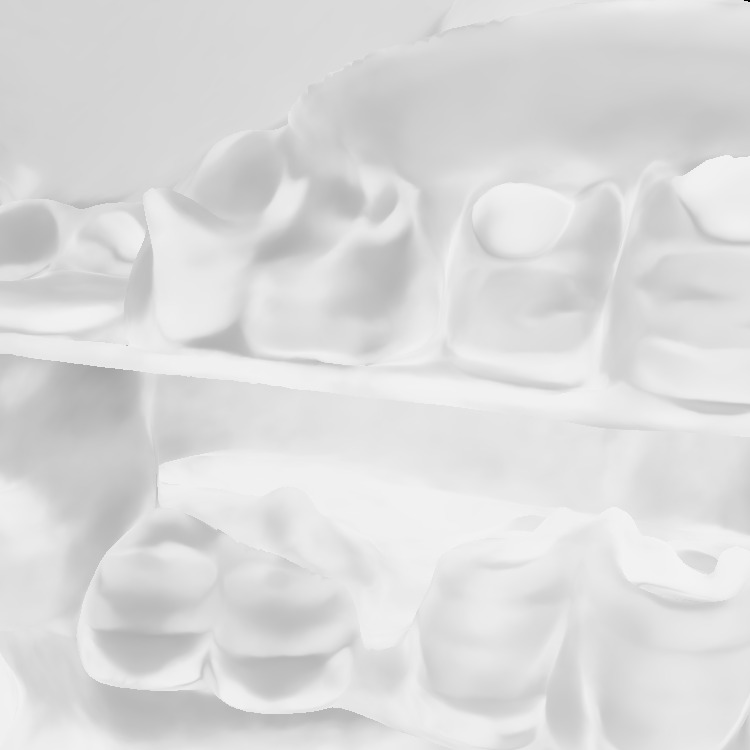} &
\includegraphics[width=\width]{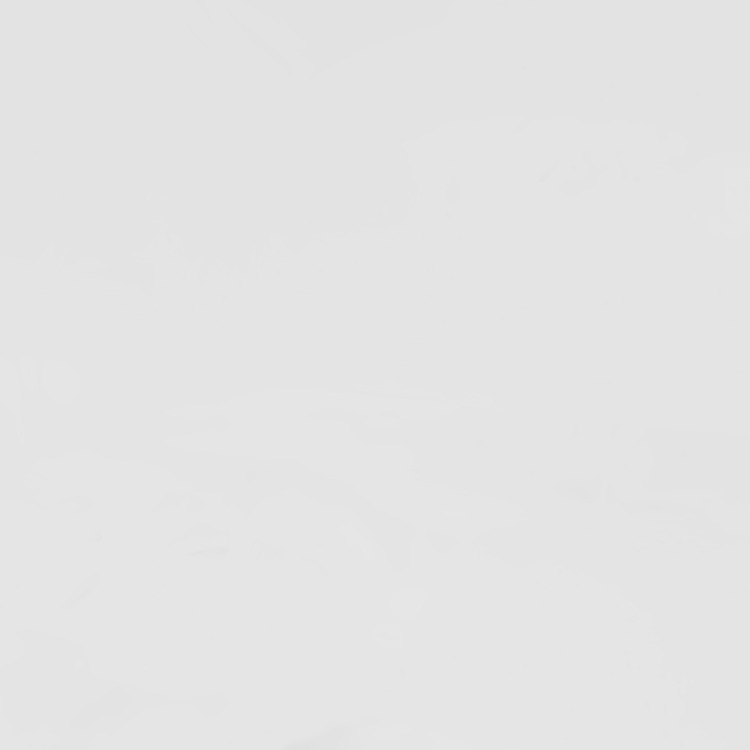} &
\includegraphics[width=\width]{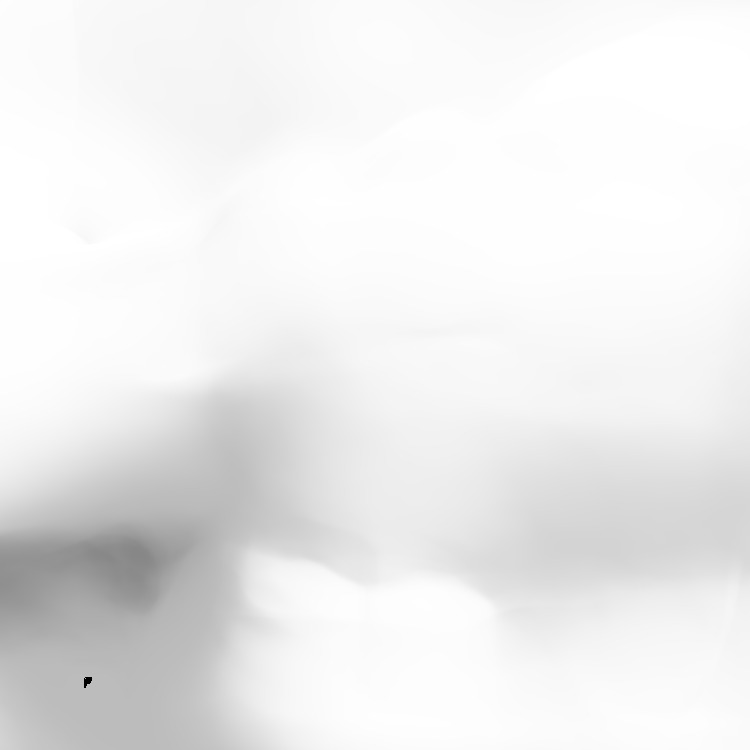} &
\includegraphics[width=\width]{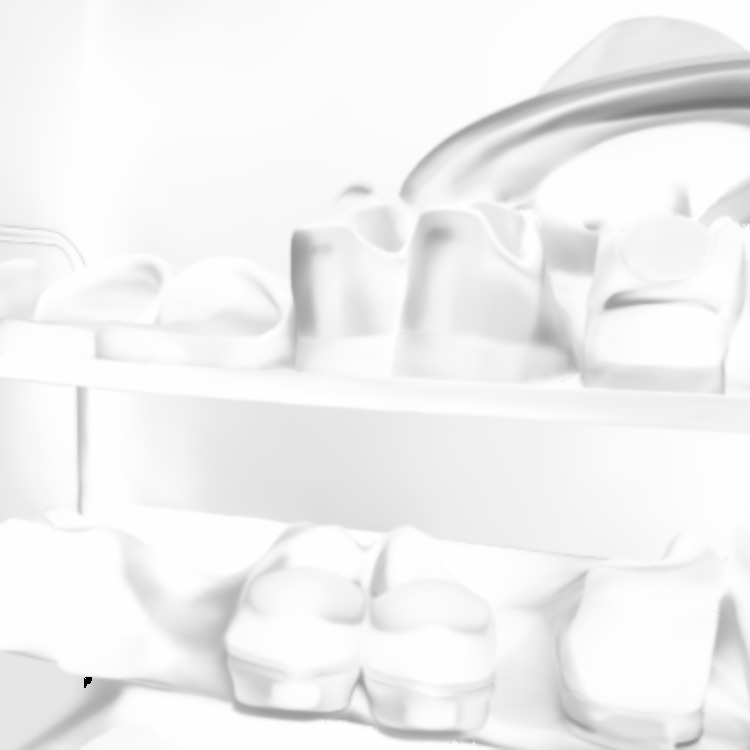} &
\includegraphics[width=\width]{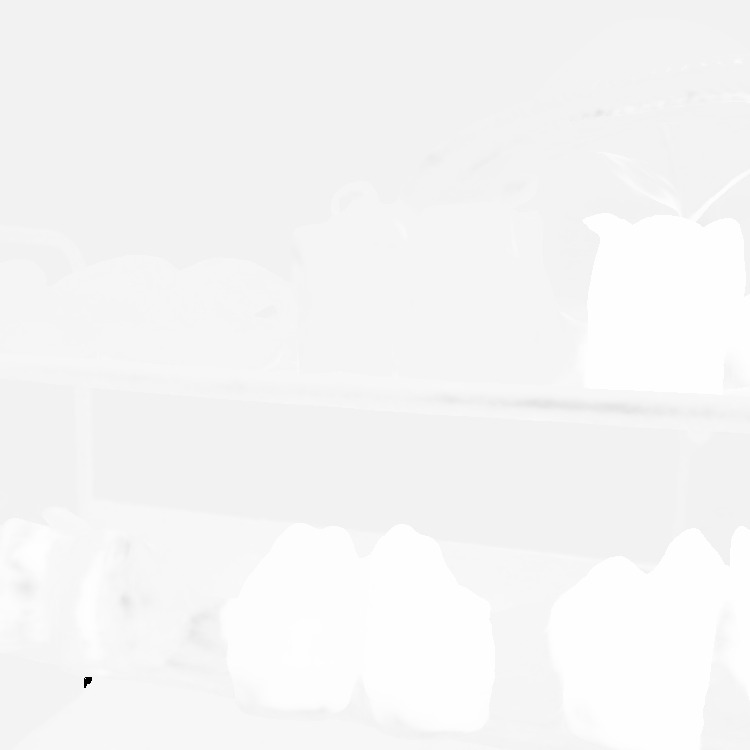} &
 \\

%% file: generated/suppl_qualitative_generated_real_1.tex
{\makebox[5pt]{\rotatebox{90}{\tiny \hspace{6pt} Table}}} &
\includegraphics[width=\width]{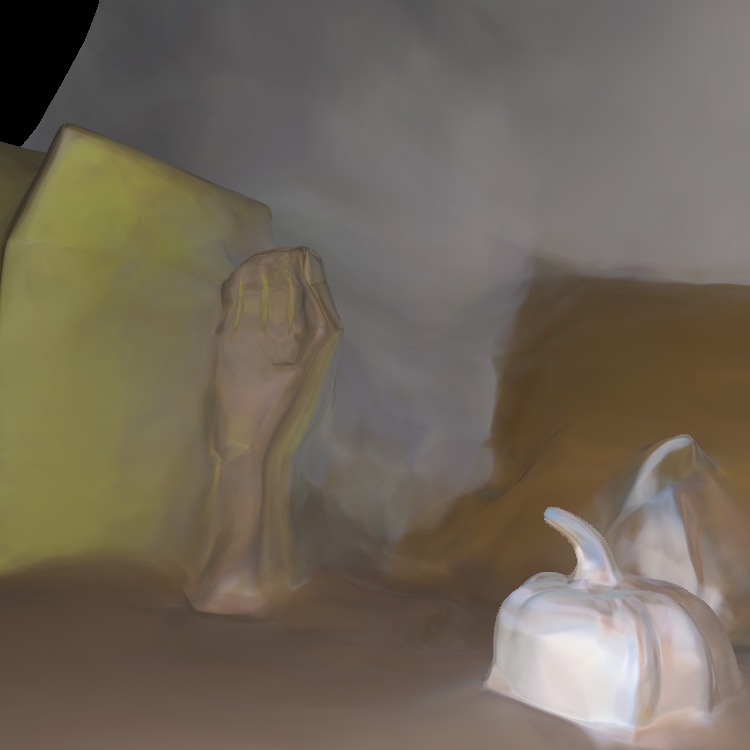} &
\includegraphics[width=\width]{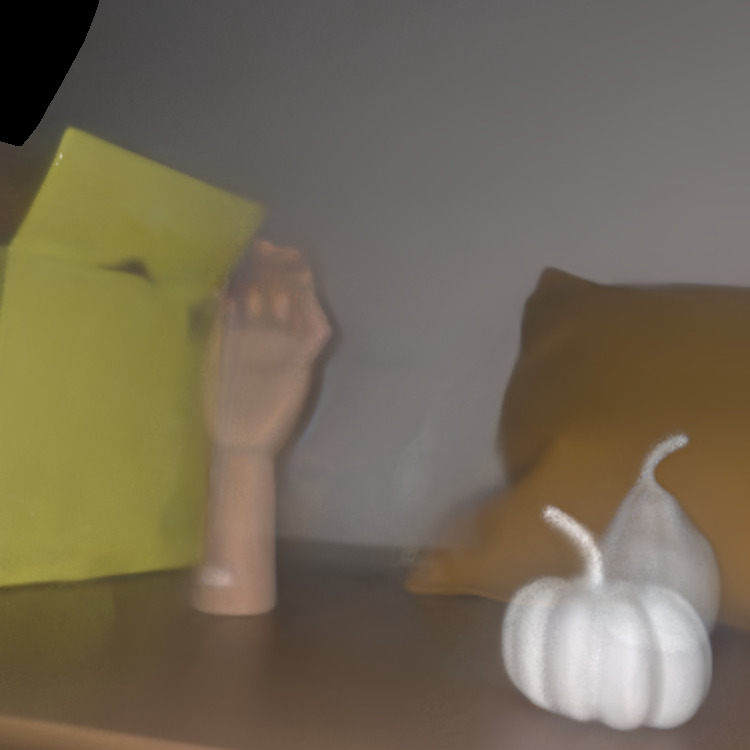} &
\includegraphics[width=\width]{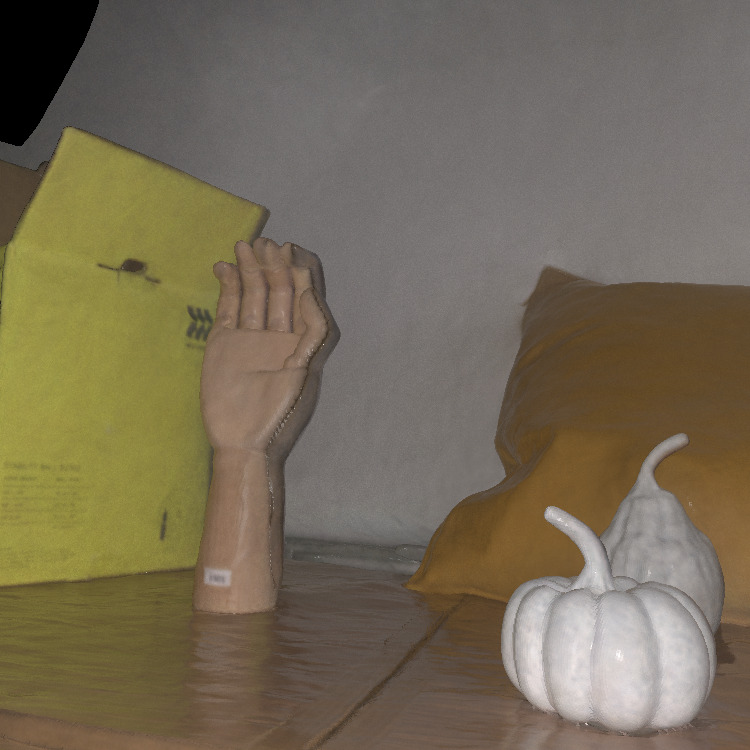} &
\includegraphics[width=\width]{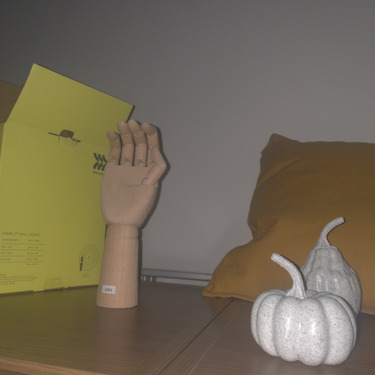} \\
{\makebox[5pt]{\rotatebox{90}{\tiny Albedo}}} &
\includegraphics[width=\width]{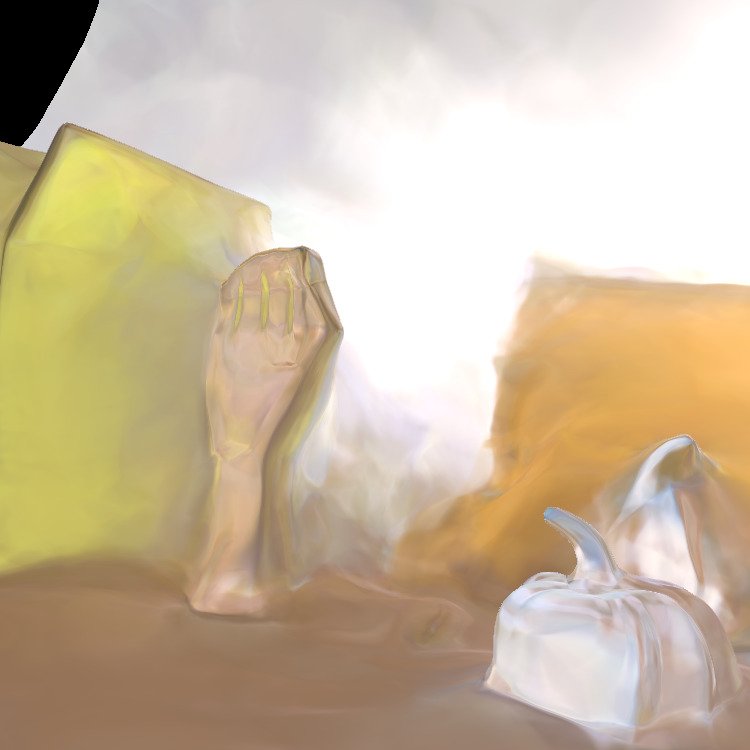} &
\includegraphics[width=\width]{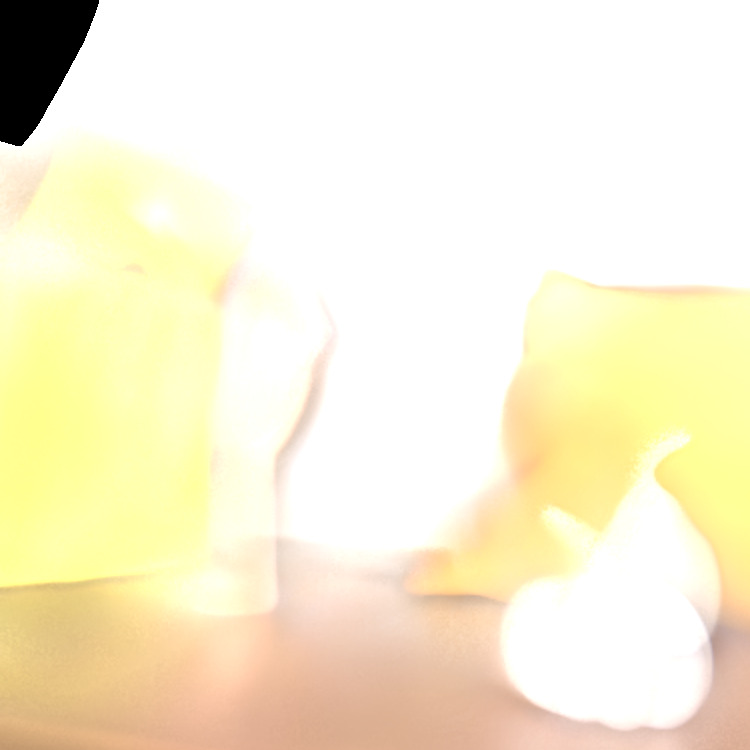} &
\includegraphics[width=\width]{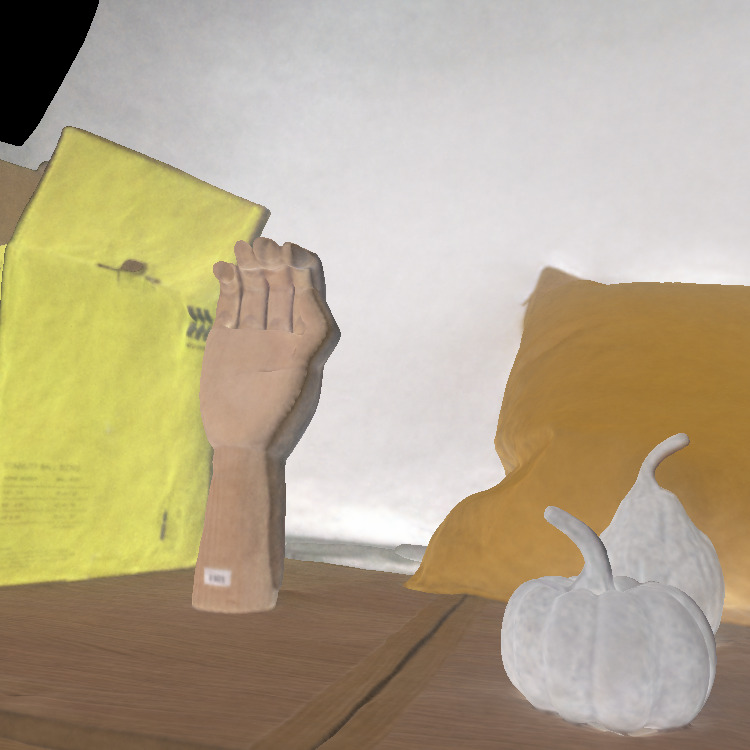} &
 \\
{\makebox[5pt]{\rotatebox{90}{\tiny Roughness}}} &
\includegraphics[width=\width]{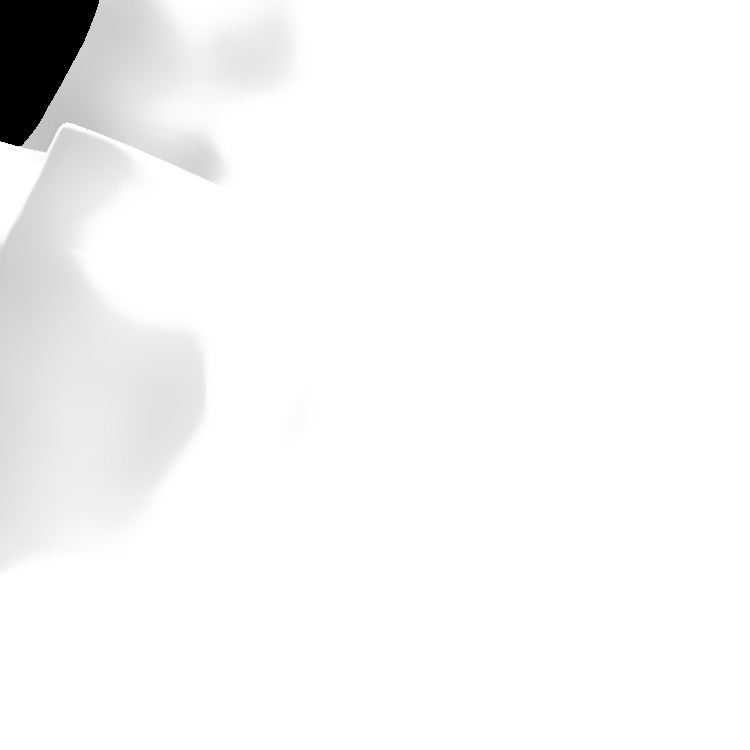} &
\includegraphics[width=\width]{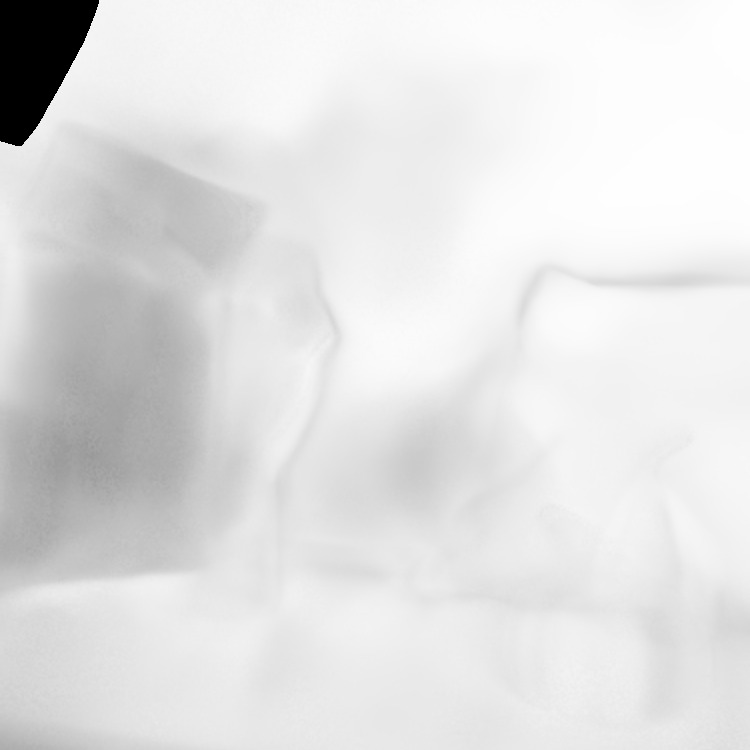} &
\includegraphics[width=\width]{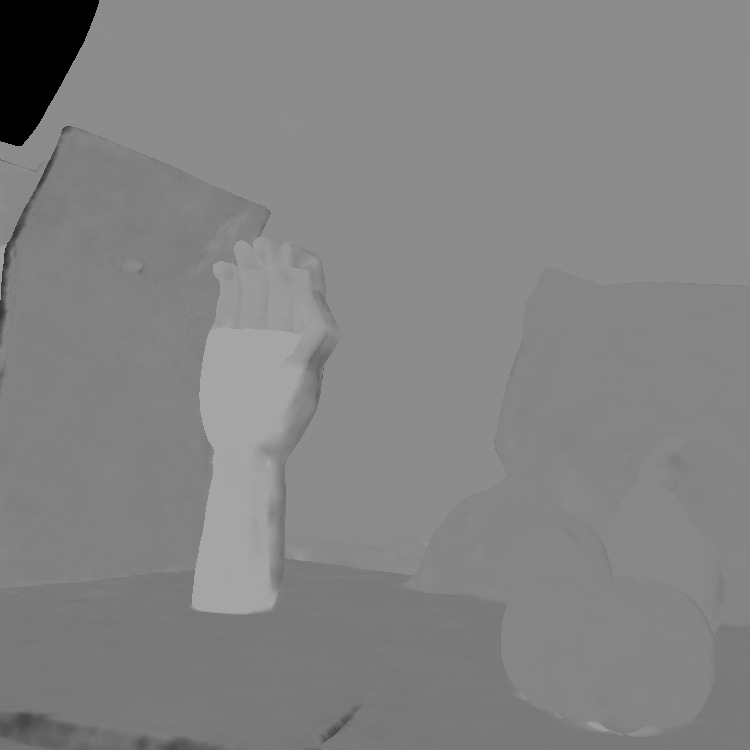} &
 \\
{\makebox[5pt]{\rotatebox{90}{\tiny \hspace{6pt} Table}}} &
\includegraphics[width=\width]{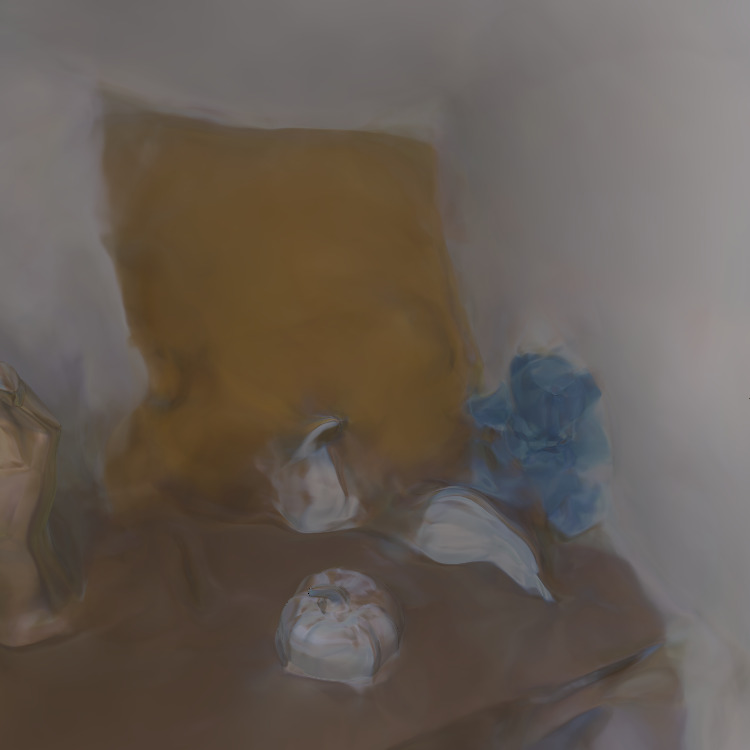} &
\includegraphics[width=\width]{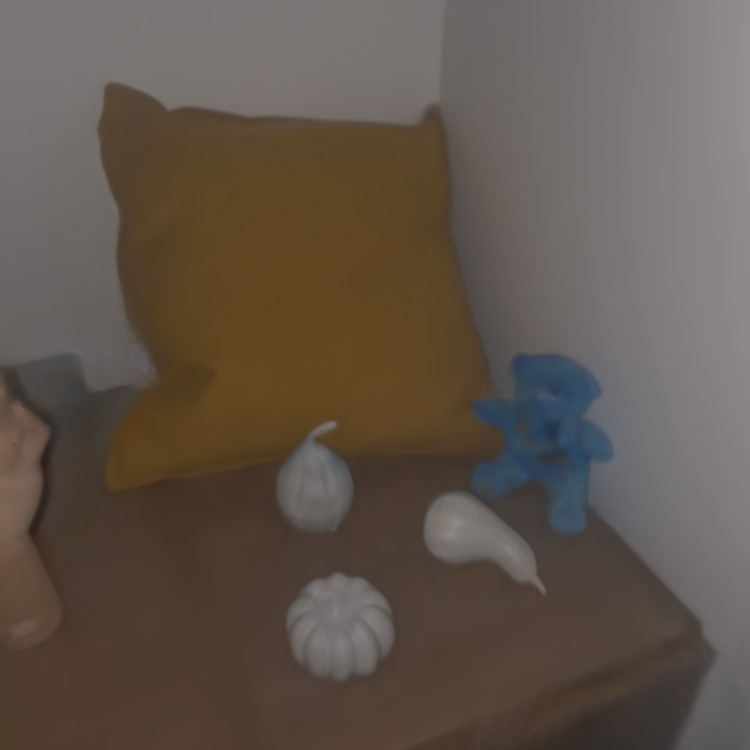} &
\includegraphics[width=\width]{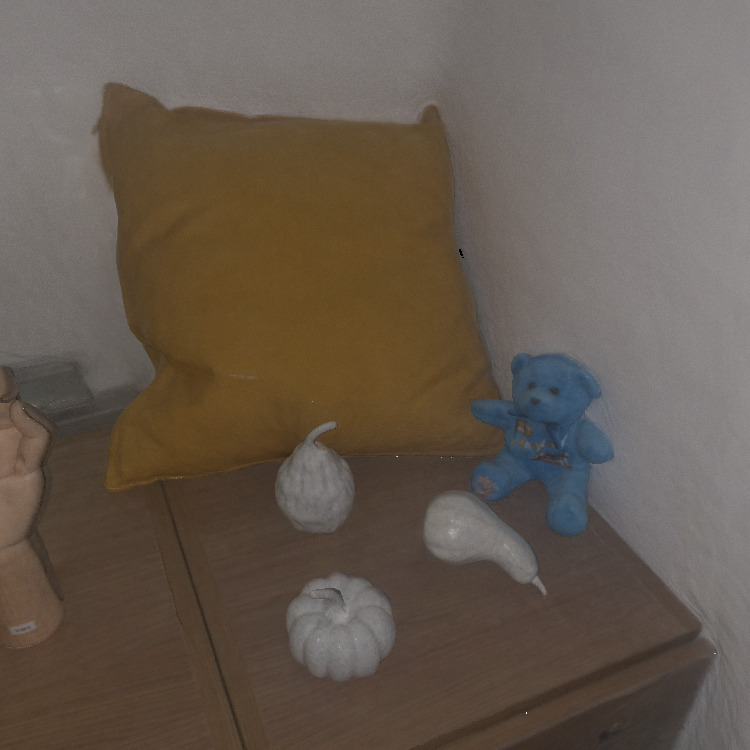} &
\includegraphics[width=\width]{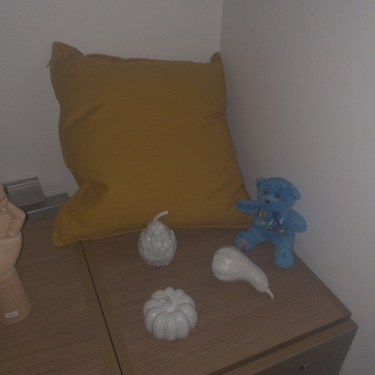} \\
{\makebox[5pt]{\rotatebox{90}{\tiny Albedo}}} &
\includegraphics[width=\width]{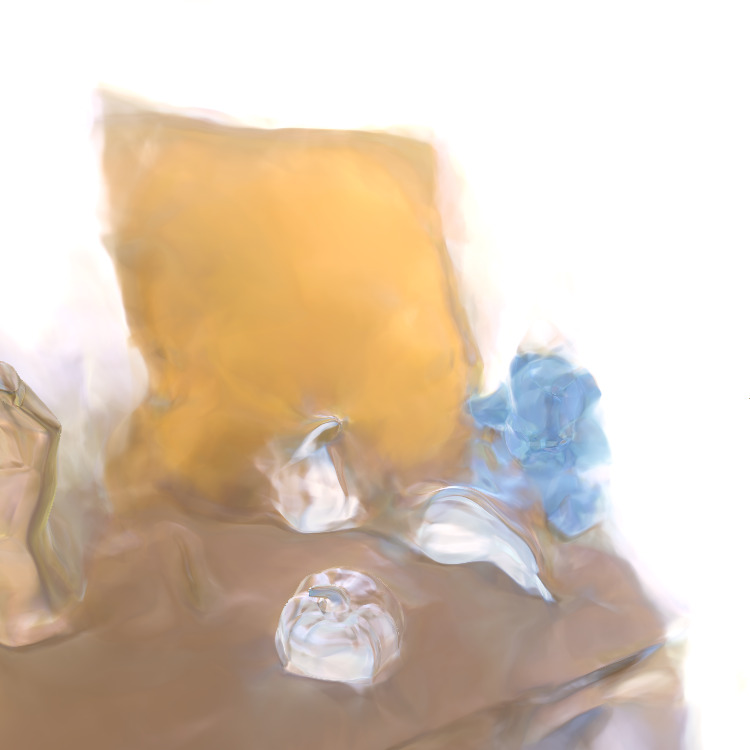} &
\includegraphics[width=\width]{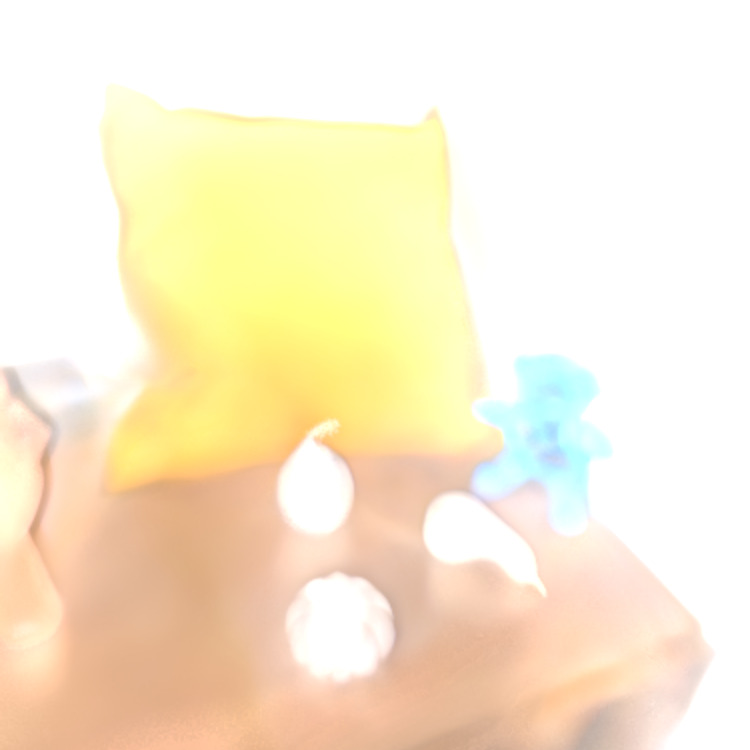} &
\includegraphics[width=\width]{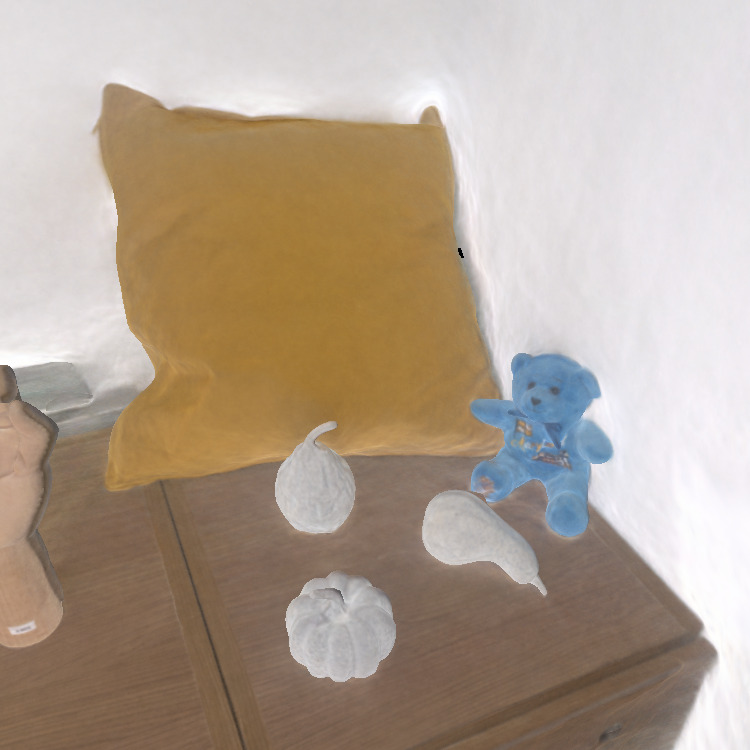} &
 \\
{\makebox[5pt]{\rotatebox{90}{\tiny Roughness}}} &
\includegraphics[width=\width]{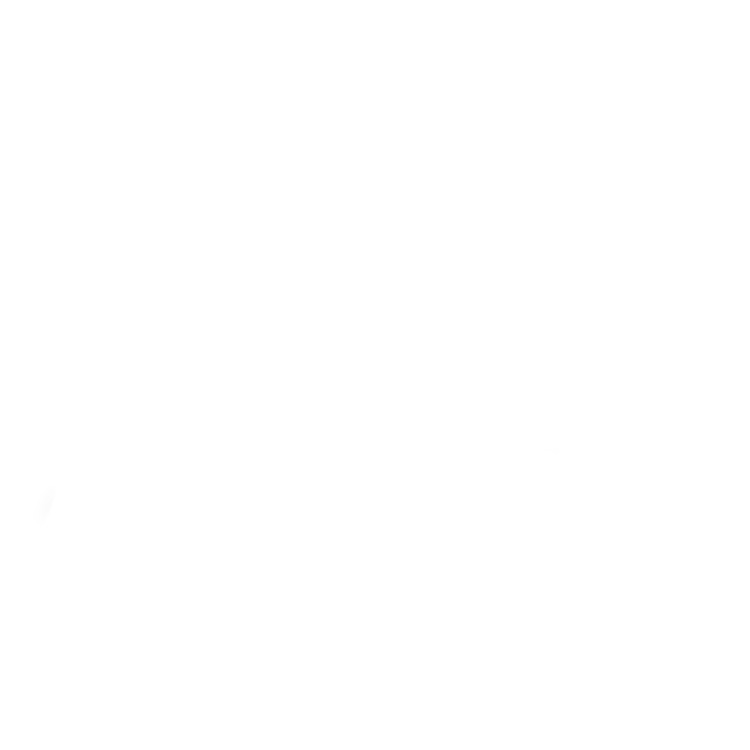} &
\includegraphics[width=\width]{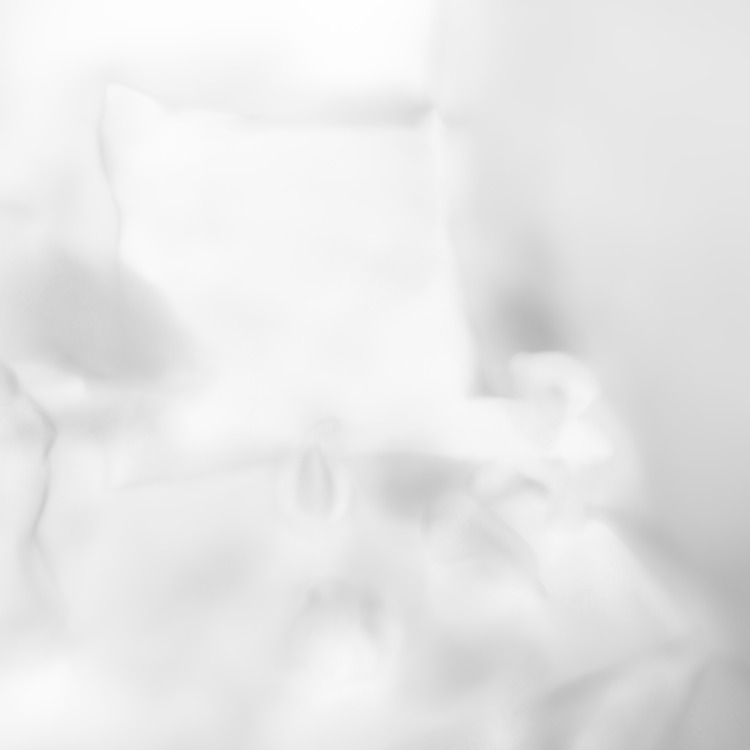} &
\includegraphics[width=\width]{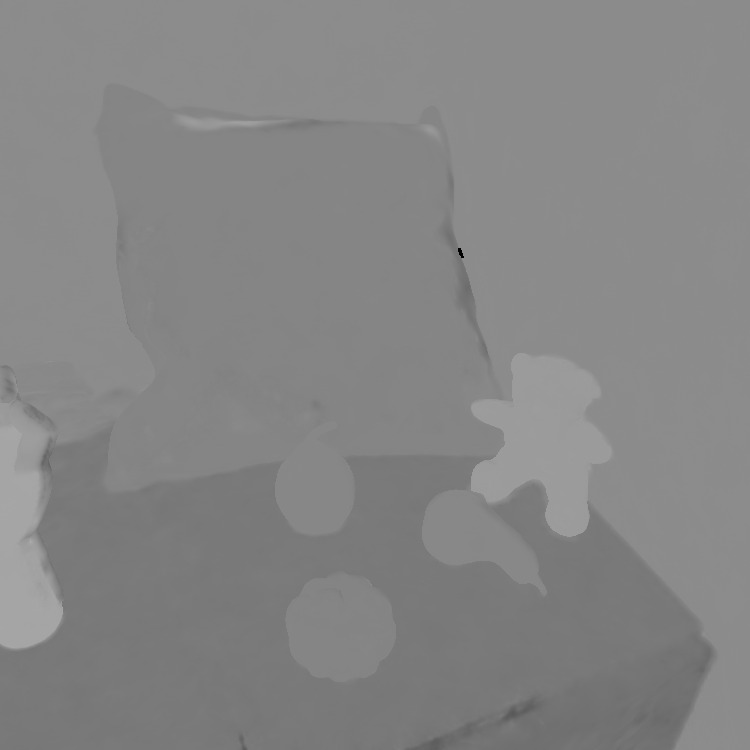} &
 \\

%% file: generated/suppl_qualitative_generated_real_2.tex
{\makebox[5pt]{\rotatebox{90}{\tiny \hspace{3pt} Sill}}} &
\includegraphics[width=\width]{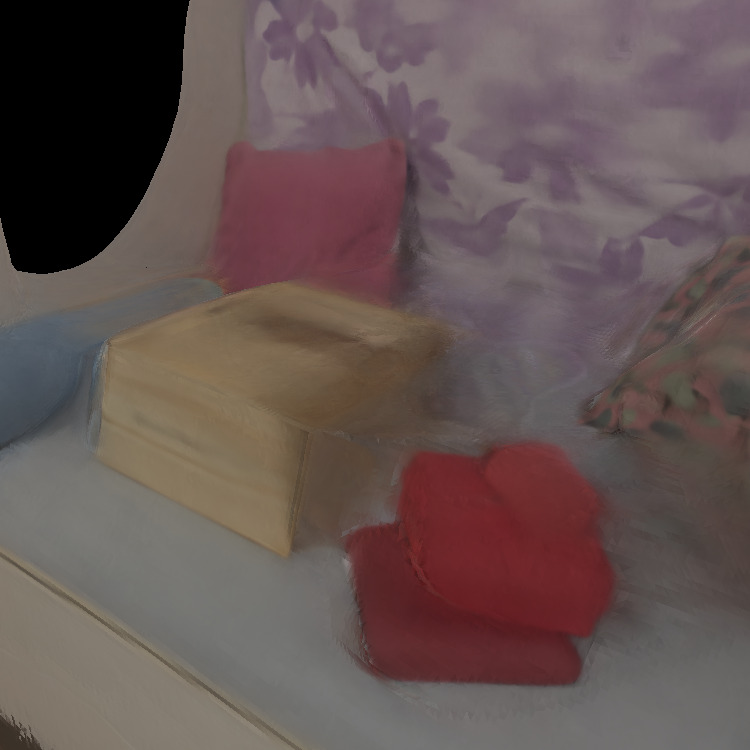} &
\includegraphics[width=\width]{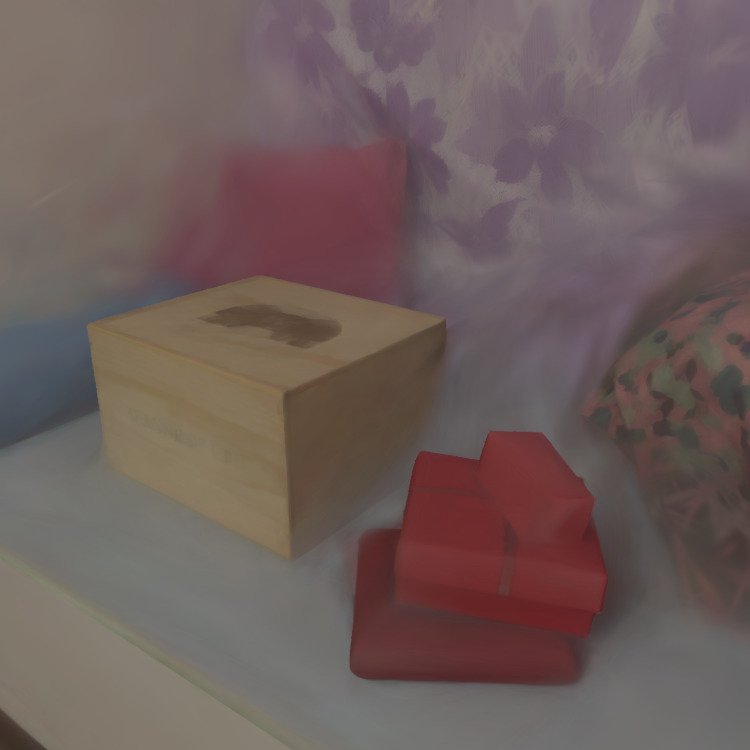} &
\includegraphics[width=\width]{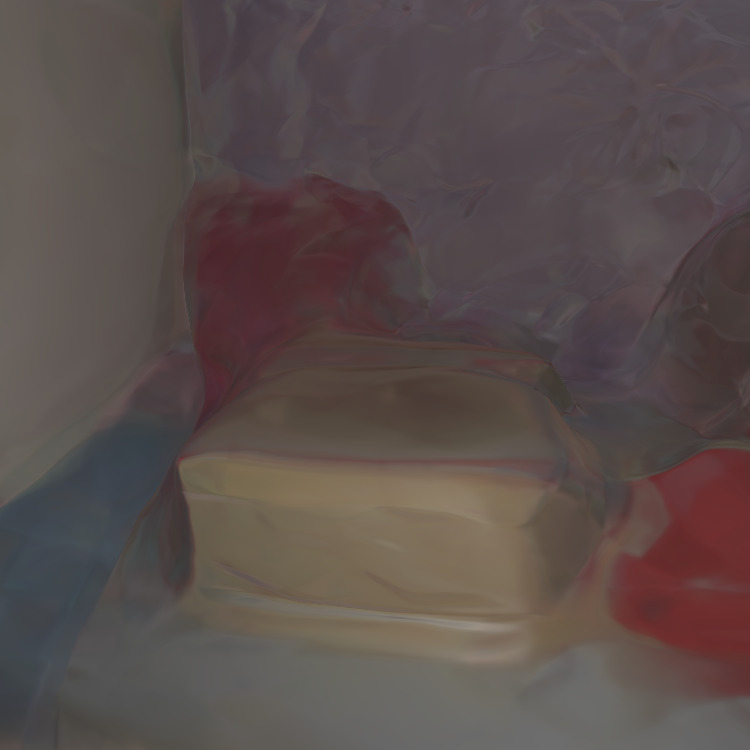} &
\includegraphics[width=\width]{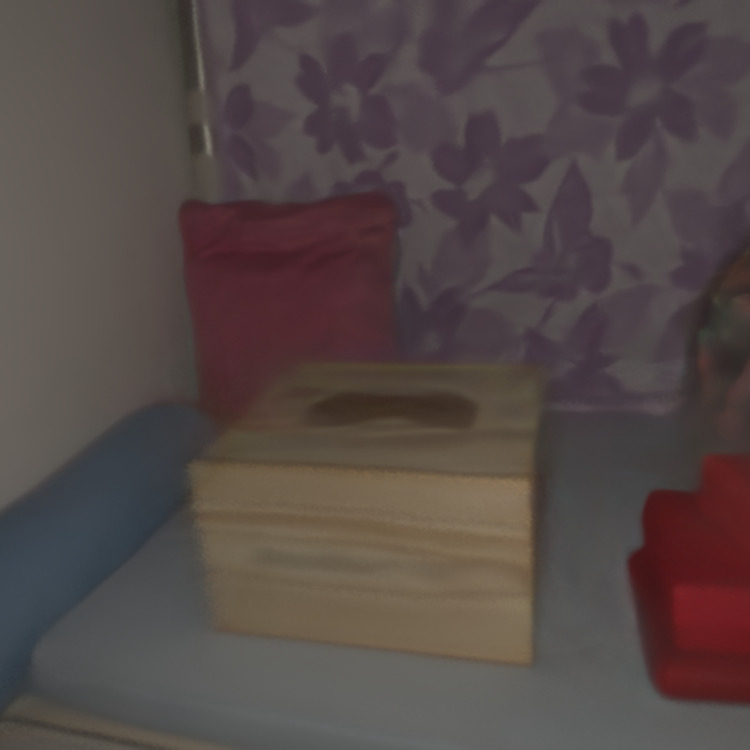} &
\includegraphics[width=\width]{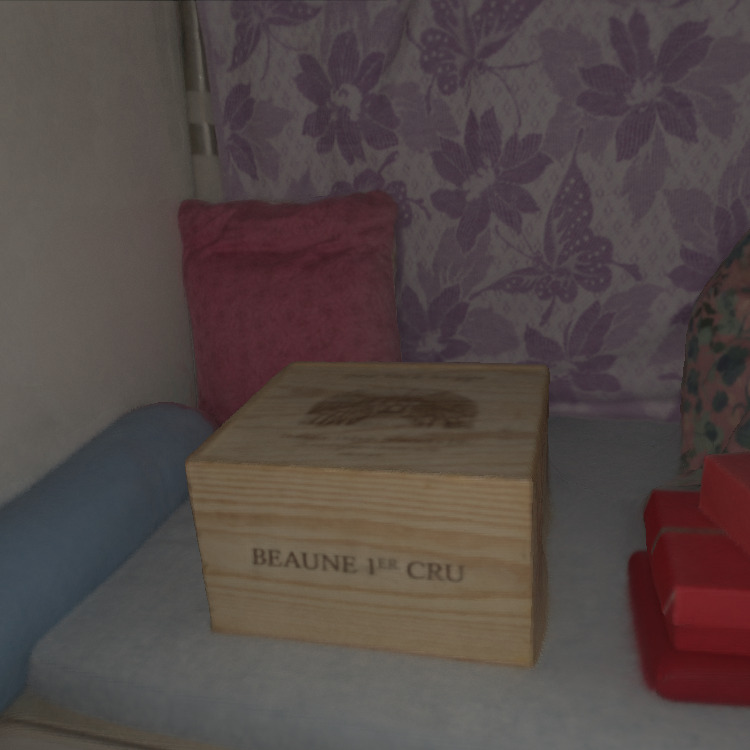} &
\includegraphics[width=\width]{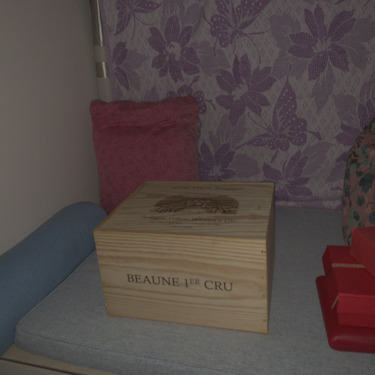} \\
{\makebox[5pt]{\rotatebox{90}{\tiny Albedo}}} &
\includegraphics[width=\width]{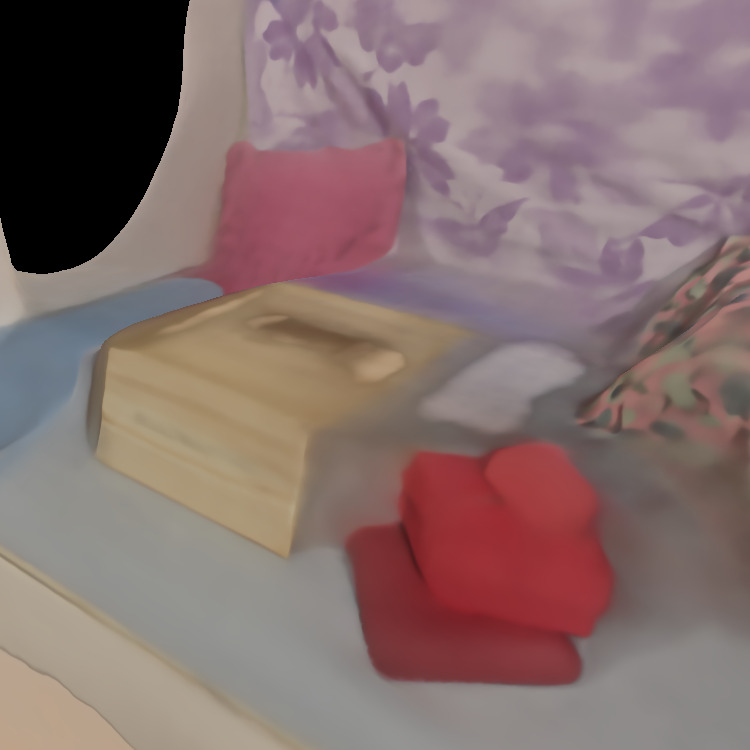} &
\includegraphics[width=\width]{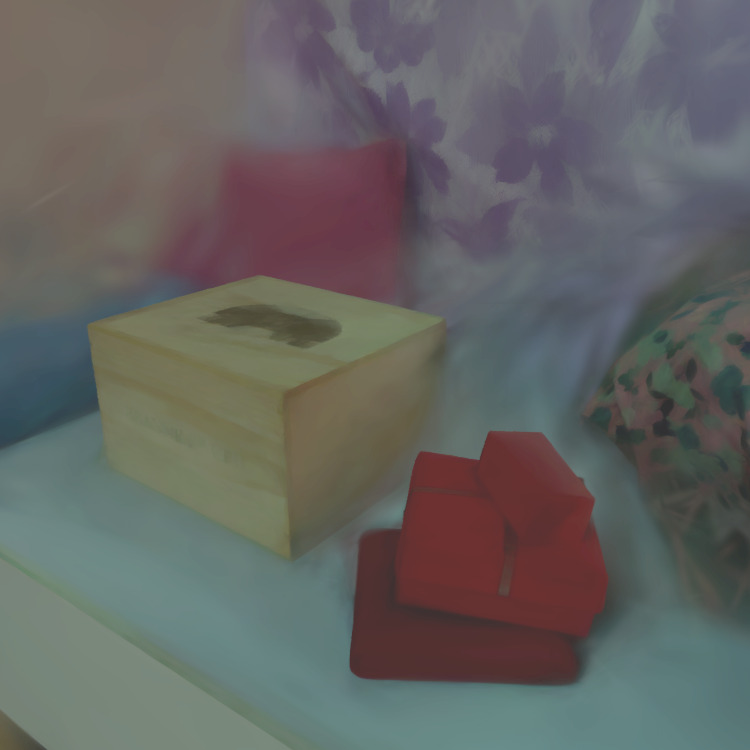} &
\includegraphics[width=\width]{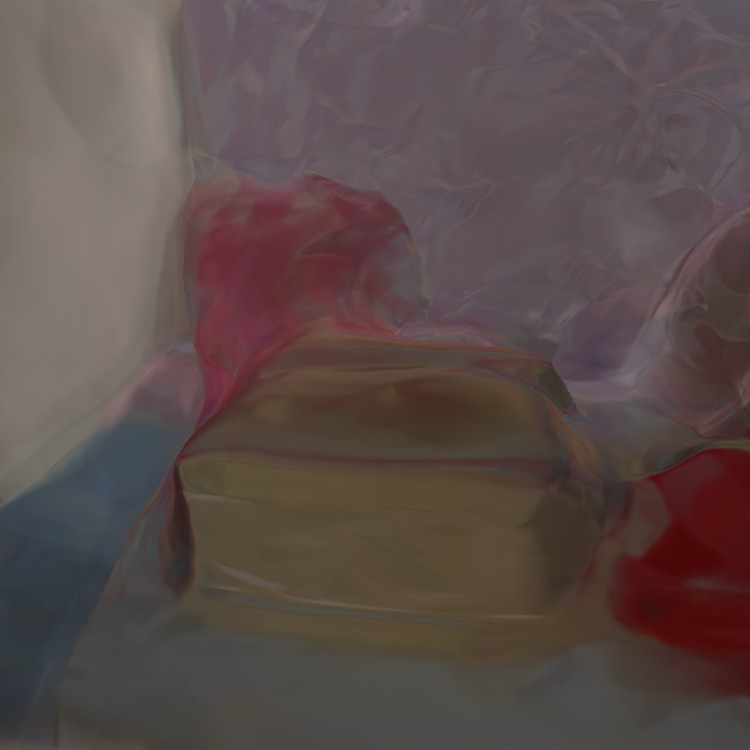} &
\includegraphics[width=\width]{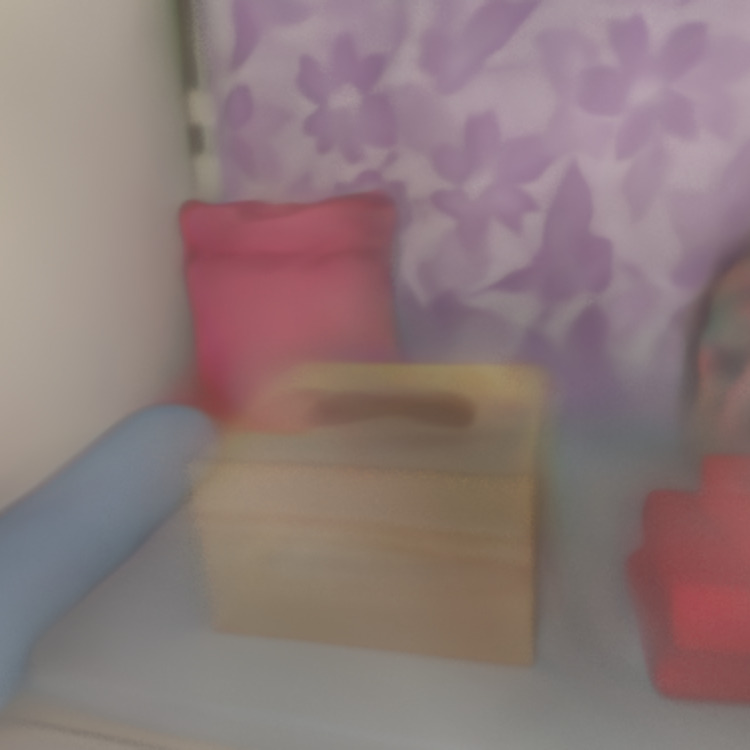} &
\includegraphics[width=\width]{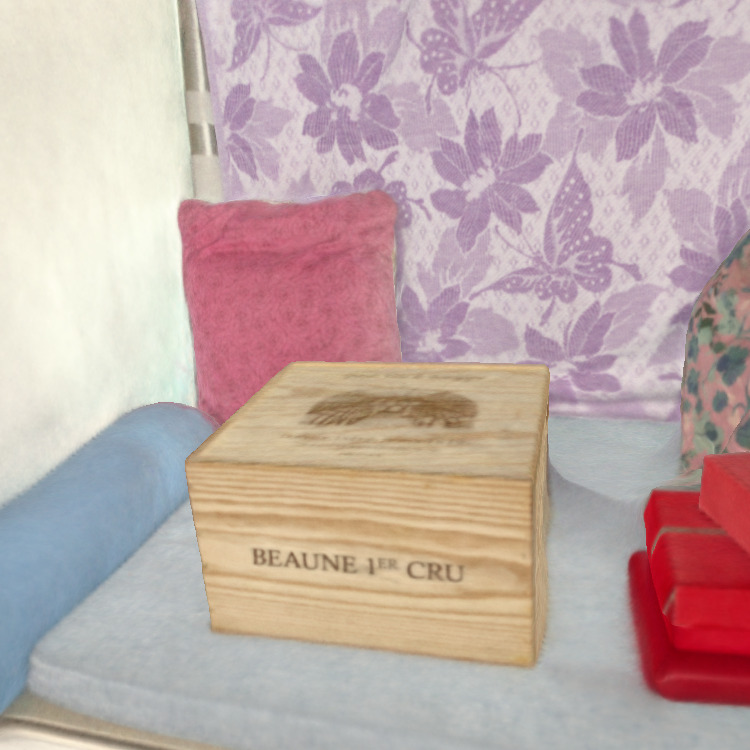} &
 \\
{\makebox[5pt]{\rotatebox{90}{\tiny Roughness}}} &
\includegraphics[width=\width]{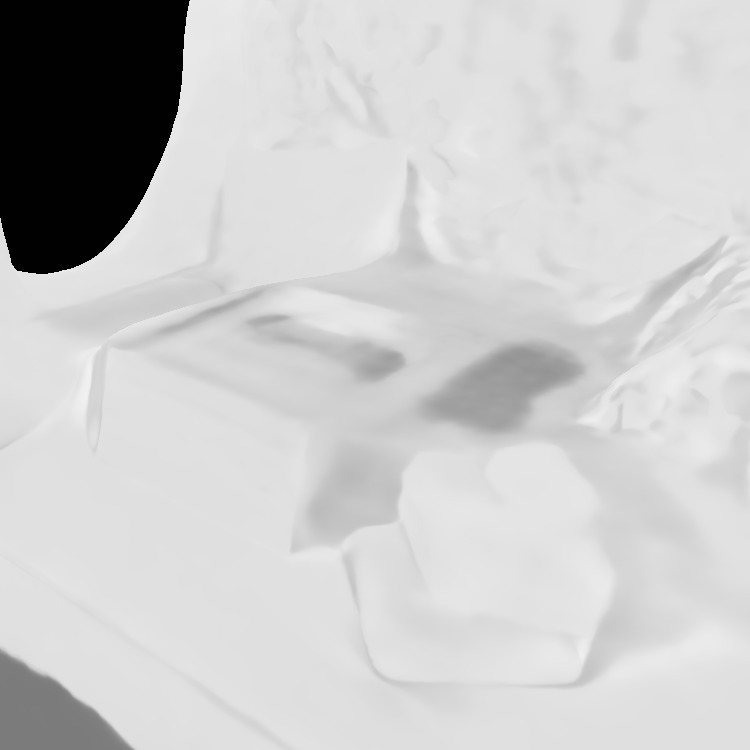} &
\includegraphics[width=\width]{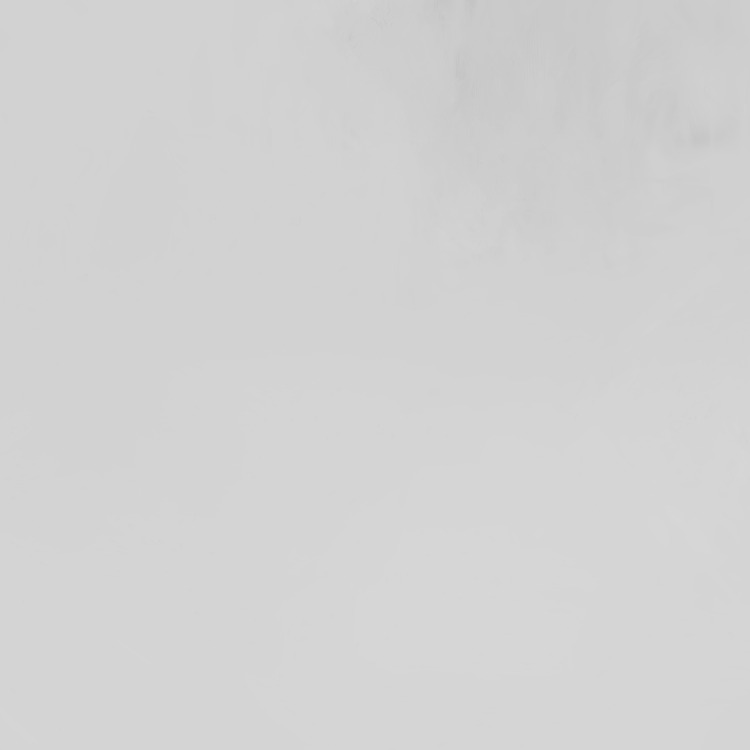} &
\includegraphics[width=\width]{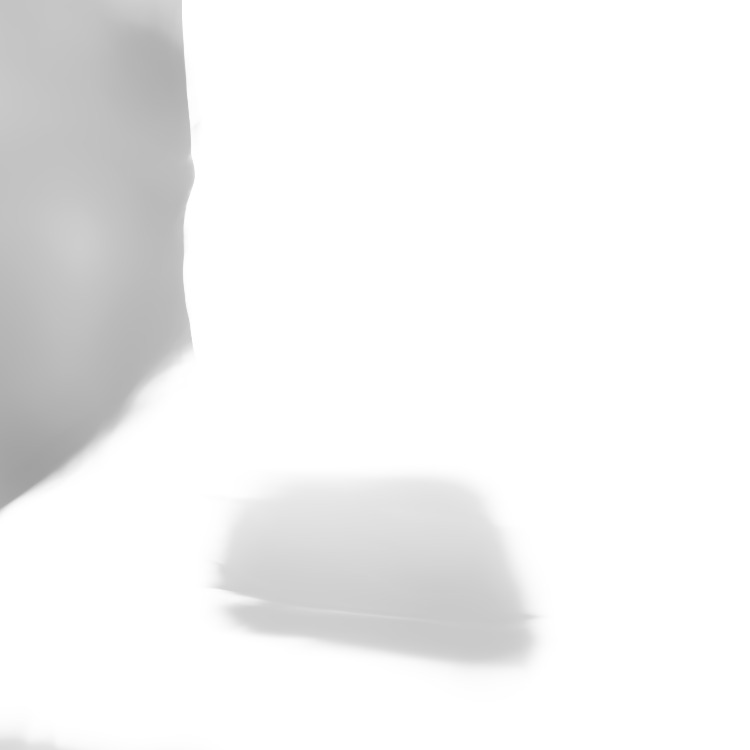} &
\includegraphics[width=\width]{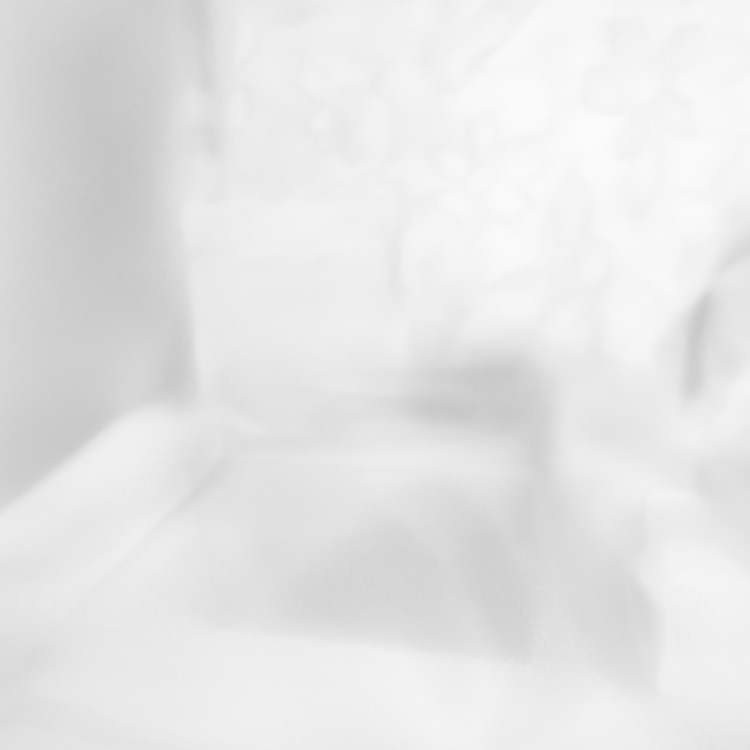} &
\includegraphics[width=\width]{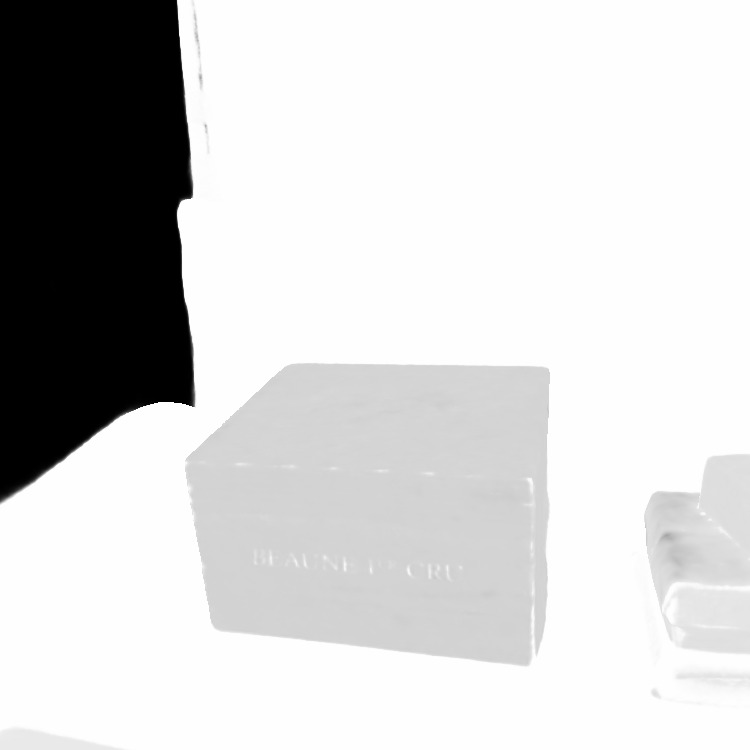} &
 \\
{\makebox[5pt]{\rotatebox{90}{\tiny \hspace{3pt} Sill}}} &
\includegraphics[width=\width]{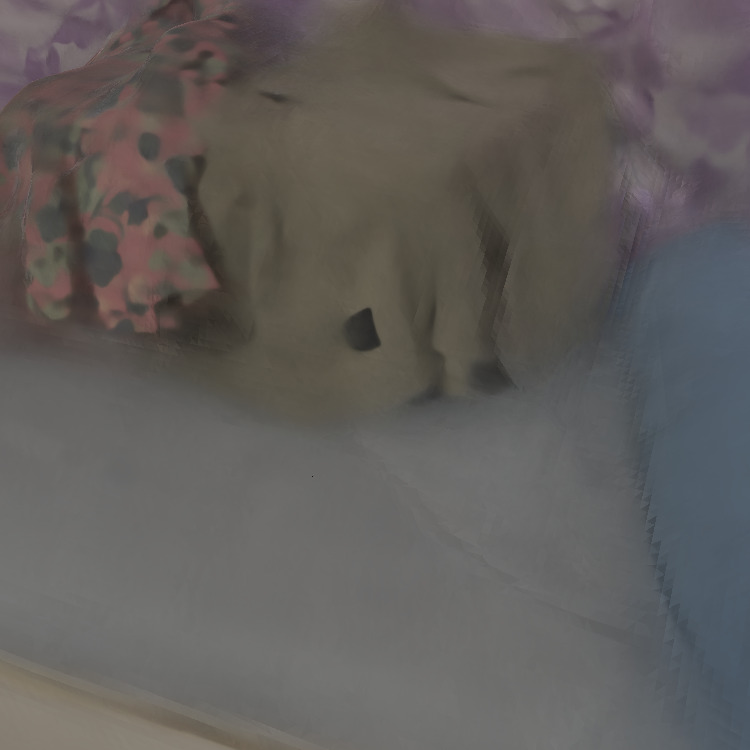} &
\includegraphics[width=\width]{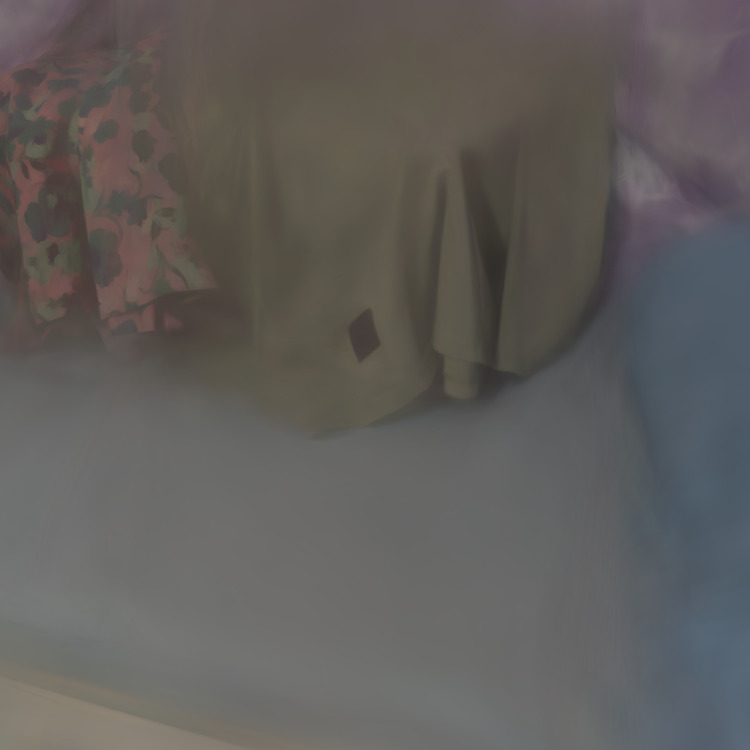} &
\includegraphics[width=\width]{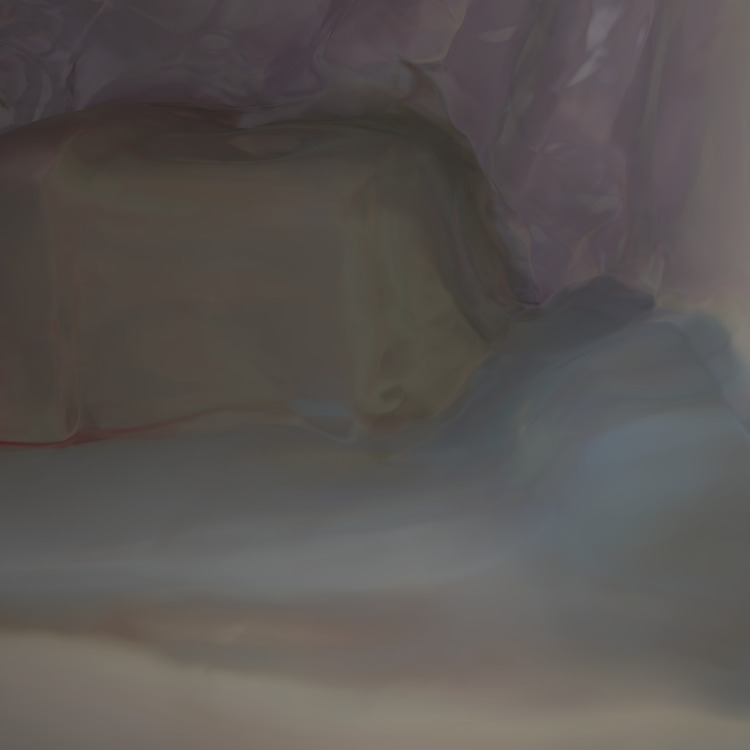} &
\includegraphics[width=\width]{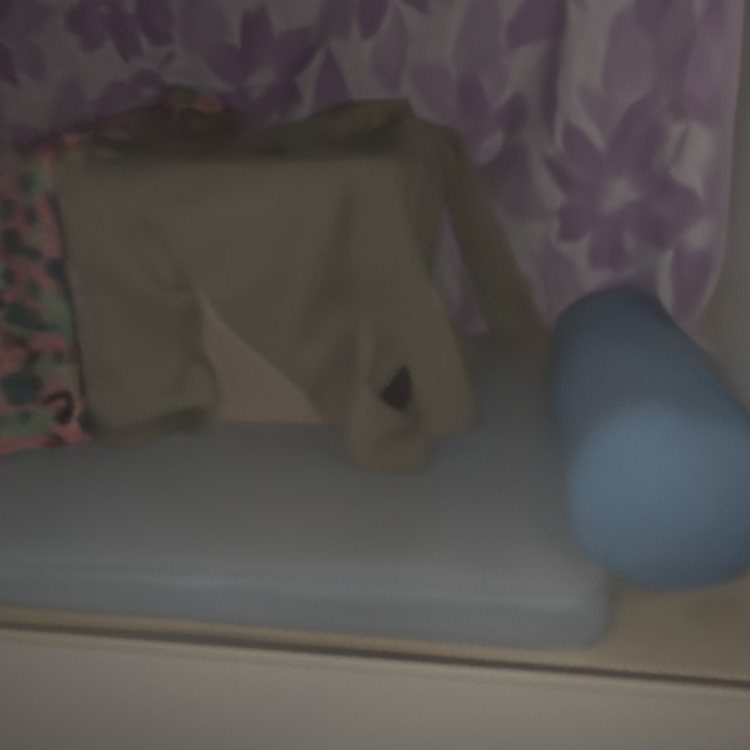} &
\includegraphics[width=\width]{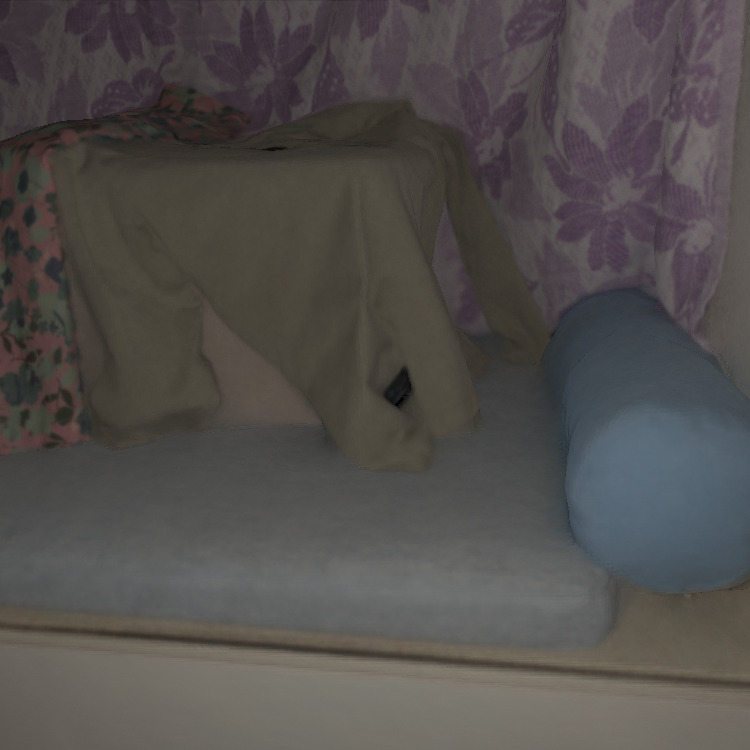} &
\includegraphics[width=\width]{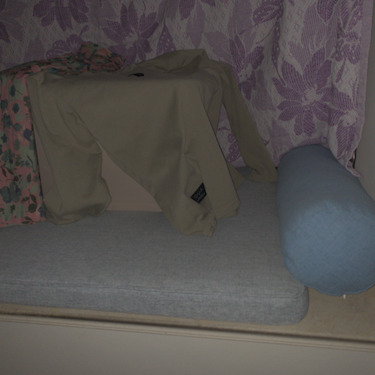} \\
{\makebox[5pt]{\rotatebox{90}{\tiny Albedo}}} &
\includegraphics[width=\width]{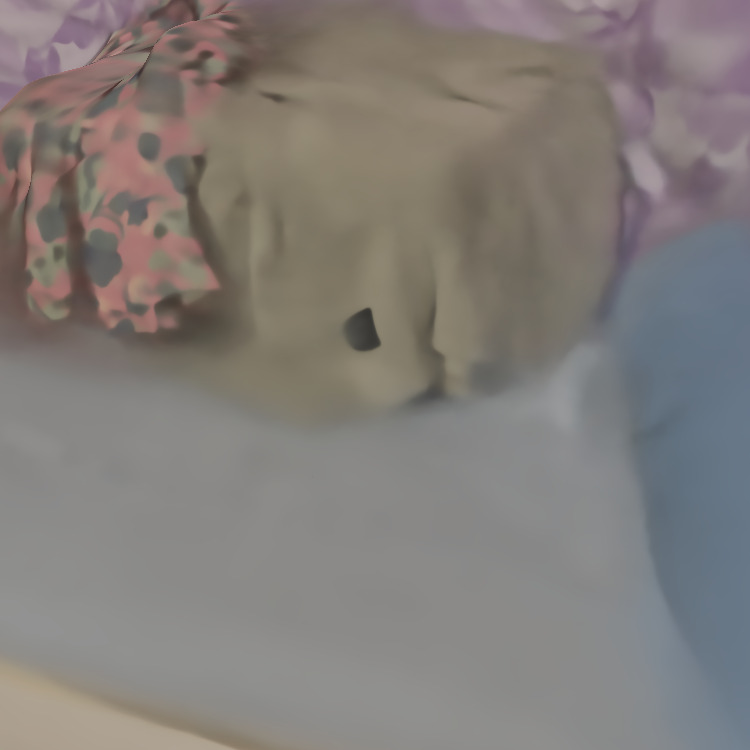} &
\includegraphics[width=\width]{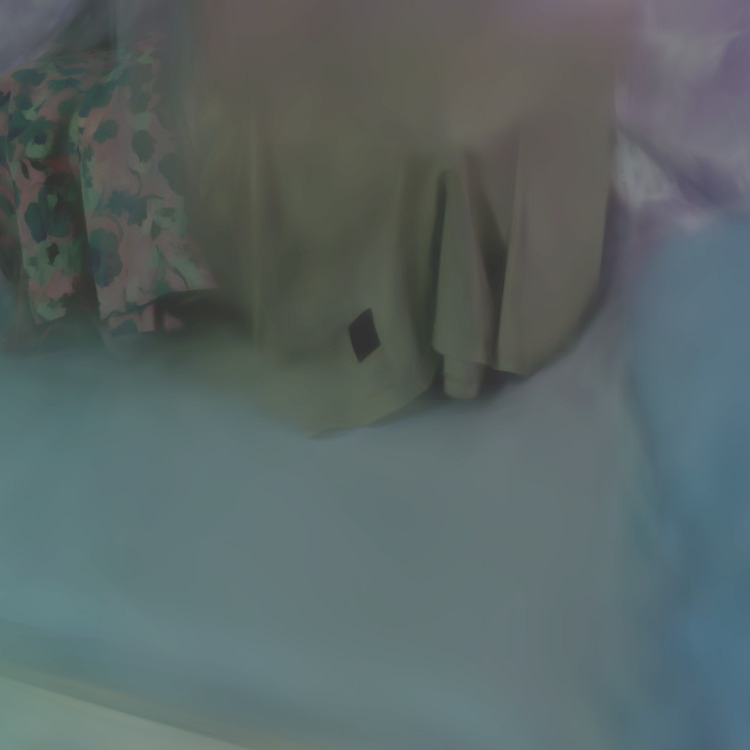} &
\includegraphics[width=\width]{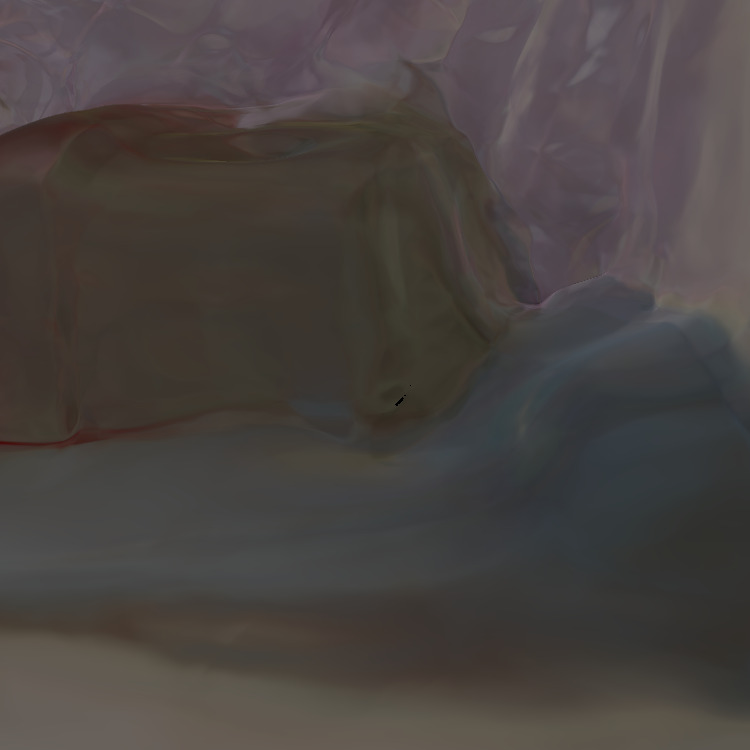} &
\includegraphics[width=\width]{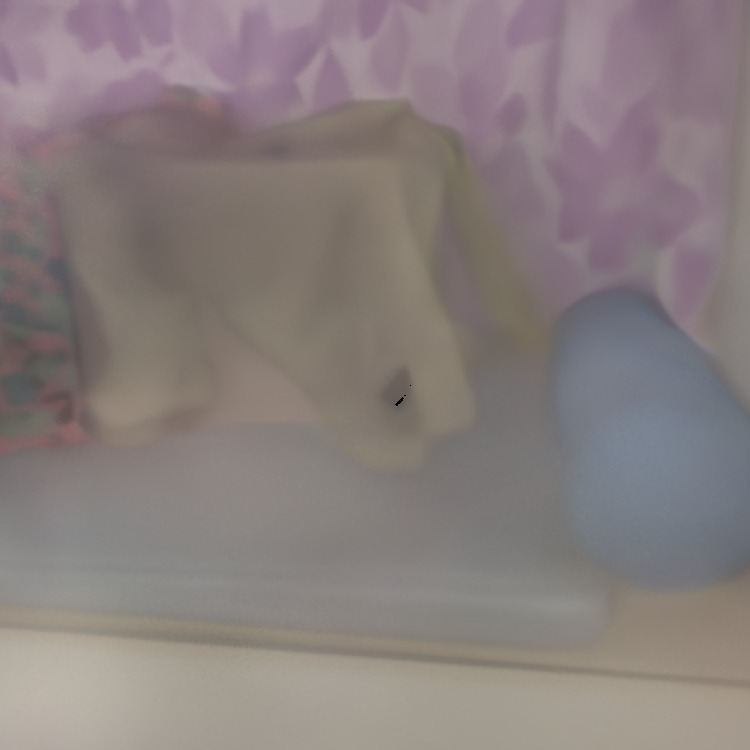} &
\includegraphics[width=\width]{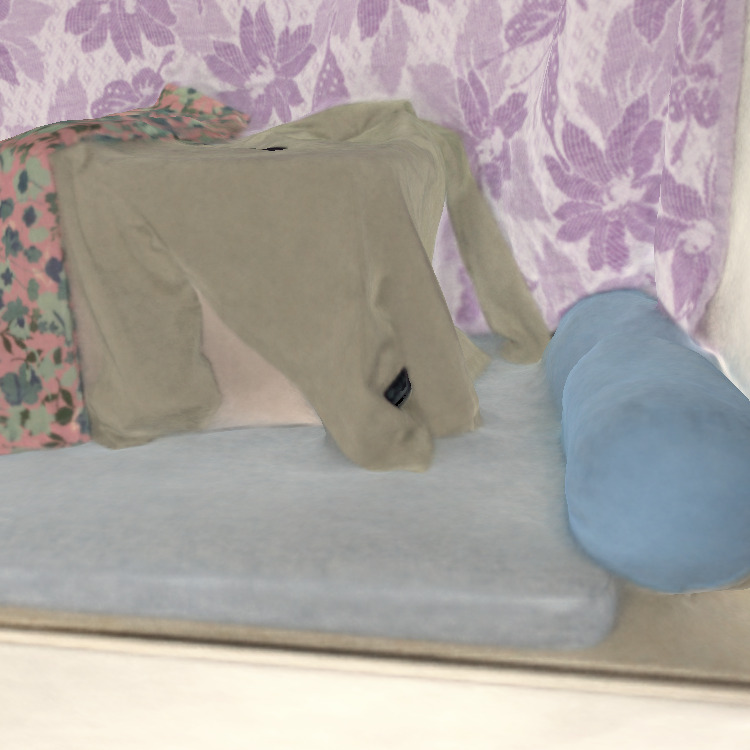} &
 \\
{\makebox[5pt]{\rotatebox{90}{\tiny Roughness}}} &
\includegraphics[width=\width]{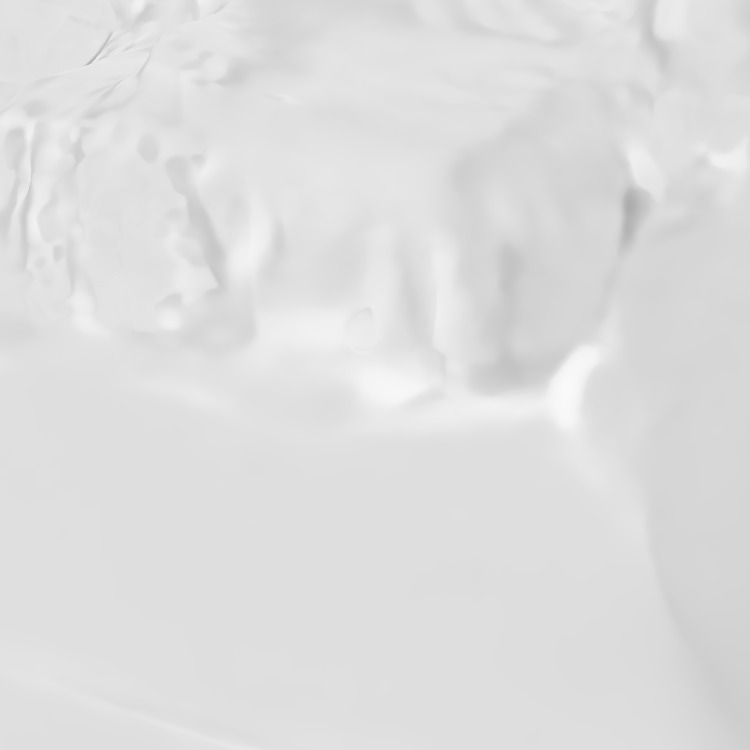} &
\includegraphics[width=\width]{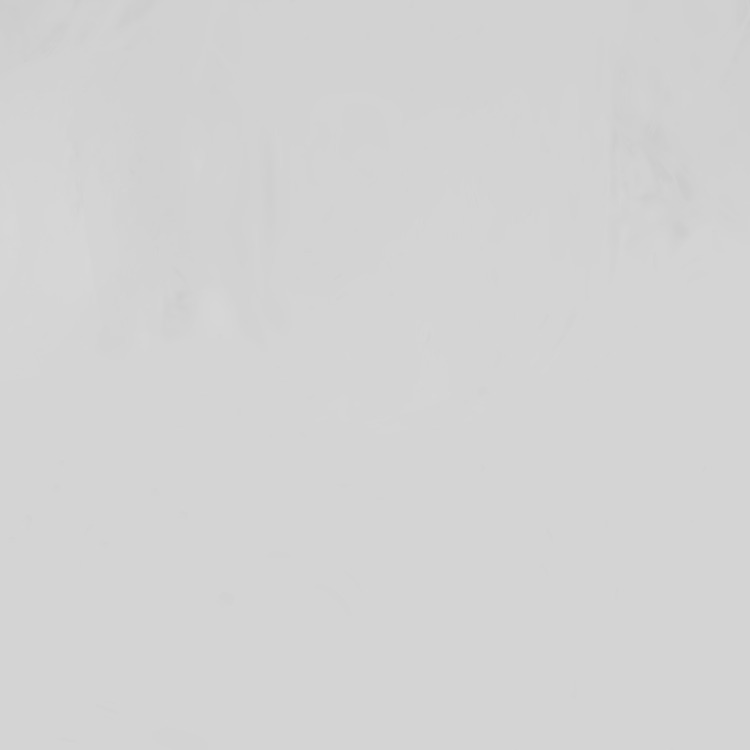} &
\includegraphics[width=\width]{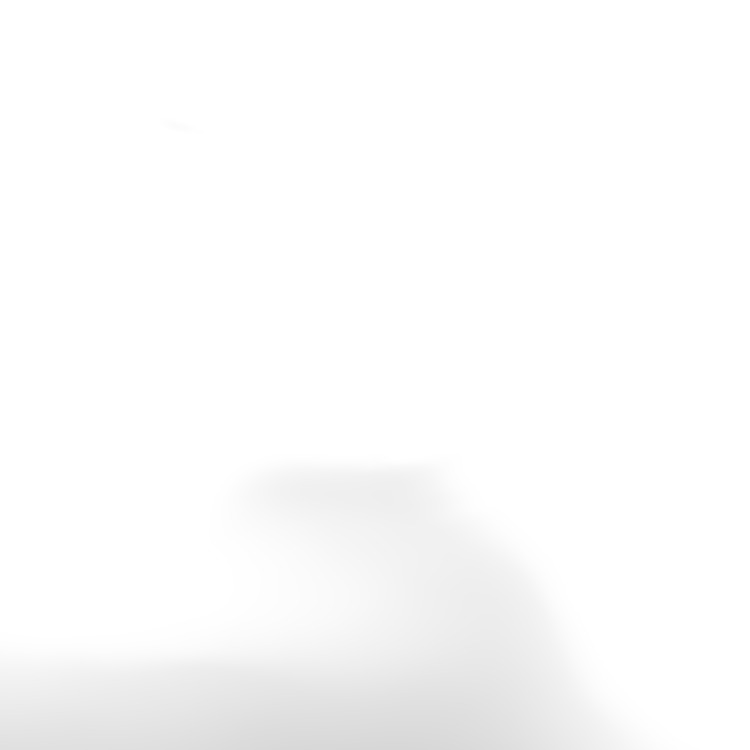} &
\includegraphics[width=\width]{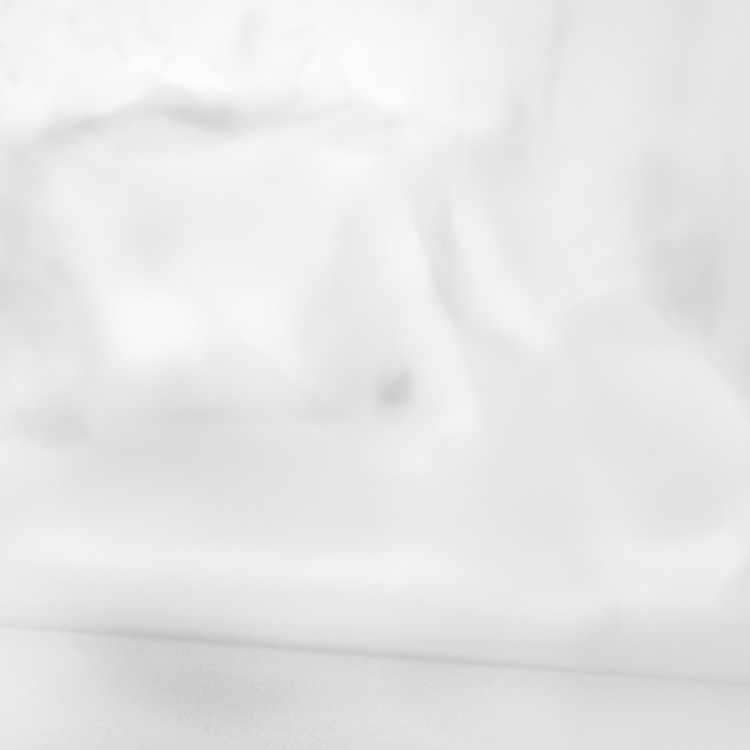} &
\includegraphics[width=\width]{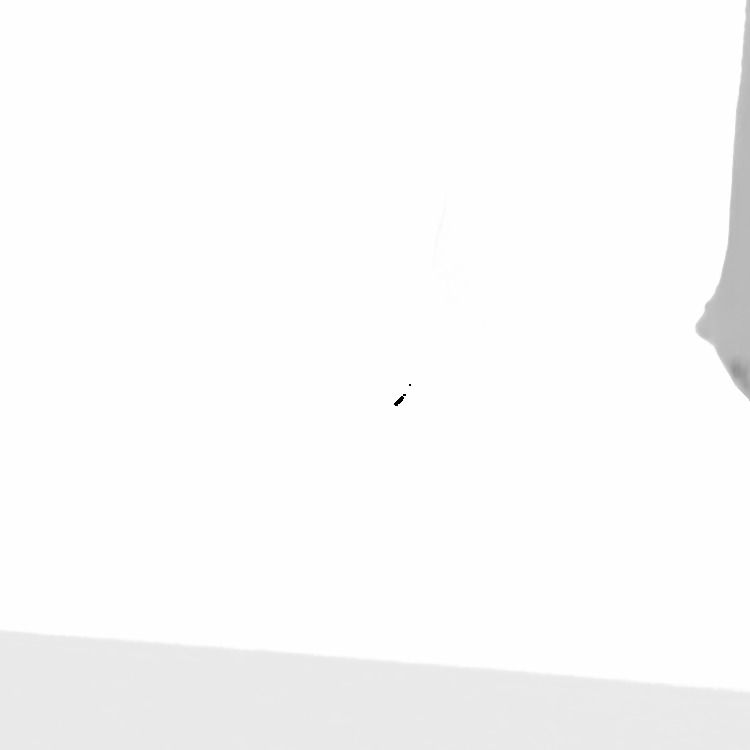} &
 \\

%% file: generated/suppl_qualitative_generated_real_3.tex
{\makebox[5pt]{\rotatebox{90}{\tiny \hspace{0pt} Coffee Table}}} &
\includegraphics[width=\width]{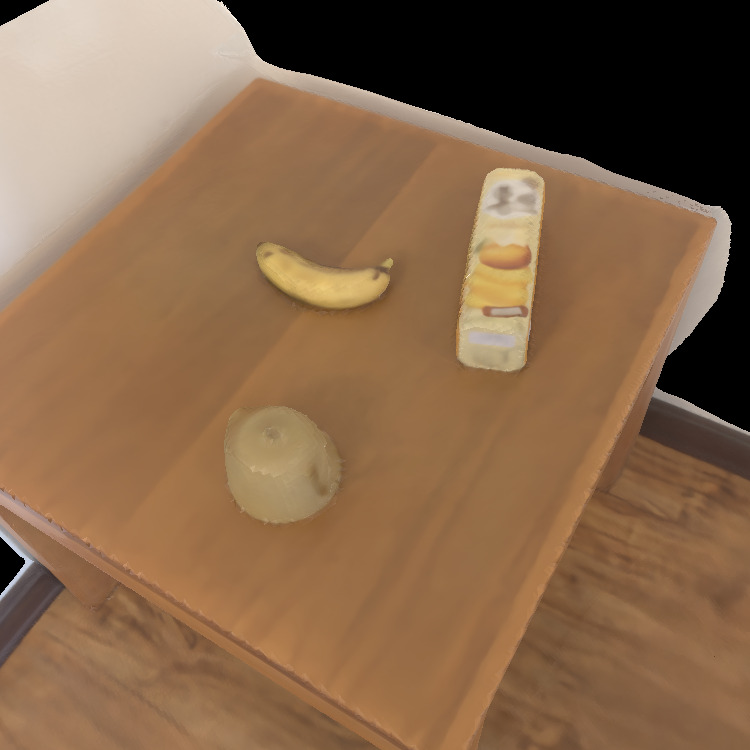} &
\includegraphics[width=\width]{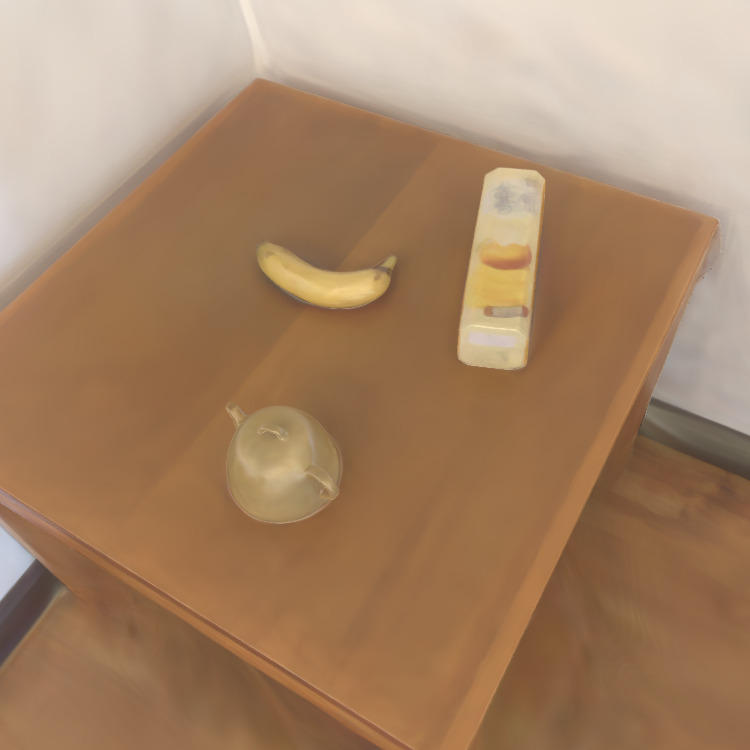} &
\includegraphics[width=\width]{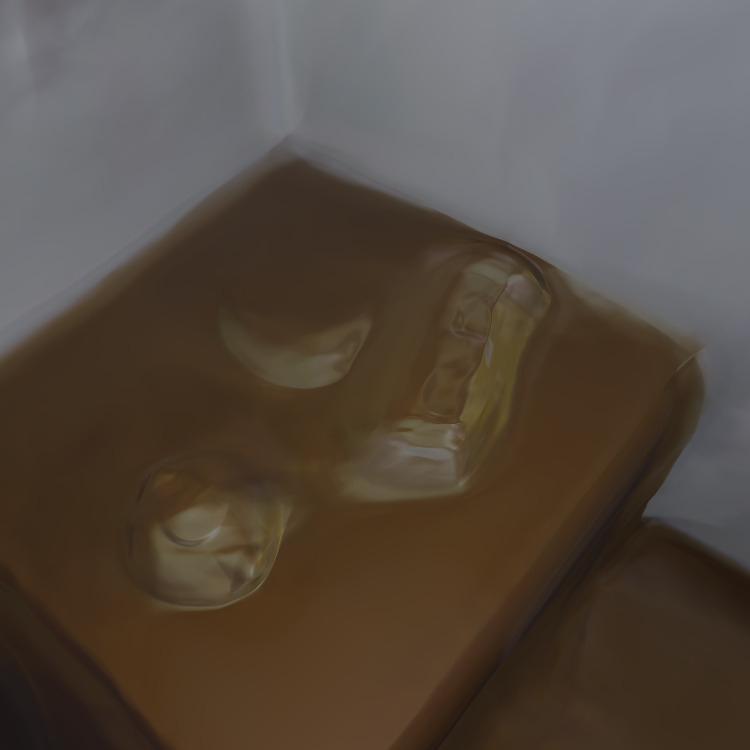} &
\includegraphics[width=\width]{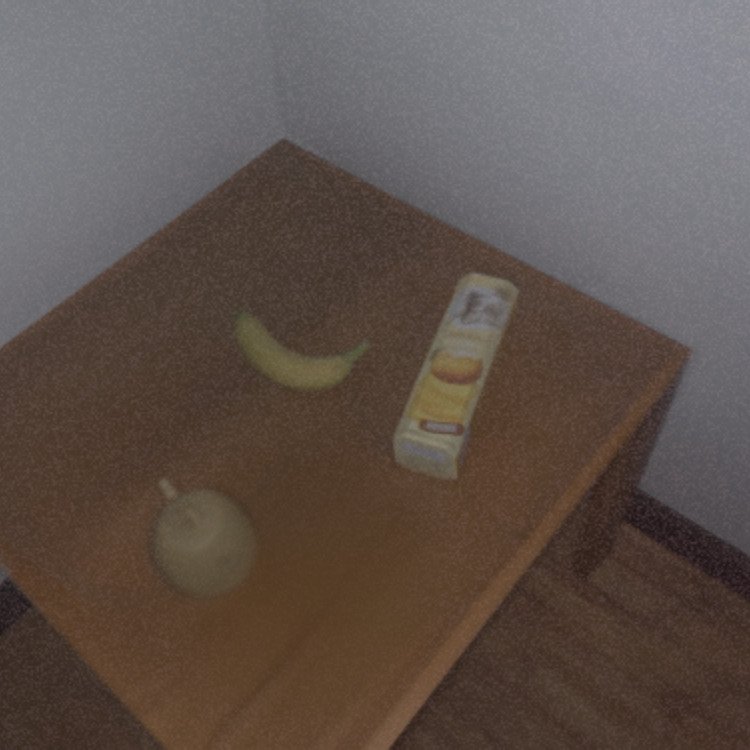} &
\includegraphics[width=\width]{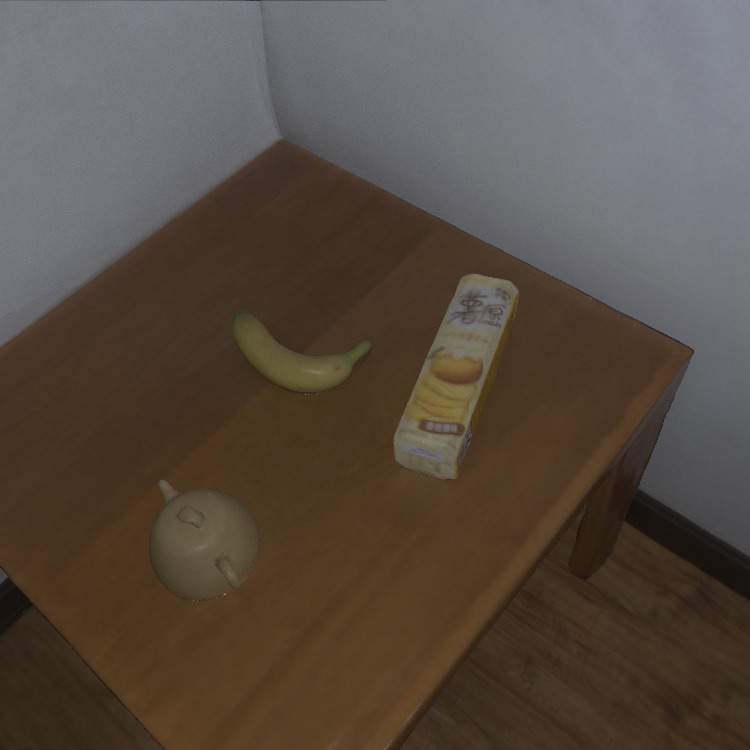} &
\includegraphics[width=\width]{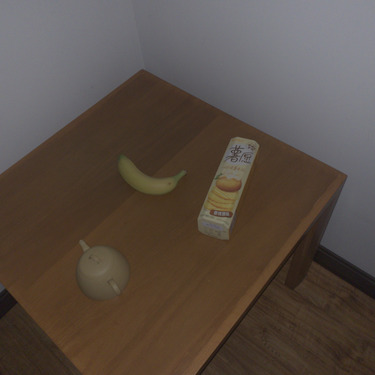} \\
{\makebox[5pt]{\rotatebox{90}{\tiny Albedo}}} &
\includegraphics[width=\width]{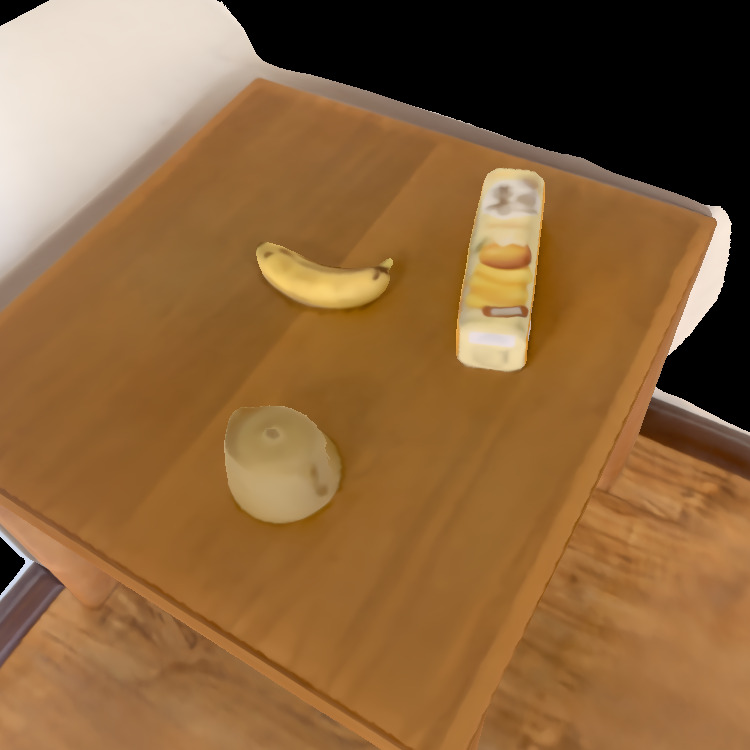} &
\includegraphics[width=\width]{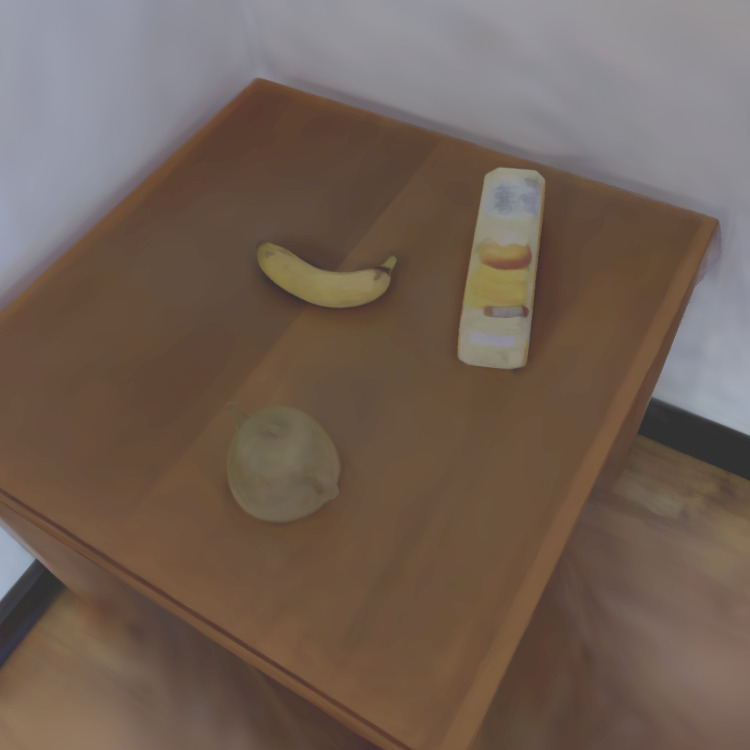} &
\includegraphics[width=\width]{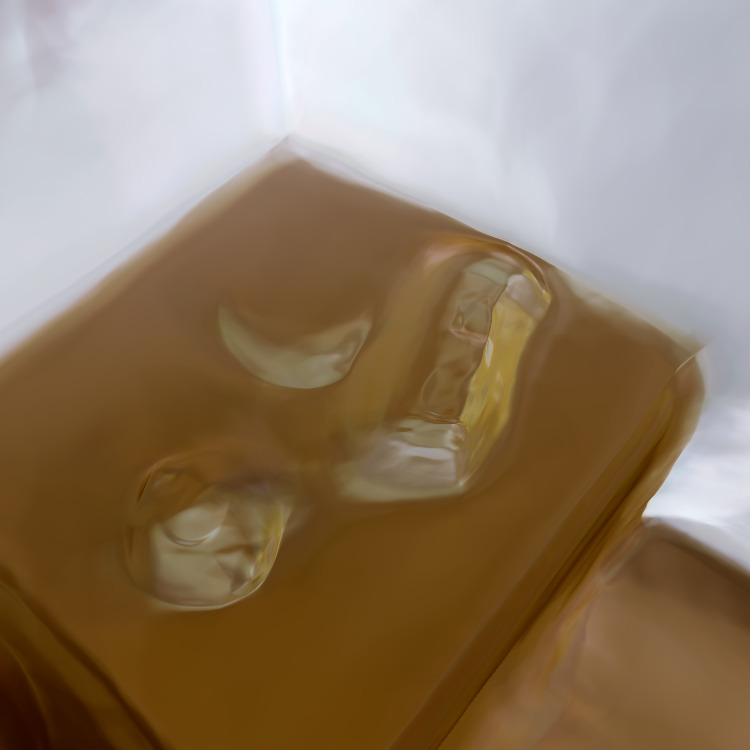} &
\includegraphics[width=\width]{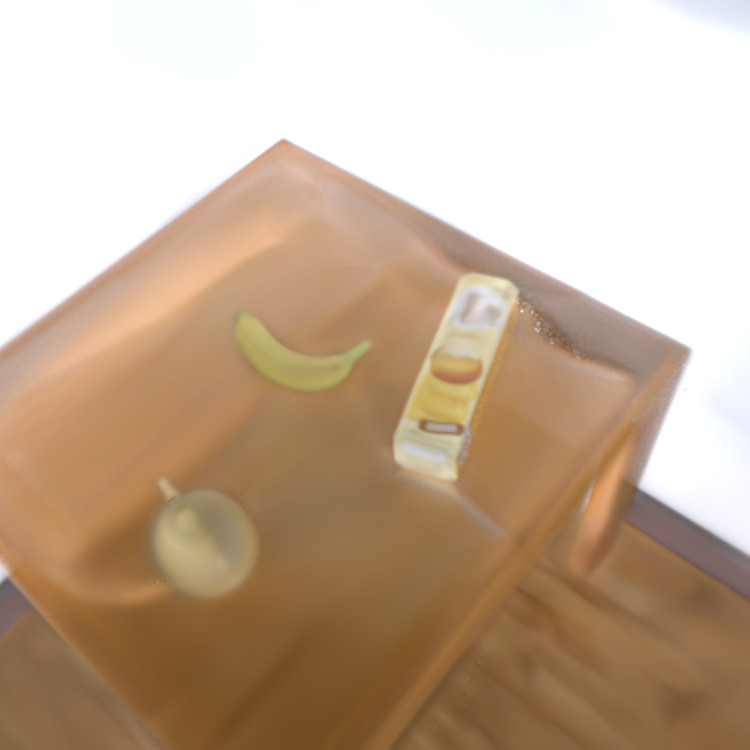} &
\includegraphics[width=\width]{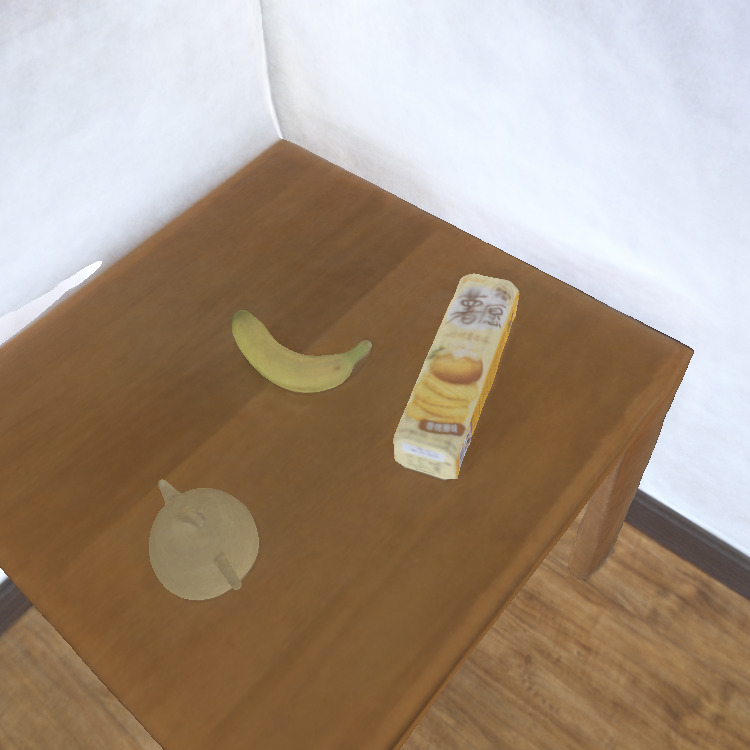} &
 \\
{\makebox[5pt]{\rotatebox{90}{\tiny Roughness}}} &
\includegraphics[width=\width]{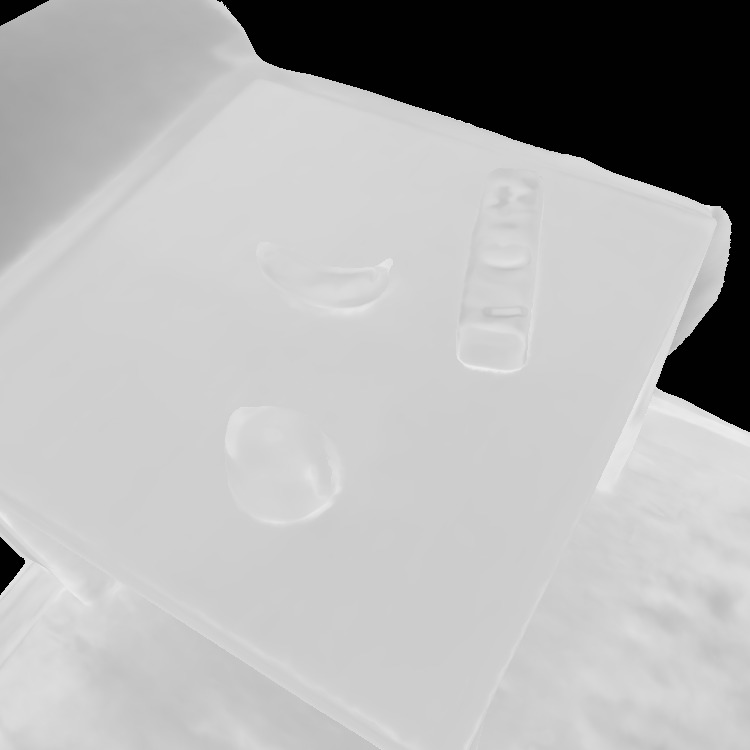} &
\includegraphics[width=\width]{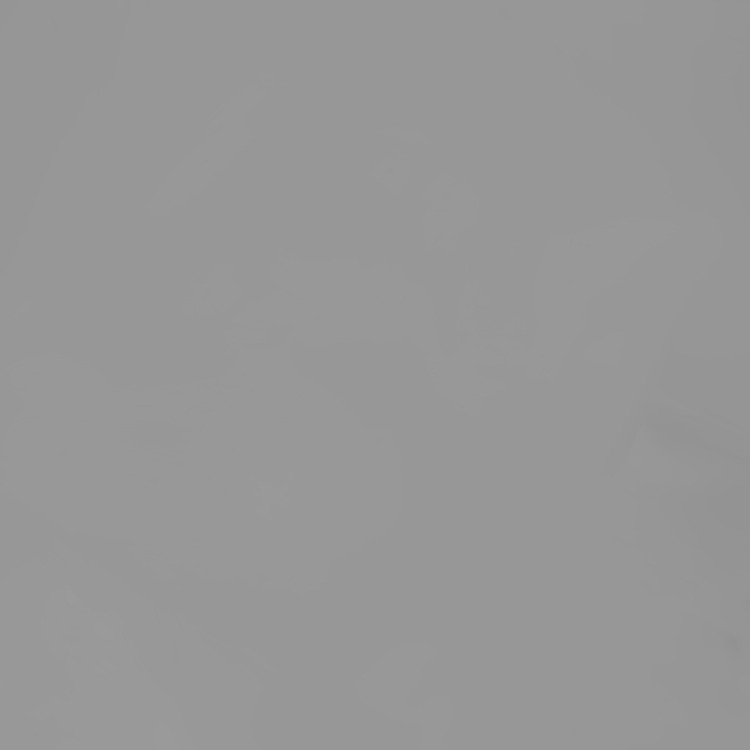} &
\includegraphics[width=\width]{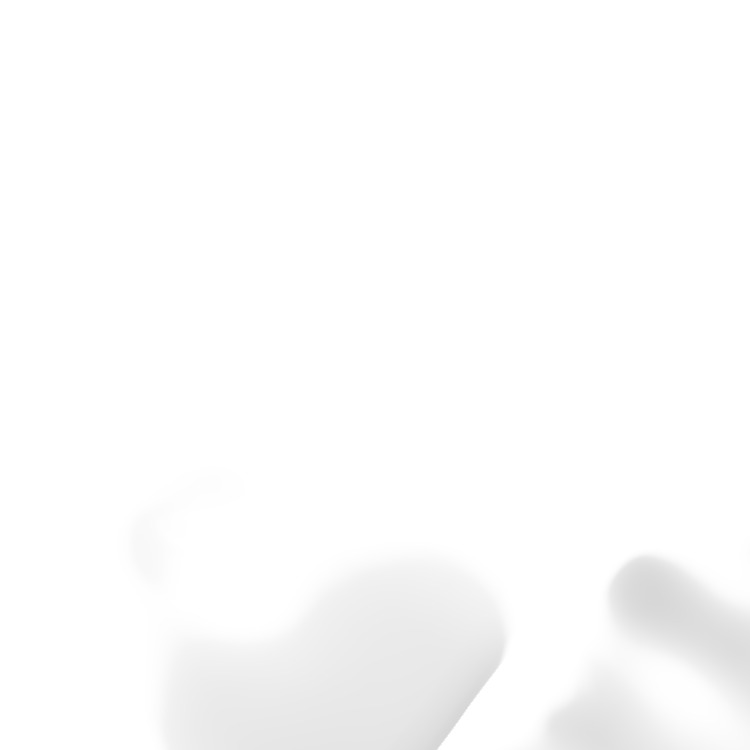} &
\includegraphics[width=\width]{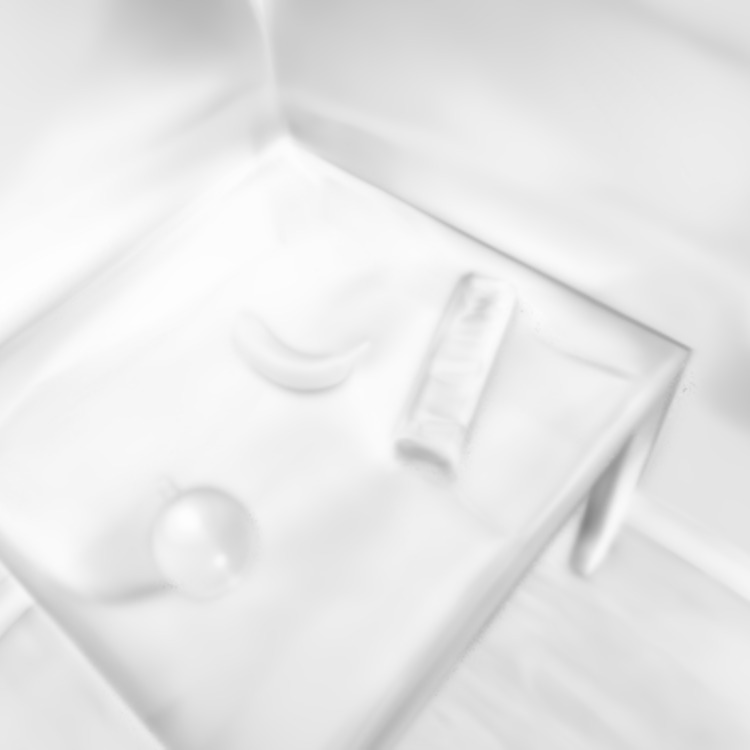} &
\includegraphics[width=\width]{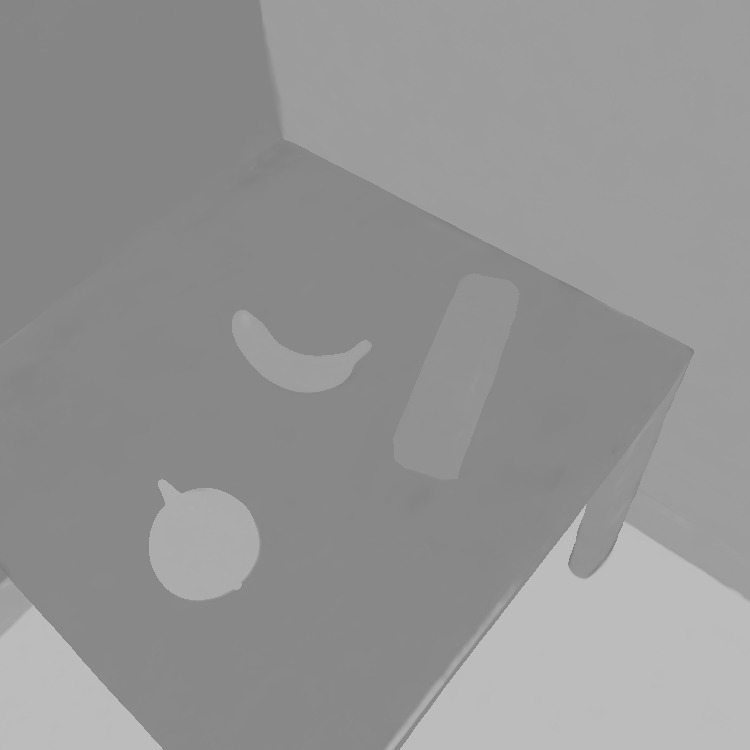} &
 \\
{\makebox[5pt]{\rotatebox{90}{\tiny \hspace{0pt} Coffee Table}}} &
\includegraphics[width=\width]{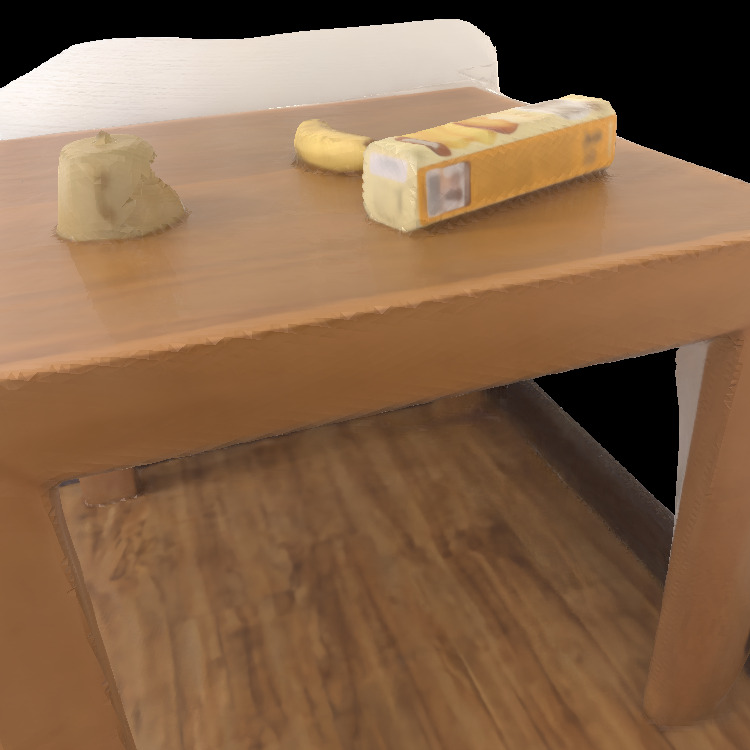} &
\includegraphics[width=\width]{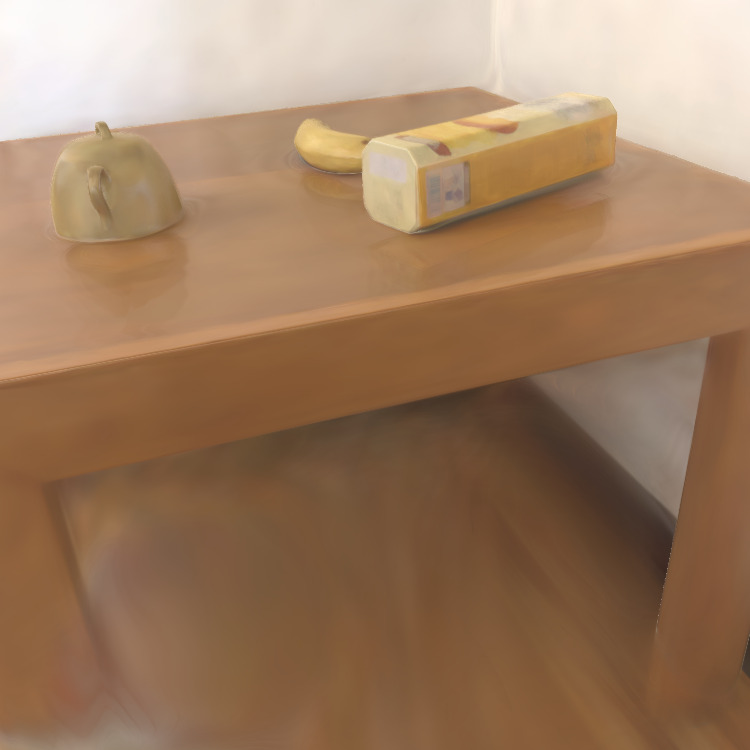} &
\includegraphics[width=\width]{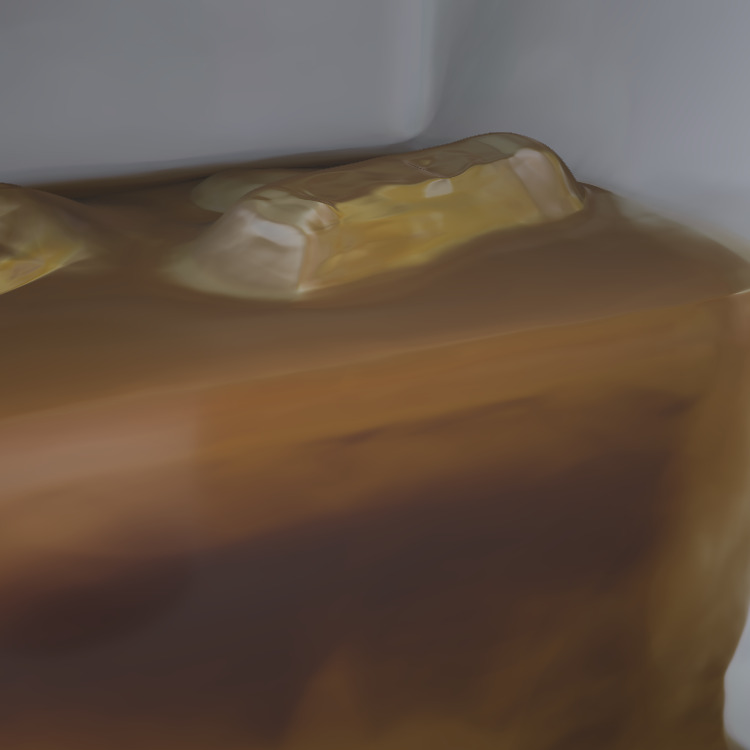} &
\includegraphics[width=\width]{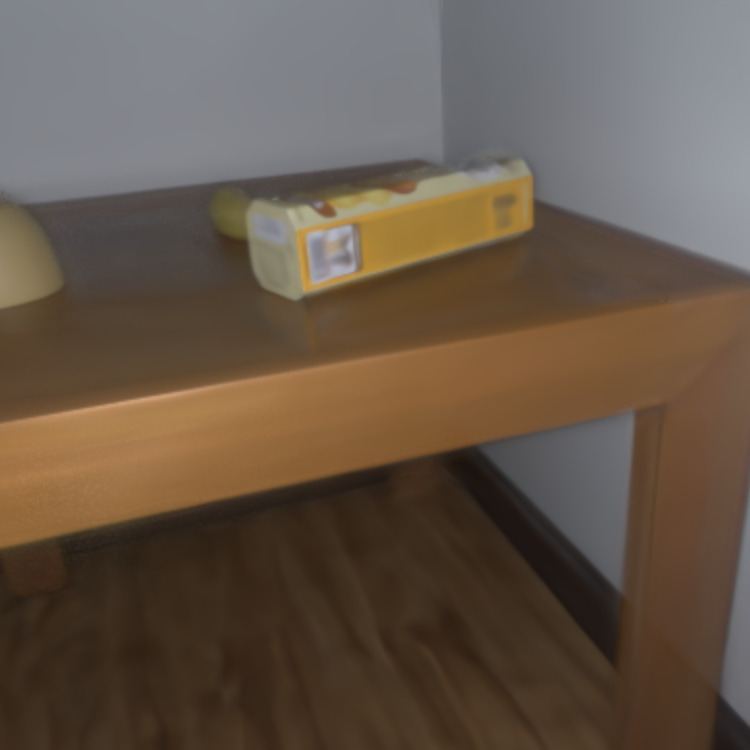} &
\includegraphics[width=\width]{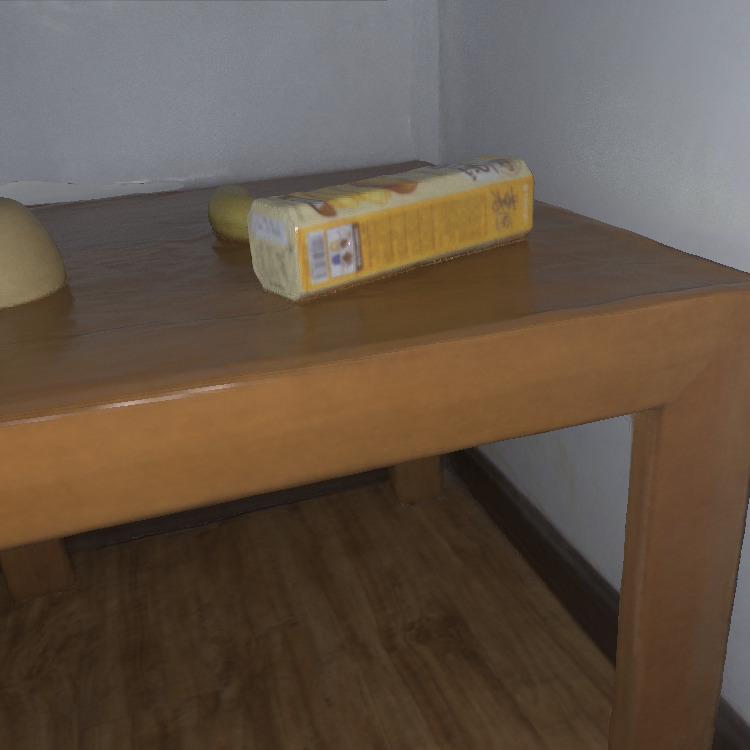} &
\includegraphics[width=\width]{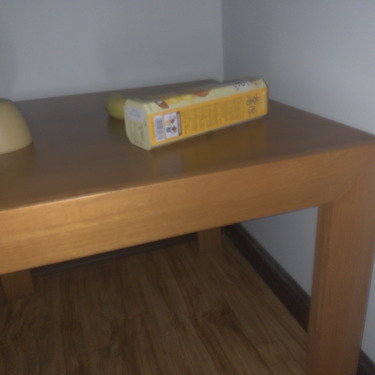} \\
{\makebox[5pt]{\rotatebox{90}{\tiny Albedo}}} &
\includegraphics[width=\width]{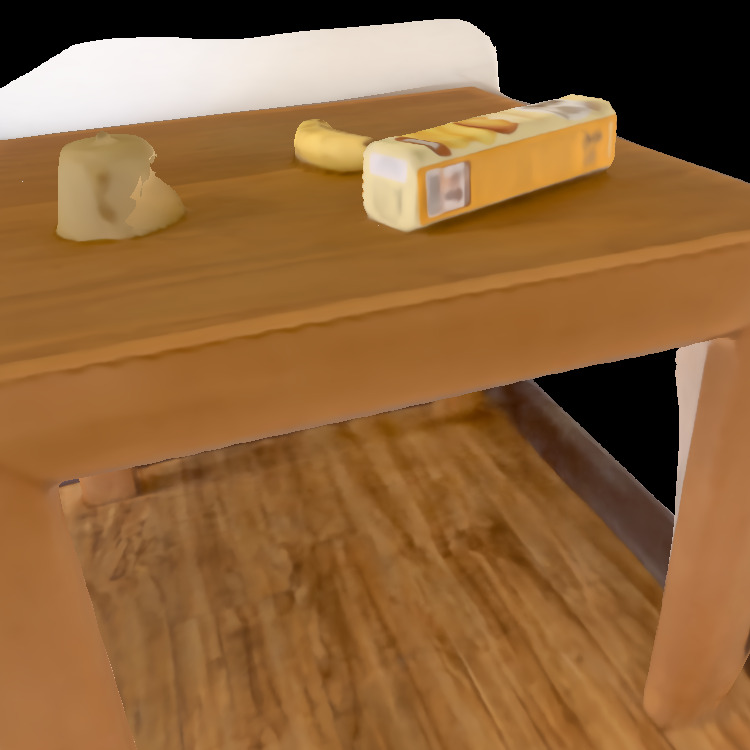} &
\includegraphics[width=\width]{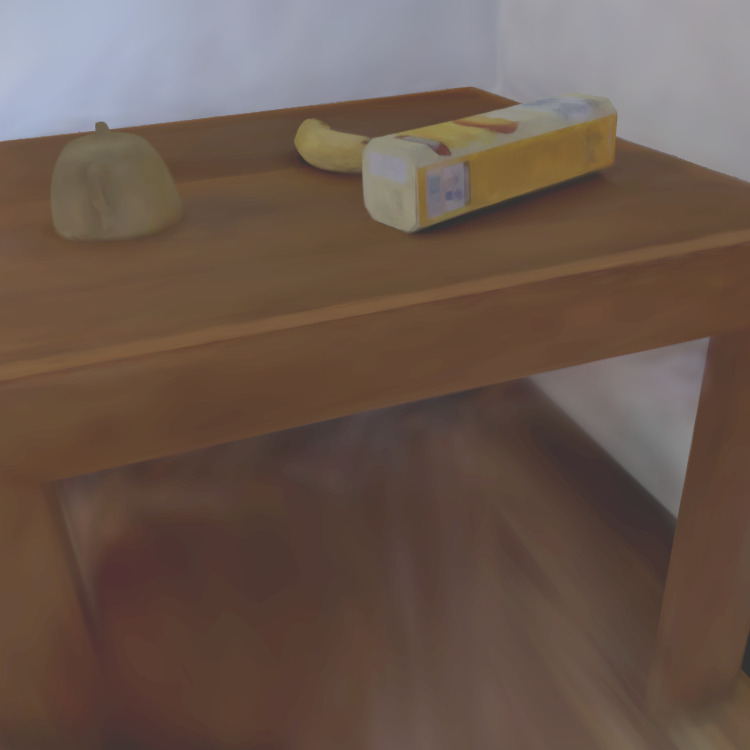} &
\includegraphics[width=\width]{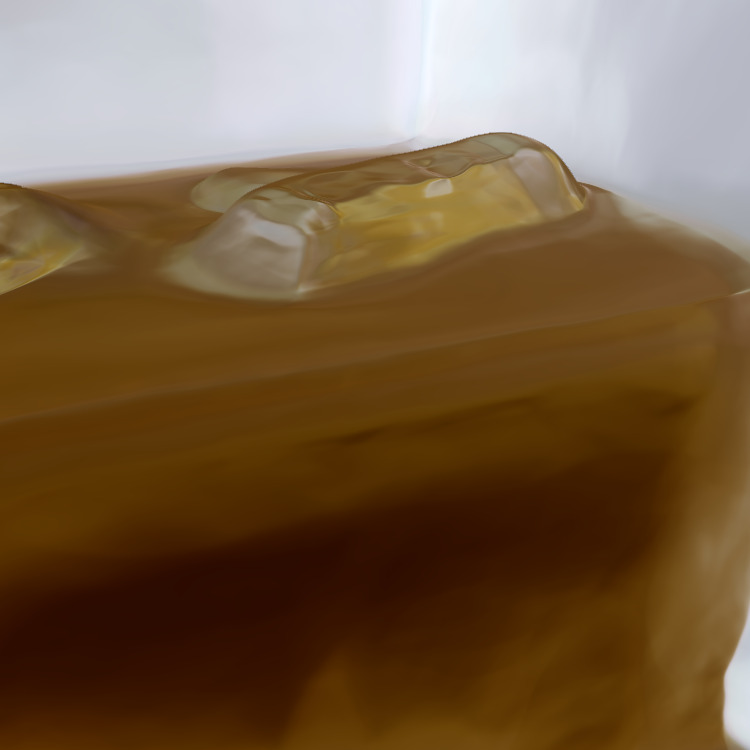} &
\includegraphics[width=\width]{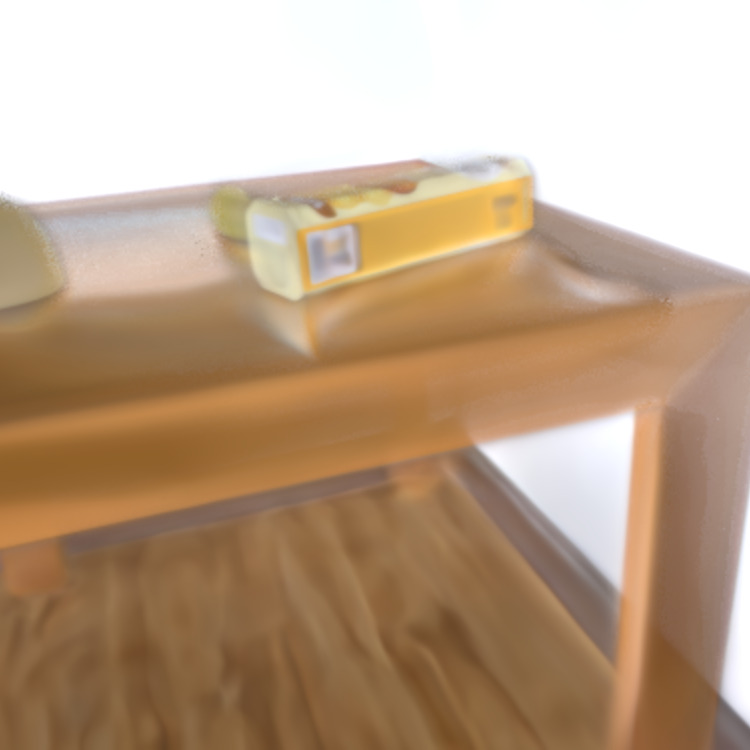} &
\includegraphics[width=\width]{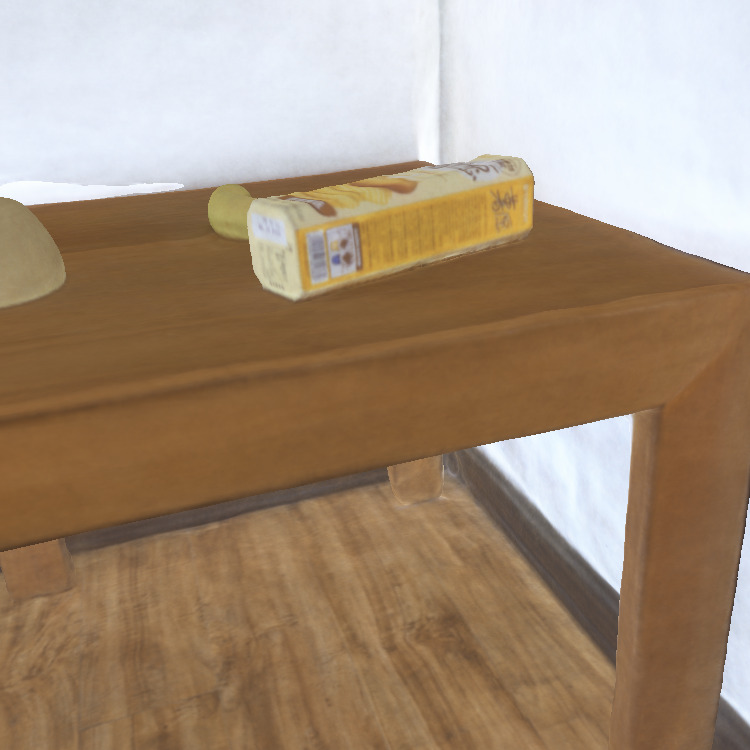} &
 \\
{\makebox[5pt]{\rotatebox{90}{\tiny Roughness}}} &
\includegraphics[width=\width]{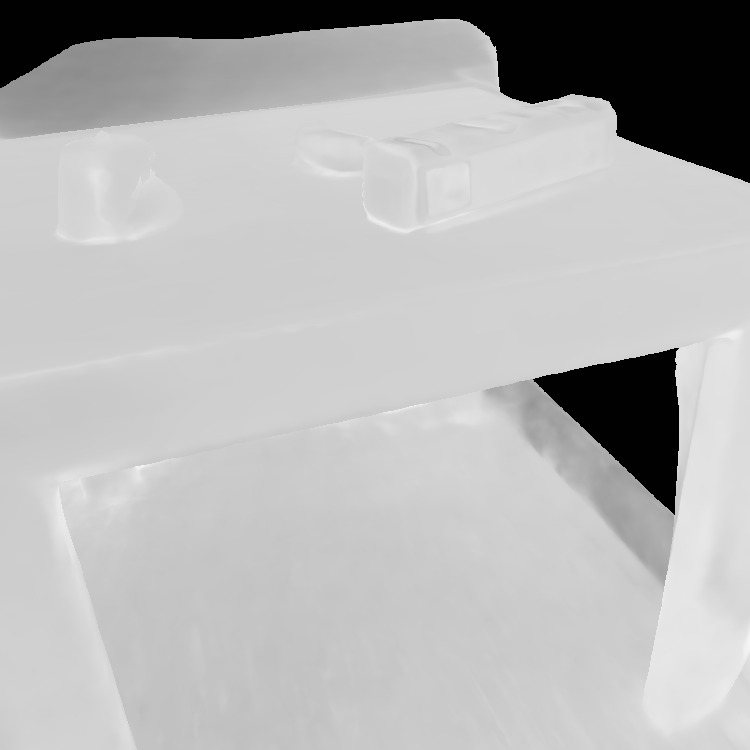} &
\includegraphics[width=\width]{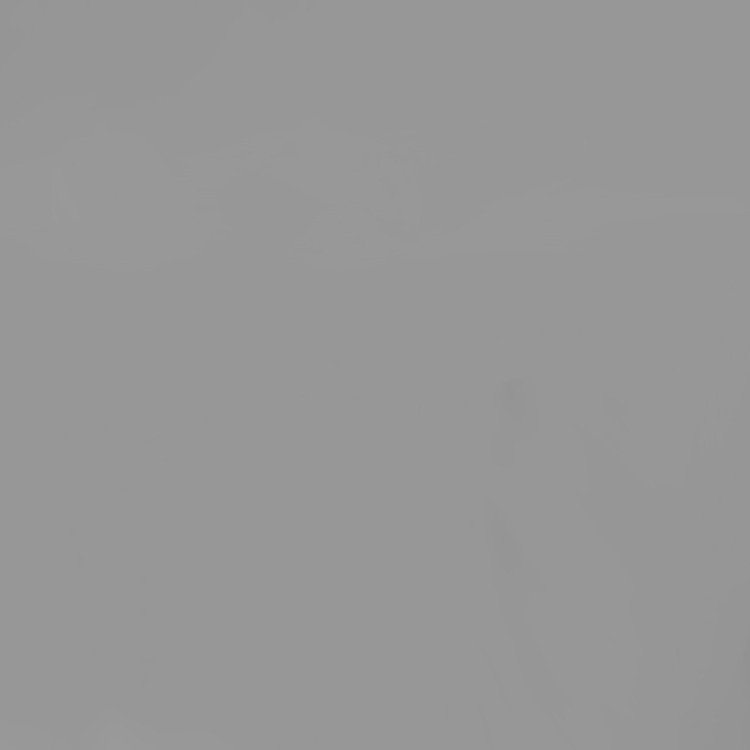} &
\includegraphics[width=\width]{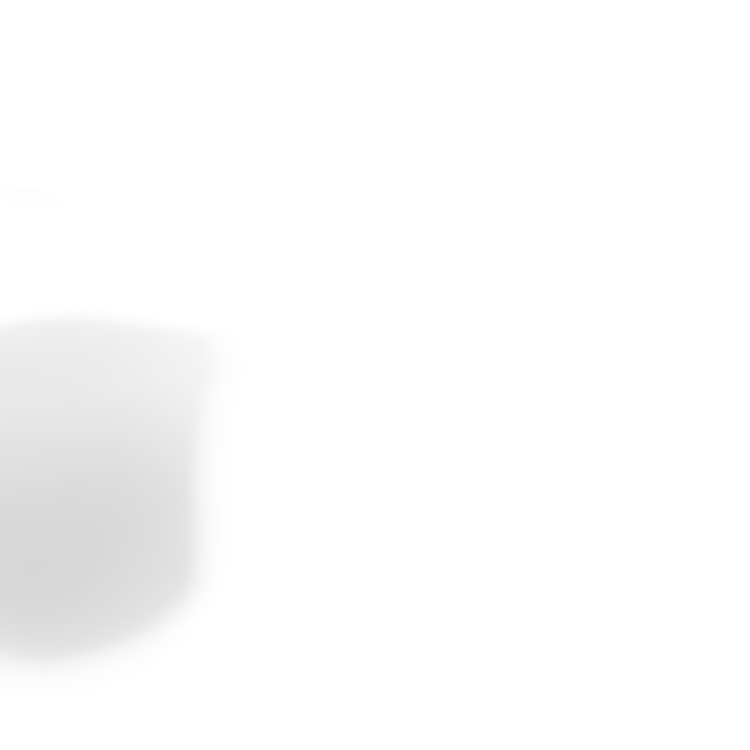} &
\includegraphics[width=\width]{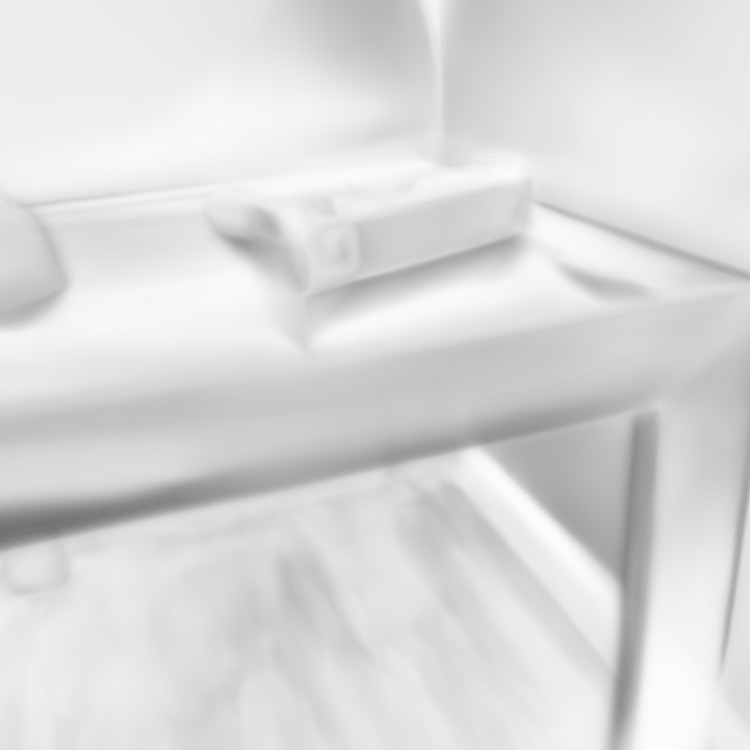} &
\includegraphics[width=\width]{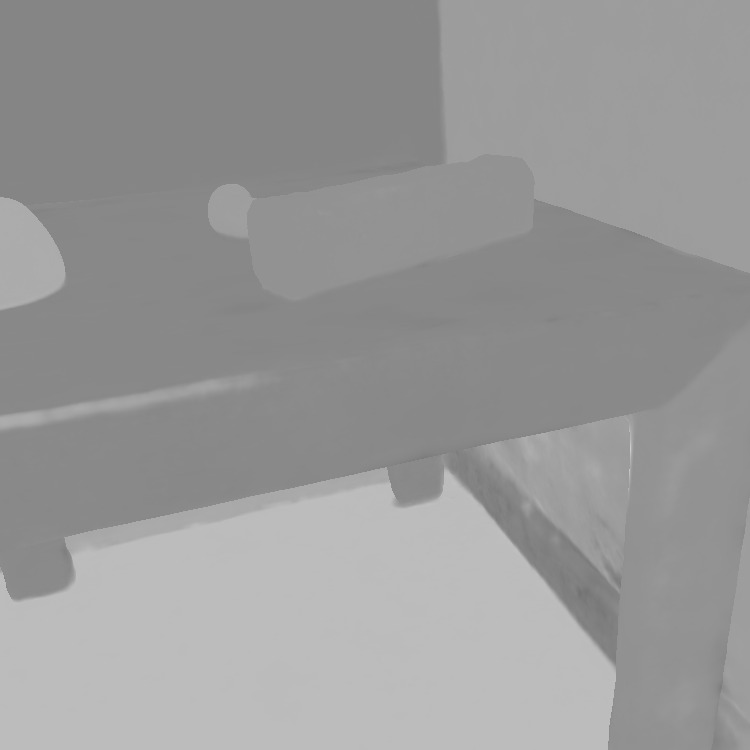} &
 \\

%% file: main.bbl
\begin{thebibliography}{62}
\providecommand{\natexlab}[1]{#1}
\providecommand{\url}[1]{\texttt{#1}}
\expandafter\ifx\csname urlstyle\endcsname\relax
  \providecommand{\doi}[1]{doi: #1}\else
  \providecommand{\doi}{doi: \begingroup \urlstyle{rm}\Url}\fi

\bibitem[Azinovic et~al.(2019)Azinovic, Li, Kaplanyan, and
  NieBner]{azinovicInversePathTracing2019a}
Dejan Azinovic, Tzu-Mao Li, Anton Kaplanyan, and Matthias NieBner.
\newblock Inverse {{Path Tracing}} for {{Joint Material}} and {{Lighting
  Estimation}}.
\newblock In \emph{2019 {{IEEE}}/{{CVF Conference}} on {{Computer Vision}} and
  {{Pattern Recognition}} ({{CVPR}})}, pages 2442--2451, Long Beach, CA, USA,
  2019. IEEE.

\bibitem[Bi et~al.(2020)Bi, Xu, Sunkavalli, Kriegman, and
  Ramamoorthi]{biDeep3DCapture2020}
Sai Bi, Zexiang Xu, Kalyan Sunkavalli, David Kriegman, and Ravi Ramamoorthi.
\newblock Deep {{3D Capture}}: {{Geometry}} and {{Reflectance From Sparse
  Multi-View Images}}.
\newblock In \emph{2020 {{IEEE}}/{{CVF Conference}} on {{Computer Vision}} and
  {{Pattern Recognition}} ({{CVPR}})}, pages 5959--5968, Seattle, WA, USA,
  2020. IEEE.

\bibitem[Boss et~al.()Boss, Engelhardt, Kar, Li, Sun, Barron, Lensch, and
  Jampani]{bossSAMURAIShapeMaterial}
Mark Boss, Andreas Engelhardt, Abhishek Kar, Yuanzhen Li, Deqing Sun,
  Jonathan~T Barron, Hendrik P~A Lensch, and Varun Jampani.
\newblock {{SAMURAI}}: {{Shape And Material}} from {{Unconstrained Real-world
  Arbitrary Image}} collections.

\bibitem[Boss et~al.(2021{\natexlab{a}})Boss, Braun, Jampani, Barron, Liu, and
  Lensch]{bossNeRDNeuralReflectance2021}
Mark Boss, Raphael Braun, Varun Jampani, Jonathan~T. Barron, Ce Liu, and
  Hendrik~P.A. Lensch.
\newblock {{NeRD}}: {{Neural Reflectance Decomposition}} from {{Image
  Collections}}.
\newblock In \emph{2021 {{IEEE}}/{{CVF International Conference}} on {{Computer
  Vision}} ({{ICCV}})}, pages 12664--12674, Montreal, QC, Canada,
  2021{\natexlab{a}}. IEEE.

\bibitem[Boss et~al.(2021{\natexlab{b}})Boss, Jampani, Braun, Liu, Barron, and
  Lensch]{bossNeuralPILNeuralPreIntegrated2021}
Mark Boss, Varun Jampani, Raphael Braun, Ce Liu, Jonathan Barron, and
  Hendrik~PA Lensch.
\newblock Neural-{{PIL}}: {{Neural Pre-Integrated Lighting}} for {{Reflectance
  Decomposition}}.
\newblock In \emph{Advances in {{Neural Information Processing Systems}}},
  pages 10691--10704. Curran Associates, Inc., 2021{\natexlab{b}}.

\bibitem[Burley()]{burleyPhysicallyBasedShading}
Brent Burley.
\newblock Physically {{Based Shading}} at {{Disney}}.

\bibitem[Cheng et~al.(2023)Cheng, Li, and
  Li]{chengWildLightInthewildInverse2023}
Ziang Cheng, Junxuan Li, and Hongdong Li.
\newblock {{WildLight}}: {{In-the-wild Inverse Rendering}} with a
  {{Flashlight}}.
\newblock In \emph{2023 {{IEEE}}/{{CVF Conference}} on {{Computer Vision}} and
  {{Pattern Recognition}} ({{CVPR}})}, pages 4305--4314, Vancouver, BC, Canada,
  2023. IEEE.

\bibitem[Chung et~al.()Chung, Choi, and
  Baek]{chungDifferentiableInverseRendering}
Hoon-Gyu Chung, Seokjun Choi, and Seung-Hwan Baek.
\newblock Differentiable {{Inverse Rendering}} with {{Interpretable Basis
  BRDFs}}.

\bibitem[Cui et~al.()Cui, Gu, Shi, Tan, and
  Kautz]{cuiPolarimetricMultiViewStereo}
Zhaopeng Cui, Jinwei Gu, Boxin Shi, Ping Tan, and Jan Kautz.
\newblock Polarimetric {{Multi-View Stereo}}.

\bibitem[Engelhardt et~al.(2024)Engelhardt, Raj, Boss, Zhang, Kar, Li, Sun,
  Brualla, Barron, Lensch, and Jampani]{engelhardtSHINOBIShapeIllumination2024}
Andreas Engelhardt, Amit Raj, Mark Boss, Yunzhi Zhang, Abhishek Kar, Yuanzhen
  Li, Deqing Sun, Ricardo~Martin Brualla, Jonathan~T. Barron, Hendrik P.~A.
  Lensch, and Varun Jampani.
\newblock {{SHINOBI}}: {{Shape}} and {{Illumination}} using {{Neural Object
  Decomposition}} via {{BRDF Optimization In-the-wild}}.
\newblock In \emph{2024 {{IEEE}}/{{CVF Conference}} on {{Computer Vision}} and
  {{Pattern Recognition}} ({{CVPR}})}, pages 19636--19646, Seattle, WA, USA,
  2024. IEEE.

\bibitem[Ge et~al.(2023)Ge, Hu, Zhao, Liu, and
  Chen]{geRefNeuSAmbiguityReducedNeural2023}
Wenhang Ge, Tao Hu, Haoyu Zhao, Shu Liu, and Ying-Cong Chen.
\newblock Ref-{{NeuS}}: {{Ambiguity-Reduced Neural Implicit Surface Learning}}
  for {{Multi-View Reconstruction}} with {{Reflection}}.
\newblock In \emph{2023 {{IEEE}}/{{CVF International Conference}} on {{Computer
  Vision}} ({{ICCV}})}, pages 4228--4237, Paris, France, 2023. IEEE.

\bibitem[Georgoulis et~al.(2014)Georgoulis, Proesmans, and
  Van~Gool]{georgoulisTacklingShapesBRDFs2014}
Stamatios Georgoulis, Marc Proesmans, and Luc Van~Gool.
\newblock Tackling {{Shapes}} and {{BRDFs Head-On}}.
\newblock In \emph{2014 2nd {{International Conference}} on {{3D Vision}}},
  pages 267--274, Tokyo, 2014. IEEE.

\bibitem[Goldman and Chen(2005)]{goldmanVignetteExposureCalibration2005}
D.B. Goldman and Jiun-Hung Chen.
\newblock Vignette and exposure calibration and compensation.
\newblock In \emph{Tenth {{IEEE International Conference}} on {{Computer
  Vision}} ({{ICCV}}'05) {{Volume}} 1}, pages 899--906 Vol. 1, 2005.

\bibitem[Gu et~al.(2025)Gu, Wei, Zeng, Yao, and
  Zhang]{guIRGSInterReflectiveGaussian2025}
Chun Gu, Xiaofei Wei, Zixuan Zeng, Yuxuan Yao, and Li Zhang.
\newblock {{IRGS}}: {{Inter-Reflective Gaussian Splatting}} with {{2D Gaussian
  Ray Tracing}}, 2025.

\bibitem[Guo et~al.(2019)Guo, Lincoln, Davidson, Busch, Yu, Whalen, Harvey,
  {Orts-Escolano}, Pandey, Dourgarian, Tang, Tkach, Kowdle, Cooper, Dou,
  Fanello, Fyffe, Rhemann, Taylor, Debevec, and
  Izadi]{guoRelightablesVolumetricPerformance2019}
Kaiwen Guo, Peter Lincoln, Philip Davidson, Jay Busch, Xueming Yu, Matt Whalen,
  Geoff Harvey, Sergio {Orts-Escolano}, Rohit Pandey, Jason Dourgarian, Danhang
  Tang, Anastasia Tkach, Adarsh Kowdle, Emily Cooper, Mingsong Dou, Sean
  Fanello, Graham Fyffe, Christoph Rhemann, Jonathan Taylor, Paul Debevec, and
  Shahram Izadi.
\newblock The relightables: Volumetric performance capture of humans with
  realistic relighting.
\newblock \emph{ACM Transactions on Graphics}, 38\penalty0 (6):\penalty0 1--19,
  2019.

\bibitem[Hadadan et~al.(2021)Hadadan, Chen, and
  Zwicker]{hadadanNeuralRadiosity2021}
Saeed Hadadan, Shuhong Chen, and Matthias Zwicker.
\newblock Neural radiosity.
\newblock \emph{ACM Transactions on Graphics}, 40\penalty0 (6):\penalty0 1--11,
  2021.

\bibitem[Hadadan et~al.(2023)Hadadan, Lin, Nov{\'a}k, Rousselle, and
  Zwicker]{hadadanInverseGlobalIllumination2023}
Saeed Hadadan, Geng Lin, Jan Nov{\'a}k, Fabrice Rousselle, and Matthias
  Zwicker.
\newblock Inverse {{Global Illumination}} using a {{Neural Radiometric Prior}}.
\newblock In \emph{{{ACM SIGGRAPH}} 2023 {{Conference Proceedings}}}, pages
  1--11, New York, NY, USA, 2023. Association for Computing Machinery.

\bibitem[Hasselgren et~al.()Hasselgren, Hofmann, and
  Munkberg]{hasselgrenShapeLightMaterial}
Jon Hasselgren, Nikolai Hofmann, and Jacob Munkberg.
\newblock Shape, {{Light}}, and {{Material Decomposition}} from {{Images}}
  using {{Monte Carlo Rendering}} and {{Denoising}}.

\bibitem[Higo et~al.(2009)Higo, Matsushita, Joshi, and
  Ikeuchi]{higoHandheldPhotometricStereo2009}
Tomoaki Higo, Yasuyuki Matsushita, Neel Joshi, and Katsushi Ikeuchi.
\newblock A hand-held photometric stereo camera for 3-{{D}} modeling.
\newblock In \emph{2009 {{IEEE}} 12th {{International Conference}} on
  {{Computer Vision}}}, pages 1234--1241, 2009.

\bibitem[Jin et~al.()Jin, Liu, Xu, Zhang, Han, Bi, Zhou, Xu, and
  Su]{jinTensoIRTensorialInverse}
Haian Jin, Isabella Liu, Peijia Xu, Xiaoshuai Zhang, Songfang Han, Sai Bi,
  Xiaowei Zhou, Zexiang Xu, and Hao Su.
\newblock {{TensoIR}}: {{Tensorial Inverse Rendering}}.

\bibitem[Kajiya(1986)]{kajiyaRenderingEquation1986}
James~T. Kajiya.
\newblock The rendering equation.
\newblock \emph{SIGGRAPH Comput. Graph.}, 20\penalty0 (4):\penalty0 143--150,
  1986.

\bibitem[Li et~al.(2020{\natexlab{a}})Li, Zhou, Wu, Shi, Diao, and
  Tan]{liMultiViewPhotometricStereo2020}
Min Li, Zhenglong Zhou, Zhe Wu, Boxin Shi, Changyu Diao, and Ping Tan.
\newblock Multi-{{View Photometric Stereo}}: {{A Robust Solution}} and
  {{Benchmark Dataset}} for {{Spatially Varying Isotropic Materials}}.
\newblock \emph{IEEE Transactions on Image Processing}, 29:\penalty0
  4159--4173, 2020{\natexlab{a}}.

\bibitem[Li et~al.(2020{\natexlab{b}})Li, Shafiei, Ramamoorthi, Sunkavalli, and
  Chandraker]{liInverseRenderingComplex2020}
Zhengqin Li, Mohammad Shafiei, Ravi Ramamoorthi, Kalyan Sunkavalli, and
  Manmohan Chandraker.
\newblock Inverse {{Rendering}} for {{Complex Indoor Scenes}}: {{Shape}},
  {{Spatially-Varying Lighting}} and {{SVBRDF From}} a {{Single Image}}.
\newblock In \emph{2020 {{IEEE}}/{{CVF Conference}} on {{Computer Vision}} and
  {{Pattern Recognition}} ({{CVPR}})}, pages 2472--2481, Seattle, WA, USA,
  2020{\natexlab{b}}. IEEE.

\bibitem[Lindenberger et~al.()Lindenberger, Sarlin, Larsson, and
  Pollefeys]{lindenbergerPixelPerfectStructureFromMotionFeaturemetric}
Philipp Lindenberger, Paul-Edouard Sarlin, Viktor Larsson, and Marc Pollefeys.
\newblock Pixel-{{Perfect Structure-From-Motion With Featuremetric
  Refinement}}.

\bibitem[Liu et~al.(2023)Liu, Wang, Lin, Long, Wang, Liu, Komura, and
  Wang]{liuNeRONeuralGeometry2023}
Yuan Liu, Peng Wang, Cheng Lin, Xiaoxiao Long, Jiepeng Wang, Lingjie Liu, Taku
  Komura, and Wenping Wang.
\newblock {{NeRO}}: {{Neural Geometry}} and {{BRDF Reconstruction}} of
  {{Reflective Objects}} from {{Multiview Images}}.
\newblock \emph{ACM Trans. Graph.}, 42\penalty0 (4):\penalty0 114:1--114:22,
  2023.

\bibitem[LLC(2025)]{llcLibRawLibRaw2025}
LibRaw LLC.
\newblock {{LibRaw}}/{{LibRaw}}, 2025.

\bibitem[Luan et~al.(2021)Luan, Zhao, Bala, and
  Dong]{luanUnifiedShapeSVBRDF2021a}
Fujun Luan, Shuang Zhao, Kavita Bala, and Zhao Dong.
\newblock Unified {{Shape}} and {{SVBRDF Recovery}} using {{Differentiable
  Monte Carlo Rendering}}.
\newblock 2021.

\bibitem[Lyu et~al.(2022)Lyu, Tewari, Leimk{\"u}hler, Habermann, and
  Theobalt]{lyuNeuralRadianceTransfer2022a}
Linjie Lyu, Ayush Tewari, Thomas Leimk{\"u}hler, Marc Habermann, and Christian
  Theobalt.
\newblock Neural {{Radiance Transfer Fields}} for {{Relightable Novel-View
  Synthesis}} with {{Global Illumination}}.
\newblock In \emph{Computer {{Vision}} -- {{ECCV}} 2022}, pages 153--169.
  Springer Nature Switzerland, Cham, 2022.

\bibitem[Mildenhall et~al.(2021)Mildenhall, Srinivasan, Tancik, Barron,
  Ramamoorthi, and Ng]{mildenhallNeRFRepresentingScenes2021}
Ben Mildenhall, Pratul~P. Srinivasan, Matthew Tancik, Jonathan~T. Barron, Ravi
  Ramamoorthi, and Ren Ng.
\newblock {{NeRF}}: Representing scenes as neural radiance fields for view
  synthesis.
\newblock \emph{Commun. ACM}, 65\penalty0 (1):\penalty0 99--106, 2021.

\bibitem[Mildenhall et~al.(2022)Mildenhall, Hedman, {Martin-Brualla},
  Srinivasan, and Barron]{mildenhallNeRFDarkHigh2022}
Ben Mildenhall, Peter Hedman, Ricardo {Martin-Brualla}, Pratul~P. Srinivasan,
  and Jonathan~T. Barron.
\newblock {{NeRF}} in the {{Dark}}: {{High Dynamic Range View Synthesis}} from
  {{Noisy Raw Images}}.
\newblock In \emph{2022 {{IEEE}}/{{CVF Conference}} on {{Computer Vision}} and
  {{Pattern Recognition}} ({{CVPR}})}, pages 16169--16178, New Orleans, LA,
  USA, 2022. IEEE.

\bibitem[M{\"u}ller et~al.(2021)M{\"u}ller, Rousselle, Nov{\'a}k, and
  Keller]{mullerRealtimeNeuralRadiance2021}
Thomas M{\"u}ller, Fabrice Rousselle, Jan Nov{\'a}k, and Alexander Keller.
\newblock Real-time neural radiance caching for path tracing.
\newblock \emph{ACM Trans. Graph.}, 40\penalty0 (4):\penalty0 36:1--36:16,
  2021.

\bibitem[Munkberg et~al.(2022)Munkberg, Chen, Hasselgren, Evans, Shen, Muller,
  Gao, and Fidler]{munkbergExtractingTriangular3D2022}
Jacob Munkberg, Wenzheng Chen, Jon Hasselgren, Alex Evans, Tianchang Shen,
  Thomas Muller, Jun Gao, and Sanja Fidler.
\newblock Extracting {{Triangular 3D Models}}, {{Materials}}, and {{Lighting
  From Images}}.
\newblock In \emph{2022 {{IEEE}}/{{CVF Conference}} on {{Computer Vision}} and
  {{Pattern Recognition}} ({{CVPR}})}, pages 8270--8280, New Orleans, LA, USA,
  2022. IEEE.

\bibitem[Nam et~al.(2018)Nam, Lee, Gutierrez, and
  Kim]{namPracticalSVBRDFAcquisition2018}
Giljoo Nam, Joo~Ho Lee, Diego Gutierrez, and Min~H. Kim.
\newblock Practical {{SVBRDF}} acquisition of {{3D}} objects with unstructured
  flash photography.
\newblock \emph{ACM Trans. Graph.}, 37\penalty0 (6):\penalty0 267:1--267:12,
  2018.

\bibitem[Pandey et~al.(2021)Pandey, Escolano, Legendre, H{\"a}ne, Bouaziz,
  Rhemann, Debevec, and Fanello]{pandeyTotalRelightingLearning2021}
Rohit Pandey, Sergio~Orts Escolano, Chloe Legendre, Christian H{\"a}ne, Sofien
  Bouaziz, Christoph Rhemann, Paul Debevec, and Sean Fanello.
\newblock Total relighting: Learning to relight portraits for background
  replacement.
\newblock \emph{ACM Transactions on Graphics}, 40\penalty0 (4):\penalty0 1--21,
  2021.

\bibitem[Parker et~al.(2010)Parker, Bigler, Dietrich, Friedrich, Hoberock,
  Luebke, McAllister, McGuire, Morley, Robison, and
  Stich]{parkerOptiXGeneralPurpose2010}
Steven~G. Parker, James Bigler, Andreas Dietrich, Heiko Friedrich, Jared
  Hoberock, David Luebke, David McAllister, Morgan McGuire, Keith Morley,
  Austin Robison, and Martin Stich.
\newblock {{OptiX}}: A general purpose ray tracing engine.
\newblock \emph{ACM Transactions on Graphics}, 29\penalty0 (4):\penalty0 1--13,
  2010.

\bibitem[Pharr et~al.(2023)Pharr, Jakob, and
  Humphreys]{pharrPhysicallyBasedRendering2023}
Matt Pharr, Wenzel Jakob, and Greg Humphreys.
\newblock \emph{Physically Based Rendering: From Theory to Implementation}.
\newblock The MIT Press, Cambridge London, fourth edition edition, 2023.

\bibitem[Ravi et~al.(2025)Ravi, Gabeur, Hu, Hu, Ryali, Ma, Khedr, R{\"a}dle,
  Rolland, Gustafson, Mintun, Girshick, Doll{\'a}r, and
  Feichtenhofer]{raviSAM2SEGMENT2025}
Nikhila Ravi, Valentin Gabeur, Yuan-Ting Hu, Ronghang Hu, Chaitanya Ryali,
  Tengyu Ma, Haitham Khedr, Roman R{\"a}dle, Chloe Rolland, Laura Gustafson,
  Eric Mintun, Ross Girshick, Piotr Doll{\'a}r, and Christoph Feichtenhofer.
\newblock {{SAM}} 2: {{SEGMENT ANYTHING IN IMAGES AND VIDEOS}}.
\newblock 2025.

\bibitem[Riechert(2025)]{riechertLetmaikRawpy2025}
Maik Riechert.
\newblock Letmaik/rawpy, 2025.

\bibitem[Schmitt et~al.()Schmitt, Donne, Riegler, Koltun, and
  Geiger]{schmittJointEstimationPose}
Carolin Schmitt, Simon Donne, Gernot Riegler, Vladlen Koltun, and Andreas
  Geiger.
\newblock On {{Joint Estimation}} of {{Pose}}, {{Geometry}} and {{svBRDF From}}
  a {{Handheld Scanner}}.

\bibitem[Schmitt et~al.(2020)Schmitt, Donne, Riegler, Koltun, and
  Geiger]{schmittJointEstimationPose2020}
Carolin Schmitt, Simon Donne, Gernot Riegler, Vladlen Koltun, and Andreas
  Geiger.
\newblock On {{Joint Estimation}} of {{Pose}}, {{Geometry}} and {{svBRDF From}}
  a {{Handheld Scanner}}.
\newblock In \emph{2020 {{IEEE}}/{{CVF Conference}} on {{Computer Vision}} and
  {{Pattern Recognition}} ({{CVPR}})}, pages 3490--3500, Seattle, WA, USA,
  2020. IEEE.

\bibitem[Sengupta et~al.(2019)Sengupta, Gu, Kim, Liu, Jacobs, and
  Kautz]{senguptaNeuralInverseRendering2019}
Soumyadip Sengupta, Jinwei Gu, Kihwan Kim, Guilin Liu, David~W. Jacobs, and Jan
  Kautz.
\newblock Neural {{Inverse Rendering}} of an {{Indoor Scene}} from a {{Single
  Image}}.
\newblock \emph{arXiv:1901.02453 [cs]}, 2019.

\bibitem[Sun et~al.(2023)Sun, Cai, Li, Yan, Zhang, Marshall, Huang, Zhao, and
  Dong]{sunNeuralPBIRReconstructionShape2023}
Cheng Sun, Guangyan Cai, Zhengqin Li, Kai Yan, Cheng Zhang, Carl Marshall,
  Jia-Bin Huang, Shuang Zhao, and Zhao Dong.
\newblock Neural-{{PBIR Reconstruction}} of {{Shape}}, {{Material}}, and
  {{Illumination}}.
\newblock In \emph{2023 {{IEEE}}/{{CVF International Conference}} on {{Computer
  Vision}} ({{ICCV}})}, pages 18000--18010, Paris, France, 2023. IEEE.

\bibitem[Tuytel et~al.()Tuytel, {Greg Zaal}, et~al.]{tuytelPolyHavenPoly}
Rob Tuytel, {Greg Zaal}, et~al.
\newblock Poly {{Haven}} {$\bullet$} {{Poly Haven}}.
\newblock https://polyhaven.com/.

\bibitem[Wang et~al.(2024)Wang, Hu, Zhu, and
  Lau]{wangInverseRenderingGlossy2024}
Haoyuan Wang, Wenbo Hu, Lei Zhu, and Rynson~W.H. Lau.
\newblock Inverse {{Rendering}} of {{Glossy Objects}} via the {{Neural
  Plenoptic Function}} and {{Radiance Fields}}.
\newblock In \emph{2024 {{IEEE}}/{{CVF Conference}} on {{Computer Vision}} and
  {{Pattern Recognition}} ({{CVPR}})}, pages 19999--20008, Seattle, WA, USA,
  2024. IEEE.

\bibitem[Wang et~al.(2021)Wang, Liu, Liu, Theobalt, Komura, and
  Wang]{wangNeuSLearningNeural2021}
Peng Wang, Lingjie Liu, Yuan Liu, Christian Theobalt, Taku Komura, and Wenping
  Wang.
\newblock {{NeuS}}: {{Learning Neural Implicit Surfaces}} by {{Volume
  Rendering}} for {{Multi-view Reconstruction}}.
\newblock In \emph{Advances in {{Neural Information Processing Systems}}},
  pages 27171--27183. Curran Associates, Inc., 2021.

\bibitem[{Wenzel Jakob} et~al.(){Wenzel Jakob}, {S\'ebastien Speierer},
  {Nicolas Roussel}, {Merlin Nimier-David}, {Delio Vicini}, {Tizian Zeltner},
  {Baptiste Nicolet}, {Miguel Crespo}, {Vincent Leroy}, and {Ziyi
  Zhang}]{wenzeljakobMitsuba3Renderer}
{Wenzel Jakob}, {S\'ebastien Speierer}, {Nicolas Roussel}, {Merlin
  Nimier-David}, {Delio Vicini}, {Tizian Zeltner}, {Baptiste Nicolet}, {Miguel
  Crespo}, {Vincent Leroy}, and {Ziyi Zhang}.
\newblock Mitsuba 3 renderer.

\bibitem[Wu et~al.(2023{\natexlab{a}})Wu, Hu, Li, Zhang, Fan, and
  Yu]{wuNeFIIInverseRendering2023}
Haoqian Wu, Zhipeng Hu, Lincheng Li, Yongqiang Zhang, Changjie Fan, and Xin Yu.
\newblock {{NeFII}}: {{Inverse Rendering}} for {{Reflectance Decomposition}}
  with {{Near-Field Indirect Illumination}}.
\newblock In \emph{2023 {{IEEE}}/{{CVF Conference}} on {{Computer Vision}} and
  {{Pattern Recognition}} ({{CVPR}})}, pages 4295--4304, Vancouver, BC, Canada,
  2023{\natexlab{a}}. IEEE.

\bibitem[Wu et~al.(2023{\natexlab{b}})Wu, Chowdhury, Shanmugaraja, Jacobs, and
  Sengupta]{wuMeasuredAlbedoWild2023}
Jiaye Wu, Sanjoy Chowdhury, Hariharmano Shanmugaraja, David Jacobs, and
  Soumyadip Sengupta.
\newblock Measured {{Albedo}} in the {{Wild}}: {{Filling}} the {{Gap}} in
  {{Intrinsics Evaluation}}.
\newblock In \emph{2023 {{IEEE International Conference}} on {{Computational
  Photography}} ({{ICCP}})}, pages 1--12, 2023{\natexlab{b}}.

\bibitem[Wu et~al.(2023{\natexlab{c}})Wu, Zhu, Yaldiz, Zhu, Cai, Matai,
  Porikli, Li, Chandraker, and Ramamoorthi]{wuFactorizedInversePath2023}
Liwen Wu, Rui Zhu, Mustafa~B. Yaldiz, Yinhao Zhu, Hong Cai, Janarbek Matai,
  Fatih Porikli, Tzu-Mao Li, Manmohan Chandraker, and Ravi Ramamoorthi.
\newblock Factorized {{Inverse Path Tracing}} for {{Efficient}} and {{Accurate
  Material-Lighting Estimation}}.
\newblock In \emph{2023 {{IEEE}}/{{CVF International Conference}} on {{Computer
  Vision}} ({{ICCV}})}, pages 3825--3835, Paris, France, 2023{\natexlab{c}}.
  IEEE.

\bibitem[Yariv et~al.(2020)Yariv, Kasten, Moran, Galun, Atzmon, Ronen, and
  Lipman]{yarivMultiviewNeuralSurface2020}
Lior Yariv, Yoni Kasten, Dror Moran, Meirav Galun, Matan Atzmon, Basri Ronen,
  and Yaron Lipman.
\newblock Multiview {{Neural Surface Reconstruction}} by {{Disentangling
  Geometry}} and {{Appearance}}.
\newblock In \emph{Advances in {{Neural Information Processing Systems}}},
  pages 2492--2502. Curran Associates, Inc., 2020.

\bibitem[Yariv et~al.(2021)Yariv, Gu, Kasten, and
  Lipman]{yarivVolumeRenderingNeural2021}
Lior Yariv, Jiatao Gu, Yoni Kasten, and Yaron Lipman.
\newblock Volume {{Rendering}} of {{Neural Implicit Surfaces}}.
\newblock In \emph{Advances in {{Neural Information Processing Systems}}},
  pages 4805--4815. Curran Associates, Inc., 2021.

\bibitem[Zeng et~al.(2023)Zeng, Chen, Dong, Peers, Wu, and
  Tong]{zengRelightingNeuralRadiance2023}
Chong Zeng, Guojun Chen, Yue Dong, Pieter Peers, Hongzhi Wu, and Xin Tong.
\newblock Relighting {{Neural Radiance Fields}} with {{Shadow}} and {{Highlight
  Hints}}.
\newblock In \emph{{{ACM SIGGRAPH}} 2023 {{Conference Proceedings}}}, pages
  1--11, New York, NY, USA, 2023. Association for Computing Machinery.

\bibitem[Zhang et~al.(2023{\natexlab{a}})Zhang, Yao, Li, Liu, Fang, McKinnon,
  Tsin, and Quan]{zhangNeILFInterReflectableLight2023}
Jingyang Zhang, Yao Yao, Shiwei Li, Jingbo Liu, Tian Fang, David McKinnon,
  Yanghai Tsin, and Long Quan.
\newblock {{NeILF}}++: {{Inter-Reflectable Light Fields}} for {{Geometry}} and
  {{Material Estimation}}.
\newblock In \emph{2023 {{IEEE}}/{{CVF International Conference}} on {{Computer
  Vision}} ({{ICCV}})}, pages 3578--3587, Paris, France, 2023{\natexlab{a}}.
  IEEE.

\bibitem[Zhang et~al.(2021{\natexlab{a}})Zhang, Luan, Wang, Bala, and
  Snavely]{zhangPhySGInverseRendering2021a}
Kai Zhang, Fujun Luan, Qianqian Wang, Kavita Bala, and Noah Snavely.
\newblock {{PhySG}}: {{Inverse Rendering}} with {{Spherical Gaussians}} for
  {{Physics-based Material Editing}} and {{Relighting}}.
\newblock In \emph{2021 {{IEEE}}/{{CVF Conference}} on {{Computer Vision}} and
  {{Pattern Recognition}} ({{CVPR}})}, pages 5449--5458, Nashville, TN, USA,
  2021{\natexlab{a}}. IEEE.

\bibitem[Zhang et~al.(2022{\natexlab{a}})Zhang, Luan, Li, and
  Snavely]{zhangIRONInverseRendering2022a}
Kai Zhang, Fujun Luan, Zhengqi Li, and Noah Snavely.
\newblock {{IRON}}: {{Inverse Rendering}} by {{Optimizing Neural SDFs}} and
  {{Materials}} from {{Photometric Images}}.
\newblock In \emph{2022 {{IEEE}}/{{CVF Conference}} on {{Computer Vision}} and
  {{Pattern Recognition}} ({{CVPR}})}, pages 5555--5564, New Orleans, LA, USA,
  2022{\natexlab{a}}. IEEE.

\bibitem[Zhang et~al.(2021{\natexlab{b}})Zhang, Fanello, Tsai, Sun, Xue,
  Pandey, {Orts-Escolano}, Davidson, Rhemann, Debevec, Barron, Ramamoorthi, and
  Freeman]{zhangNeuralLightTransport2021}
Xiuming Zhang, Sean Fanello, Yun-Ta Tsai, Tiancheng Sun, Tianfan Xue, Rohit
  Pandey, Sergio {Orts-Escolano}, Philip Davidson, Christoph Rhemann, Paul
  Debevec, Jonathan~T. Barron, Ravi Ramamoorthi, and William~T. Freeman.
\newblock Neural {{Light Transport}} for {{Relighting}} and {{View Synthesis}}.
\newblock \emph{ACM Transactions on Graphics}, 40\penalty0 (1):\penalty0 1--17,
  2021{\natexlab{b}}.

\bibitem[Zhang et~al.(2021{\natexlab{c}})Zhang, Srinivasan, Deng, Debevec,
  Freeman, and Barron]{zhangNeRFactorNeuralFactorization2021}
Xiuming Zhang, Pratul~P. Srinivasan, Boyang Deng, Paul Debevec, William~T.
  Freeman, and Jonathan~T. Barron.
\newblock {{NeRFactor}}: Neural factorization of shape and reflectance under an
  unknown illumination.
\newblock \emph{ACM Trans. Graph.}, 40\penalty0 (6):\penalty0 237:1--237:18,
  2021{\natexlab{c}}.

\bibitem[Zhang et~al.(2022{\natexlab{b}})Zhang, Sun, He, Fu, Jia, and
  Zhou]{zhangModelingIndirectIllumination2022}
Yuanqing Zhang, Jiaming Sun, Xingyi He, Huan Fu, Rongfei Jia, and Xiaowei Zhou.
\newblock Modeling {{Indirect Illumination}} for {{Inverse Rendering}}.
\newblock In \emph{2022 {{IEEE}}/{{CVF Conference}} on {{Computer Vision}} and
  {{Pattern Recognition}} ({{CVPR}})}, pages 18622--18631, New Orleans, LA,
  USA, 2022{\natexlab{b}}. IEEE.

\bibitem[Zhang et~al.(2023{\natexlab{b}})Zhang, Xu, Yu, Ye, Jing, Wang, Yu, and
  Yang]{zhangNeMFInverseVolume2023}
Youjia Zhang, Teng Xu, Junqing Yu, Yuteng Ye, Yanqing Jing, Junle Wang, Jingyi
  Yu, and Wei Yang.
\newblock {{NeMF}}: {{Inverse Volume Rendering}} with {{Neural Microflake
  Field}}.
\newblock In \emph{2023 {{IEEE}}/{{CVF International Conference}} on {{Computer
  Vision}} ({{ICCV}})}, pages 22862--22872, Paris, France, 2023{\natexlab{b}}.
  IEEE.

\bibitem[Zhao et~al.(2023)Zhao, Lichy, Perrin, Frahm, and
  Sengupta]{zhaoMVPSNetFastGeneralizable2023}
Dongxu Zhao, Daniel Lichy, Pierre-Nicolas Perrin, Jan-Michael Frahm, and
  Soumyadip Sengupta.
\newblock {{MVPSNet}}: {{Fast Generalizable Multi-view Photometric Stereo}}.
\newblock In \emph{2023 {{IEEE}}/{{CVF International Conference}} on {{Computer
  Vision}} ({{ICCV}})}, pages 12491--12502, Paris, France, 2023. IEEE.

\bibitem[Zhou et~al.(2013)Zhou, Wu, and
  Tan]{zhouMultiviewPhotometricStereo2013}
Zhenglong Zhou, Zhe Wu, and Ping Tan.
\newblock Multi-view {{Photometric Stereo}} with {{Spatially Varying Isotropic
  Materials}}.
\newblock In \emph{2013 {{IEEE Conference}} on {{Computer Vision}} and
  {{Pattern Recognition}}}, pages 1482--1489, Portland, OR, USA, 2013. IEEE.

\bibitem[Zhu et~al.(2022)Zhu, Luan, Huo, Lin, Zhong, Xi, Wang, Bao, Zheng, and
  Tang]{zhuLearningbasedInverseRendering2022}
Jingsen Zhu, Fujun Luan, Yuchi Huo, Zihao Lin, Zhihua Zhong, Dianbing Xi, Rui
  Wang, Hujun Bao, Jiaxiang Zheng, and Rui Tang.
\newblock Learning-based {{Inverse Rendering}} of {{Complex Indoor Scenes}}
  with {{Differentiable Monte Carlo Raytracing}}.
\newblock In \emph{{{SIGGRAPH Asia}} 2022 {{Conference Papers}}}, pages 1--8,
  New York, NY, USA, 2022. Association for Computing Machinery.

\end{thebibliography}
